\pdfoutput=1
\documentclass[10pt,twocolumn,letterpaper]{article}

\usepackage{arxiv}
\usepackage{times}
\usepackage{epsfig}
\usepackage{graphicx}
\usepackage{amsmath}
\usepackage{amssymb}
\usepackage{appendix}

\arxivfinalcopy % *** Uncomment this line for the final submission

\usepackage{booktabs}
\usepackage{multirow}
\usepackage{wrapfig}

\usepackage[pagebackref=true,breaklinks=true,letterpaper=true,colorlinks,bookmarks=false]{hyperref}

\usepackage[capitalize]{cleveref}
\crefname{section}{Sec.}{Secs.}
\Crefname{section}{Section}{Sections}
\Crefname{table}{Table}{Tables}
\crefname{table}{Tab.}{Tabs.}

\ifarxivfinal\fi

\begin{document}

%%%%%%%%% TITLE
\title{Deformable Model-Driven Neural Rendering for High-Fidelity 3D Reconstruction of Human Heads Under Low-View Settings}

\author{Baixin Xu$^1$\thanks{Project page: \href{https://github.com/xubaixinxbx/3dheads}{https://github.com/xubaixinxbx/3dheads}.} \quad
Jiarui Zhang$^2$ \quad 
Kwan-Yee Lin $^{3,4}$ \quad
Chen Qian $^5$\quad
Ying He$^1\thanks{Corresponding author: Y. He (yhe@ntu.edu.sg).}$\\
$^1$ S-Lab, Nanyang Technological University \quad
$^2$ Peking University \quad\\
$^3$ The Chinese University of Hong Kong \quad
$^4$ Shanghai Artificial Intelligence Laboratory \quad\\
$^5$ SenseTime Research \\
}

\maketitle
\ifarxivfinal\thispagestyle{empty}\fi

%%%%%%%%% ABSTRACT
\begin{abstract}

Reconstructing 3D human heads in low-view settings presents  technical challenges, mainly due to the pronounced risk of overfitting with limited views and high-frequency signals. To address this, we propose geometry decomposition and adopt a two-stage, coarse-to-fine training strategy, allowing for progressively capturing high-frequency geometric details. We represent 3D human heads using the zero level-set of a combined signed distance field, comprising a smooth template, a non-rigid deformation, and a high-frequency displacement field. The template captures features that are independent of both identity and expression and is co-trained with the deformation network across multiple individuals with sparse and randomly selected views. The displacement field, capturing individual-specific details, undergoes separate training for each person. Our network training does not require 3D supervision or object masks. Experimental results demonstrate the effectiveness and robustness of our geometry decomposition and two-stage training strategy. Our method outperforms existing neural rendering approaches in terms of reconstruction accuracy and novel view synthesis under low-view settings. Moreover, the pre-trained template serves a good initialization for our model when encountering unseen individuals. 
\end{abstract}

\section{Introduction}
\label{sec:intro}
Accurately modeling and rendering human heads is crucial for various digital human-related applications as the head is one of the most distinguishing features that helps us identify individuals. Neural implicit functions~\cite{oechsle2021unisurf,wang2021neus,yariv2021volume,wang2022hf,fu2022geo,zheng2022avatar,darmon2022improving} have recently emerged as a promising technique for synthesizing novel views of complex objects, including human heads, by learning a continuous function from multi-view images without being tied to a specific resolution. However, training such deep learning models often requires a large number of images as input, which can be costly and computationally inefficient. Moreover, a single implicit field may not generalize well to unseen heads, particularly under setting of low or sparse views.

This paper aims to develop a robust method for learning neural implicit functions that can accurately reconstruct 3D human heads with high-fidelity geometry and appearance from low-view inputs, thereby reducing the need of extensive data collection and annotation. To achieve this goal, we learn a signed distance field (SDF) that consists of a smooth template, a non-rigid deformation, and a high-frequency displacement map. The template represents identity-independent and expression-neutral features, while the deformation and displacement maps encode identity-dependent geometric details that are trained for each specific individual. We represent 3D human heads as the zero level-set of the composed SDF. Our training involves two stages. The first stage takes the whole set of persons as input, and learns an identity-independent and expression-neutral template head and a non-rigid deformation between each observed head and the template head. In the second stage, we learn an identity-dependent displacement map for further refining the geometry. To train the proposed SDF without any 3D supervision, we adopt a volume rendering scheme~\cite{oechsle2021unisurf,yariv2021volume,wang2021neus} to minimize the difference between the rendered colors and the ground-truth colors. We also adopt regularization terms for the SDF, the deformation, the displacement map and the latent codes.

We evaluate our approach on both senior and young persons and demonstrate that it is robust and effective in learning SDFs for both types of inputs, resulting in 3D surfaces with high-fidelity geometry. Our experiments show that our method outperforms existing methods in terms of reconstruction accuracy and visual quality. We also demonstrate that our method can generalize to unseen identities by using the pre-trained template as a good initialization. 

  \begin{figure*}[!htbp]
      \centering
      \includegraphics[height=1.6in,trim={20 0 10 0},clip]{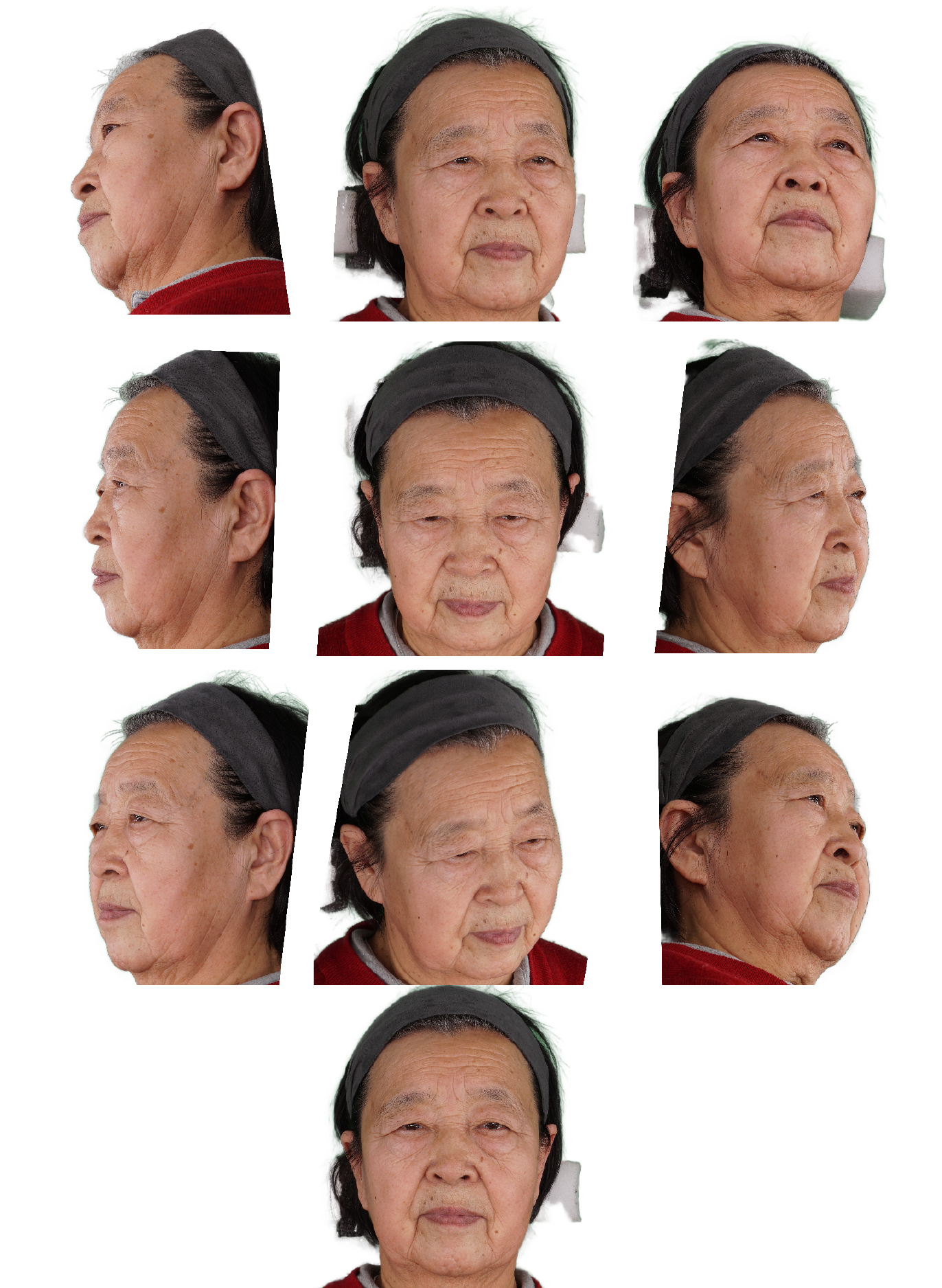} \hspace{-0.1in}
      \includegraphics[height=1.6in, trim={250 0 310 0},clip]{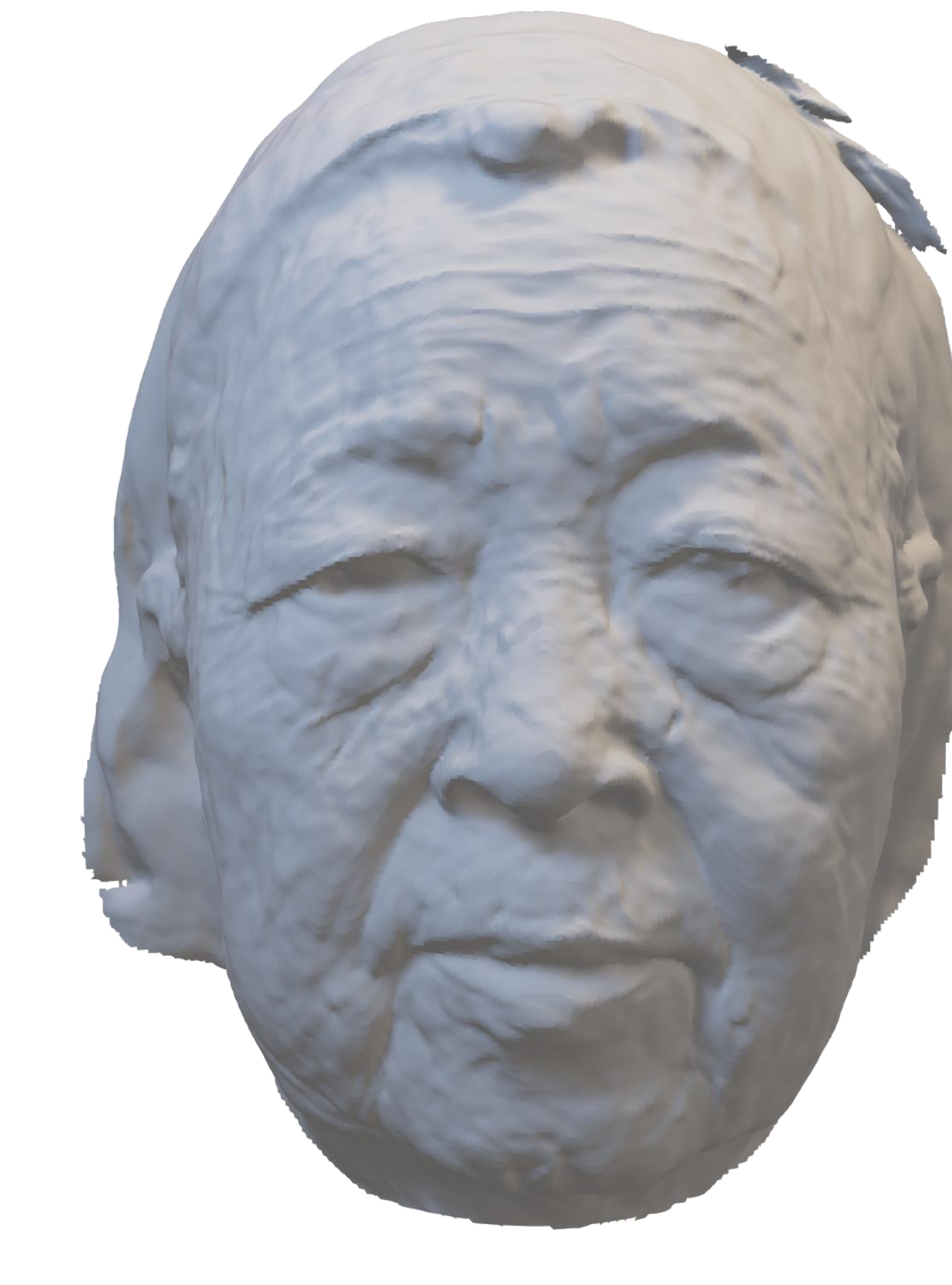}
      \includegraphics[height=1.6in,trim={200 0 120 20},clip]{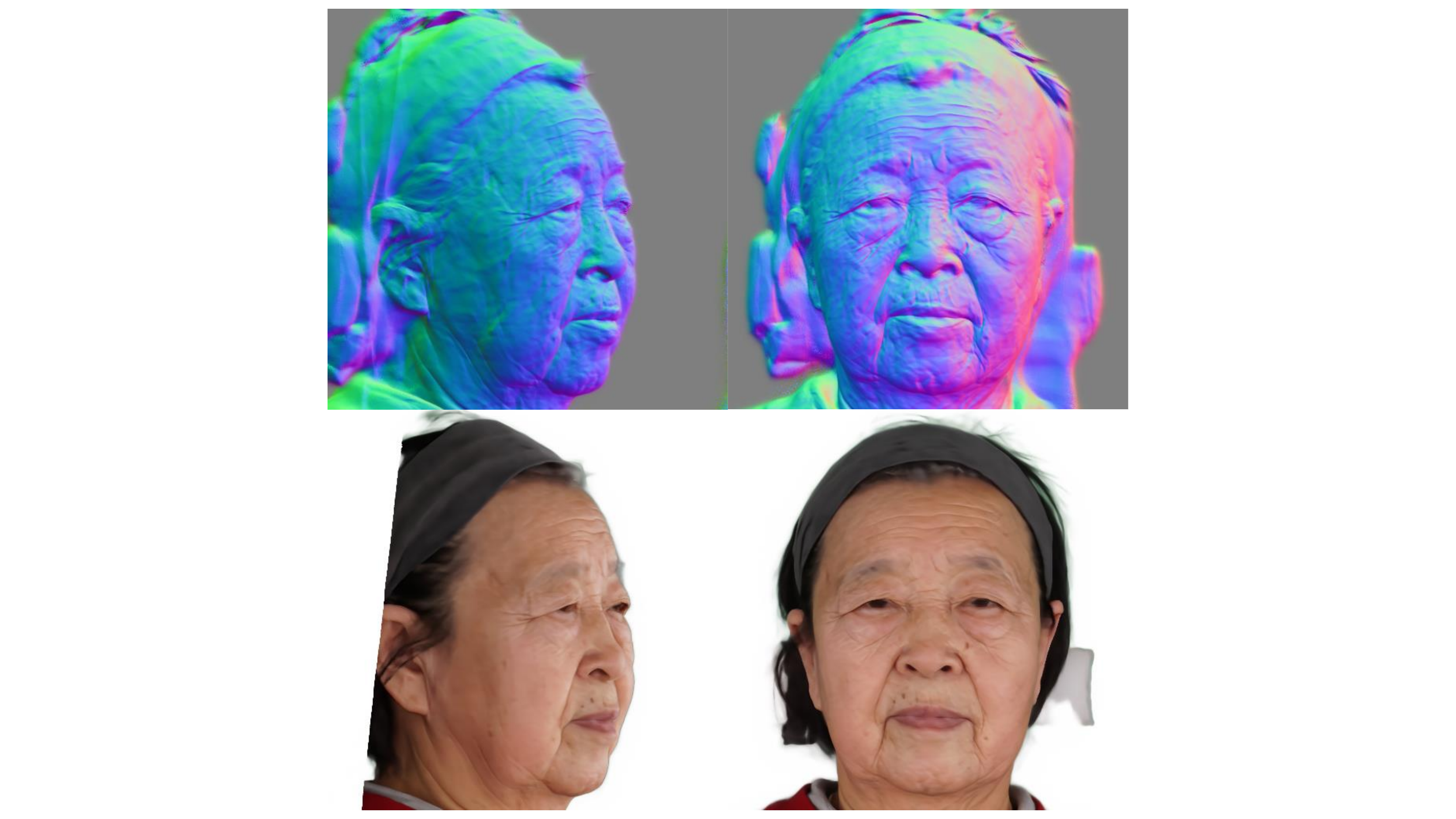}\hspace{-0.2in}
      \includegraphics[height=1.6in,trim={60 0 120 20},clip]{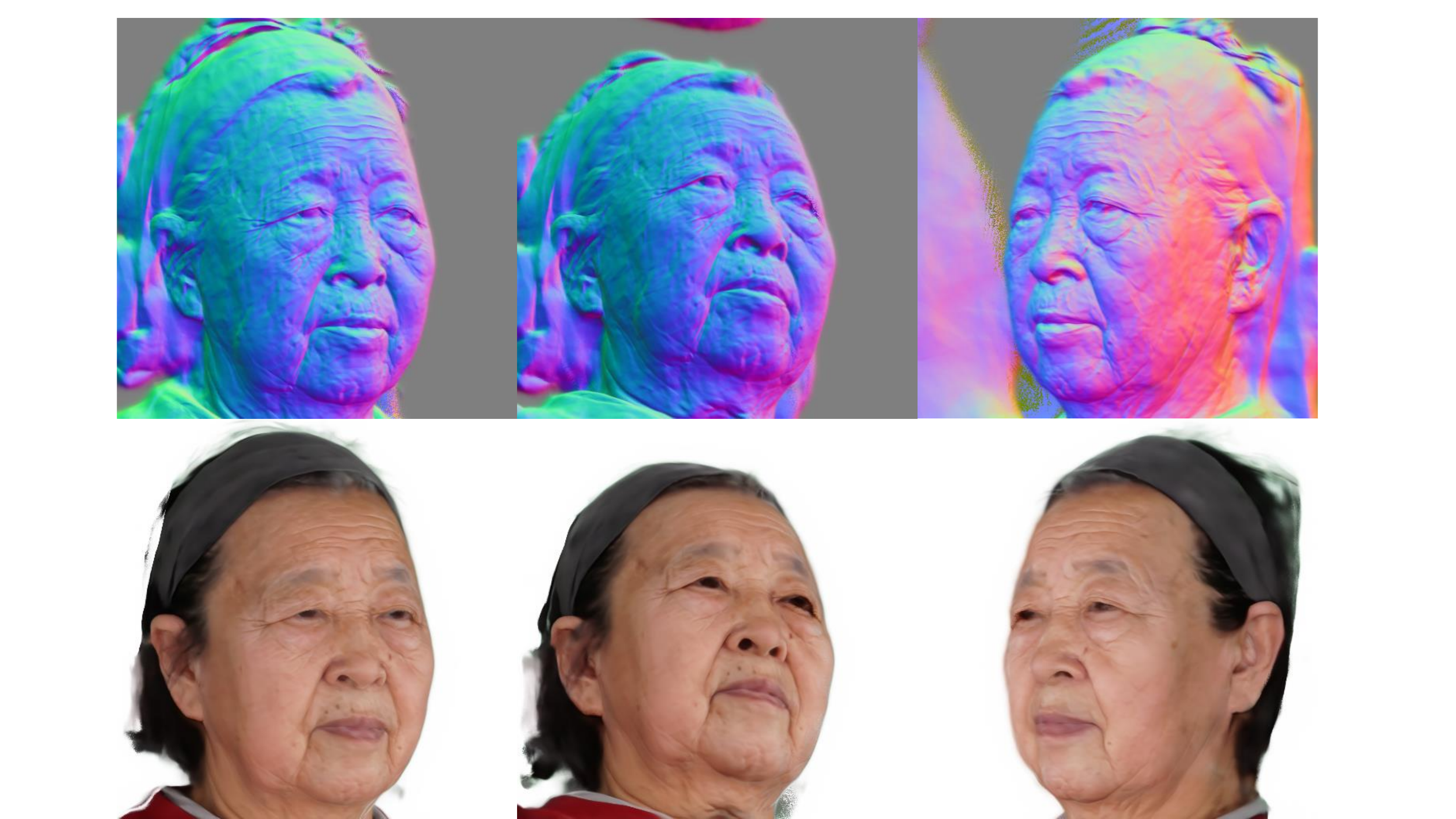}\\     
      \makebox[1.1in]{Input}
      \makebox[1.1in]{3D head}
      \makebox[1.6in]{Training views} 
      \makebox[2.5in]{Novel views}\\
      %\vspace{-0.1in}
      \caption{An example of a 3D head reconstructed by our method using 10 input views. Refer also to the supplementary material for a video demonstration.
      }
      \label{fig:teaser}
  \end{figure*}

\section{Related Work}
\label{sec:related work}

\noindent\textbf{Multi-view 3D reconstruction} is a classical problem in computer vision, and traditional approaches can be divided into voxel-based~\cite{seitz1999photorealistic, kutulakos2000theory, de1999poxels} and point-based methods~\cite{galliani2015massively, barnes2009patchmatch,schonberger2016pixelwise,schonberger2016structure}. 
Voxel-based methods divide the 3D space into voxels and determine which ones belong to the object. However, due to the cubic space complexity, these methods are computationally expensive and may not be suitable for reconstructing complex objects. Point-based methods are memory efficient, but they lack connectivity information, which can lead to incomplete or inaccurate reconstructions.

\noindent\textbf{Neural radiance fields}~\cite{mildenhall2021nerf,zhang2020nerf++, barron2021mip} have demonstrated remarkable results in representing complex 3D scenes from only 2D images as input. However, due to the discrete nature of the radiance field, their geometric reconstructions often suffer from inaccuracies. To improve the capability of geometry reconstruction, several techniques have been proposed to replace the density-based representation by a geometry-oriented representation. For example, UNISURF~\cite{oechsle2021unisurf} uses occupancy fields, while NeuS~\cite{wang2021neus} and VolSDF~\cite{yariv2021volume} use signed distance fields. 

\noindent\textbf{Neural implicit representations} have garnered significant attention in 3D reconstruction due to their representation power and memory efficiency~\cite{park2019deepsdf, sitzmann2019scene}. DVR~\cite{niemeyer2020differentiable} and IDR~\cite{yariv2020multiview} focus on making the surface rendering pipeline differentiable. These methods often require masks to distinguish objects from the background. Recent works focus on improving the applicability and representation capability of neural implicit functions.
Geo-NeuS~\cite{fu2022geo} introduces multi-view constraints from structure from motion to encourage SDF networks to avoid geometry ambiguity under rendering loss. D2IM-Net~\cite{li2021d2im} adopts an implicit displacement field for recovering geometric details from a single input image. 
IDF~\cite{yifan2021geometry} represents complex surfaces as a smooth base surface and a displacement mapping.
HF-NeuS~\cite{wang2022hf} decomposes the SDF into a base function and a displacement function with a coarse-to-fine strategy to gradually increase high-frequency details. DIF~\cite{deng2021deformed} deforms a target object to match the template shape and employs a correction field for handling topological changes.

\noindent\textbf{3D morphable face model} (3DMM)~\cite{blanz1999morphable} has been extensively used in 3D face reconstruction and animation. It is a statistical model that represents the shape and texture of a human face using a low-dimensional vector space. 
The combination of 3DMM with deep learning has proven effective in producing high-quality 3D faces~\cite{li2022implicit,zheng2022imface}, however it often requires 3D supervision to achieve good performance. Recent works also aim to improve the representation capability of 3DMM to the entire head, using multi-view images~\cite{yenamandra2021i3dmm,hong2022headnerf,zhuang2022mofanerf, wang2022morf} and monocular videos~\cite{zheng2022avatar}.
We refer the readers to a comprehensive survey of 3D morphable model and recent developments~\cite{egger20203d}.

\section{Method}
\label{sec:method}

\subsection{Overview}

We have a set of RGB portrait images $\mathcal{I} = \{I_{i}\}$ for $m$ individuals. Each individual is captured from  $k$ different viewpoints covering the front, left, and right sides. Additionally, each image is  associated with camera parameters $\pi_i$. Our goal is to learn a signed distance field for the 3D geometry of the human head and a radiance field for colors for each individual. To obtain high-fidelity geometry and appearance, our method decomposes the geometric representation of 3D human heads into identity-independent and -dependent components. 

The ID-independent component is a smooth base surface that represents the common geometric characteristics of human heads and serves as a template. Since the ID-independent and expression-neutral template is the standard reference for all individuals. 
The ID-dependent components include 
a non-rigid deformation that maps each individual head to the template and a displacement field that encodes high-frequency geometric details such as wrinkles and small facial features that are particular to a given person. 

\begin{figure*}[htbp]
    \centering
    \includegraphics[width=\textwidth]{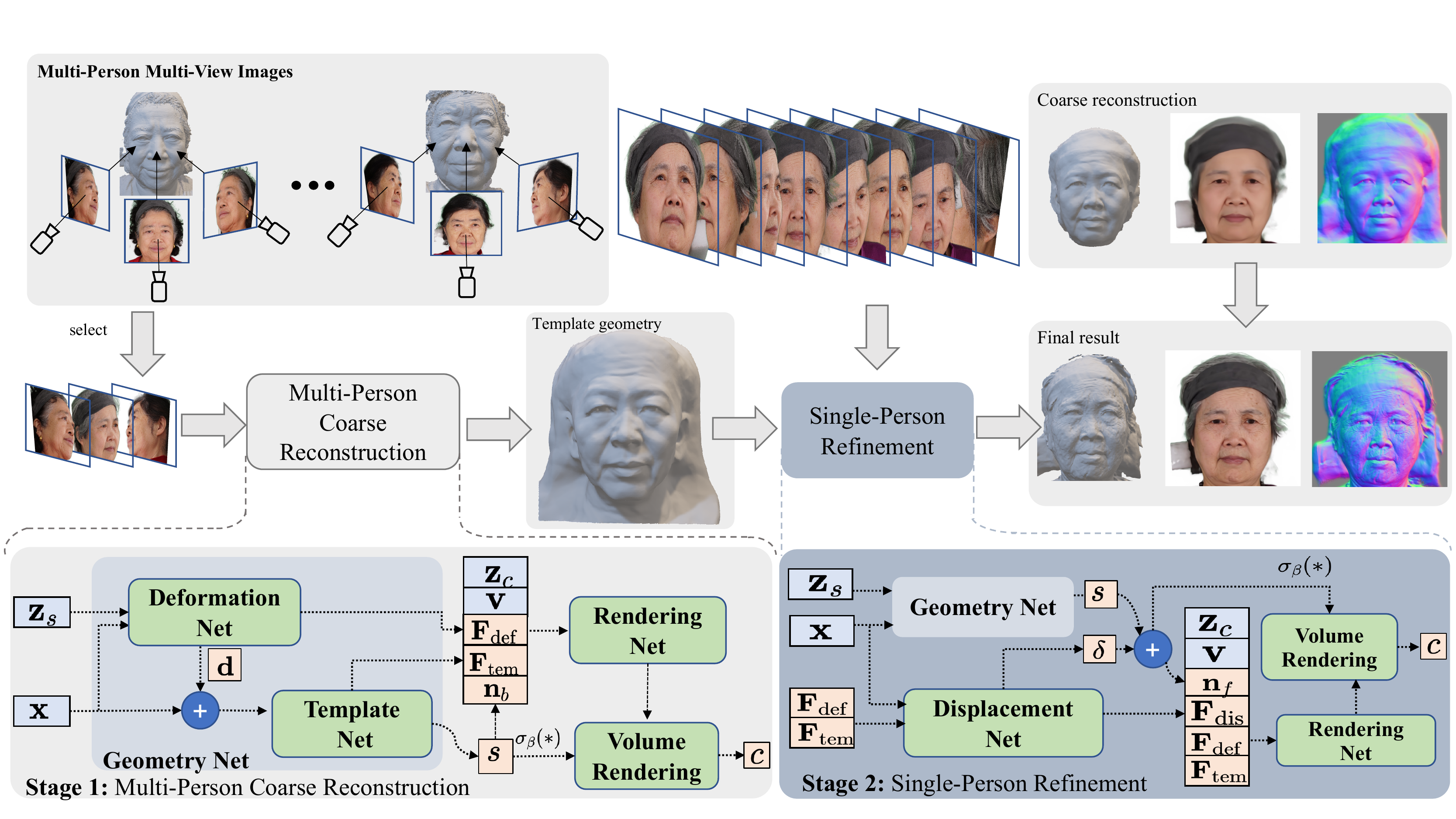}
    %\vspace{-0.1in} 3.2inch for single column
    \caption{Data flow and network architecture. Our model reconstructs 3D geometry and colors of human heads in a two-stage, coarse-to-fine manner. In the first stage, we optimize the Template Network and the Deformation Network, along with the latent code space and Rendering Network, across different identities to obtain an identity-independent and expression neutral template. In the second stage, we introduce a Displacement Network and further optimize it with a set of portrait images of a specific subject. }
    \label{fig:pipelineOur}
\end{figure*}

As shown in Figure~\ref{fig:pipelineOur}, our method employs a two-stage training framework for reconstructing 3D geometry and colors in a coarse-to-fine manner. The first stage is to train a geometry network for learning the template and a point-wise non-rigid deformation between an individual head and  the template head. In the second stage, we train a displacement network for learning identity-related fine geometric details to further improve the reconstruction quality. 

\subsection{Multi-Person Coarse Reconstruction}

\noindent\textbf{Geometry network.} To learn a template head from multiple individuals, we follow i3DMM~\cite{yenamandra2021i3dmm} to design the Geometry Network $f_{\text{geo}}$ with two components: a Template Network $f_{\text{tem}}$ and a Deformation Network $f_{\text{def}}$. The Template Network is designed to obtain common features across all identities, and its output is a mean face of all identities used in training. 
On the other hand, the Deformation Network aims to find correspondences from each specific subject to the mean face and learn features associated with the identity.

To model each identity separately, we introduce a shape code $\mathbf{z}_{s}$ and a color code $\mathbf{z}_{c}$, which are learned in the shape and color embedding spaces, respectively. Specifically, with the shape code $\mathbf{z}_{s}$
representing an identity, the Geometry Network $f_{\text{geo}}$ learns a global SDF for each 3D point $\mathbf{x} \in \mathbb{R}^{3}$ and generates a geometric feature $\mathbf{F}_{\text{geo}}$ for $\mathbf{x}$ that will be used to learn the displacement map and radiance information.
Then the surface normal $\mathbf{n}_b$, viewing direction $\mathbf{v}$, and geometric feature $\mathbf{F}_{\text{geo}}$ at $\mathbf{x}$ are fed into the Rendering Network $f_{\text{ren}}$, which predicts the radiance for the ray. The output of Stage 1 is an identity-independent and expression-neutral template. 

Specifically, given a query point $\mathbf{x} \in \mathbb{R}^3$ in the observation space and an identity-related latent code for geometry $\mathbf{z}_s\in\mathbb{R}^{128}$, the Deformation Network $f_\text{def}$ predicts an offset vector $\mathbf{d} \in \mathbb{R}^3 $ that maps $\bf x$ to the template space and also outputs an ID-dependent feature $\mathbf{F}_{\text{def}} \in \mathbb{R}^{192}$
\begin{equation}
     f_{\text{def}}(\mathbf{x}, \mathbf{z}_{s})=(\mathbf{d}, \mathbf{F}_{\text{def}})\label{eq:deform}.
\end{equation}    
Then the Template Network returns a signed distance value $s \in \mathbb{R}$ for the deformed position $\mathbf{x+d}$ in the template space and a template feature $\mathbf{F}_{\text{tem}} \in \mathbf{R}^{64}$ for the deformed position $\mathbf{x}+\mathbf{d}$
\begin{equation}
f_{\text{tem}}(\mathbf{x} + \mathbf{d})=(s, \mathbf{F}_{\text{tem}}).\label{eq:template}
\end{equation}
The identity-independent template feature $\mathbf{F}_{\text{tem}}$ will be concatenated with the 
ID-dependent deformation feature $\mathbf{F}_{\text{def}}$ and used in the subsequent Rendering Network.
Putting it all together, we formulate the Geometry Network as follows
\begin{equation}
    f_{\text{geo}} = f_{\text{tem}} \circ f_{\text{def}}(\mathbf{x}, \mathbf{z}_{s}) \label{eq:f-geo-first-stage}.
\end{equation}

\noindent\textbf{Rendering network.} To learn the SDF $s$ from a set of 2D images $\mathcal{I}$ with camera parameters, we cast rays from the camera position $\bf o$ to each pixel of the input images. Consider a ray $\mathbf{r}(t)=\mathbf{o}+t\mathbf{v}$, where $\bf v$ is a unit vector and $t\geq 0$ is the distance from the camera center.
The Rendering Network, denoted as $f_{\text{ren}}$, 
is responsible for computing the radiance at any query point $\mathbf{x}\in \mathbf{r}(t)$. 

To consider the variety of textures among different persons in the dataset, the network also takes an ID-dependent latent code for color $\mathbf{z}_c$, the template feature $\mathbf{F}_{\text{tem}}$, the deformation feature $\mathbf{F}_{\text{def}}$, and the normal of the base surface $\mathbf{n}_b$ as input~\cite{liu2021editing} and outputs the radiance $c\in\mathbb{R}^{3}$
\begin{align}
    f_{\text{ren}}(\mathbf{z}_{c}, \mathbf{x}, \mathbf{v}, \text{concat}(\mathbf{F}_{\text{def}},\mathbf{F}_{\text{tem}}), \mathbf{n}_b)=c,\label{eq:render1}
\end{align}
where $\mathbf{n}_b=\nabla s$ is the gradient of the SDF of the base surface given an individual, representing the normal direction.

To compute the radiance, we first transform the SDF into an S-density $\sigma$ as in~\cite{yariv2021volume}
\begin{align}
    \sigma(\mathbf{x}) = \alpha \Phi_\beta(-s(\mathbf{x})),\label{eq:s-density}
\end{align}
where $\alpha$ and $\beta$ are learnable parameters and $\Phi_\beta$ is the cumulative distribution function of the Laplace distribution with zero mean and $\beta$ scale. We query the radiance $c_i$ and the signed distance $s_i$ for a set of samples $\{\mathbf{r}(t_i)\}$ along $\bf r$. The color of the pixel $C(\mathbf{r})$ on the image is obtained by accumulating all the samples along the ray
\begin{displaymath}
    C(r) = \sum_{i} T_{i}\left(1-\exp \left(-\sigma_{i} u_i\right)\right) c_{i},
\end{displaymath}
where $c_i$ is the radiance computed by the Rendering Network, $T_{i} = \exp \left(-\sum_{j=1}^{i-1} \sigma_{j}u_j\right)$ is the transparency indicating the probability that the ray travels from $0$ to $t_i$ without hitting any other particle, and $u_i = t_{i+1}-t_i$ is the distance between adjacent samples.

\subsection{Single-Person Refinement}
% \subsection

With the pre-trained template head available, Stage 2 of our method aims to refine the SDF to learn fine geometric details for a specific individual. 
Notice that the template together with the non-rigid deformation computed in Stage 1 does not suit our purpose. The reason is two-folded. Firstly, the deformation alone does not provide sufficient degrees of freedom for modeling high-frequency details. Secondly, the template represents the mean shape of all individuals in the dataset and does not carry ID-dependent features for a specific individual. For example, some individuals have a scarf while others do not.  

To address this issue, we propose a high-frequency displacement map that captures the ID-dependent geometric details of each specific individual. To achieve this, we introduce the Displacement Network, which takes both the ID-independent template feature $\mathbf{F}_{\text{tem}}$ and the ID-dependent deformation feature $\mathbf{F}_{\text{def}}$ as inputs, along with the query position $\mathbf{x}$.

The network then outputs a displacement $\delta \in \mathbb{R}$ and a displacement feature $\mathbf{F}_{\text{dis}} \in \mathbb{R}^{64}$, which is ID-dependent
\begin{equation}
    f_{\text{dis}}(\mathbf{x}, \mathbf{F}_{\text{tem}}, \mathbf{F}_{\text{def}})=(\delta, \mathbf{F}_{\text{dis}}),\label{eq:displace}.
\end{equation}
We use the displacement to update the signed distance. Specifically, for the query point $\mathbf{x}$ in the observation space, the updated signed distance is
\begin{equation}
    \hat{s}(\mathbf{x}) = s(\mathbf{x})+\delta(\mathbf{x}) = f_{\text{tem}}(\mathbf{x}+\mathbf{d}) + \delta(\mathbf{x}).\label{eq:final-sdf}
\end{equation}
The displacement $\delta$ is a non-zero value if the query point $\bf x$ is on the region with fine geometric details, such as wrinkles; otherwise, for points $\bf x$ are on a relatively smooth region, such as cheek, $\delta$ is close to zero. Since we expect the template undergoing the ID-dependent non-rigid deformation recovers most of the shape and the displacement is to add fine details, $\delta$ should be small and smooth. We regularize $\delta$ by using both the absolute value term $|\delta|$ to control the significance and a total variation (TV) term $|\nabla\delta|$ to ensure the smoothness. 

The displacement feature $\mathbf{F}_{\text{dis}}$ is designed to further enrich the ID-dependent features of the individual and help capture more fine-grained geometric details. Therefore, the overall feature $\mathbf{F}_{\text{all}}$ is a concatenation of the deformation feature $\mathbf{F}_{\text{def}}$, the template feature $\mathbf{F}_{\text{tem}}$ and the displacement feature $\mathbf{F}_{\text{dis}}$, i.e., 
\begin{equation}\mathbf{F}_{\text{all}}=\text{concat}(\mathbf{F}_{\text{def}},\mathbf{F}_{\text{tem}},\mathbf{F}_{\text{dis}}).
\label{eqn:allfeatures}
\end{equation}

Note that the Rendering Network $f_{\text{ren}}$ is used in both stages to train the template and the final shape separately. In Stage 1, we feed the deformation and the template features $\text{concat}(\mathbf{F}_{\text{def}}, \mathbf{F}_{\text{tem}})$ into it for computing the color for the specific human head. In Stage 2 we feed the overall feature \begin{equation}
 f_{\text{ren}}(\mathbf{z}_{c}, \mathbf{x}, \mathbf{v}, \mathbf{F}_{\text{all}}, \mathbf{n}_f)=c\label{eq:render2}
 \end{equation}
 into the network to obtain the radiance for a specific individual,
 where $\mathbf{n}_f=\nabla\hat{s}$ is the normal direction of the final surface. Our ablation study confirms that the displacement map is effective in reconstructing ID-dependent details or features that are absent from the template.

\subsection{Training}

All the components, $f_{\text{def}}$, $f_{\text{tem}}$, $f_{\text{dis}}$ and $f_{\text{ren}}$, are multi-layer perceptrons. The Deformation Network, the Template Network and the Displacement Network consist of 4, 8, and 4 layers, respectively. The Rendering Network is used in both stages and they are slightly different. In Stage 1, it has 4 layers with positional encoding of 6 and 4 frequencies for point coordinates and viewing directions, respectively. To deal with high-frequency details in Stage 2, we add 2 more layers for the Rendering Network and also increase the number of frequencies by 2 for both point locations and views. We also use a skip-connection to concatenate the input to the 3rd layer in order to strength the relationship between the input variables and the output radiance. All hidden layers have a width of 256.

Our training does not involve 3D supervision, and uses only the pixel colors to guide the training. The color loss is 
\begin{equation}
\mathcal{L}_{\text{col}} = \lambda_1\|C - C_{\text{gt}} \|_{1},\label{eq:color_loss}\\
\end{equation}
where $C_{\text{gt}}$ is the ground-truth color.

We adopt the following regularization terms. Since all human heads are of similar geometry and sizes, the deformation $\mathbf{d}$ should be smooth and not be too large. So we define the deformation loss term as
\begin{align}
     \mathcal{L}_\text{def} &= \lambda_{2}\| \mathbf{d} \|_{2} + \lambda_{3} \| \nabla \mathbf{d}\|_{2}\label{eq:deform_loss}
\end{align}
by regularizing the magnitudes of the offset vector and its gradient.
 
The template $s$ is a signed distance field, whose gradient satisfies the Eikonal equation $\|\nabla s\|_2=1$, therefore we use the Eikonal term as in ~\cite{gropp2020implicit}
\begin{equation}
     \mathcal{L}_{\text{eik}} = \lambda_{4} \left(\|\nabla s\|_{2}-1\right)^{2}\label{eq:eik} 
\end{equation}
to regularize the SDF.

To regularize the displacement map $\delta$, we consider the following situation:
let $\mathbf{p}$ be a point on the deformed template, i.e., 
$s(\mathbf{p})=0$. Denote by $\mathbf{p}'$ the corresponding point of $\bf p$ such that 1) it is along the normal direction of $\bf p$ 
\begin{displaymath}    \mathbf{p}'=\mathbf{p}+\lambda \frac{\nabla s(\mathbf{p})}{\|\nabla s(\mathbf{p})\|}, 
\end{displaymath}
where $\lambda\in\mathbb{R}$ is the signed distance between $\mathbf{p}'$ and $\bf p$, indicating how far we should move $\bf p$ along the normal; 
and 2) it is on the final surface, i.e., 
$$
s(\mathbf{p}')+\delta(\mathbf{p}')=0.
$$
Using Taylor expansion, we obtain
$$
s(\mathbf{p}')=\lambda \|\nabla s(\mathbf{p})\|,
$$
which implies that the displacement $\delta$ satisfies
$$
\delta(\mathbf{p}')=-\lambda\|\nabla s\|.
$$
We expect that the distance between $\bf p$ and $\mathbf{p}'$ is small and the displacement itself is also smooth. Therefore, we define the following displacement loss term 
\begin{equation}
 \mathcal{L}_{\text{dis}} = \lambda_{5} |\delta| + \lambda_{6} \| \nabla \delta \|_1.\label{eq:displace_loss}
\end{equation}
The first term in Equation (\ref{eq:displace_loss}) restricts the size of the displacement, while the second term, which is the total variation (TV)~\cite{osher2003image} of the displacement map $\delta$,  is to keep $\delta$ smooth while preserving fine details. 

Following DIF~\cite{deng2021deformed} and DeepSDF~\cite{park2019deepsdf}, we regularize the latent codes by assuming a Gaussian distribution $$\mathcal{L}_{\text{cod}}=\lambda_{6} \left(\|\mathbf{z}_{s}\|_{2} + \|\mathbf{z}_{c}\|_{2}\right).$$ 

Putting it all together, we define the loss function as follows:
\begin{align}
 \mathcal{L} &= \mathcal{L}_{\text{col}} +  \mathcal{L}_{\text{eik}} +  \mathcal{L}_{\text{def}} + \mathcal{L}_{\text{dis}} + \mathcal{L}_{\text{cod}}. 
\end{align}
In our implementation, we empirically set the weights as 
$\lambda_1=0.01$ and $\lambda_2=\cdots=\lambda_6=0.001$.

\subsection{Properties}
Our network offers two unique properties.

Firstly, limited viewpoints often result in missing/insufficient information, which can significantly downgrade the quality of existing methods. Our approach addresses this challenge through geometry decomposition. By training the template on individuals within the same group, who share significant geometric similarities, we obtain a smooth mean shape that represents the main geometric features of human heads. Furthermore, the randomly selected views of different individuals can complement each other to optimize the template geometry. As a result, the pre-trained template serves as a good initialization for stage 2 training, facilitating the learning of high-frequency details. 

Secondly, we have observed that learning both the smooth base surface and high-frequency details simultaneously in a low-view setting is highly unstable. For example, HF-NeuS frequently fails to accurately reconstruct geometry under the setting of only 10 views. In contrast, our method trains the network in a two-stage, coarse-to-fine manner, which effectively increases robustness. By establishing a solid foundation with the pre-trained template in stage 1, it is much easier to learn the high-frequency, ID-dependent details (such as wrinkles and scarfs) in stage 2. 

The proposed geometry decomposition and coarse-to-fine training enable our method to effectively complement missing information and produce high-quality reconstructions even in challenging low-view settings.

\subsection{Discussions}
While there are similar neural rendering models, such as DIF~\cite{deng2021deformed}, HF-NeuS~\cite{wang2022hf} and IND~\cite{li2022implicit}, our method differs from them in both design principles and application domains.

The Implicit Neural Deformation (IND) method~\cite{li2022implicit} consists of a template network, a deformation network and a rendering network. Using 3D supervision of a prior 3D face model, it can reconstruct 3D faces from sparse views, typically 2 to 4, but it is limited to only reconstructing the facial region. Since IND does not model high-frequency signals, the reconstructed facial regions do not carry additional high-frequency information beyond the template. Another significant difference is the way geometric features are fed to the rendering network. In IND, the geometric feature is the output of the template network, which cannot distinguish identities. Our method decomposes the geometric feature for each point into a template feature, which is identity independent and shared across multiple persons, a deformation feature, and a displacement feature, which are identity dependent. The feature decomposition property enables us to recover the unique geometric characteristics associated with a specific identity that are unavailable on the template, for example, scarves and wrinkles. IND is not able to do that. Furthermore, since obtaining 3D priors is often difficult in practice, our method, which does not rely on 3D supervision, is more flexible.

The Deformed Implicit Field (DIF) method~\cite{deng2021deformed} is a shape modeling approach that utilizes a template, a non-rigid deformation and a correction field to model a group of 3D shapes with similar characteristics. Unlike our method, DIF aims at reconstructing 3D shapes from point clouds, and uses the correction field to update the topology of the template SDF and regularizes it with an $L_1$ loss term for controlling its size. In contrast, our method utilizes a displacement map for modeling high-frequency details, and regularizes it using both $L_1$ and total variation losses. Additionally, our method does reconstruction from multiview images rather than point clouds.

High-frequency NeuS~\cite{wang2022hf} improves the quality of surface reconstruction in NeuS~\cite{wang2021neus} by introducing a high-frequency displacement function. It learns frequencies at multiple scales and gradually increases the frequency content in a coarse-to-fine manner. However, unlike our method, it does not use template and trains the network in a single stage. Although it performs well for dense view inputs, we observed that HF-NeuS is unstable under low-view settings, as missing information in the input images leads to the network focusing more on improving appearance rather than learning correct geometry. In contrast, our method tackles the challenge through geometry decomposition and a two-stage training strategy, allowing us to produce high-quality 3D reconstruction with low-view inputs. 

We qualitatively compare our method with a few other relevant neural face models and neural rendering approaches in Table~\ref{tab:comp_method}. Our method is unique in that it requires neither object masks nor 3D supervision, and it can reconstruct high-quality 3D heads with fine details from 2D images under a low-view setting. 

\begin{table}[ht]
\setlength\tabcolsep{2pt}
\centering
\begin{small}
\begin{tabular}{c|c|c|c|c|c|c}
\hline
Method & Task & Scenario & 3DS          &  D & LV  & M\\
\hline

i3DMM~\cite{yenamandra2021i3dmm} & 3D $\rightarrow$ 3D & Head & Yes & No & - & - \\
\hline
HeadNeRF~\cite{hong2022headnerf} & 2D $\rightarrow$ 3D & Head & No & No & No &No \\
\hline
FaceVerse~\cite{wang2022faceverse} & 2D $\rightarrow$ 3D & Face & Yes & Yes & No & No\\
\hline
ImFace~\cite{zheng2022imface} & 3D $\rightarrow$ 3D & Face & Yes & Yes & - & -\\
\hline
PhyDIR~\cite{zhang2022physically} & 2D $\rightarrow$ 3D & Face & No & Yes & No & No\\
\hline
SparseNeuS~\cite{long2022sparseneus} & 2D $\rightarrow$ 3D & General & No & No & Yes & No \\
\hline
RGBDNeRF~\cite{yuan2022neural} & 2.5D $\rightarrow$ 3D & General & No & No & Yes & No\\
\hline
IND~\cite{li2022implicit} & 2D $\rightarrow$ 3D & Face & Yes & No & Yes & Yes\\
\hline
H3D-Net~\cite{ramon2021h3d} & 2D $\rightarrow$ 3D & Head & Yes & Yes & Yes & Yes\\
\hline
Ours & 2D $\rightarrow$ 3D & Head & No & Yes & Yes & No \\
\hline
\end{tabular}
\end{small}
\caption{Qualitative comparison with other related neural rendering approaches. ``3DS'': 3D supervision; ``D'': reconstructing fine details; ''LV'': low-view setting; ``M'': requiring object masks.}
% \vspace{-0.1in}
\label{tab:comp_method}
\end{table}

\section{Experiments}
\noindent\textbf{Datasets.} The Portrait Relighting Dataset~\cite{wang2020single} contains 438 subjects ranging in age from 17 to 69 years old. Each subject was photographed from 30 different views around their frontal face and the images are at a resolution of 512$\times$512. The dataset includes associated camera parameters, poses and masks for easy foreground extraction. High-fidelity 3D geometry and textures were reconstructed using commercial software PhotoScan, which are considered ground-truth. For our experiments, we selected two subsets of 15 individuals each from the dataset: PR-Senior, which consists of senior individuals with rich geometric features such as wrinkles, and PR-Young, which consists of randomly selected younger individuals from the remaining dataset. We used these subsets to evaluate our proposed method.

\noindent\textbf{Implementation details.} We implemented our model using PyTorch ~\cite{Paszke2019PyTorchAI} and
used the Adam optimizer~\cite{DBLP:journals/corr/KingmaB14} to update the learnable parameters with an initial learning rate $5\times 10^{-4}$ in an exponential decay strategy. The latent codes for shape and color are of dimensions $\mathbf{z}_{s}$, $\mathbf{z}_{c} \in \mathbb{R}^{128}$. We follow VolSDF~\cite{yariv2021volume} to set the other hyper-parameters, such as the learnable parameters $\alpha$ and $\beta$ in Equation (\ref{eq:s-density}) and the number of samples on a ray. In the first stage, we trained the model for the 30 subjects with 10 randomly selected views\footnote{ The \textbf{randomly} selected views for each individual plays a crucial role in Stage 1 training, as these views complement each other for providing sufficient information to cover the whole head.} for 5K epochs.
 %\textcolor{red}{, taking approximately 2 days on a single A100 GPU.} \textcolor{blue}{is it necessary to tell the timing?} 
 In the second stage, we trained the model for 300K iterations for each specific person from 10 to 20 views. 

\begin{table*}[!h]
\setlength\tabcolsep{2pt}
\scriptsize
\centering
\label{tab:face_eval}
\begin{tabular}{c|ccc|ccc|ccc|ccc|ccc|ccc}
\toprule
\multirow{3}{*}{Method} &\multicolumn{9}{c|}{PR-Senior} &\multicolumn{9}{c}{PR-Young}\\
\cmidrule[0.5pt](rl){2-19}
& \multicolumn{3}{c|}{10 views}  & \multicolumn{3}{c|}{15 views}&\multicolumn{3}{c|}{20 views} 
& \multicolumn{3}{c|}{10 views}  & \multicolumn{3}{c|}{15 views}&\multicolumn{3}{c}{20 views} \\
\cmidrule[0.5pt](rl){2-19}
&CD & PSNR$_\text{t}$  & PSNR$_\text{n}$ & CD  & PSNR$_\text{t}$ & PSNR$_\text{n}$  & CD & PSNR$_\text{t}$  & PSNR$_\text{n}$ &CD & PSNR$_\text{t}$  & PSNR$_\text{n}$ & CD  & PSNR$_\text{t}$ & PSNR$_\text{n}$  & CD & PSNR$_\text{t}$  & PSNR$_\text{n}$ \\
\midrule
NeuS~\cite{wang2021neus} &1.795&34.55&23.93&2.626&34.44&28.50&0.956&34.37&28.96
&1.478&35.22&23.85&1.107&35.05&26.10&1.236&34.93&26.85 \\
VolSDF~\cite{yariv2021volume} &3.336&33.56&23.41  &2.483&33.46&26.07&2.279&33.41&28.09
&2.282&33.84&22.24&2.251&33.94&25.26&1.805&33.95&25.94\\
HF-NeuS~\cite{wang2022hf} &2.471&\textbf{34.81}&21.24&1.740&\textbf{34.77}&25.25&0.866&\textbf{34.70}&27.26
&2.724&\textbf{35.51}&20.29&1.885&\textbf{35.24}&21.75&0.831&\textbf{35.33}&26.50\\
Ours &\textbf{0.785}&33.78&\textbf{27.89}&\textbf{0.689}&33.82&\textbf{28.73}&\textbf{0.708}&33.55&\textbf{30.60}
&\textbf{0.930}&34.09&\textbf{26.80}&\textbf{0.784}&34.16&\textbf{27.75}&\textbf{0.818}&33.94&\textbf{30.13}\\
\bottomrule
\end{tabular}
\caption{Quantitative results for the PR-Senior and PR-Young datasets. To make fair comparison, we train all models using the same epochs. Since 3D ground truths are only available for facial regions, we crop the reconstructed 3D meshes before calculating Chamfer distances ($\downarrow$). We also apply face masks to the rendered images for calculating PSNR ($\uparrow$). The subscripts t and n denote the training and novel views, respectively. The Camber distances are measured in units of $10^{-4}$. Note that VolSDF and HF-NeuS fail to reconstruct proper geometry for 6 and 35 out of the 90 tests (30 models, 3 different view settings), respectively. We removed these models when calculating the Chamfer distances. NeuS and ours successfully reconstruct all models. See the supplementary material for more results.}
\end{table*}

\begin{figure}
    \centering
    \rotatebox{90}{\begin{tiny}{\textbf{training view}}\end{tiny}}
    \includegraphics[width=0.18\columnwidth]{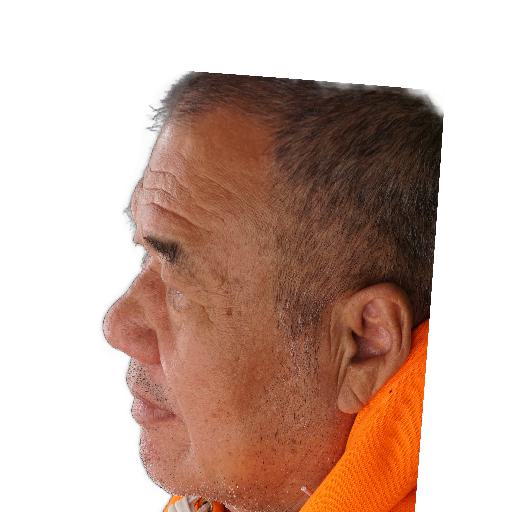}
    \includegraphics[width=0.18\columnwidth]{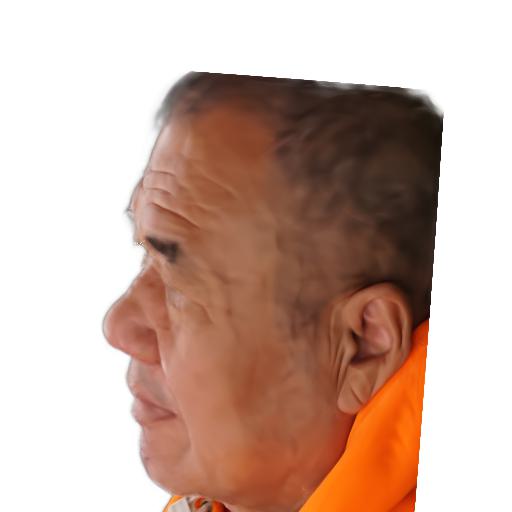}
    \includegraphics[width=0.18\columnwidth]{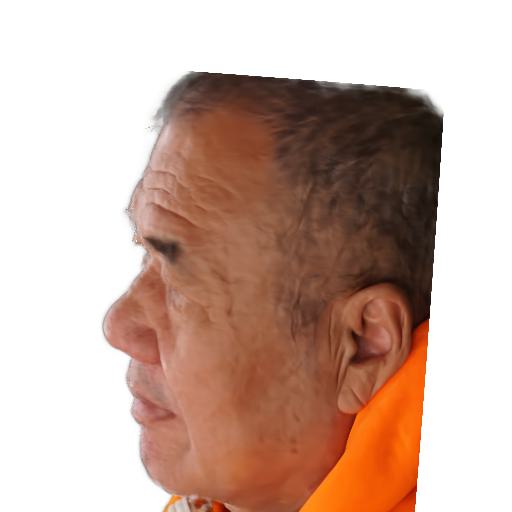}
    \includegraphics[width=0.18\columnwidth]{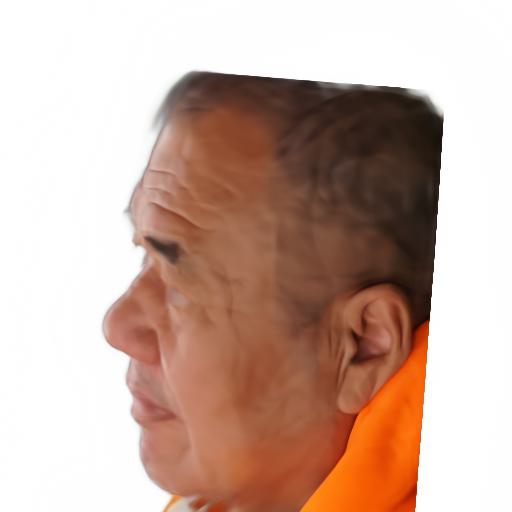}
    \includegraphics[width=0.18\columnwidth]{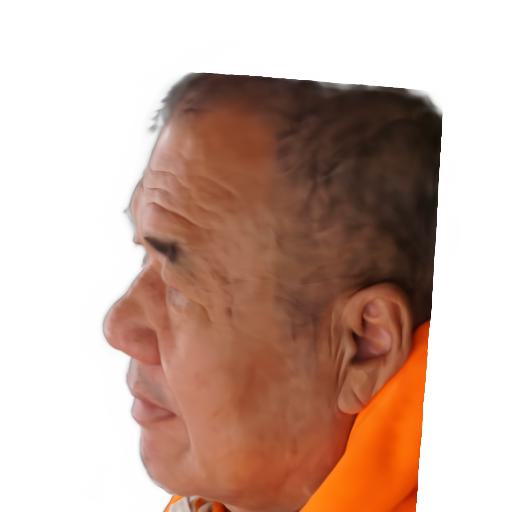}
    \\
    %\rotatebox{90}{\scriptsize train view normal}
    \rotatebox{90}{\scriptsize}
    \includegraphics[width=0.18\columnwidth]{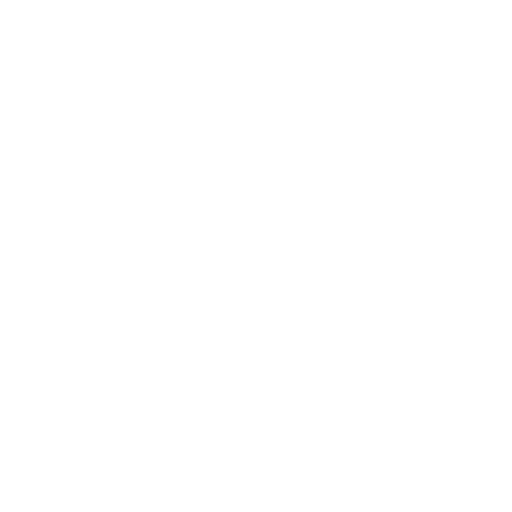}
    \includegraphics[width=0.18\columnwidth]{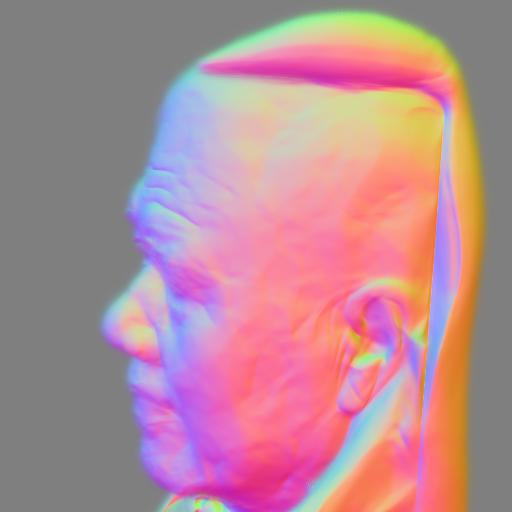}
    \includegraphics[width=0.18\columnwidth]{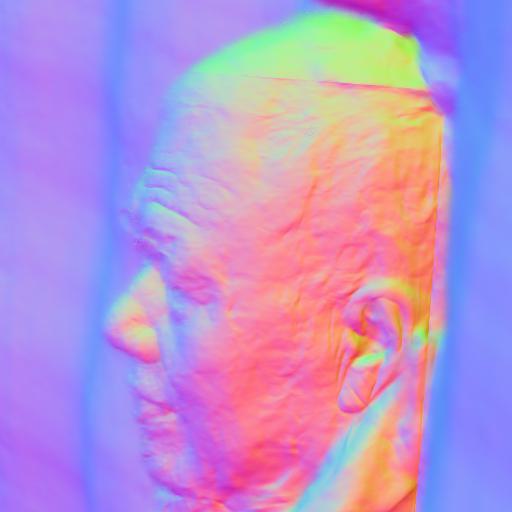}
    \includegraphics[width=0.18\columnwidth]{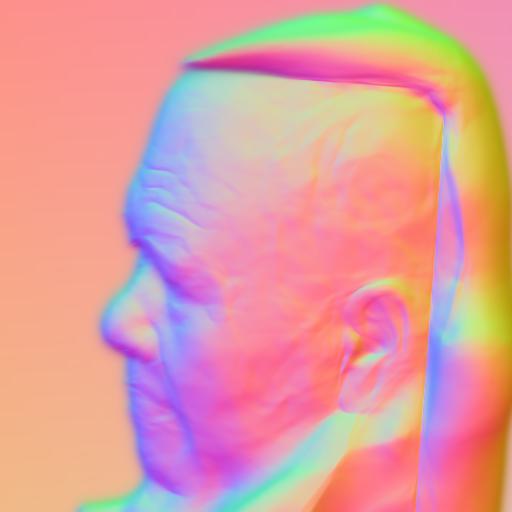}
    \includegraphics[width=0.18\columnwidth]{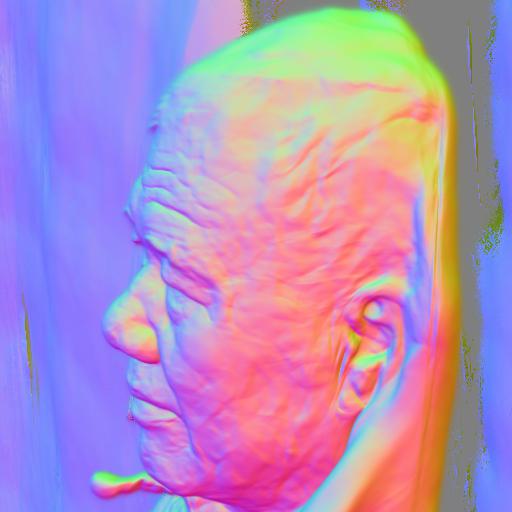}
    \\
    \rotatebox{90}{\begin{scriptsize}\textbf{novel view}\end{scriptsize}}\includegraphics[width=0.18\columnwidth]{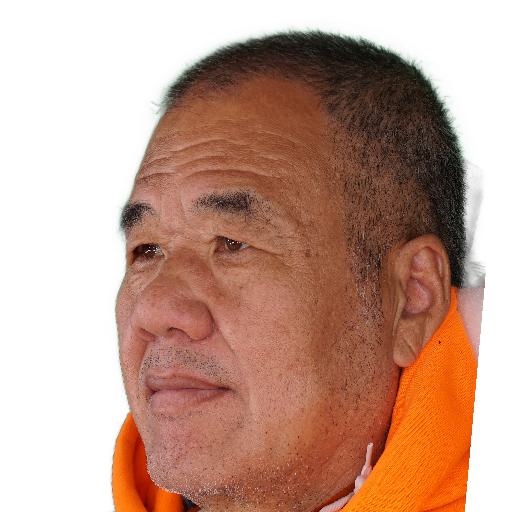}
    \includegraphics[width=0.18\columnwidth]{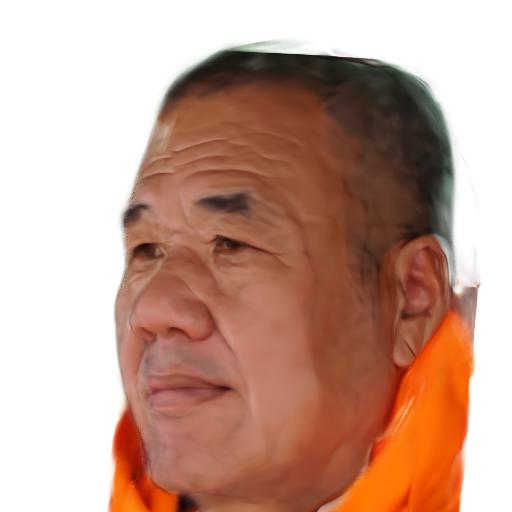}
    \includegraphics[width=0.18\columnwidth]{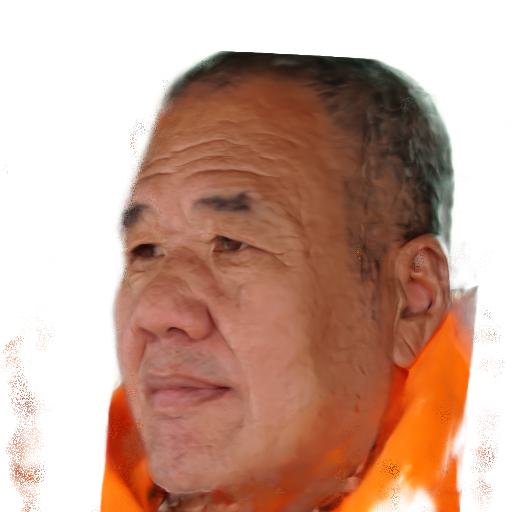}
    \includegraphics[width=0.18\columnwidth]{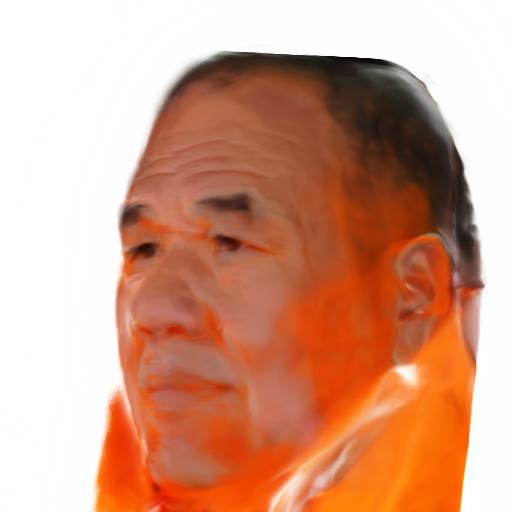}
    \includegraphics[width=0.18\columnwidth]{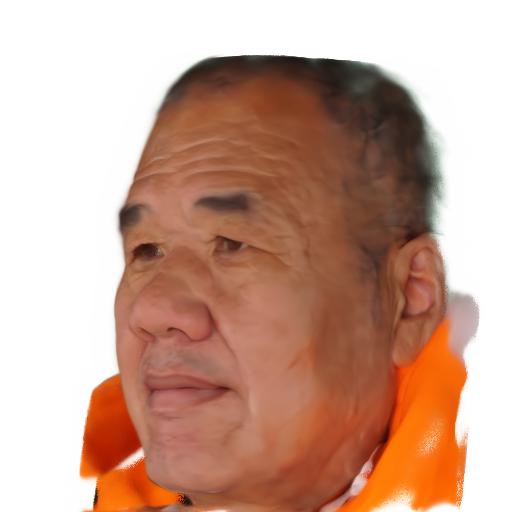}
    \\
    %\rotatebox{90}{\scriptsize novel view normal}
    \rotatebox{90}{\scriptsize}
    \includegraphics[width=0.18\columnwidth]{results/template_effects/gt_571_blank.jpg}
    \includegraphics[width=0.18\columnwidth]{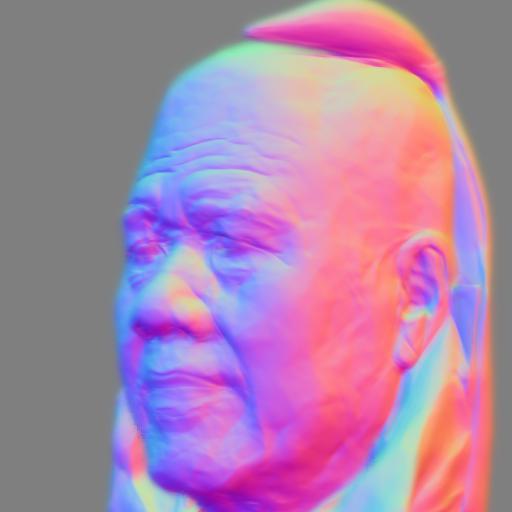}
    \includegraphics[width=0.18\columnwidth]{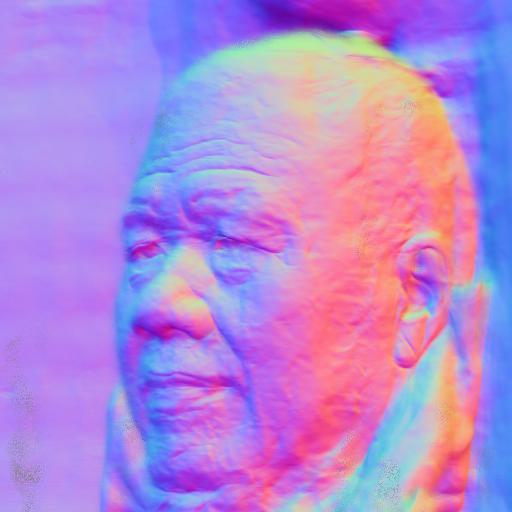}
    \includegraphics[width=0.18\columnwidth]{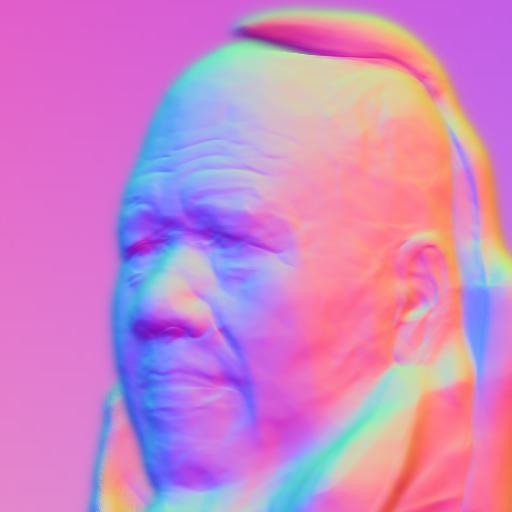}
    \includegraphics[width=0.18\columnwidth]{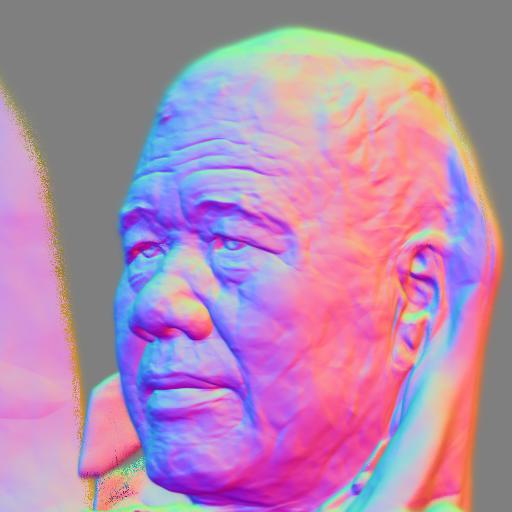}
    \\
    \makebox[0.18\columnwidth]{\scriptsize GT}
    \makebox[0.18\columnwidth]{\scriptsize NeuS}
    \makebox[0.18\columnwidth]{\scriptsize HF-NeuS}
    \makebox[0.18\columnwidth]{\scriptsize VolSDF}
    \makebox[0.18\columnwidth]{\scriptsize Ours}
    \label{fig:template_effect}
    \caption{Our method leverages a template in neural rendering, which makes the trained model more resilient to noise. Even when some of the training views have incomplete information about the human head, our method is able to complete the shape with reasonable geometry. Furthermore, our novel view result is closer to the ground truth than that of existing methods, demonstrating the effectiveness of using a template. Model: 571, 10 views, 4 are incomplete due to unusual viewing angles and cropping.
    }
    \vspace{-0.1in}
    \end{figure}

\begin{figure}[htb] \centering
    \rotatebox{90}{\begin{tiny}\textbf{training view}\end{tiny}}
    \includegraphics[width=0.18\columnwidth]{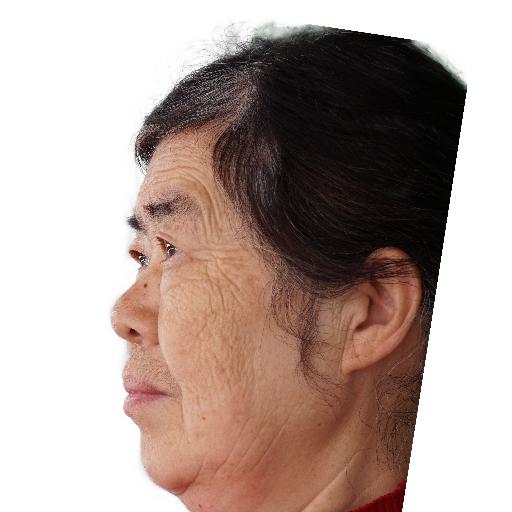}
    \includegraphics[width=0.18\columnwidth]{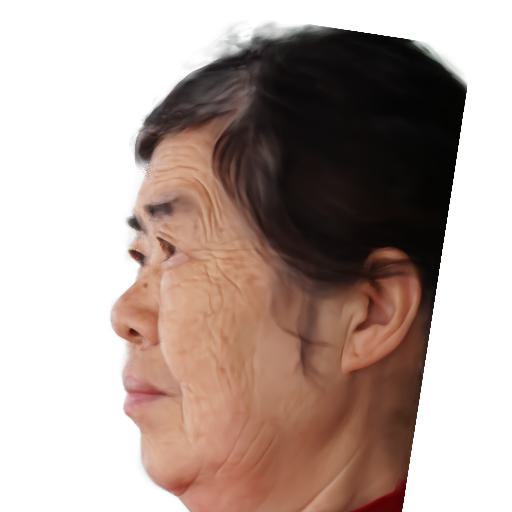}
    \includegraphics[width=0.18\columnwidth]{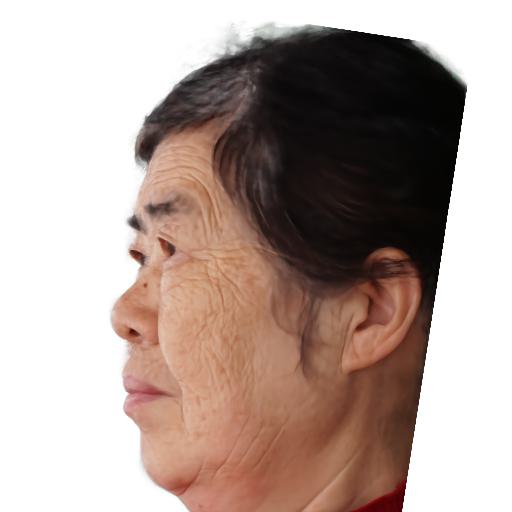}
    \includegraphics[width=0.18\columnwidth]{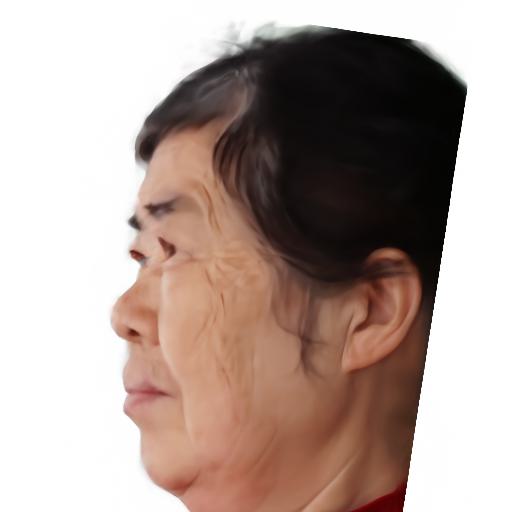}
    \includegraphics[width=0.18\columnwidth]{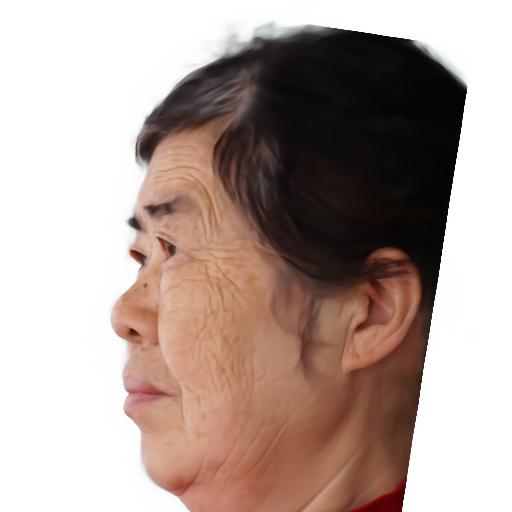}
    \\
    \includegraphics[width=0.18\columnwidth]{results/template_effects/gt_571_blank.jpg}
    \includegraphics[width=0.18\columnwidth]{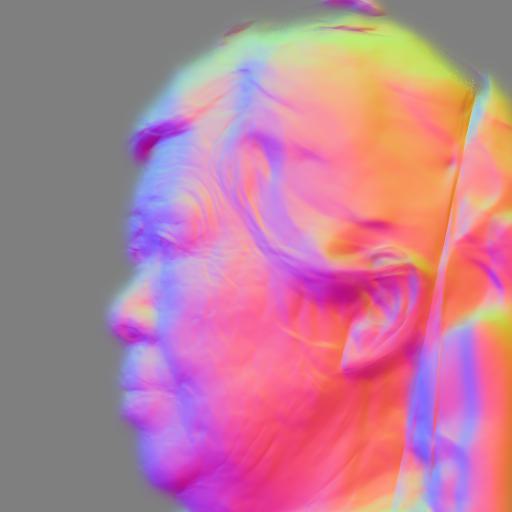}
    \includegraphics[width=0.18\columnwidth]{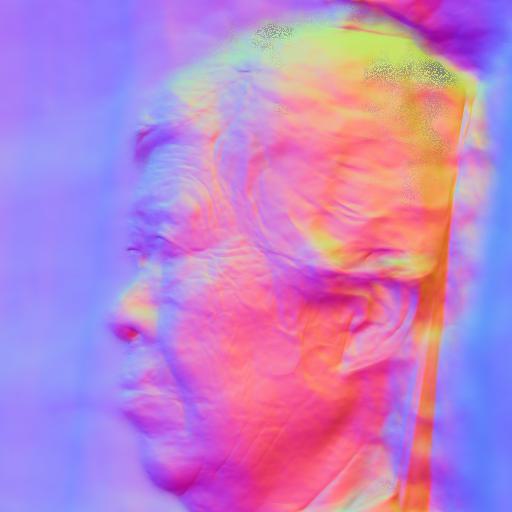}
    \includegraphics[width=0.18\columnwidth]{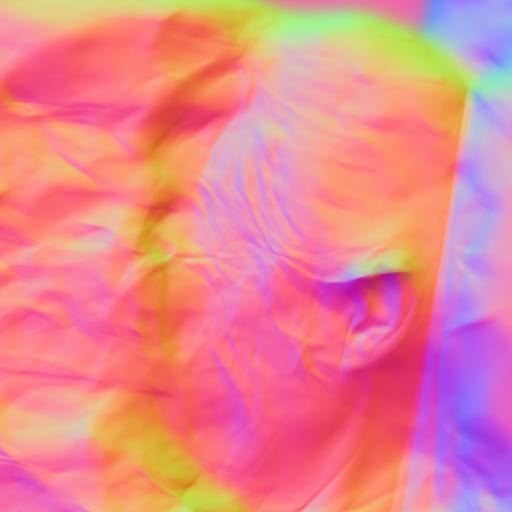}
    \includegraphics[width=0.18\columnwidth]{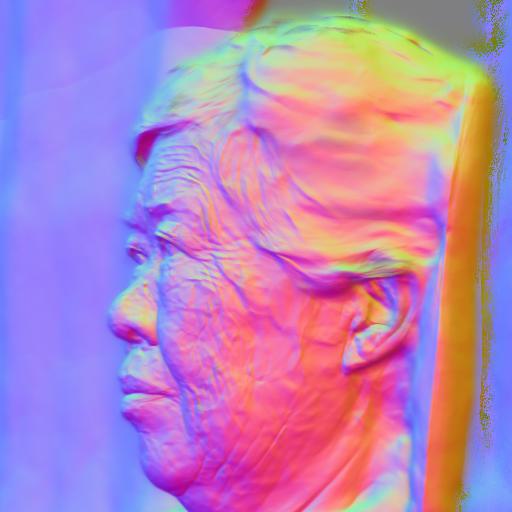}
    \\
    \rotatebox{90}{\begin{scriptsize}\textbf{novel view}\end{scriptsize}}
    \includegraphics[width=0.18\columnwidth]{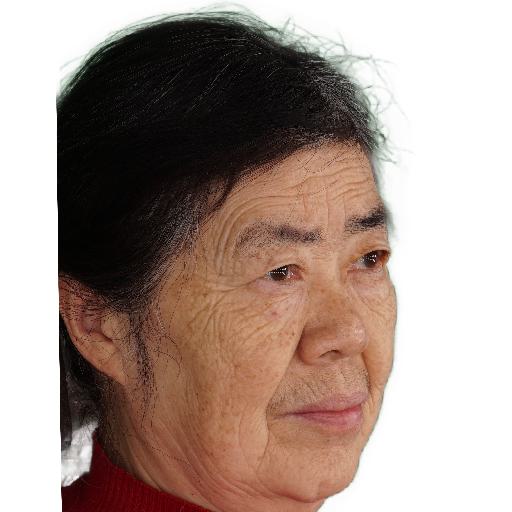}
    \includegraphics[width=0.18\columnwidth]{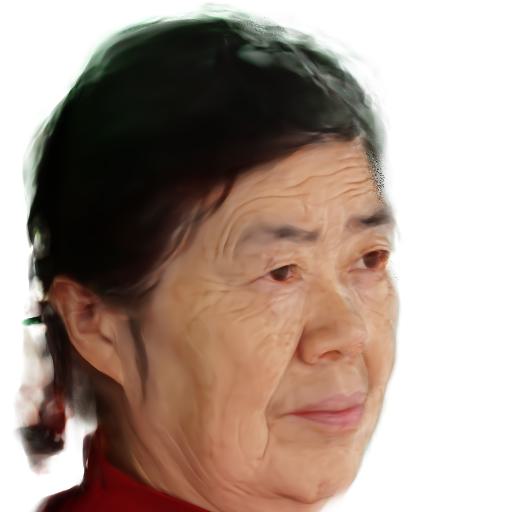}
    \includegraphics[width=0.18\columnwidth]{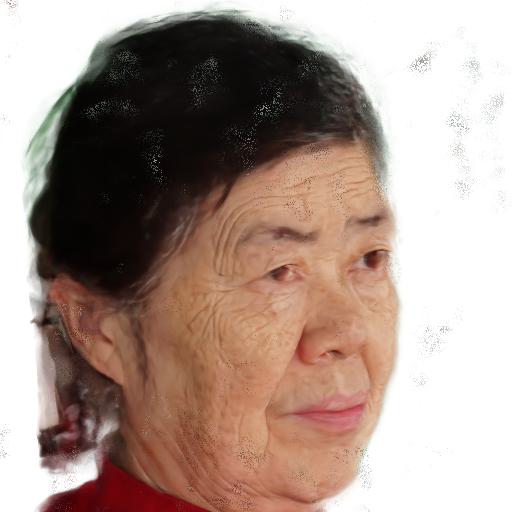}
    \includegraphics[width=0.18\columnwidth]{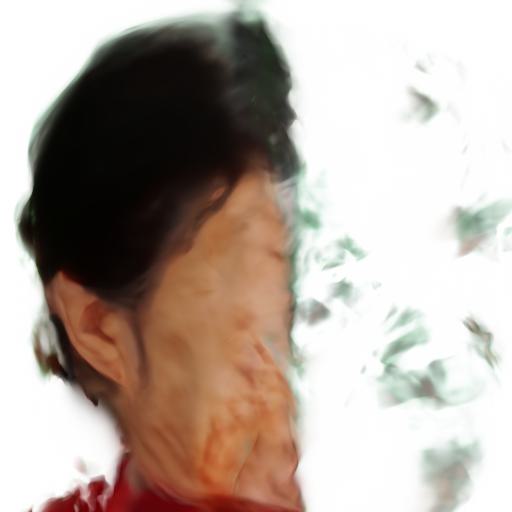}
    \includegraphics[width=0.18\columnwidth]{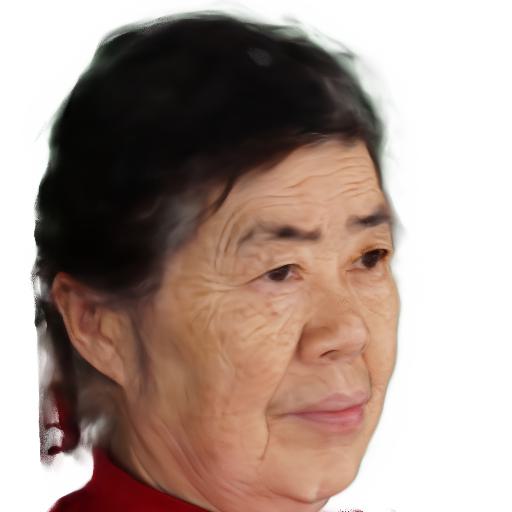}
    \\
    \includegraphics[width=0.18\columnwidth]{results/template_effects/gt_571_blank.jpg}
    \includegraphics[width=0.18\columnwidth]{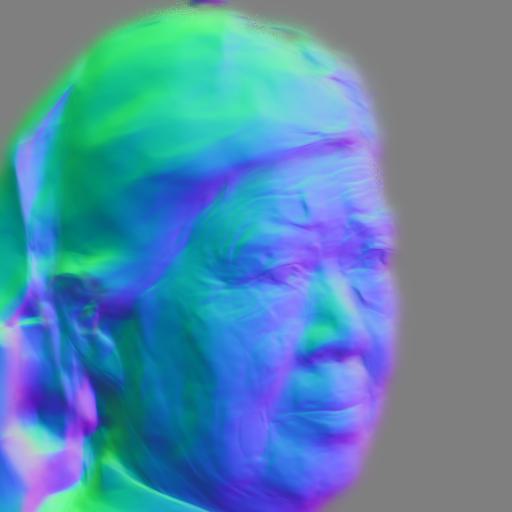}
    \includegraphics[width=0.18\columnwidth]{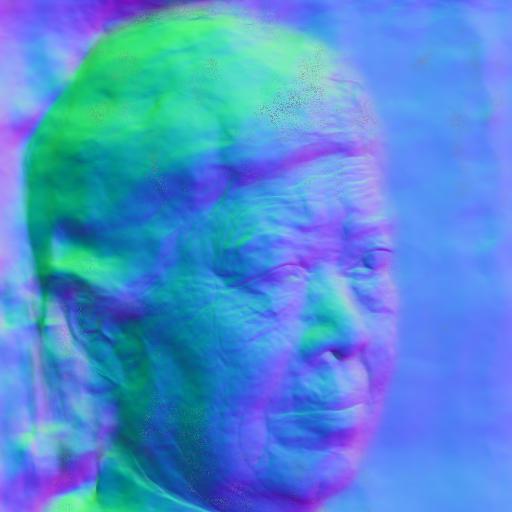}
    \includegraphics[width=0.18\columnwidth]{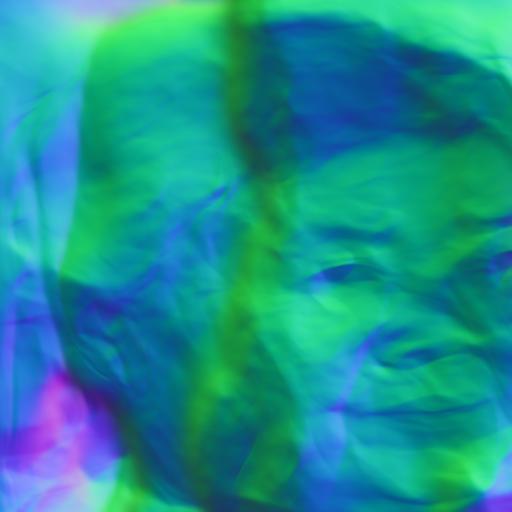}
    \includegraphics[width=0.18\columnwidth]{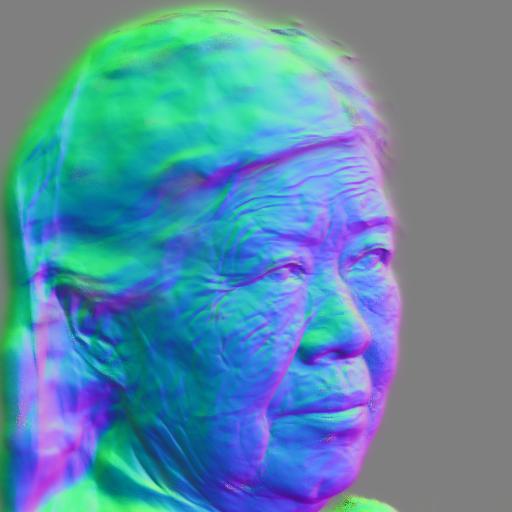}
    \\
    \makebox[0.18\columnwidth]{\scriptsize GT}
    \makebox[0.18\columnwidth]{\scriptsize NeuS}
    \makebox[0.18\columnwidth]{\scriptsize HF-NeuS}
    \makebox[0.18\columnwidth]{\scriptsize VolSDF}
    \makebox[0.18\columnwidth]{\scriptsize Ours}
    \caption{Our method excels in reconstructing fine details compared to other methods, thanks to the use of a displacement map for capturing high-frequency signals. We demonstrate the effectiveness of our approach in reconstructing fine details in a training view result (top) and a novel view result (bottom) here. Model: 377, 10 views.
    \vspace{-0.1in}
    } 
    \label{fig:10view_detail_facial}
\end{figure}

\noindent\textbf{Baselines.} We conducted a comparative evaluation of our method with state-of-the-art neural rendering methods, including NeuS~\cite{wang2021neus}, VolSDF~\cite{yariv2021volume}, and HF-NeuS~\cite{wang2022hf}, on PR-Dataset. We used the official implementations provided by the authors for all methods, and set the same 1,024 rays in all experiments to ensure a fair comparison. 

\noindent\textbf{Quality measures.} To assess the performance of our method and other approaches, we utilized the marching cubes algorithm~\cite{lorensen1987marching} to extract the zero-level set from the computed signed distance field. We cropped both the ground-truth and predicted meshes to focus on the facial region of interest, and computed the Chamfer distance (CD) between them to evaluate geometric quality. Additionally, we applied the face masks to the rendered RGB images to calculate the PSNR for the facial regions for evaluating visual appearance. The results are reported in Tables \ref{tab:face_eval}.

\noindent\textbf{Geometry accuracy.} Both NeuS and our method were able to succesffully reconstruct the geometry of 3D heads for all testing models, while VolSDF and HF-NeuS failed to reconstruct the geometry of human heads in in a significant number of examples under the setting of 10 to 15 views. 
Therefore, we \textbf{excluded} these failed examples when calculating the Chamfer distances for VolSDF and HF-NeuS. Since their rendered images are still visually acceptable for training views, we used all results for computing the PSNR metrics. Our method consistently outperformed all other methods in terms of geometry quality on both the PR-Senior and PR-Young datasets by a large margin with 10 to 15 views as input, demonstrating our method's ability to effectively overcome the challenge of missing/insufficient information in low-view settings. With 20 views as input, the gap became less significant, but our method still performed better than other methods. 

\noindent\textbf{Visual appearance.} We also observed that although our method did not achieve the highest PSNR score for rendering results on the \textbf{training} views, it outperformed the other methods for rendering \textbf{novel} views. This is because novel view synthesis is more dependent on the accuracy of the underlying 3D geometry than training views. Our method's superior geometry quality leads to better novel view synthesis results, demonstrating the effectiveness of our approach.

\noindent\textbf{Robustness.} While HF-NeuS also adopts a displacement field, it typically achieves the best performance with a sufficient number of views. However, as the number of views decreases, the 3D reconstruction quality often degrades significantly. For instance, under the setting of 10 views, HF-NeuS failed to reconstruct 3D geometry for 19 persons out of the 30 individuals in the PR-Senior and -Young datasets, resulting in poor novel view synthesized results. Although increasing the views to 20 improved the reconstruction quality considerably, HF-NeuS still could not reconstruct the geometry accurately for Models 487, 548, and 608. In contrast, our method is robust and performs consistently well under the same low-view settings, thanks to the use of a pre-trained template and the two-stage training strategy.

\noindent\textbf{Analysis.} 
Our method outperformed the existing methods in terms of geometry measures and novel view synthesis results in low-view settings. This can be attributed to three reasons. Firstly, in the first stage of our method, we trained the template using a multi-person dataset. Since the selection of viewpoints for the subjects is random, the different views can complement each other, resulting in better facial geometry for the template. Secondly, utilizing a pre-trained template increases the robustness of our model and enables it to resist a certain degree of ``noise'' caused by the limited number of viewpoints in the second stage of our method. For instance, when the images in the training set have certain disturbances to the reconstructed object, such as the cropping at the top of the head and clothing covering the neck as shown in Figure~\ref{fig:template_effect}, it can mislead other methods due to the lack of information, thereby producing incorrect geometry to complete the missing parts. To improve accuracy, the existing methods require more views to figure out the missing or occluded parts. In contrast, our method successfully modeled the top of the head and the neck of the character only 10 views, thanks to the use of the template, which provides a reasonable guide. Thirdly, in the case of senior individuals with rich wrinkles, the displacement field plays a crucial role in modeling high-frequency geometry, resulting in better reconstruction of fine geometric details, as demonstrated in Figure \ref{fig:10view_detail_facial}.

\begin{figure}[htbp]
\setlength\tabcolsep{0pt}
\centering
\begin{scriptsize}
\begin{tabular}{ccccc}
    \centering
        \rotatebox{90}{\textbf{training view}} &

\includegraphics[width=0.1\textwidth]{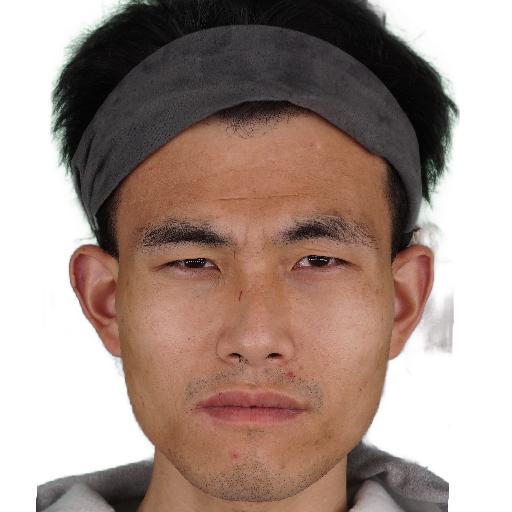} &
    \includegraphics[width=0.1\textwidth]{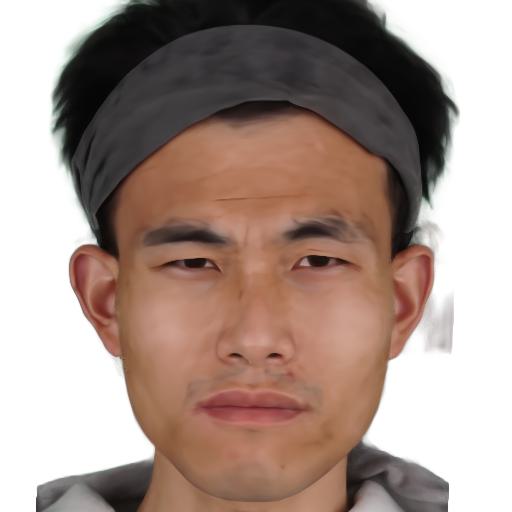} &
    \includegraphics[width=0.1\textwidth]{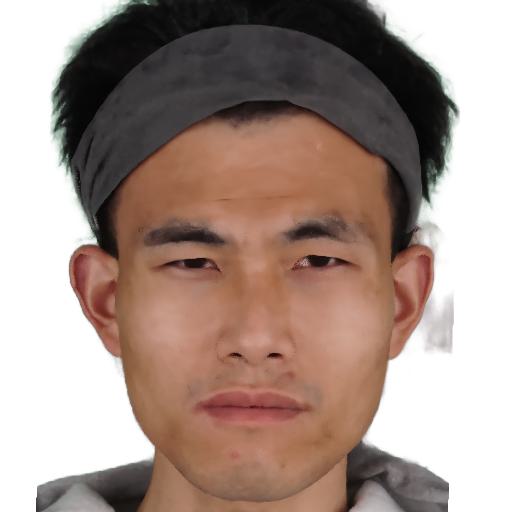} &
    \includegraphics[width=0.1\textwidth]{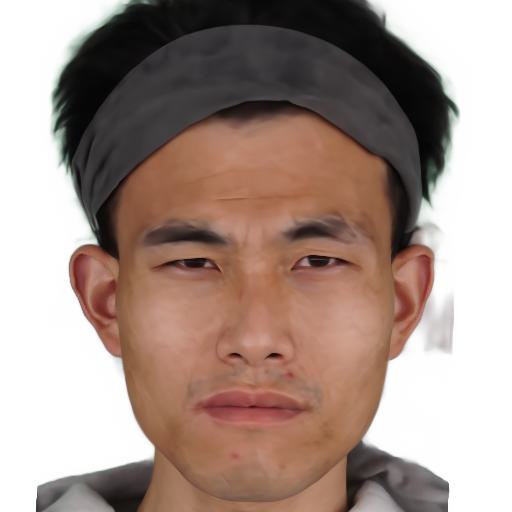}
    \\
  &
    \includegraphics[width=0.1\textwidth]{./results/template_effects/gt_571_blank.jpg} &
    \includegraphics[width=0.1\textwidth]{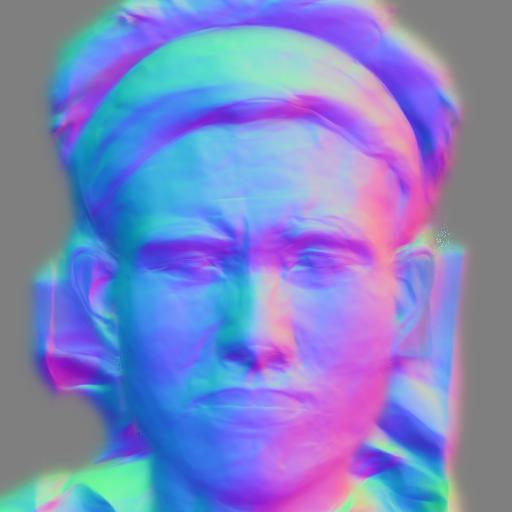} &
    \includegraphics[width=0.1\textwidth]{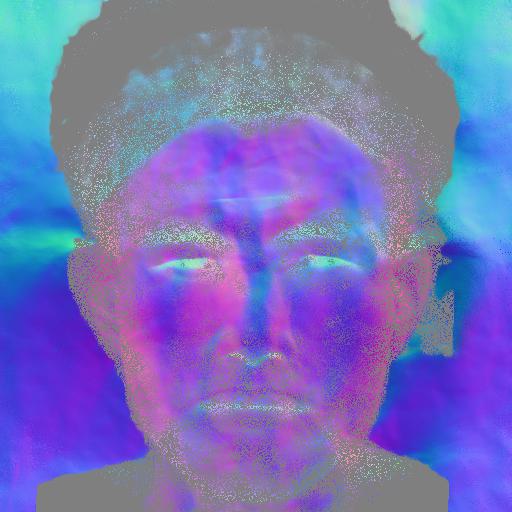} &
    \includegraphics[width=0.1\textwidth]{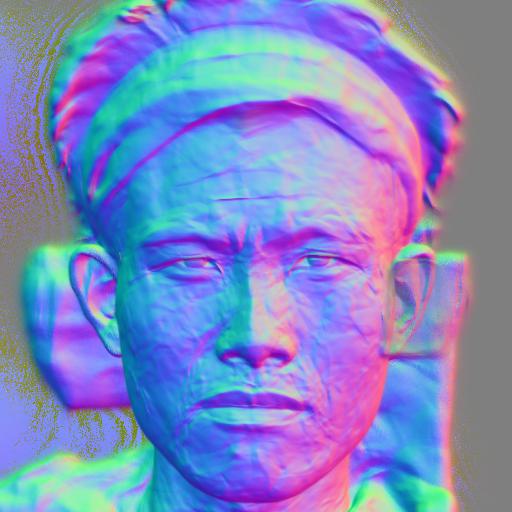}
    \\
    \rotatebox{90}{\textbf{novel view}}&
    \includegraphics[width=0.1\textwidth]{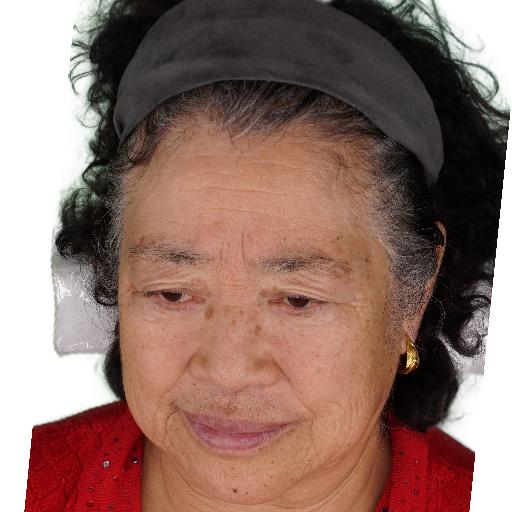} &
    \includegraphics[width=0.1\textwidth]{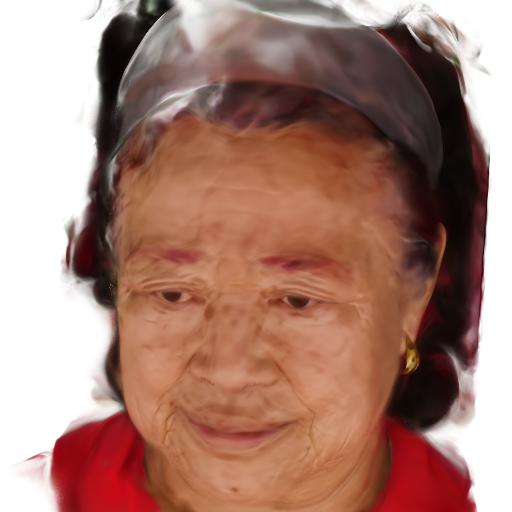} &
    \includegraphics[width=0.1\textwidth]{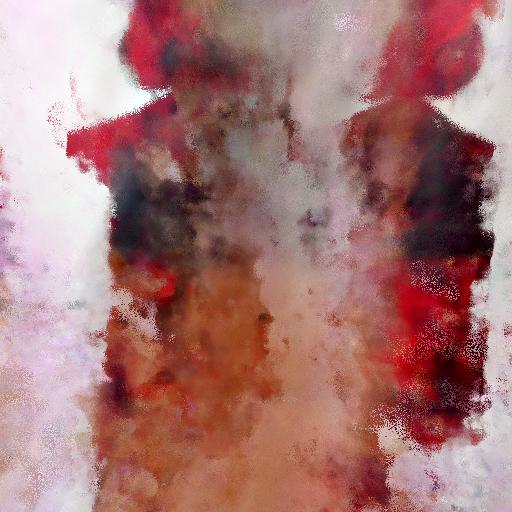} &
    \includegraphics[width=0.1\textwidth]{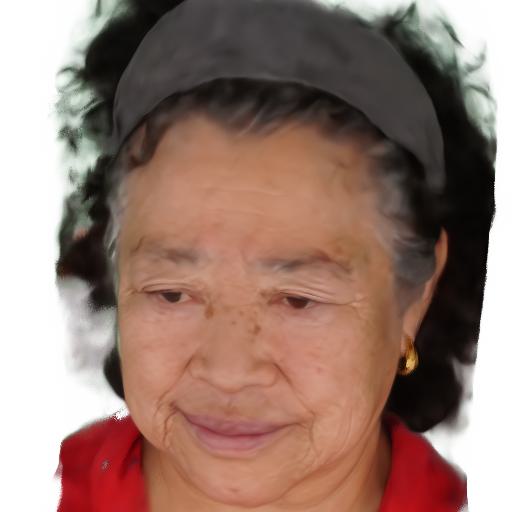}
    \\
    &\includegraphics[width=0.1\textwidth]{./results/template_effects/gt_571_blank.jpg} &
    \includegraphics[width=0.1\textwidth]{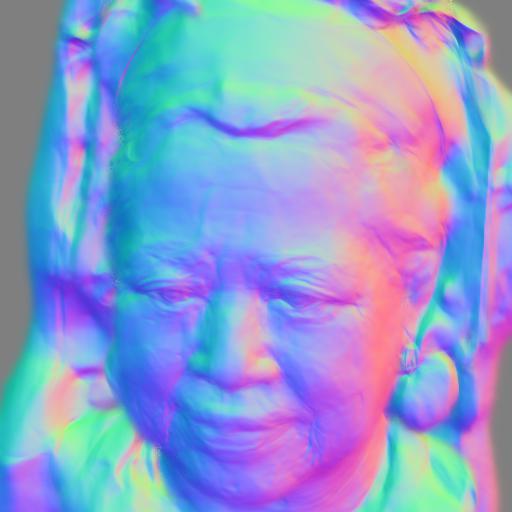} &
    \includegraphics[width=0.1\textwidth]{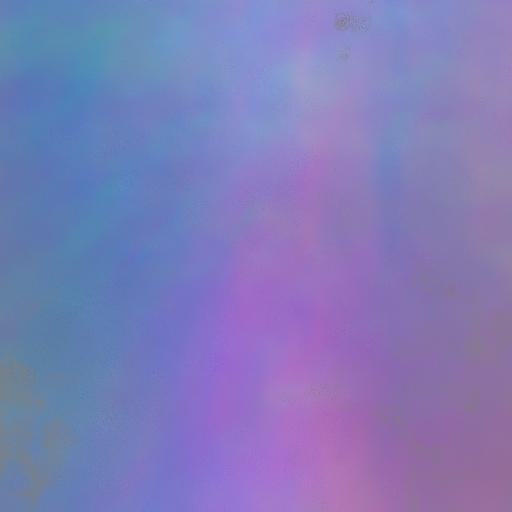} &
    \includegraphics[width=0.1\textwidth]{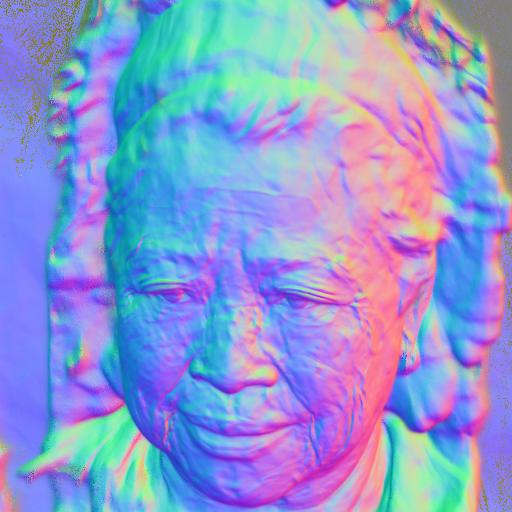}
    \\
    & 
    (a) GT & (b) NeuS & (c) HF-NeuS & (d) Ours\\
    \end{tabular}
    \end{scriptsize}
    \caption{
    HF-NeuS simultaneously learns the base surface and high-frequency details, making it unstable under a low-view setting due to the insufficient information provided in the images. This instability leads to incorrect novel view results. NeuS can reconstruct the head geometry fairly well but struggles to fully recover fine details and can yield low-quality results for novel views. In contrast, our proposed method uses geometry decomposition and coarse-to-fine training, which effectively overcomes these challenges and produces accurate results in low-view settings. Models: 548 (top) and 376 (bottom).}
    \vspace{-0.2in}
    \label{fig:hfs_fail}
\end{figure}

\vspace{-0.15in}
\begin{table}[htbp]
    \centering
    \begin{tabular}{c|c|c|c}
    \hline
        Model & CD ($10^{-4}$) & PSNR$_\text{t}$ & PSNR$_\text{n}$  \\
        \hline
        539 & 3.861 & 32.29 & 22.69 \\
        662 & 1.643 & 33.21 & 25.42\\
        \hline
    \end{tabular}
    \caption{Performance on two unseen identities under 10 views. See Figure~\ref{fig:unseen} for the visual result on Model 539 and the supplementary material for the other model.  }
    \label{tab:unseen_metric}
\end{table}

\begin{figure}[htb] \centering
    \includegraphics[width=0.08\textwidth]{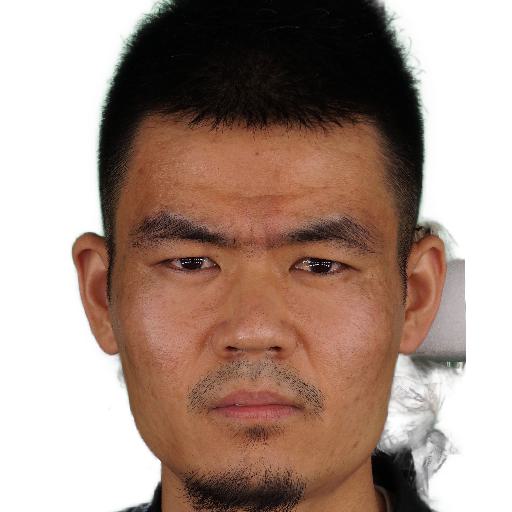}    \includegraphics[width=0.03\textwidth]{./results/template_effects/gt_571_blank.jpg}
    \includegraphics[width=0.08\textwidth]{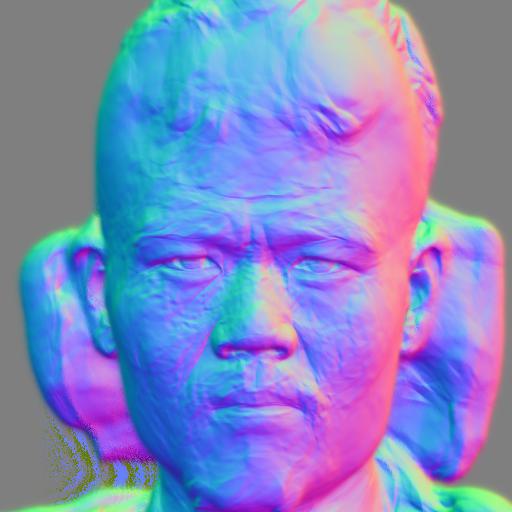}
    \includegraphics[width=0.08\textwidth]{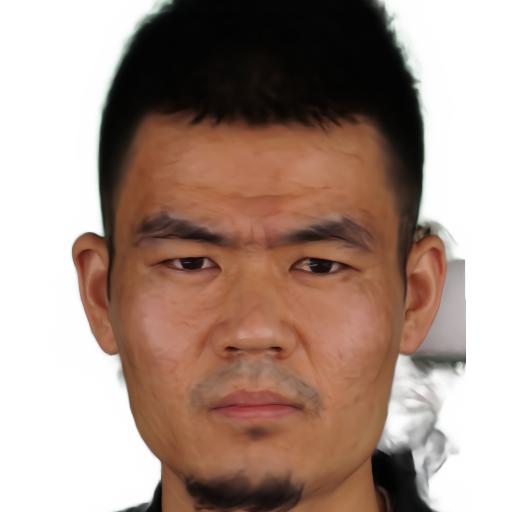}
    \includegraphics[width=0.08\textwidth]{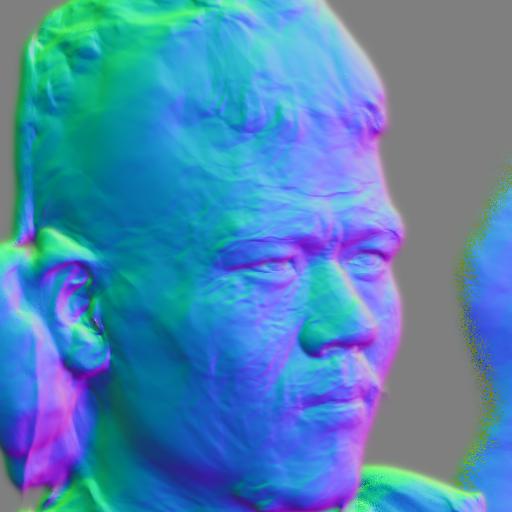}
    \includegraphics[width=0.08\textwidth]{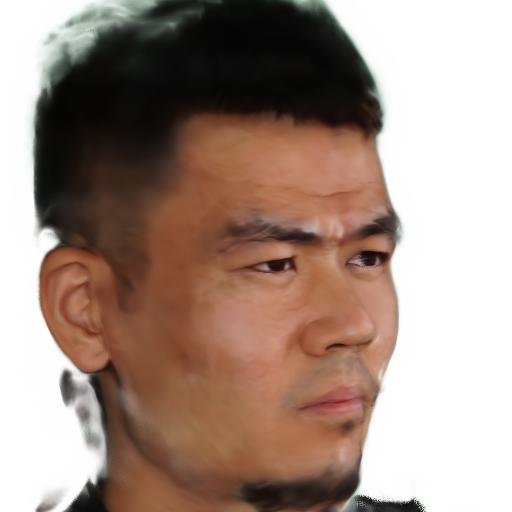}
    \\
    %\vspace{-0.1in}
    \caption{Reference (left) and rendering results (right) for an unseen individual (Model 539) under a 10-view setting.}
    \label{fig:unseen}
\end{figure}

\vspace{-0.15in}
\noindent\textbf{Unseen identities.} Our method has the ability to adapt to new individuals as the pre-trained template serves as a good initialization. Table~\ref{tab:unseen_metric} reports its performance on two unseen identities and Figure~\ref{fig:unseen} shows a visual result.

\begin{figure}[htbp]
\centering
    \includegraphics[width=0.125\textwidth]{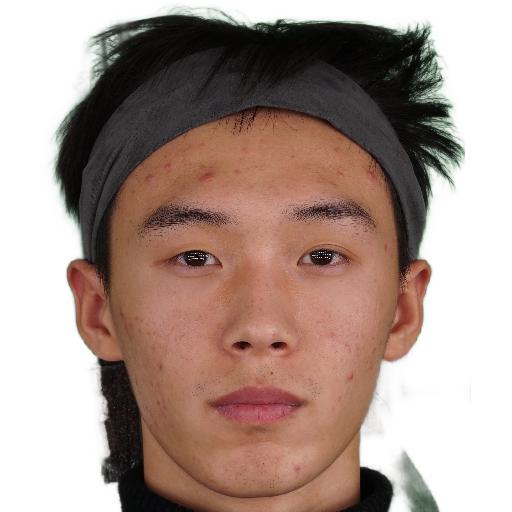}
    \includegraphics[width=0.125\textwidth]{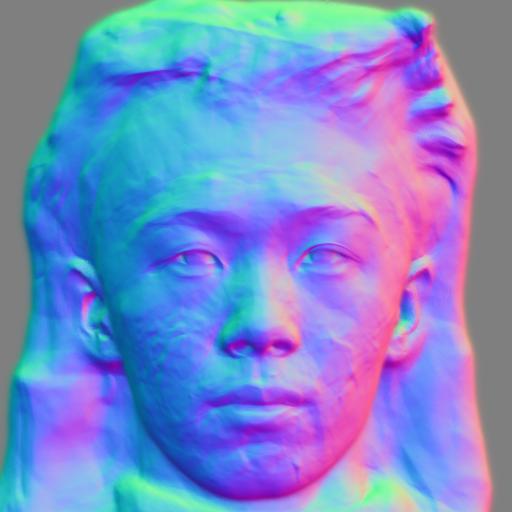}
    \includegraphics[width=0.125\textwidth]{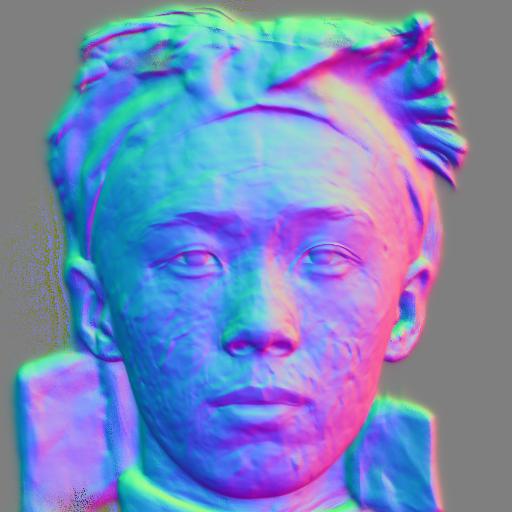}
    \\
    \includegraphics[width=0.125\textwidth]{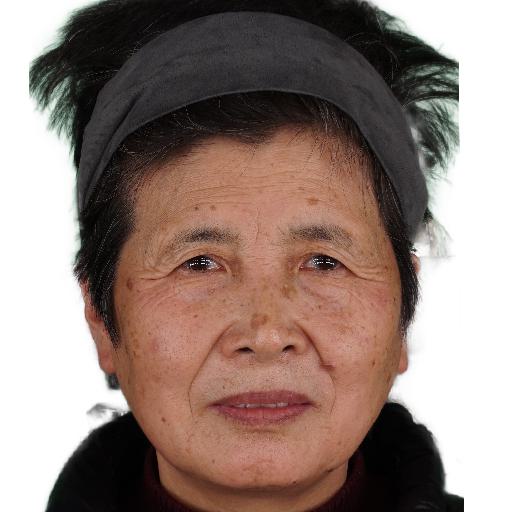}
    \includegraphics[width=0.125\textwidth]{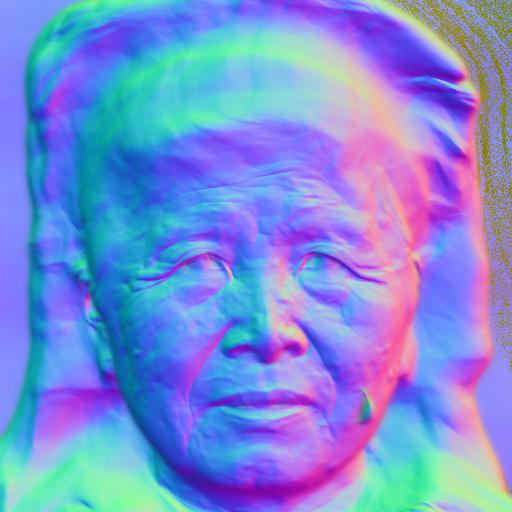}
    \includegraphics[width=0.125\textwidth]{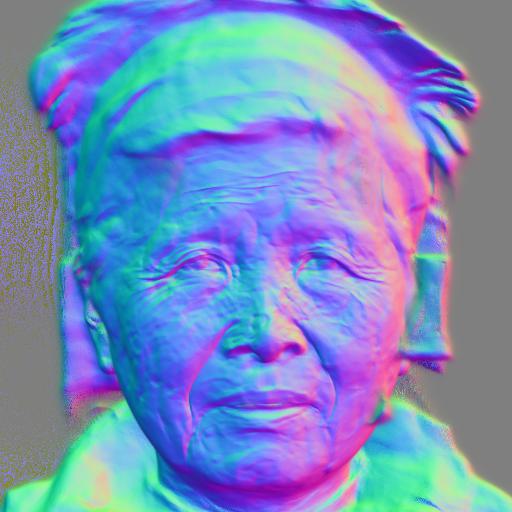}
    
    \makebox[0.125\textwidth]{\scriptsize GT}
    \makebox[0.125\textwidth]{\scriptsize w/o displacement}
    \makebox[0.125\textwidth]{\scriptsize w/ displacement}\\
    \includegraphics[width=0.125\textwidth]{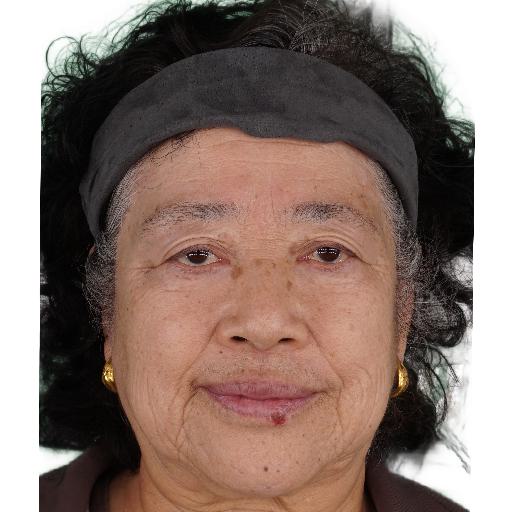}
    \includegraphics[width=0.125\textwidth]{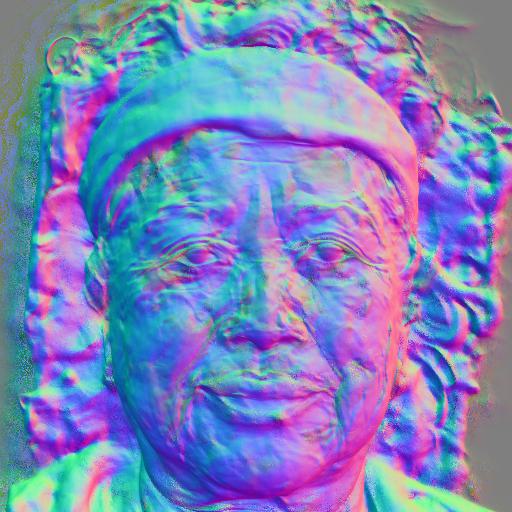}
    \includegraphics[width=0.125\textwidth]{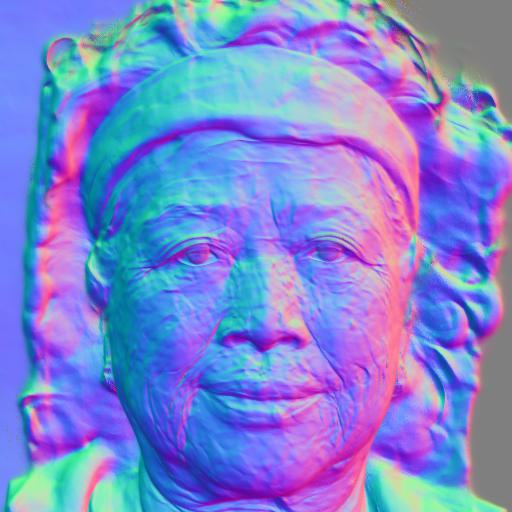}
    \\
    
    \makebox[0.125\textwidth]{\scriptsize GT}
    \makebox[0.125\textwidth]{\scriptsize w/o TV loss}
    \makebox[0.125\textwidth]{\scriptsize w/ TV loss}
    
    \caption{Ablation studies. Rows 1 \& 2: The displacement map provides an additional degree of freedom and is essential in effectively reconstructing features and fine details that are not present in the template, such as the scarf and crow's feet. Using non-rigid deformation alone is not sufficient to achieve this level of detail.  
    Row 3: The TV regularization term is effective in reducing noise while preserving sharp features and fine details. Models: 396 (top), 566 (middle) and 619 (bottom), all under a 10-view setting.}
    %在template-deformation框架下，没有displacement的模型很难重建出区别于template的特征如头巾，并且对人脸眼角处的皱重建不够精细。
    \label{fig:ablation_dis}
\end{figure}

\noindent\textbf{Ablation studies.} To evaluate the impact of the displacement field, we conducted an ablation study by training a model without the displacement field in Stage 2 and comparing it to the proposed model with the displacement field. Both models were trained using the same hyperparameters and training data under the setting of 10 views. As shown in Table~\ref{table:ablation_disp}, the proposed model with the displacement field outperforms the model without it in terms of geometry accuracy. This indicates that the displacement field plays a crucial role in improving the accuracy of our method. Figure~\ref{fig:ablation_dis} provides visual evidence for this by showing that non-rigid deformation alone cannot capture the fine details and features that are absent in the template. Therefore, the displacement map serves as a necessary supplement to enhance the accuracy of the reconstruction. We also conducted an evaluation of the impact of the TV regularization term on the reconstruction of Model 619, a senior female with rich wrinkles, under the setting of 10 views. Figure~\ref{fig:ablation_dis} (row 3) shows that the TV regularizer $\|\nabla\delta\|_1$ can effectively reduce noise while preserving fine details in the reconstruction.

\begin{figure}[htbp]
    \centering
    \includegraphics[width=0.47\textwidth]{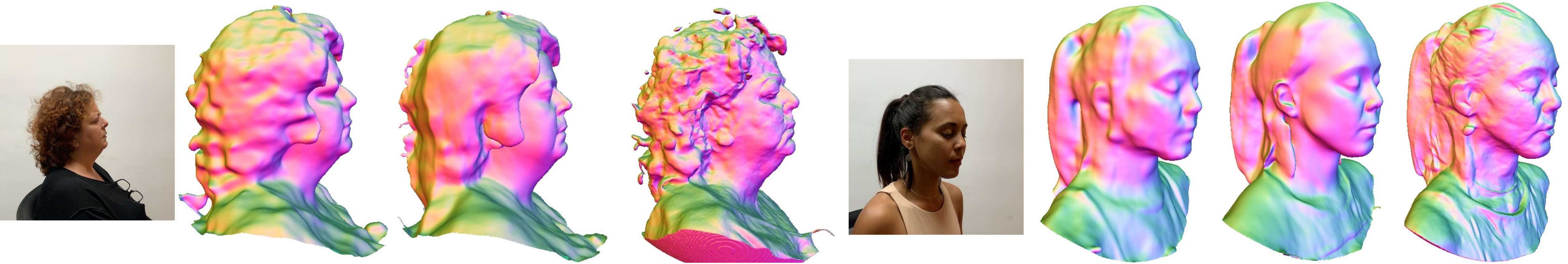}
    \caption{Results on the H3DS Dataset under a 8-view setting. Left to right: reference image, IDR~\cite{yariv2020multiview}, H3D~\cite{ramon2021h3d}, and ours. 
    }
    \label{fig:h3ds}
\end{figure}

\noindent\textbf{Other dataset.} We also tested our method on the H3DS~\cite{ramon2021h3d} dataset, consisting of 10 western individuals (5
men and 5 women) with 3D ground truth meshes. By using our pre-trained template (which contains only the head geometry) as initialization, our method successfully reconstructs all the models, including the upper body information. The reconstruction quality, measured as the average CD (mm) for the \textbf{entire head} of the 10 individuals, is IDR 8.34, H3D 6.45, and ours 5.47, demonstrating the effectiveness of our method for non-Asian individuals. Both IDR~\cite{yariv2020multiview} and H3D~\cite{ramon2021h3d} rely on accurate object masks in their differentiable rendering pipeline. H3D also uses 3D head scans from 10,000 individuals. In contrast, our method does not rely on masks or 3D supervision, providing greater flexibility. See Figure~\ref{fig:h3ds}.
\vspace{-0.15in}
\begin{table}[htbp]
\setlength\tabcolsep{1.2pt}
\centering
  \begin{tabular}{c|c|c}
    \toprule
    CD ($10^{-4}$)&PR-Senior&PR-Young\\
    \midrule
    w/ dis. &0.792&0.723\\
    \midrule
    w/o dis. &1.163&1.842 \\
    \bottomrule
  \end{tabular}
  \caption{Ablation study on the displacement field (10 views).}
  \label{table:ablation_disp}
\end{table}
\vspace{-0.3in}

\vspace{-0.1in}

\begin{figure}
\centering
\setlength\tabcolsep{0.25pt}
    \begin{scriptsize}
\begin{tabular}{ccccc}
    \centering
    \includegraphics[width=0.085\textwidth]{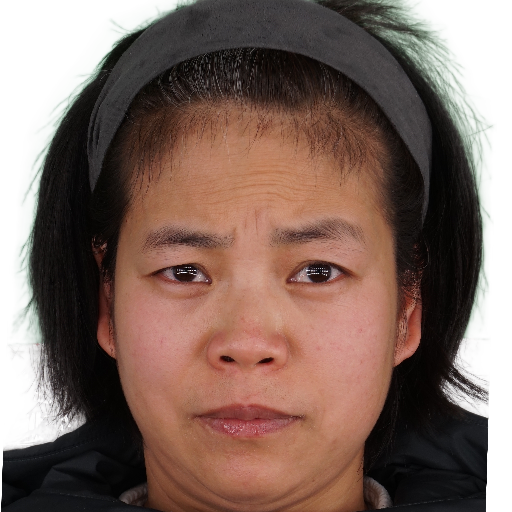} &
    \includegraphics[width=0.085\textwidth]{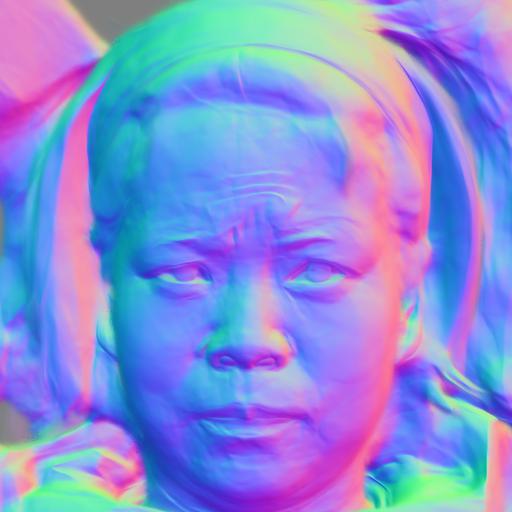} &
    \includegraphics[width=0.085\textwidth]{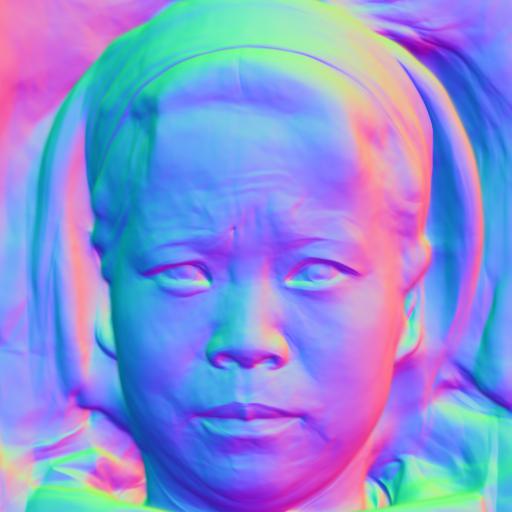} &
    \includegraphics[width=0.085\textwidth]{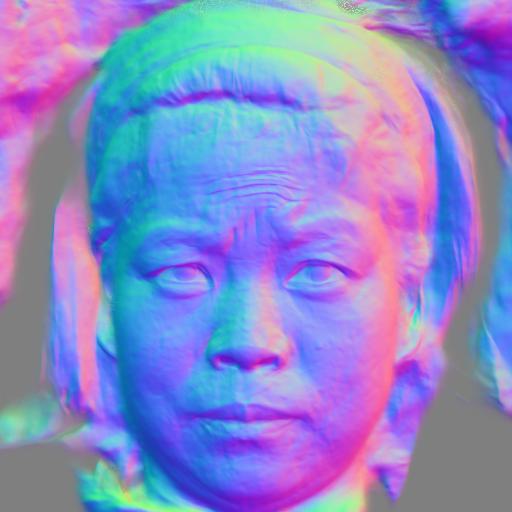} &
    \includegraphics[width=0.085\textwidth]{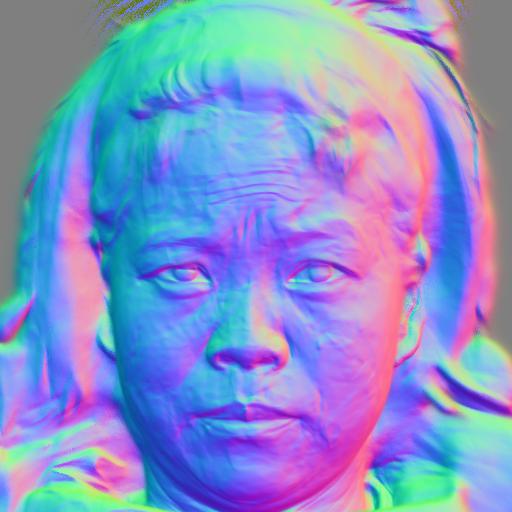}\\
    GT & NeuS & VolSDF & HF-NeuS & Ours\\
    \end{tabular}
    \end{scriptsize}
    \caption{Our method is capable of reconstructing the geometry of the entire head, in particular providing better results in the hair region of human heads compared to other methods. This is due to the fine refinement of the SDF using the displacement field. 
    Model: 429, 15 views. }
    %Hair geometry under 10 views。我们的方法在人头的头发部分展现了较精细的结果，得益于我们displacement对曲面精细的变形。
    \label{fig:hair_geometry}
\end{figure}

\begin{figure}
    \centering
    \includegraphics[width=0.09\textwidth]{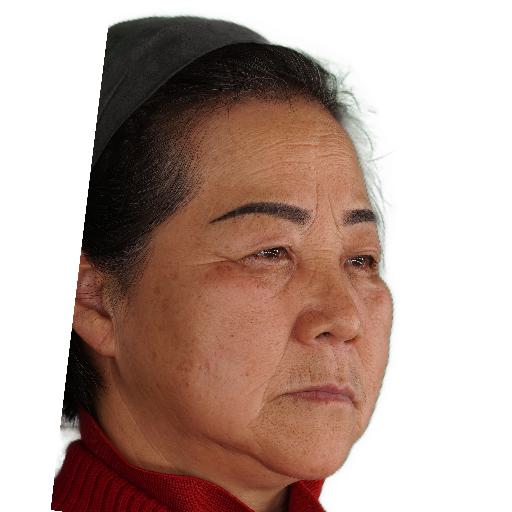}
    \includegraphics[width=0.09\textwidth]{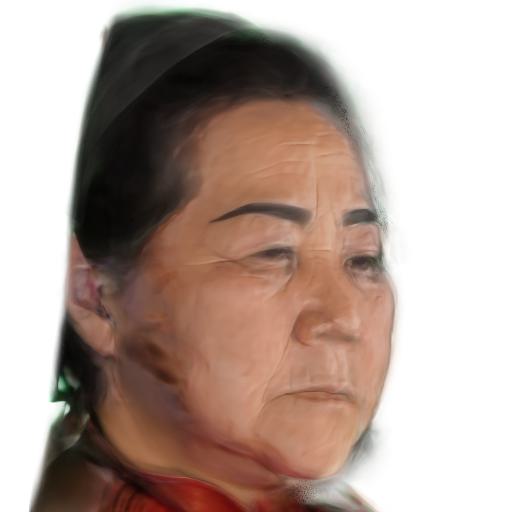}
    \includegraphics[width=0.09\textwidth]{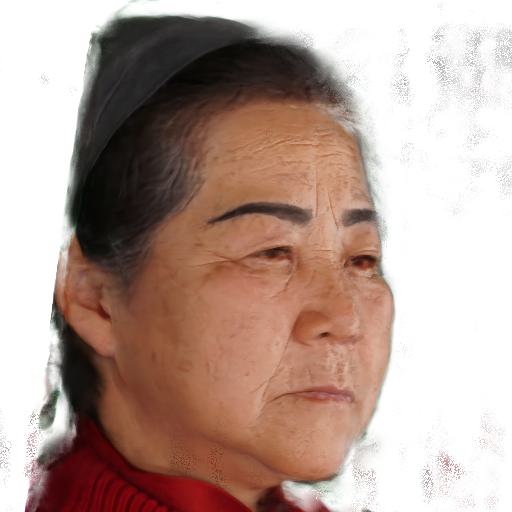}
    \includegraphics[width=0.09\textwidth]{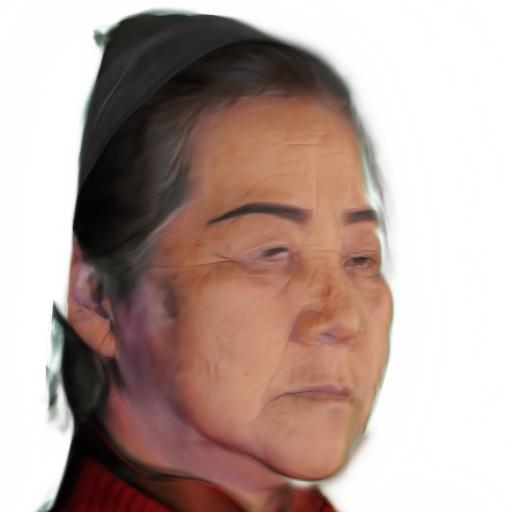}
    \includegraphics[width=0.09\textwidth]{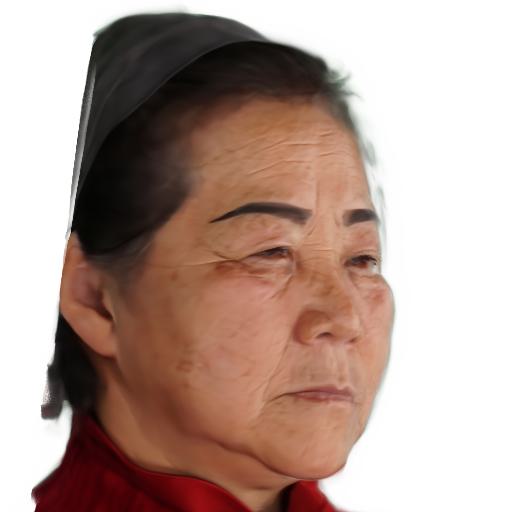}\\
    \includegraphics[width=0.09\textwidth]{./results/template_effects/gt_571_blank.jpg}
    \includegraphics[width=0.09\textwidth]{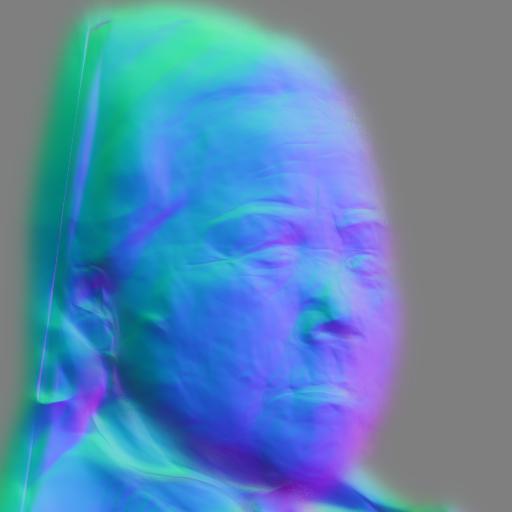}
    \includegraphics[width=0.09\textwidth]{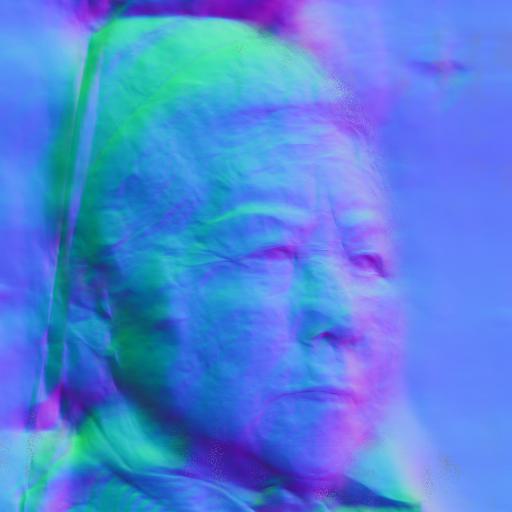}
    \includegraphics[width=0.09\textwidth]{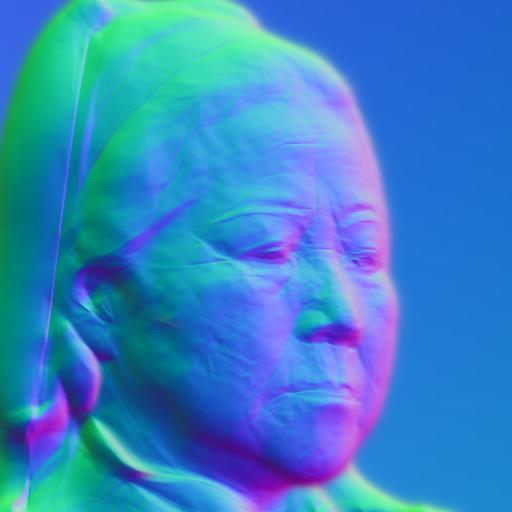}
    \includegraphics[width=0.09\textwidth]{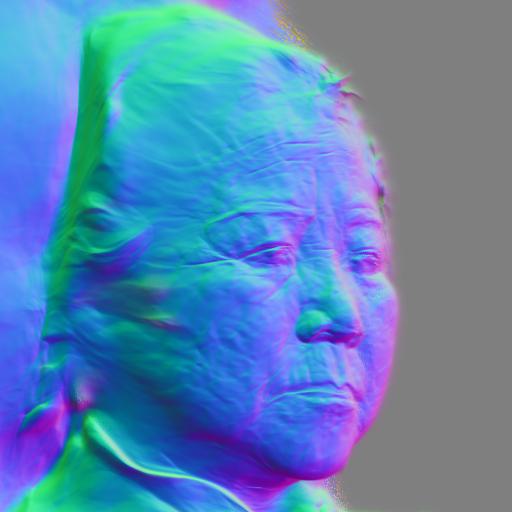}\\
    \makebox[0.6in]{\scriptsize GT}
    \makebox[0.6in]{\scriptsize NeuS}
    \makebox[0.6in]{\scriptsize HF-NeuS}
    \makebox[0.6in]{\scriptsize VolSDF}
    \makebox[0.6in]{\scriptsize Ours}
    \caption{Novel view synthesis result for Model 383 with only 5 views as input. }
    \label{fig:sparse_view}
\end{figure}

\vspace{-0.1in}

\section{Conclusion \& Future Work}
%\vspace{-0.1in}
We present a novel neural rendering model for reconstructing 3D human heads under low-view settings. By decomposing the geometry of 3D heads into an identity-independent template and two identity-dependent components (a non-rigid deformation and a displacement field), we train our network in two separate stages in a coarse-to-fine manner. Through extensive evaluation, we demonstrate that our method is robust and can accurately reconstruct 3D heads with high-quality geometry. Moreover, it outperforms state-of-the-art methods in terms of geometry accuracy and novel view synthesis with 10 to 20 views as input. The pre-trained template serves a good initialization for our model to adapt to unseen individuals.
%, which reduces the need for extensive data collection and annotation

Our approach currently models hair as a whole using implicit functions. While this is effective for capturing the overall shape of the hair (see Figure~\ref{fig:hair_geometry}), more accurate hair modeling could be achieved by using strands~\cite{rosu2022neural}. Our training time is similar to HF-NeuS, which is approximately double the training time required by VolSDF. There are promising tools such as Plenvoxels~\cite{fridovich2022plenoxels} and Plenoctree~\cite{yu2021plenoctrees} that have demonstrated substantial improvements in the training and inference time of neural radiance fields~\cite{mildenhall2021nerf}. Given this, it is highly desirable to develop similar tools for neural implicit functions, as this could significantly improve the scalability and practicality of our proposed approach.

%%%%%%%%% BODY TEXT
\begin{small}
\noindent\textbf{Acknowledgement.} We thank Prof. Dr. Feng Xu, School of Software and BNRist, Tsinghua University and SenseTime for the Portrait Relighting dataset. This study is supported under the RIE2020 Industry Alignment Fund – Industry Collaboration Projects (IAF-ICP) Funding Initiative, as well as cash and in-kind contribution from the industry partner(s). This project was also partially supported by the Ministry of Education, Singapore, under its Academic Research Fund Grants (MOE-T2EP20220-0005, RG20/20 \& RT19/22).
\end{small}
%%%%%%%%% REFERENCES
{\small
\bibliographystyle{ieee_fullname}
\bibliography{egbib}
}
\newpage
% \appendix
\setcounter{table}{0}
\setcounter{figure}{0}
\renewcommand{\thetable}{A\arabic{table}}
\renewcommand{\thefigure}{A\arabic{figure}}
\begin{appendices}
\section{Appendix}
\begin{figure*}[htbp]
    \centering
    \includegraphics[width=0.245\textwidth]{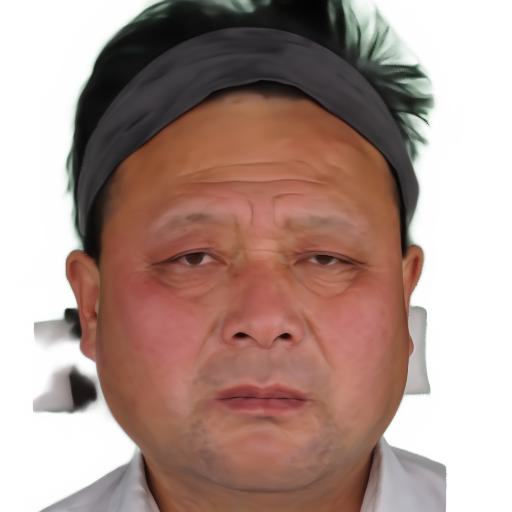}
    \includegraphics[width=0.245\textwidth]{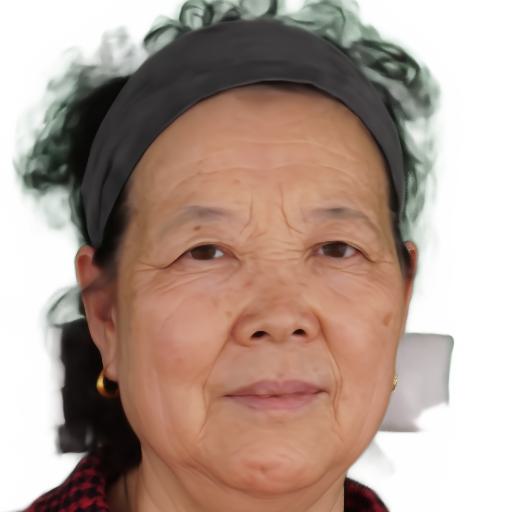}
    \includegraphics[width=0.245\textwidth]{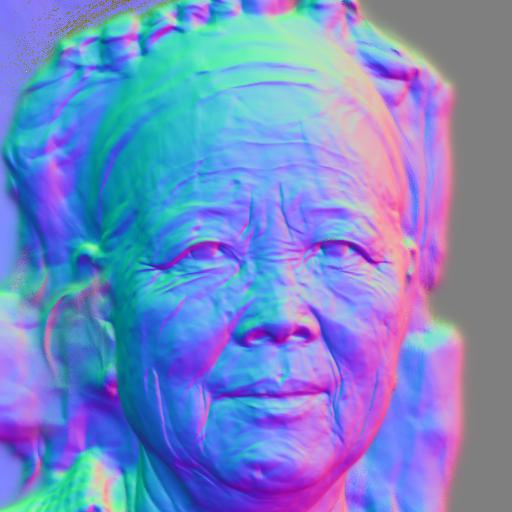}
    \includegraphics[width=0.245\textwidth]{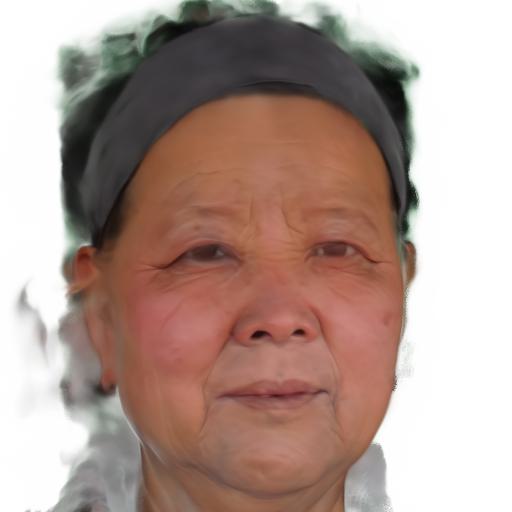}\\
    \includegraphics[width=0.245\textwidth,trim={300 170 110 290}, clip]{color_trans_20/ref_454_15.jpg}
    \includegraphics[width=0.245\textwidth,trim={300 170 110 290}, clip]{color_trans_20/src_399_15.jpg}
    \includegraphics[width=0.245\textwidth,trim={300 170 110 290}, clip]{color_trans_20/color_454_399_15_normal.jpg}
    \includegraphics[width=0.245\textwidth,trim={300 170 110 290}, clip]{color_trans_20/color_454_399_15.jpg}\\
    \includegraphics[width=0.245\textwidth]{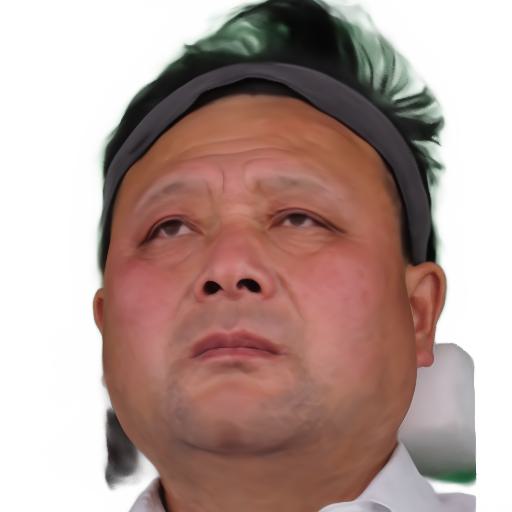}
    \includegraphics[width=0.245\textwidth]{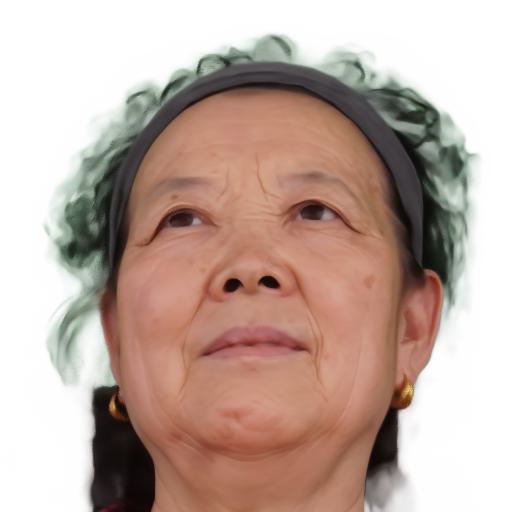}
    \includegraphics[width=0.245\textwidth]{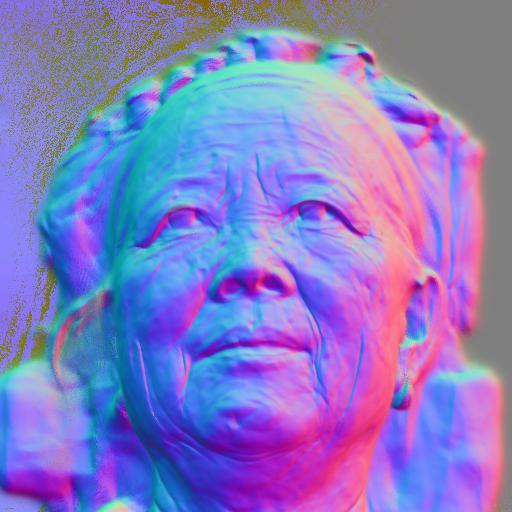}
    \includegraphics[width=0.245\textwidth]{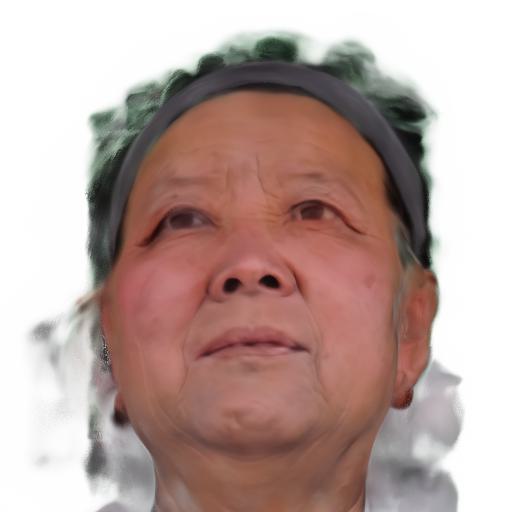}\\
    \includegraphics[width=0.245\textwidth,trim={300 200 110 260}, clip]{color_trans_20/ref_454_6.jpg}
    \includegraphics[width=0.245\textwidth,trim={300 200 110 260}, clip]{color_trans_20/src_399_6.jpg}
    \includegraphics[width=0.245\textwidth,trim={300 200 110 260}, clip]{color_trans_20/color_454_399_6_normal.jpg}
    \includegraphics[width=0.245\textwidth,trim={300 200 110 260}, clip]{color_trans_20/color_454_399_6.jpg}\\
    \includegraphics[width=0.245\textwidth]{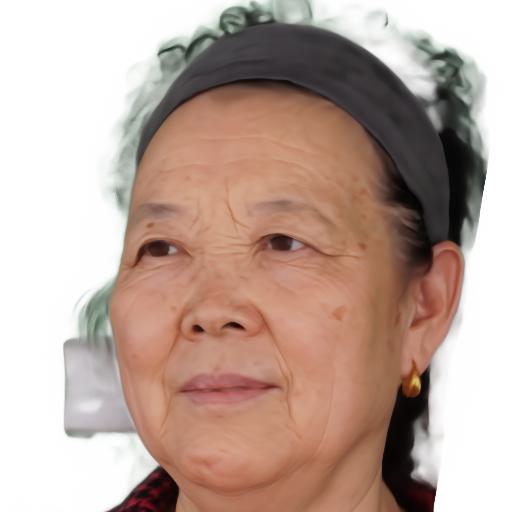}
    \includegraphics[width=0.245\textwidth]{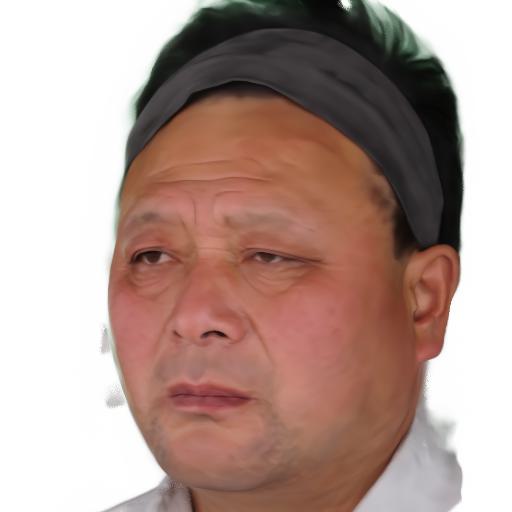}
    \includegraphics[width=0.245\textwidth]{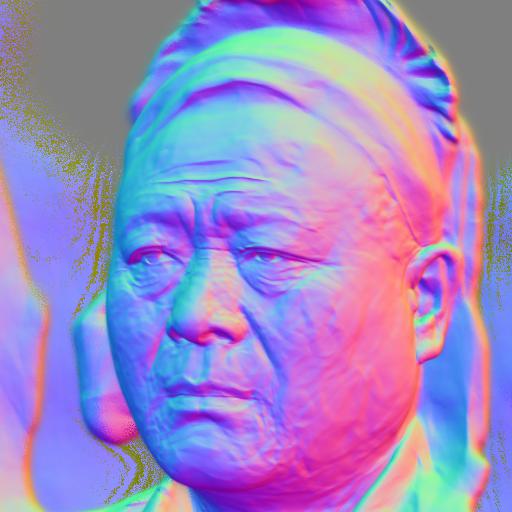}
    \includegraphics[width=0.245\textwidth]{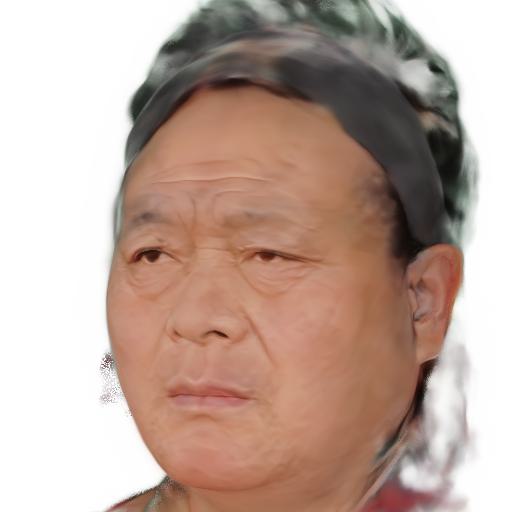}\\
    \includegraphics[width=0.245\textwidth, trim={300 170 110 280}, clip]{color_trans_20/ref_399_17.jpg}
    \includegraphics[width=0.245\textwidth, trim={300 170 110 280}, clip]{color_trans_20/src_454_17.jpg}
    \includegraphics[width=0.245\textwidth, trim={300 170 110 280}, clip]{color_trans_20/color_399_454_17_normal.jpg}
    \includegraphics[width=0.245\textwidth, trim={300 170 110 280}, clip]{color_trans_20/color_399_454_17.jpg}\\
    \makebox[0.245\textwidth]{Reference identity}
    \makebox[0.245\textwidth]{Source identity}
    \makebox[0.245\textwidth]{Source identity normal map}
    \makebox[0.245\textwidth]{Color transfer result}
    \caption{Our approach of decomposing geometry and appearance enables us to transfer the color appearance from one model to another while keeping the geometry unchanged. In this figure, we provide two examples of transferring skin colors from a reference identity to a source identity. All models are trained under 20 views. Notably, our method can preserve small geometric features, such as speckles, which are encoded in the SDF. This is in contrast to existing image-based color transfer algorithms, which cannot differentiate between geometric features and skin colors, often transferring them together.}
    \label{fig:color_trans}
\end{figure*}

In the appendix, we present 1) additional results of unseen identities and low-view inputs in
Section~\ref{subsec:unseen_sparse_sec}, which demonstrate that the pre-trained template serves as a good initialization and enables our method to adapt to new identities not available in the training dataset; 2) detailed results of our method and compare them with NeuS, HF-NeuS, and VolSDF on the PR-Senior and PR-Young datasets under a 10-view setting in Section~\ref{subsec:analysis_sec}; and 3) an application on color transfer in 
Figure~\ref{fig:color_trans} to demonstrate the flexibility and potential of our geometry decomposition.

\begin{table*}[htbp]\centering
\setlength\tabcolsep{2pt}
    \label{tab:notation}
    %\begin{small}
    %\begin{tabular}{lp{2.5in}}
    \begin{tabular}{ll}
        \hline
        Symbol & Meaning\\
        \hline
        \hline 
         $I_i$ &  the input images with camera parameters\\
         \hline
         %$\pi_i$ & camera parameters \\
         $f_{\text{geo}}$ &the Geometry Network \\
         $f_\text{tem}$  &the Template Network \\
         $f_\text{def}$ &the Deformation Network \\
         $f_{\text{ren}}$ & the Rendering Network \\
         $f_{\text{dis}}$  & the Displacement Network \\
          \hline
         $\mathbf{z}_s, \mathbf{z}_c\in\mathbb{R}^{128}$ & identity-dependent latent codes for shape and color  \\
$\mathbf{F}_{\text{def}}\in\mathbb{R}^{192}$ & identity-dependent feature associated with non-rigid deformation\\
         $\mathbf{F}_{\text{tem}}\in\mathbb{R}^{64}$ & identity-independent feature associated with the template head\\
         $\mathbf{F}_{\text{dis}}\in\mathbb{R}^{64}$ & ID-dep. geometry feature associated with displacement\\
         
         $\mathbf{F}_{\text{all}}\in\mathbb{R}^{320}$ & \small{the overall feature fed into the Rendering Network in Stage 2, which is the concatenation of $\mathbf{F}_\text{def}$,  $\mathbf{F}_\text{tem}$, and $\mathbf{F}_\text{dis}$}\\
         \hline
         $\mathbf{x}\in\mathbb{R}^3$ & a query point in the observation space\\
         $\mathbf{d}\in\mathbb{R}^3$ & an offset vector indicating the deformation from an individual to the template \\
         $\mathbf{x+d}\in\mathbb{R}^3$  & a query point in the template space\\
         $s\in\mathbb{R}$ & signed distance \\
         $\mathbf{n}_b, \mathbf{n}_f\in \mathbb{R}^3$ & normal vectors of the base and final surfaces\\
         $\delta\in\mathbb{R}$ & an implicit displacement\\
         $c\in\mathbb{R}^3$ & radiance\\
         $C\in\mathbb{R}^3$ & rgb color\\
        \bottomrule
    \end{tabular}
    \caption{Notation Table.}
\end{table*}

\subsection{Experimental Results}
\label{subsec:unseen_sparse_sec}

In this section, we present additional results of unseen identities and results of sparse views.

\subsubsection{Unseen Identities}

Our method has the ability to adapt to new individuals as the pre-trained template serves as a good initialization. 
To verify this, we consider 3 \textbf{new} identities (Models 552, 555 and 598) and each identity is associated with only 5 views. 

We adopted the pre-trained template, i.e., the one trained on 30 identities the PR-Senior and PR-Young datasets with 10 views for each identity, to learn the fine details for each identity in Stage 2.
We observed that our method also produced plausible results for the unseen identities as shown in Table~\ref{tab:unseen_sparse}. This demonstrates that the pre-trained template can adapt to new identities.

\subsubsection{Small Dataset}

\begin{figure*}
    \centering
    \rotatebox{90}{\textbf{552}}
    \includegraphics[width=0.18\textwidth]{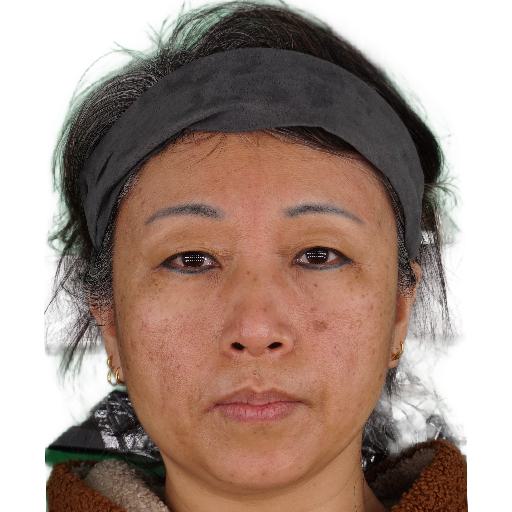}
    \includegraphics[width=0.18\textwidth]{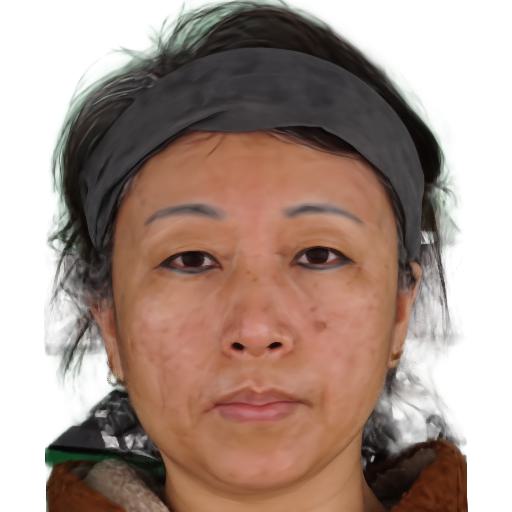}
    \includegraphics[width=0.18\textwidth]{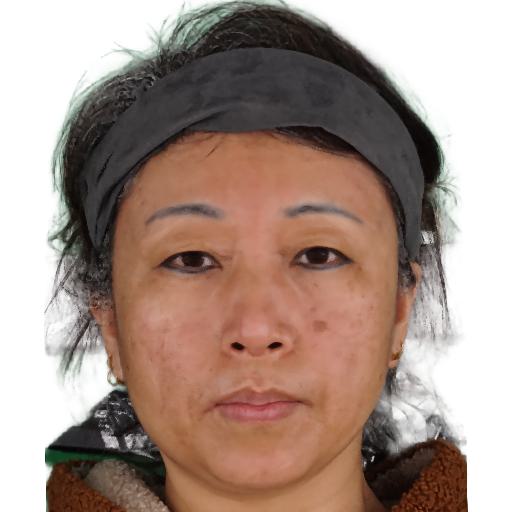}
    \includegraphics[width=0.18\textwidth]{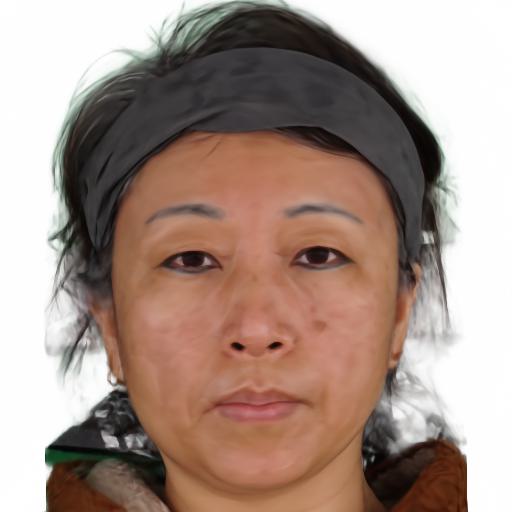}
    \includegraphics[width=0.18\textwidth]{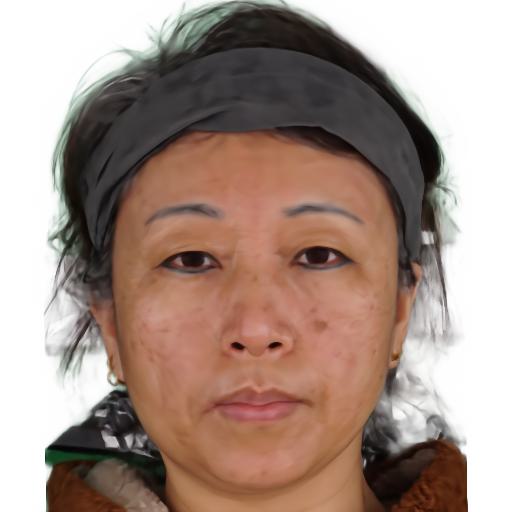}\\
    \rotatebox{90}{}
    \includegraphics[width=0.18\textwidth]{./results/template_effects/gt_571_blank.jpg}
    \includegraphics[width=0.18\textwidth]{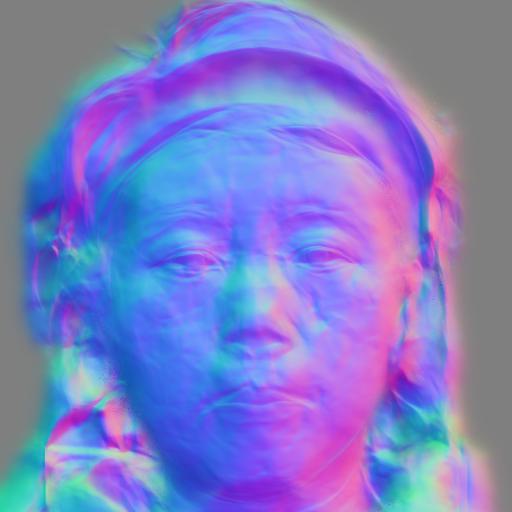}
    \includegraphics[width=0.18\textwidth]{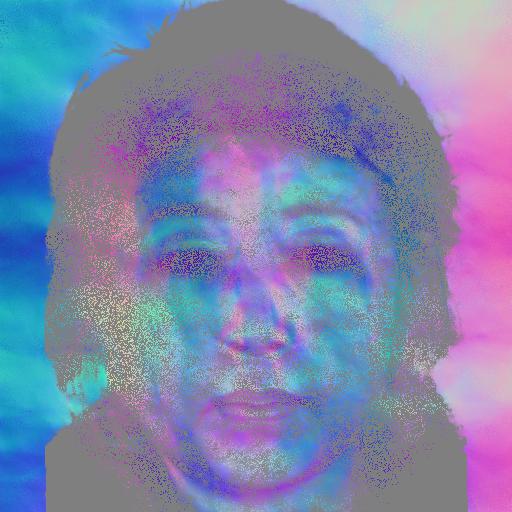}
    \includegraphics[width=0.18\textwidth]{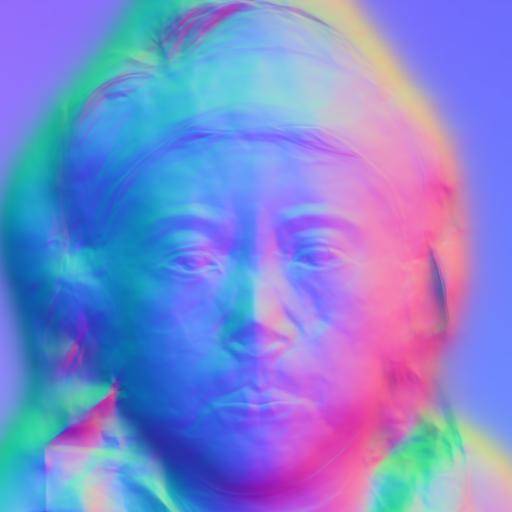}
    \includegraphics[width=0.18\textwidth]{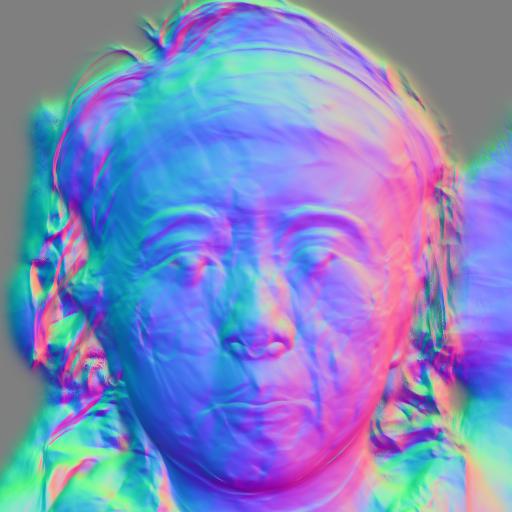}\\
    
    \rotatebox{90}{\textbf{555}}
    \includegraphics[width=0.18\textwidth]{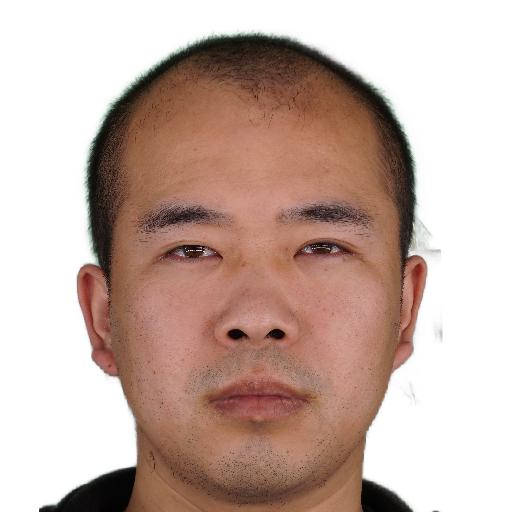}
    \includegraphics[width=0.18\textwidth]{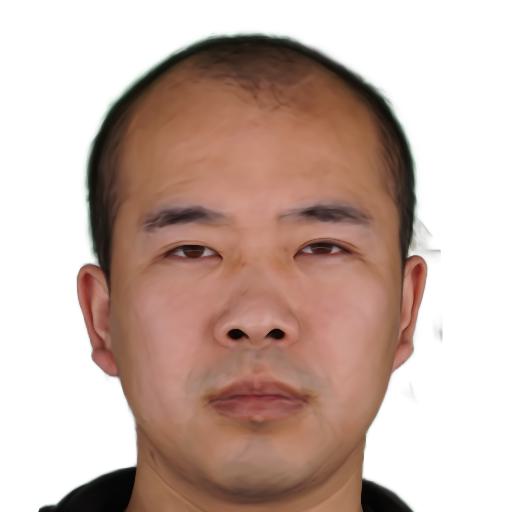}
    \includegraphics[width=0.18\textwidth]{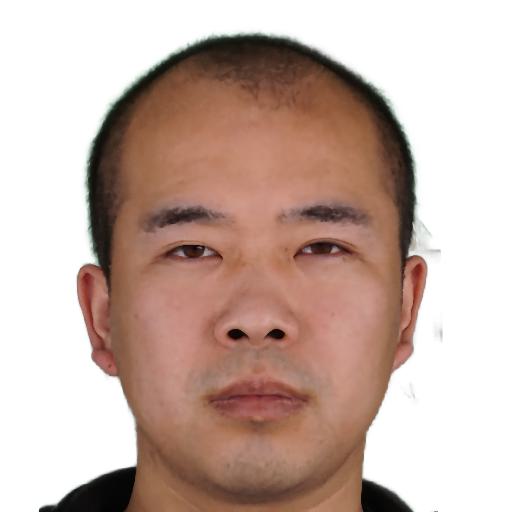}
    \includegraphics[width=0.18\textwidth]{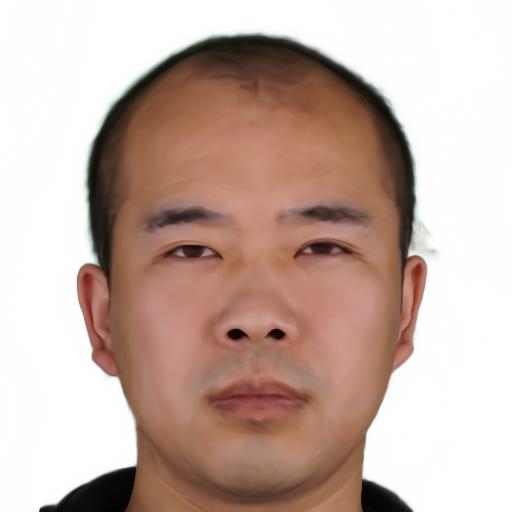}
    \includegraphics[width=0.18\textwidth]{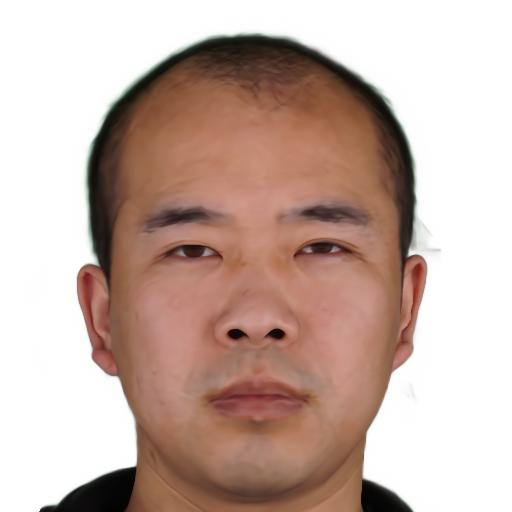}\\
    \rotatebox{90}{}
    \includegraphics[width=0.18\textwidth]{./results/template_effects/gt_571_blank.jpg}
    \includegraphics[width=0.18\textwidth]{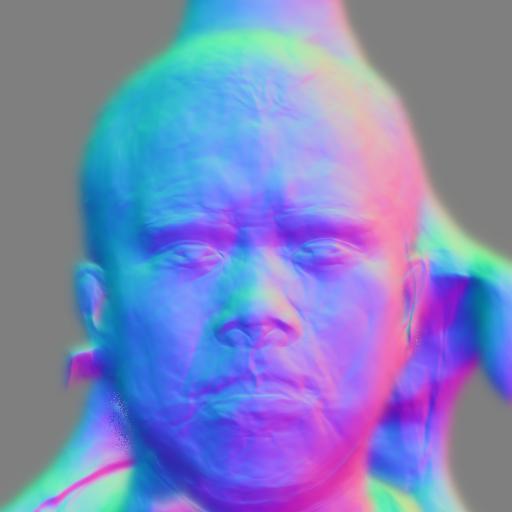}
    \includegraphics[width=0.18\textwidth]{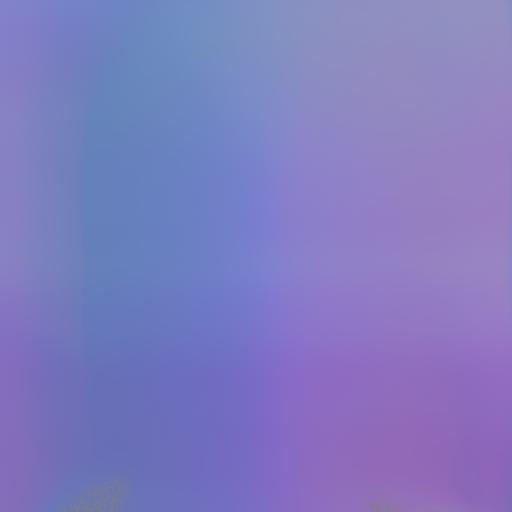}
    \includegraphics[width=0.18\textwidth]{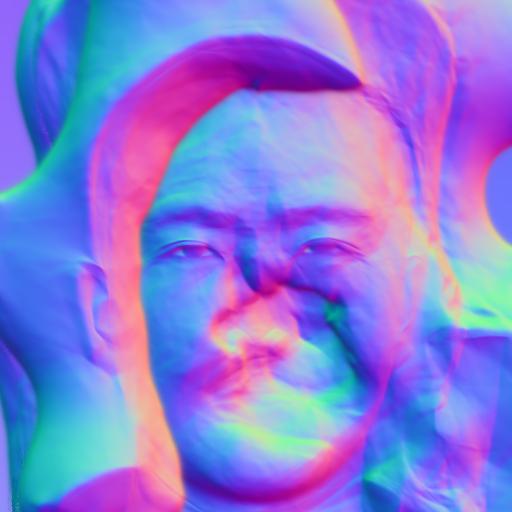}
    \includegraphics[width=0.18\textwidth]{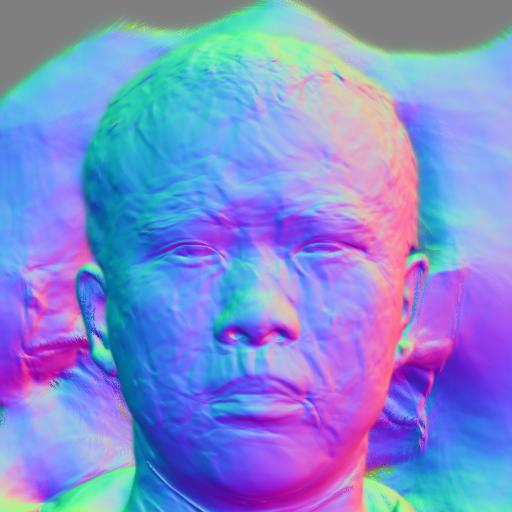}\\

    \rotatebox{90}{\textbf{598}}
    \includegraphics[width=0.18\textwidth]{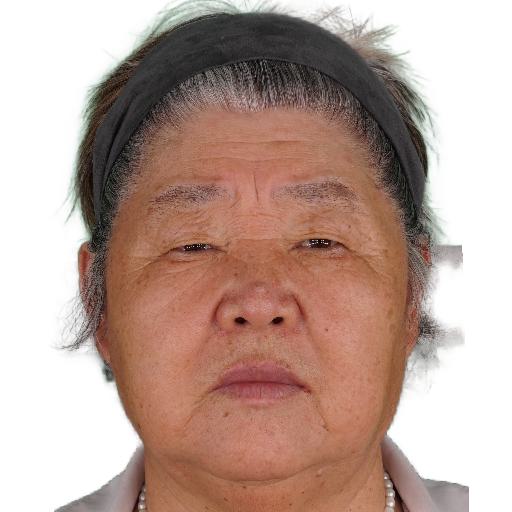}
    \includegraphics[width=0.18\textwidth]{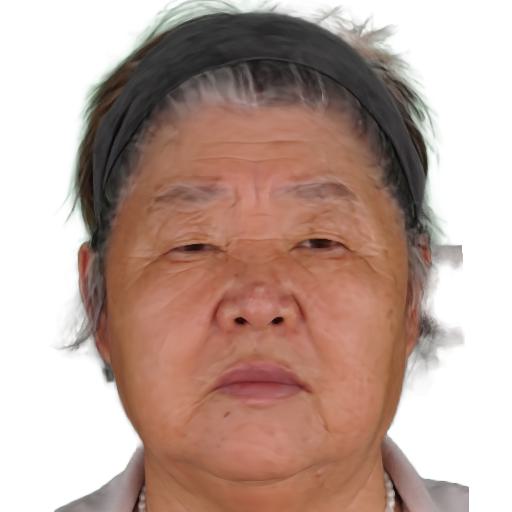}
    \includegraphics[width=0.18\textwidth]{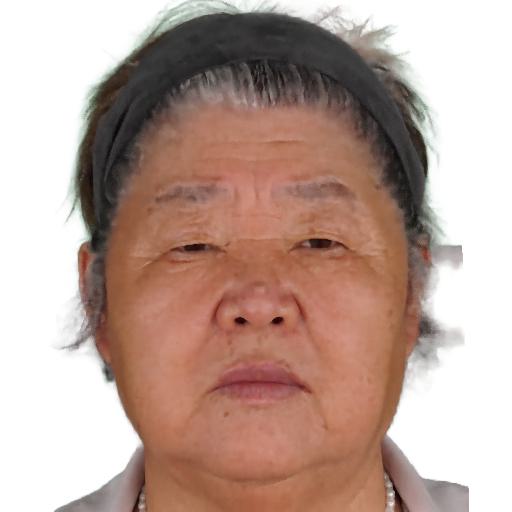}
    \includegraphics[width=0.18\textwidth]{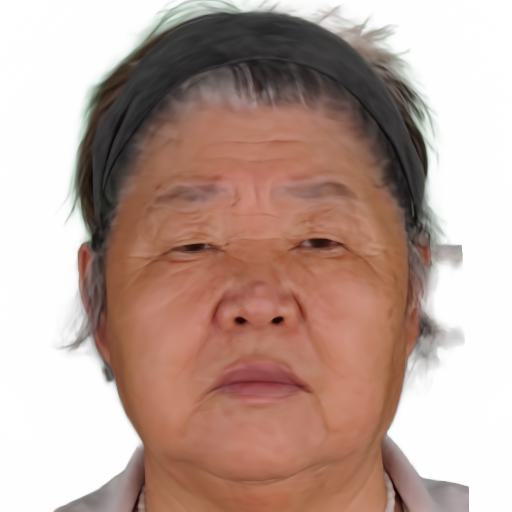}
    \includegraphics[width=0.18\textwidth]{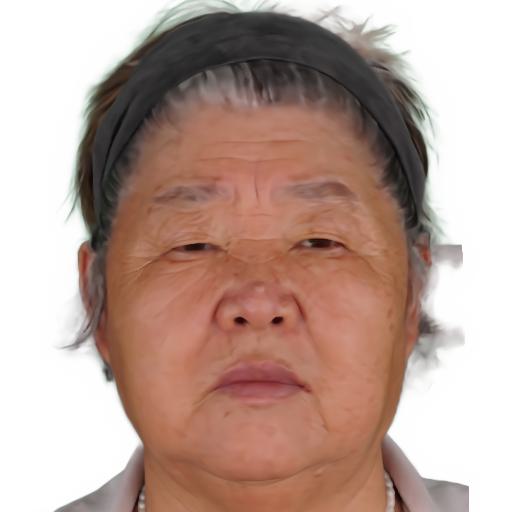}\\
    \rotatebox{90}{}
    \includegraphics[width=0.18\textwidth]{./results/template_effects/gt_571_blank.jpg}
    \includegraphics[width=0.18\textwidth]{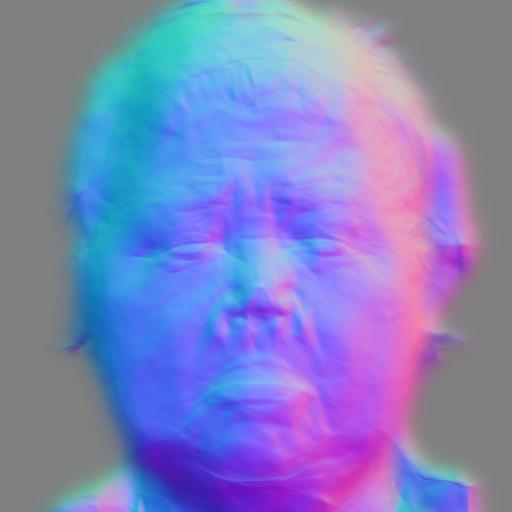}
    \includegraphics[width=0.18\textwidth]{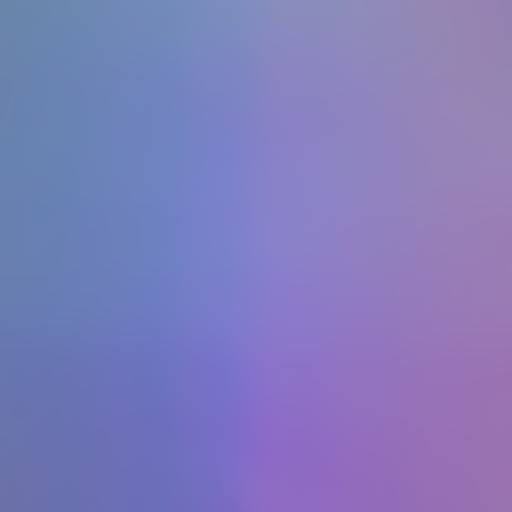}
    \includegraphics[width=0.18\textwidth]{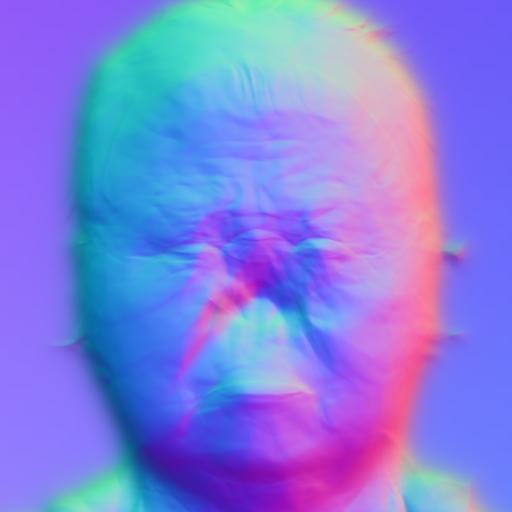}
    \includegraphics[width=0.18\textwidth]{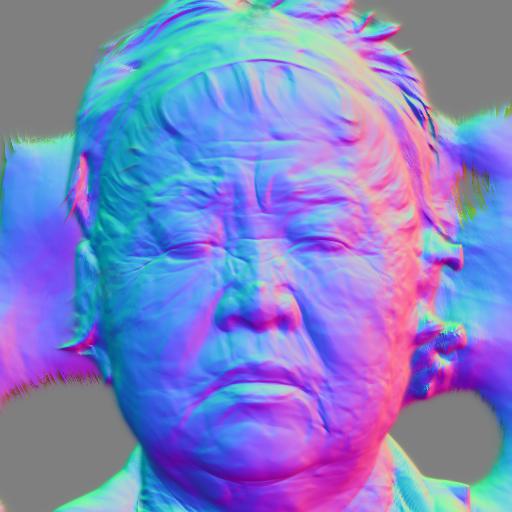}\\
    \makebox[0.18\textwidth]{\scriptsize GT}
    \makebox[0.18\textwidth]{\scriptsize NeuS}
    \makebox[0.18\textwidth]{\scriptsize HF-NeuS}
    \makebox[0.18\textwidth]{\scriptsize VolSDF}
    \makebox[0.18\textwidth]{\scriptsize Ours}
    \caption{Training view results for 3 unseen identities (552, 555, 598) with only 5 views.}
    \label{fig:unseen_result_train}
\end{figure*}

\begin{figure*}
    \centering
    \rotatebox{90}{\textbf{552}}
    \includegraphics[width=0.18\textwidth]{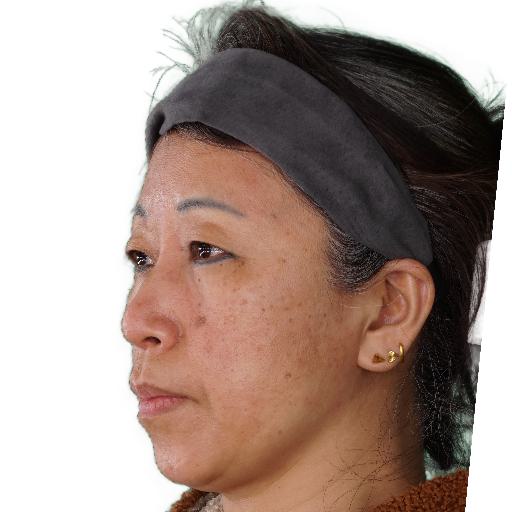}
    \includegraphics[width=0.18\textwidth]{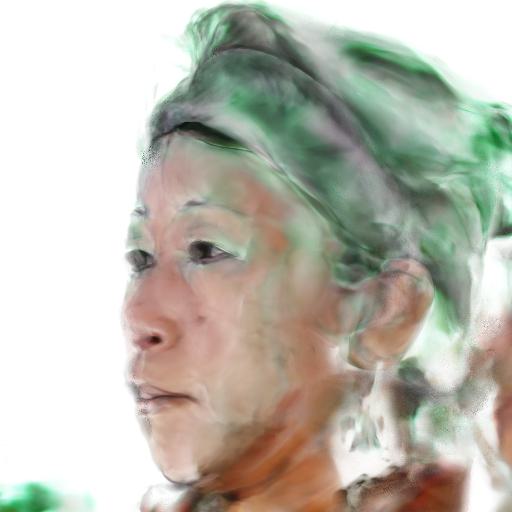}
    \includegraphics[width=0.18\textwidth]{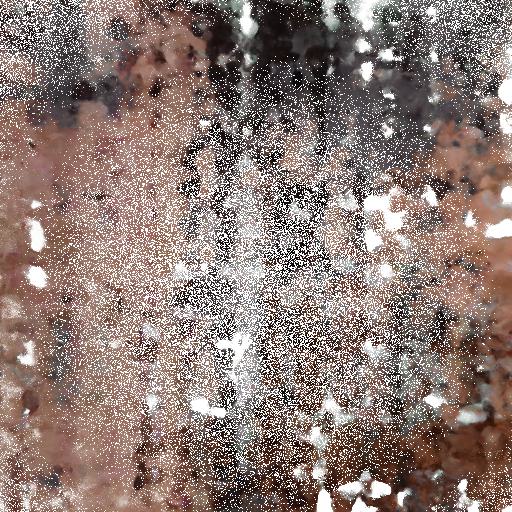}
    \includegraphics[width=0.18\textwidth]{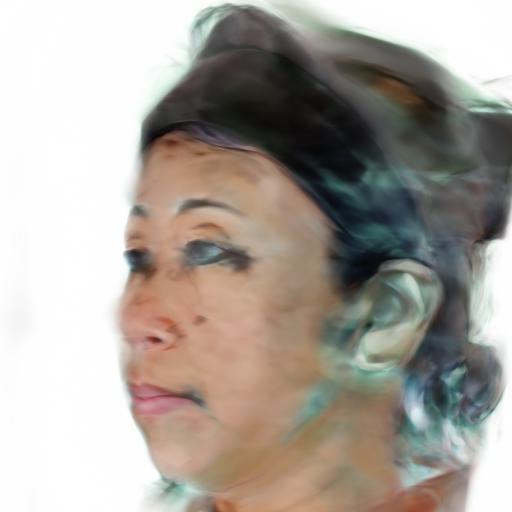}
    \includegraphics[width=0.18\textwidth]{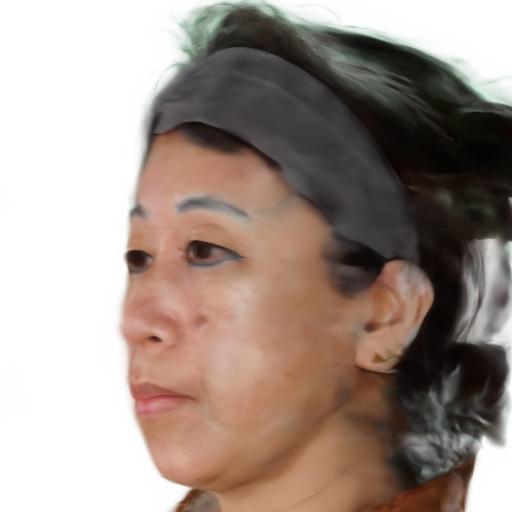}\\
    \rotatebox{90}{}
    \includegraphics[width=0.18\textwidth]{./results/template_effects/gt_571_blank.jpg}
    \includegraphics[width=0.18\textwidth]{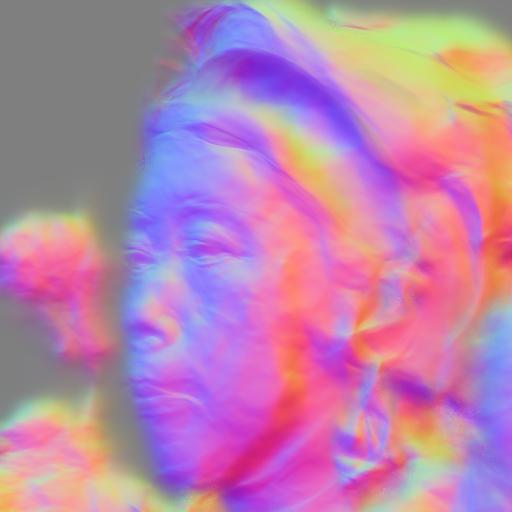}
    \includegraphics[width=0.18\textwidth]{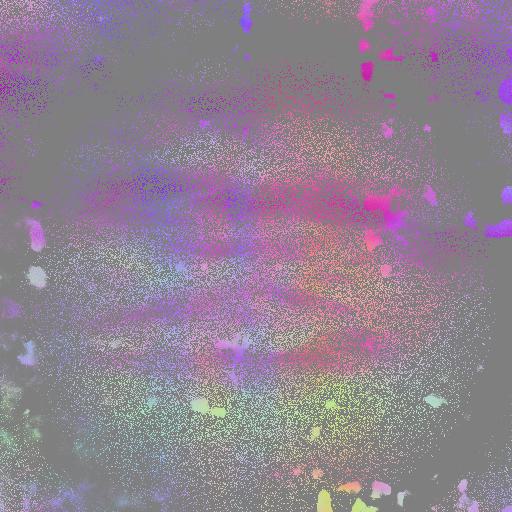}
    \includegraphics[width=0.18\textwidth]{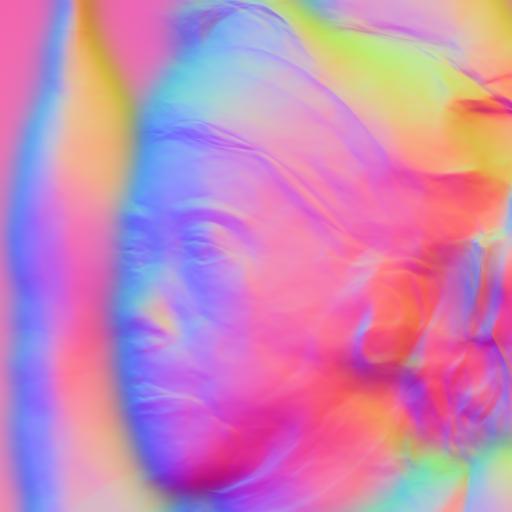}
    \includegraphics[width=0.18\textwidth]{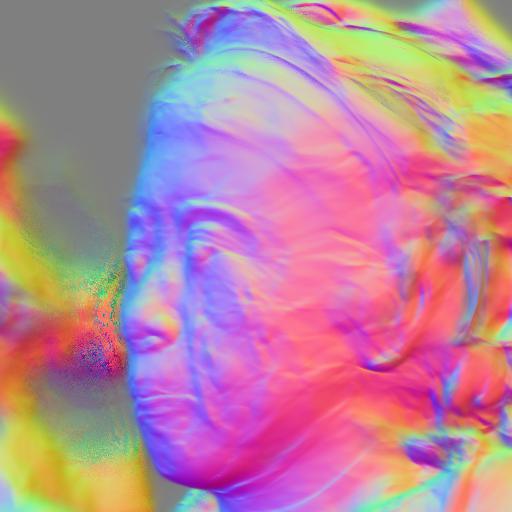}\\
    
    \rotatebox{90}{\textbf{555}}
    \includegraphics[width=0.18\textwidth]{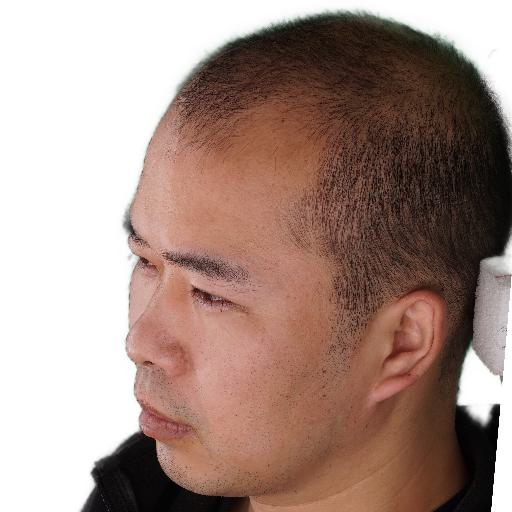}
    \includegraphics[width=0.18\textwidth]{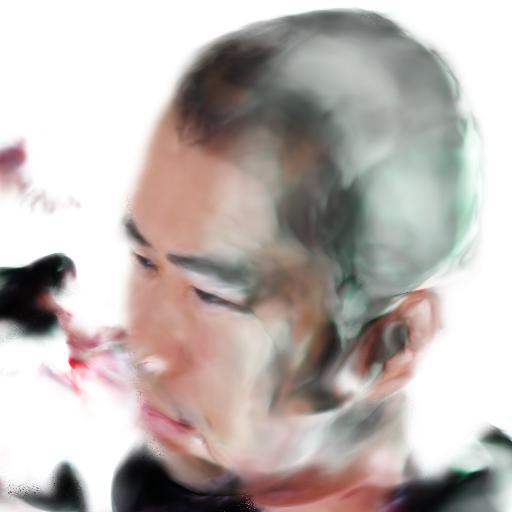}
    \includegraphics[width=0.18\textwidth]{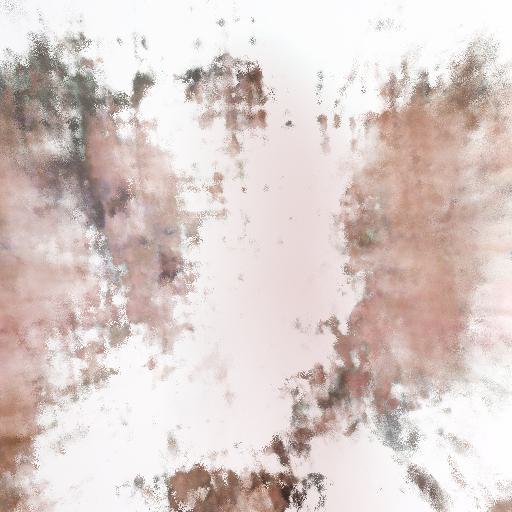}
    \includegraphics[width=0.18\textwidth]{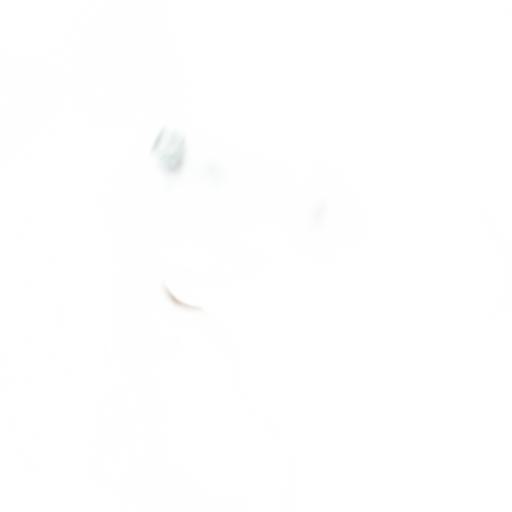}
    \includegraphics[width=0.18\textwidth]{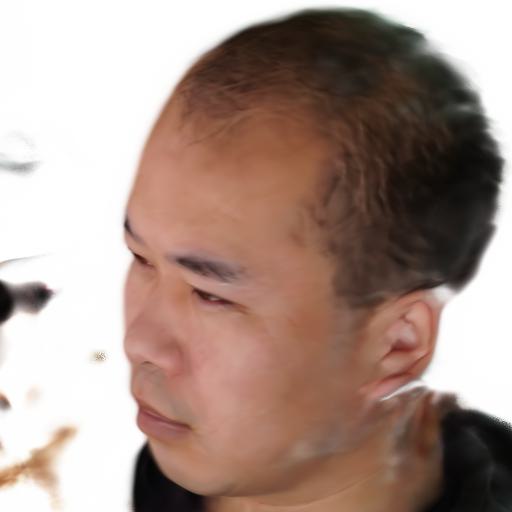}\\
    \rotatebox{90}{}
    \includegraphics[width=0.18\textwidth]{./results/template_effects/gt_571_blank.jpg}
    \includegraphics[width=0.18\textwidth]{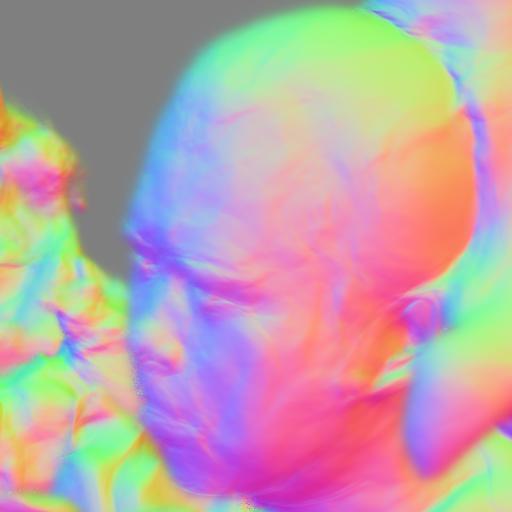}
    \includegraphics[width=0.18\textwidth]{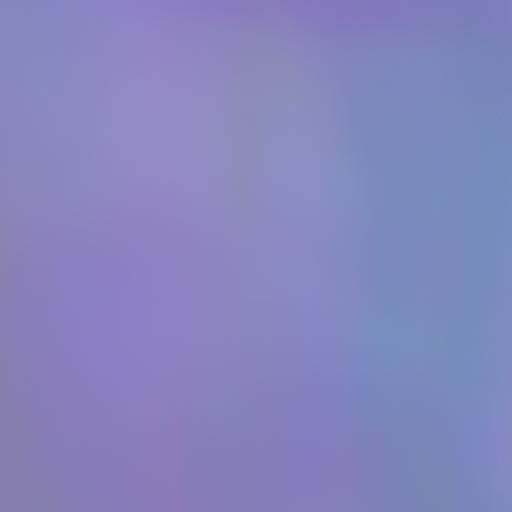}
    \includegraphics[width=0.18\textwidth]{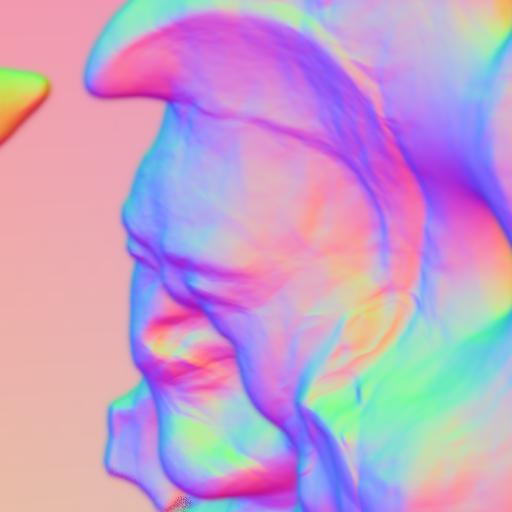}
    \includegraphics[width=0.18\textwidth]{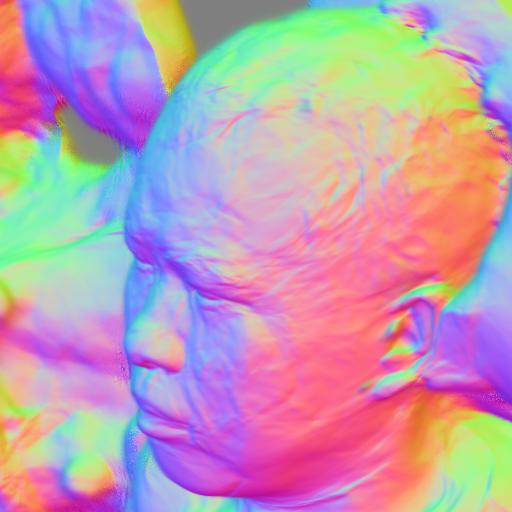}\\

    \rotatebox{90}{\textbf{598}}
    \includegraphics[width=0.18\textwidth]{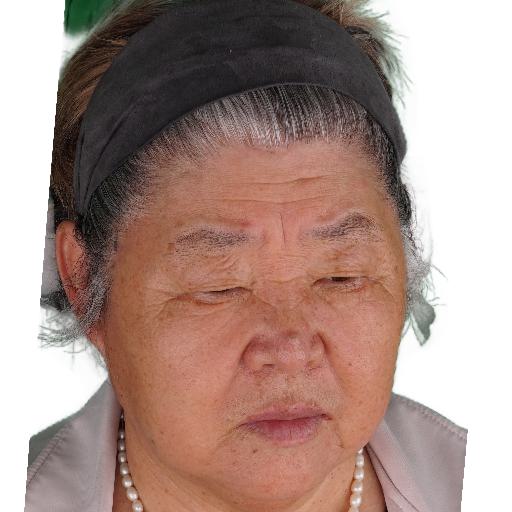}
    \includegraphics[width=0.18\textwidth]{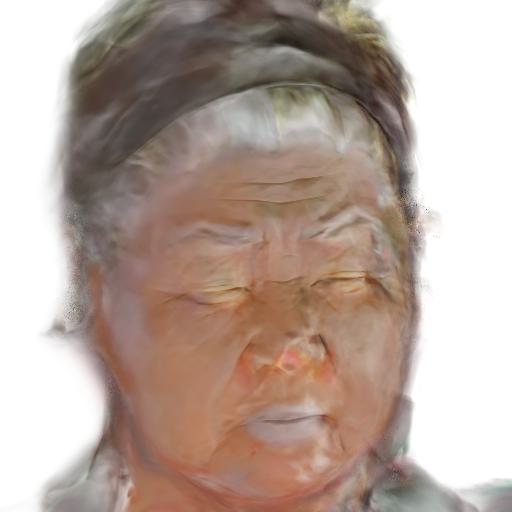}
    \includegraphics[width=0.18\textwidth]{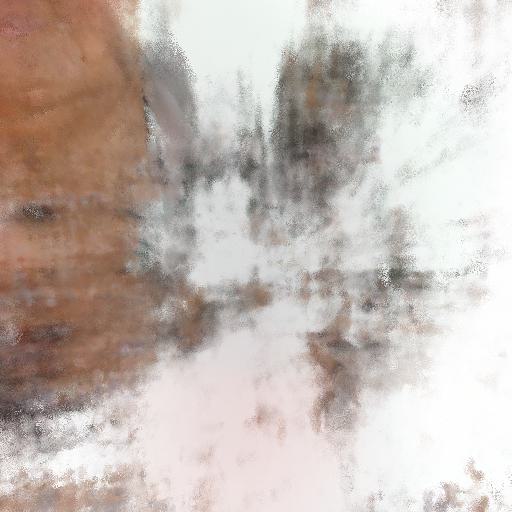}
    \includegraphics[width=0.18\textwidth]{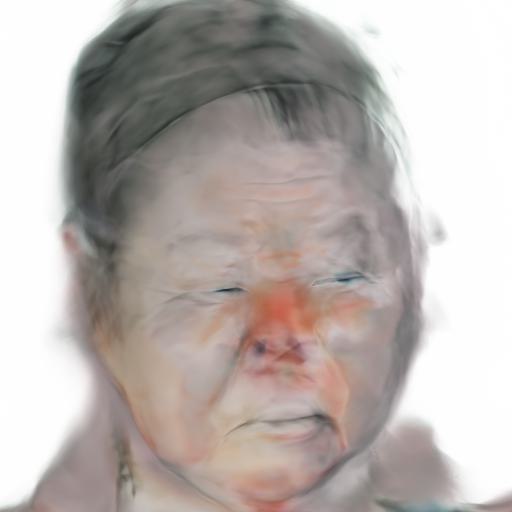}
    \includegraphics[width=0.18\textwidth]{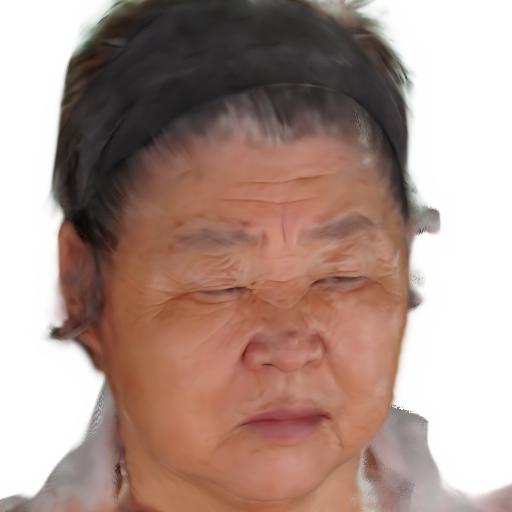}\\
    \rotatebox{90}{}
    \includegraphics[width=0.18\textwidth]{./results/template_effects/gt_571_blank.jpg}
    \includegraphics[width=0.18\textwidth]{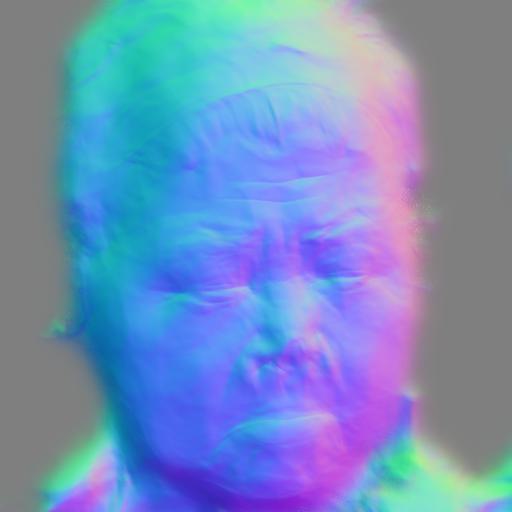}
    \includegraphics[width=0.18\textwidth]{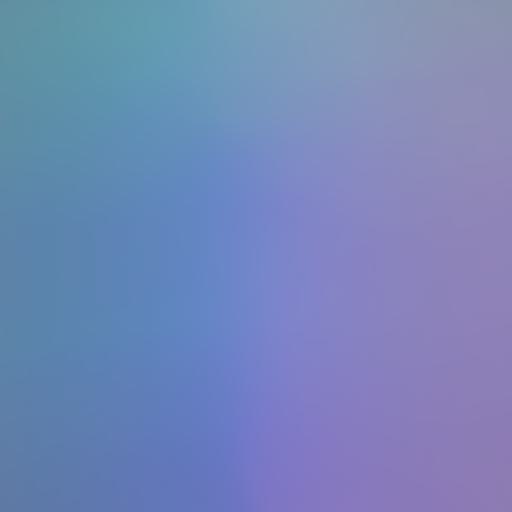}
    \includegraphics[width=0.18\textwidth]{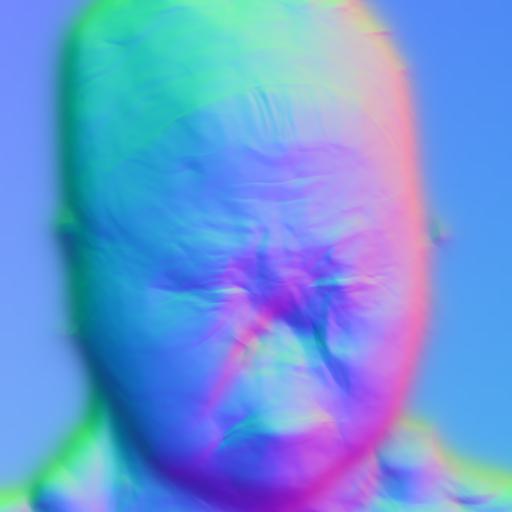}
    \includegraphics[width=0.18\textwidth]{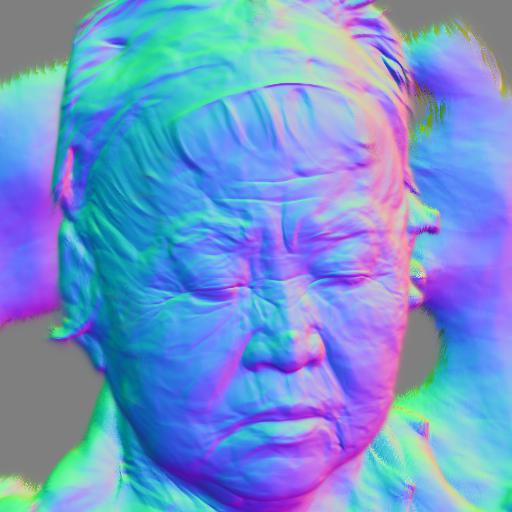}\\
    \makebox[0.18\textwidth]{\scriptsize GT}
    \makebox[0.18\textwidth]{\scriptsize NeuS}
    \makebox[0.18\textwidth]{\scriptsize HF-NeuS}
    \makebox[0.18\textwidth]{\scriptsize VolSDF}
    \makebox[0.18\textwidth]{\scriptsize Ours}
    \caption{Novel view results for 3 unseen identities (552, 555, 598) with only 5 views.}
    \label{fig:unseen_result_novel}
\end{figure*}

We also conducted another experiment using the 3 unseen identities as a small dataset. In the second experiment, we trained a \textbf{new} template using only the 15 images of the 3 new identities in Stage 1 and then used the template to learn the fine details for each identity in Stage 2.
Without a surprise, the geometry of the newly trained template is worse than that of the pre-trained template due to significantly fewer views involved in Stage 1 training. Still, our method produced a fairly good result and none of the other methods, VolSDF, NeuS and HF-NeuS, were able to reconstruct satisfactory geometry with only 5 views as input as illustrated in Figure~\ref{fig:unseen_result_train} and Figure~\ref{fig:unseen_result_novel}. 

\begin{table*}[htbp]
\setlength\tabcolsep{2pt}
\scriptsize
    \centering
    \begin{tabular}{c|ccc|ccc|ccc|ccc|ccc}
    \hline
        \multirow{2}{*}{Model} & \multicolumn{3}{c|}{NeuS} & \multicolumn{3}{c|}{HF-NeuS} & \multicolumn{3}{c|}{VolSDF} & \multicolumn{3}{c|}{Ours(Template on the 3 ids)} & \multicolumn{3}{c}{Ours(Template on 30 ids)}\\
        % \cmidrule[0.5pt](rl){2-16}
        \cline{2-16}
        &CD ($10^{-4}$) & PSNR$_\text{t}$ & PSNR$_\text{n}$ 
        &CD ($10^{-4}$) & PSNR$_\text{t}$ & PSNR$_\text{n}$ 
        &CD ($10^{-4}$) & PSNR$_\text{t}$ & PSNR$_\text{n}$
        &CD ($10^{-4}$) & PSNR$_\text{t}$ & PSNR$_\text{n}$
        &CD ($10^{-4}$) & PSNR$_\text{t}$ & PSNR$_\text{n}$\\
        \hline
        552 & 3.769&35.22&21.64&N.A.&35.40&13.36& 8.192 &33.62&23.13&1.815 &33.58&26.51& 1.197 &34.92&25.88 \\
        555 & 3.614&35.37&18.97&N.A.&35.65&12.19&243.4&33.45&11.80&1.254 &33.30&24.14& 1.071&35.04&23.39\\
        598 & 16.25&36.39&21.89&N.A.&36.30&12.81&23.76&35.63&20.27&1.056 &35.70&27.02& 1.020&35.50&27.39\\
        \hline
    \end{tabular}
    \caption{Performance on three unseen identities under 5 views. N.A. indicates no results successfully reconstructed.}
    \label{tab:unseen_sparse}
\end{table*}
\begin{figure*}[htbp]
    \centering
    \rotatebox{90}{\textbf{training view}}
    \includegraphics[width=0.18\textwidth]{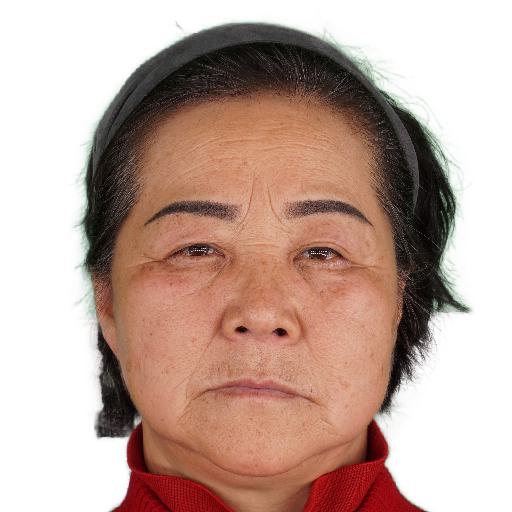}
    \includegraphics[width=0.18\textwidth]{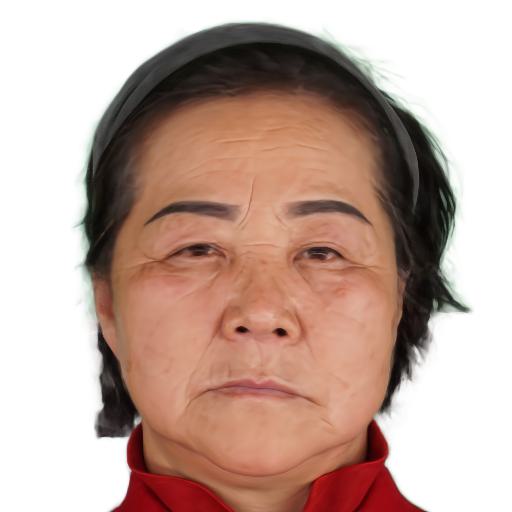}
    \includegraphics[width=0.18\textwidth]{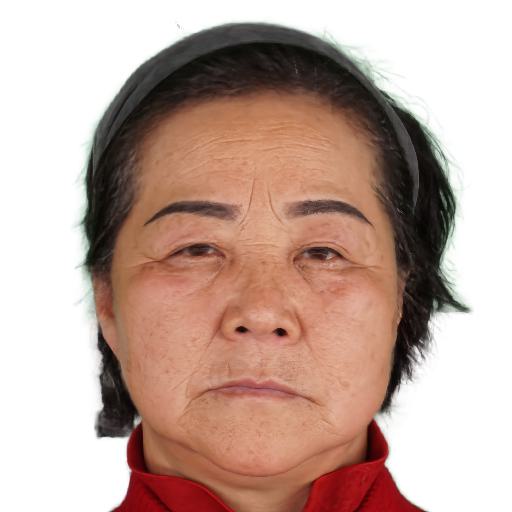}
    \includegraphics[width=0.18\textwidth]{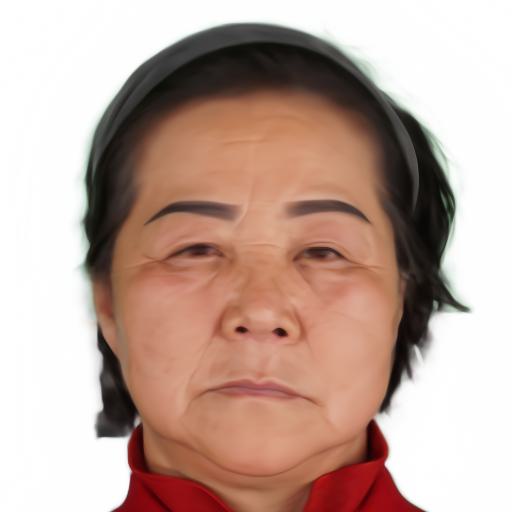}
    \includegraphics[width=0.18\textwidth]{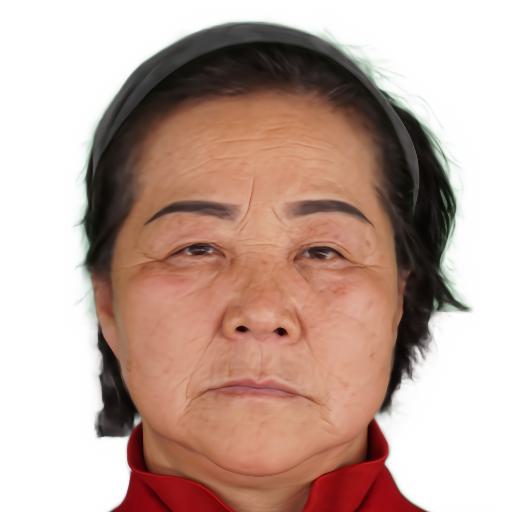}\\
    \rotatebox{90}{}
    \includegraphics[width=0.18\textwidth]{./results/template_effects/gt_571_blank.jpg}
    \includegraphics[width=0.18\textwidth]{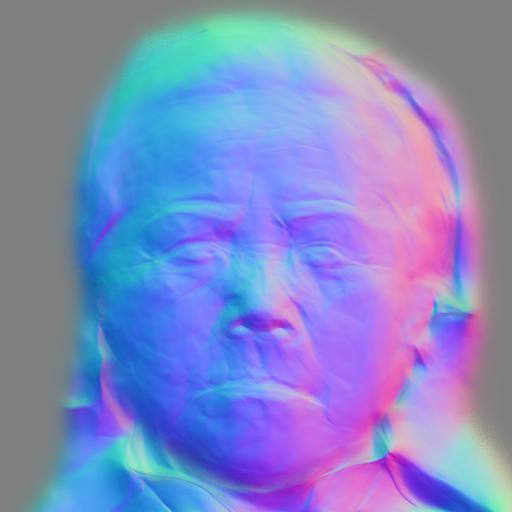}
    \includegraphics[width=0.18\textwidth]{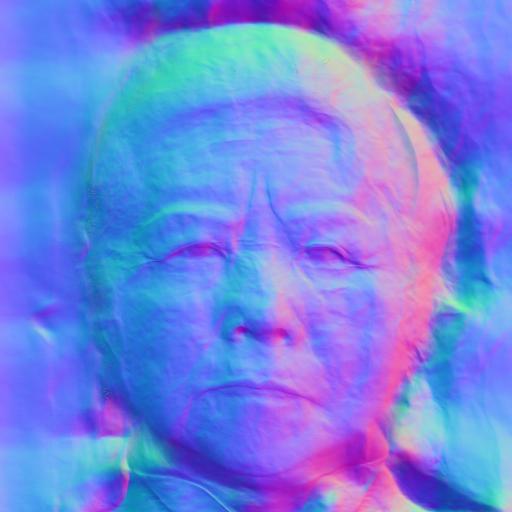}
    \includegraphics[width=0.18\textwidth]{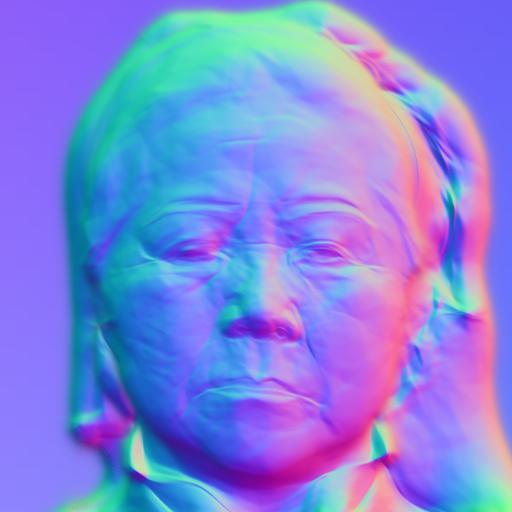}
    \includegraphics[width=0.18\textwidth]{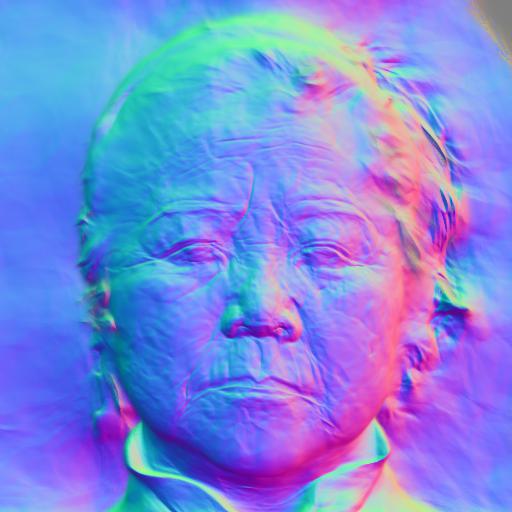}\\
    \rotatebox{90}{\textbf{novel view}}
    \includegraphics[width=0.18\textwidth]{./results/sp_view_5view/gt_383_12.jpg}
    \includegraphics[width=0.18\textwidth]{./results/sp_view_5view/neus_5_383_12_render.jpg}
    \includegraphics[width=0.18\textwidth]{./results/sp_view_5view/hfs_5_383_12_render.jpg}
    \includegraphics[width=0.18\textwidth]{./results/sp_view_5view/volsdf_5_383_12_render.jpg}
    \includegraphics[width=0.18\textwidth]{./results/sp_view_5view/ours_5_383_12_render.jpg}\\
    \rotatebox{90}{}
    \includegraphics[width=0.18\textwidth]{./results/template_effects/gt_571_blank.jpg}
    \includegraphics[width=0.18\textwidth]{./results/sp_view_5view/neus_5_383_12_normal.jpg}
    \includegraphics[width=0.18\textwidth]{./results/sp_view_5view/hfs_5_383_12_normal.jpg}
    \includegraphics[width=0.18\textwidth]{./results/sp_view_5view/volsdf_5_383_12_normal.jpg}
    \includegraphics[width=0.18\textwidth]{./results/sp_view_5view/ours_5_383_12_normal.jpg}\\
    \makebox[0.18\textwidth]{\scriptsize GT}
    \makebox[0.18\textwidth]{\scriptsize NeuS}
    \makebox[0.18\textwidth]{\scriptsize HF-NeuS}
    \makebox[0.18\textwidth]{\scriptsize VolSDF}
    \makebox[0.18\textwidth]{\scriptsize Ours}
    \caption{Results for Model 383 with only 5 views as input. The template human head was trained using 5 randomly selected views for all 30 identities of the PR-Senior and PR-Young datasets. The images of Model 383 for Stage 1 training and Stage 2 training are the same, therefore no additional views were provided. }
    \label{fig:sparse_view}
\end{figure*}

\subsubsection{Sparse Views}

This section provides further results on sparse views as shown in Figure~\ref{fig:sparse_view}. Moreover, Table~\ref{tab:unseen_sparse} also demonstrates that our approach surpasses other methods in terms of reconstructing geometry under sparse view conditions.

\subsection{Analysis}
\label{subsec:analysis_sec}

We provide a thorough evaluation of our method and the state-of-the-art methods NeuS, VolSDF, and HF-NeuS on the PR-Senior and PR-Young datasets in this supplementary material. Our analysis includes a discussion of the strengths and weaknesses of each method and a comparison of their performance under various settings. It is worth noting that when VolSDF or HF-NeuS fails to reconstruct the geometry for certain models, we exclude them from the calculation of Chamfer distances for their methods. However, we use all models when calculating the Chamfer distances for our method and NeuS, both of which can reconstruct geometry for all 30 identities.

\subsubsection{Comparison to VolSDF}

\begin{figure*}[htbp]
    \centering
    \rotatebox{90}{\textbf{371}}
    \includegraphics[width=0.09\textwidth]{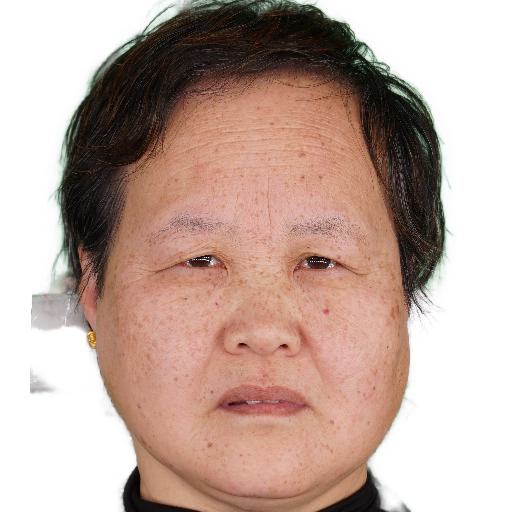}
    \includegraphics[width=0.09\textwidth]{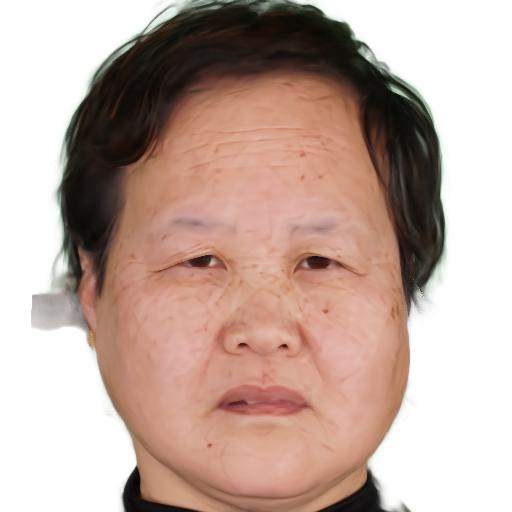}
    \includegraphics[width=0.09\textwidth]{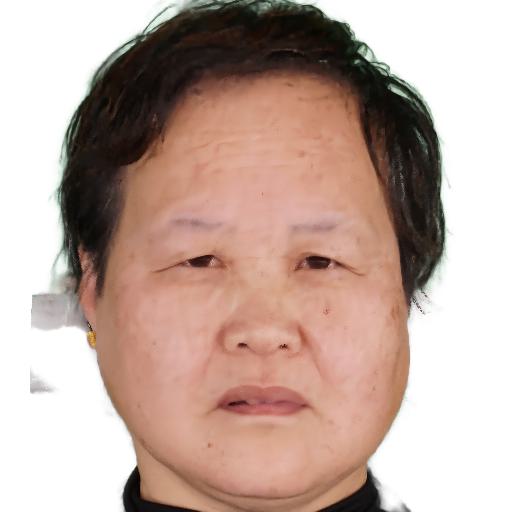}
    \includegraphics[width=0.09\textwidth]{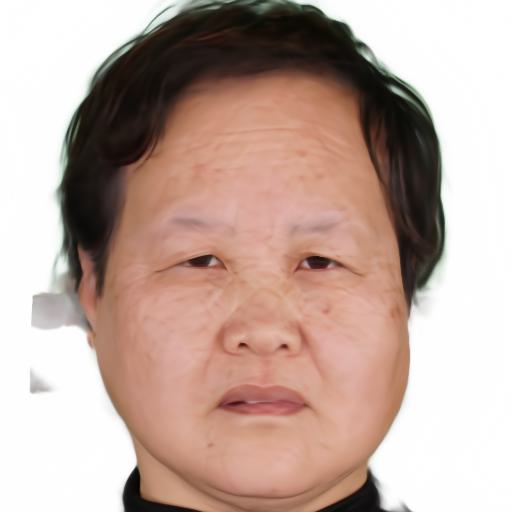}
    \includegraphics[width=0.09\textwidth]{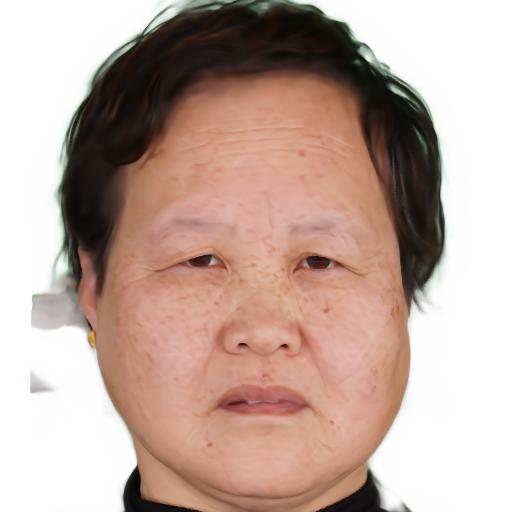}
    \includegraphics[width=0.09\textwidth]{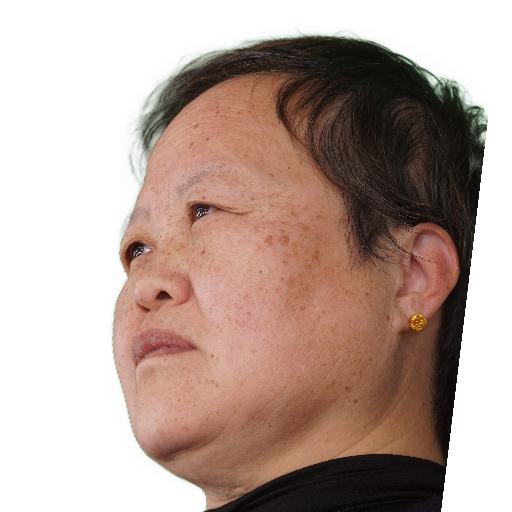}
    \includegraphics[width=0.09\textwidth]{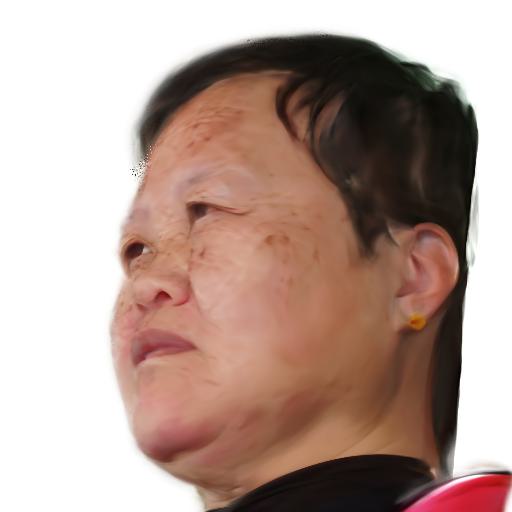}
    \includegraphics[width=0.09\textwidth]{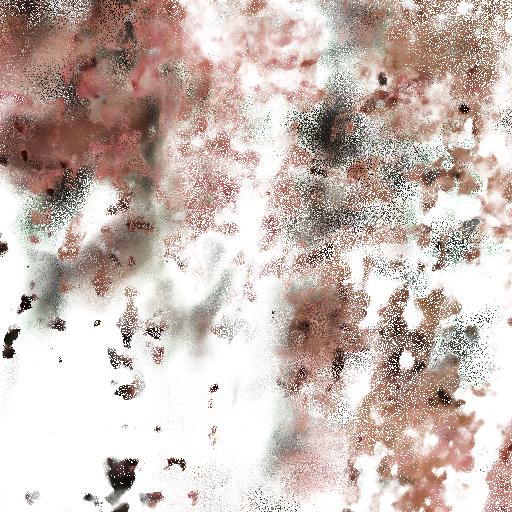}
    \includegraphics[width=0.09\textwidth]{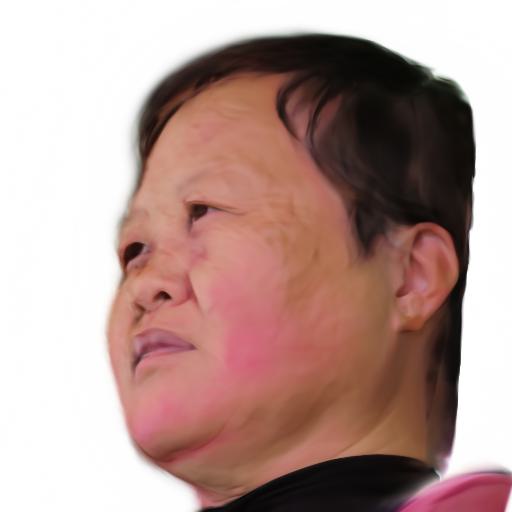}
    \includegraphics[width=0.09\textwidth]{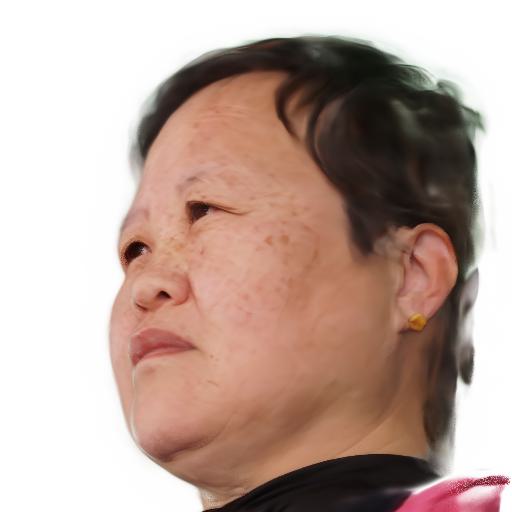}
    \rotatebox{90}{\tiny}
    \includegraphics[width=0.09\textwidth]{./results/template_effects/gt_571_blank.jpg}
    \includegraphics[width=0.09\textwidth]{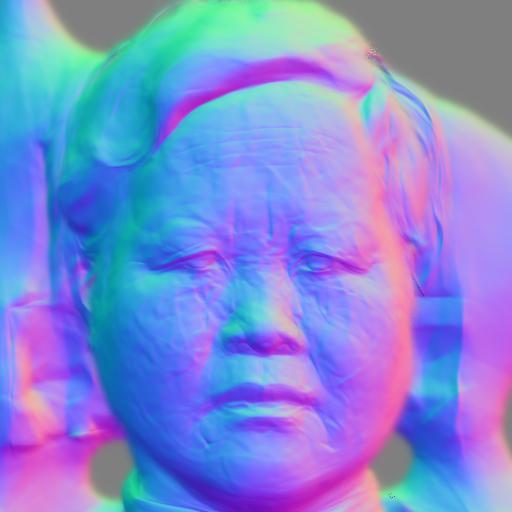}
    \includegraphics[width=0.09\textwidth]{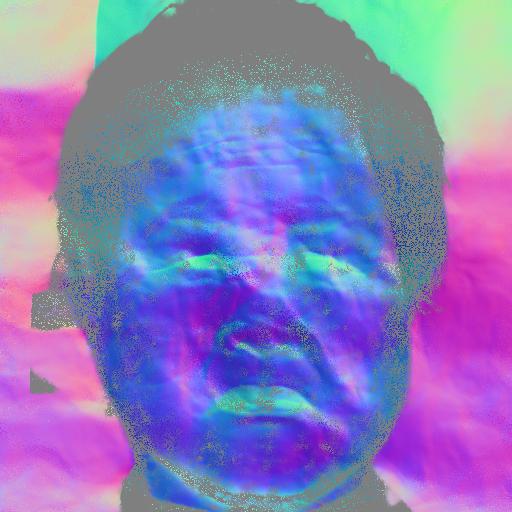}
    \includegraphics[width=0.09\textwidth]{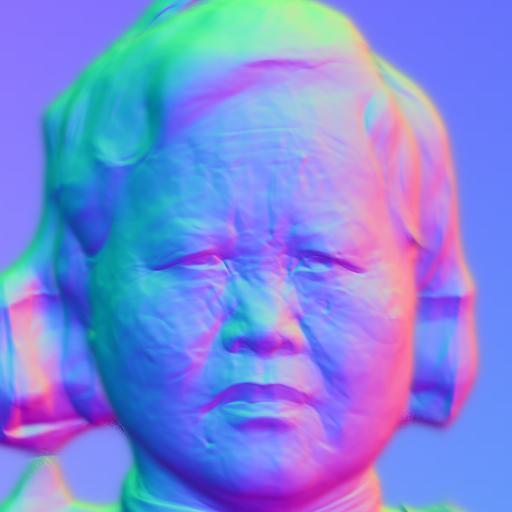}
    \includegraphics[width=0.09\textwidth]{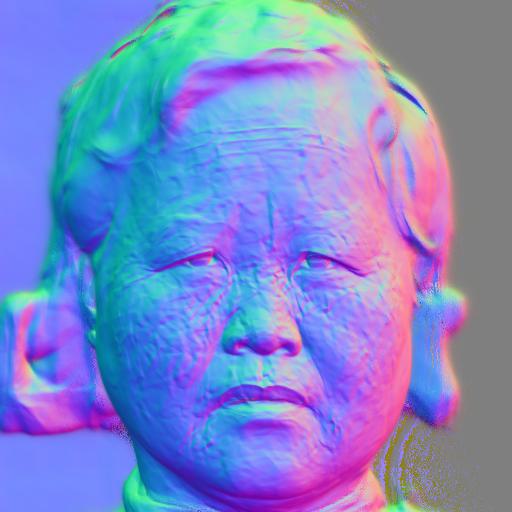}
    \includegraphics[width=0.09\textwidth]{./results/template_effects/gt_571_blank.jpg}
    \includegraphics[width=0.09\textwidth]{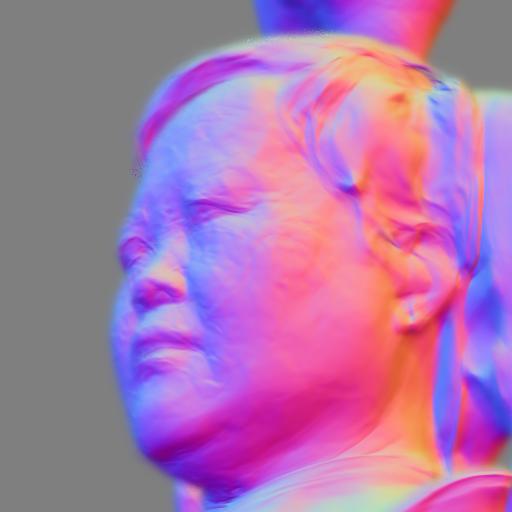}
    \includegraphics[width=0.09\textwidth]{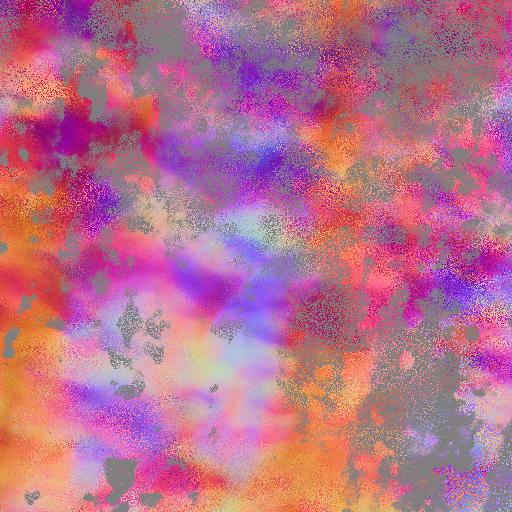}
    \includegraphics[width=0.09\textwidth]{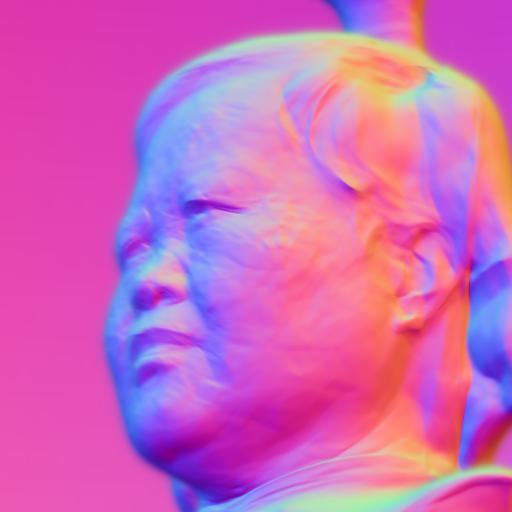}
    \includegraphics[width=0.09\textwidth]{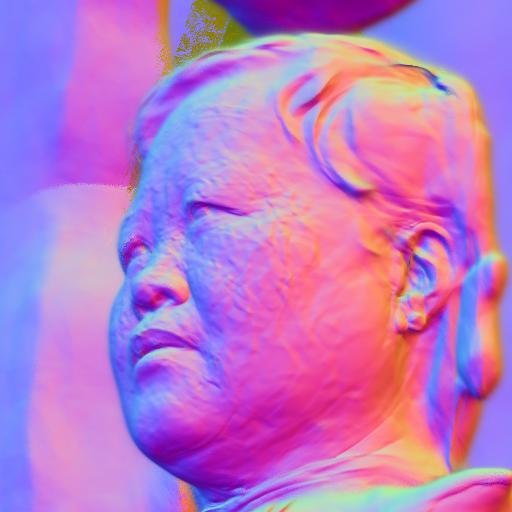}\\
    % \rotatebox{90}{\textbf{novel view}}\\
    % \rotatebox{90}{\tiny}\\
    \rotatebox{90}{\textbf{375}}
    \includegraphics[width=0.09\textwidth]{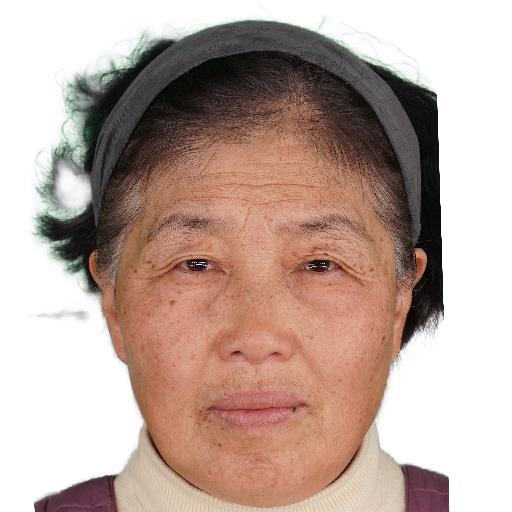}
    \includegraphics[width=0.09\textwidth]{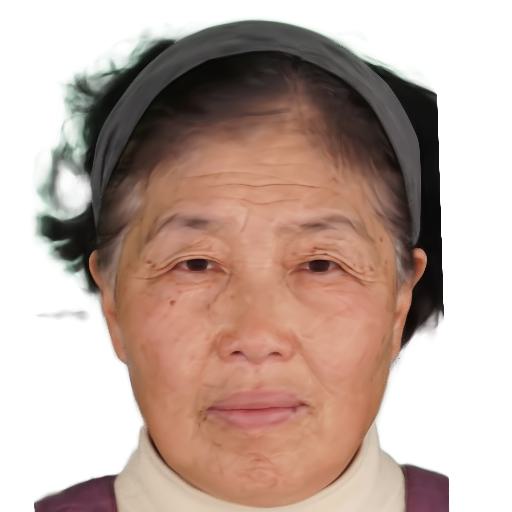}
    \includegraphics[width=0.09\textwidth]{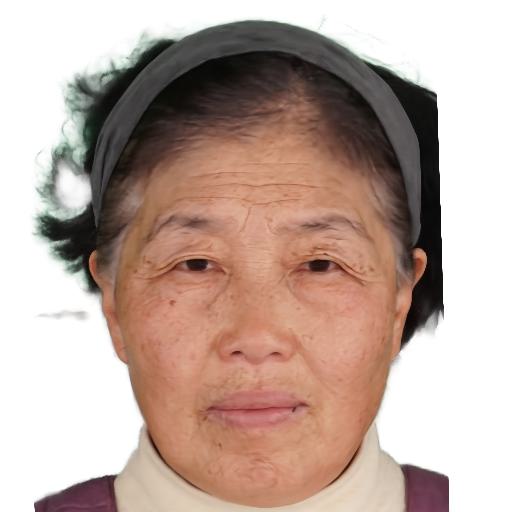}
    \includegraphics[width=0.09\textwidth]{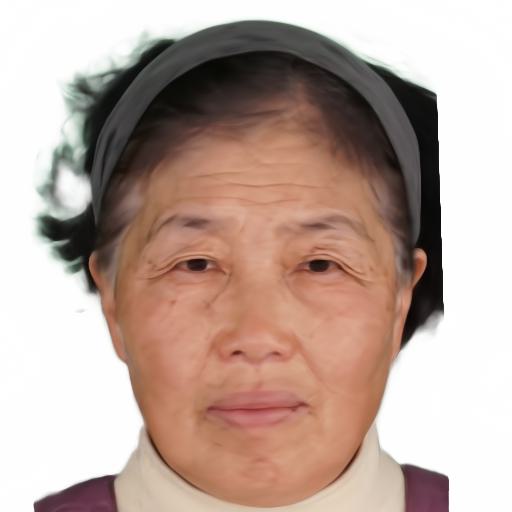}
    \includegraphics[width=0.09\textwidth]{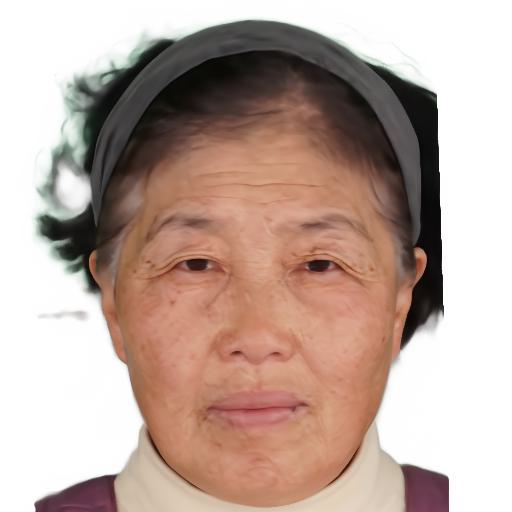}
    \includegraphics[width=0.09\textwidth]{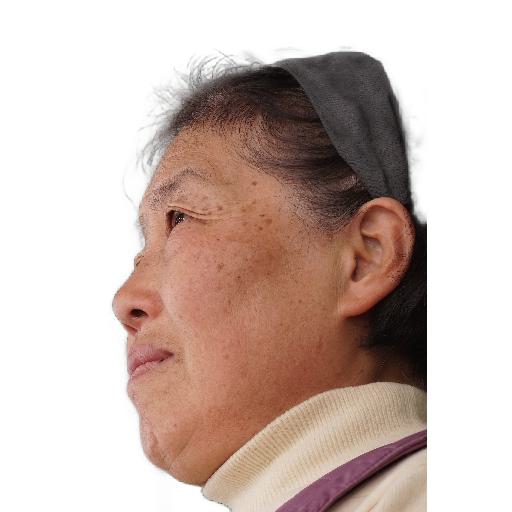}
    \includegraphics[width=0.09\textwidth]{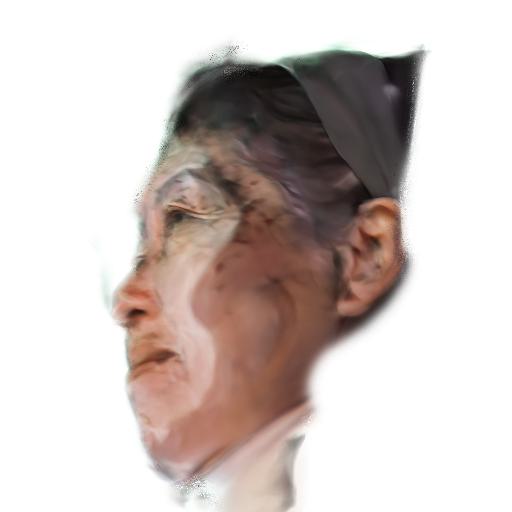}
    \includegraphics[width=0.09\textwidth]{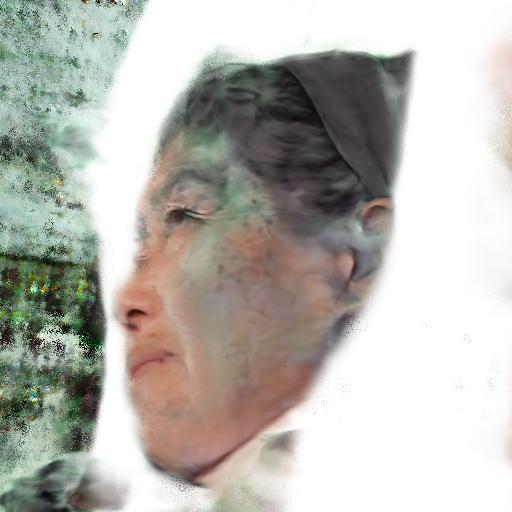}
    \includegraphics[width=0.09\textwidth]{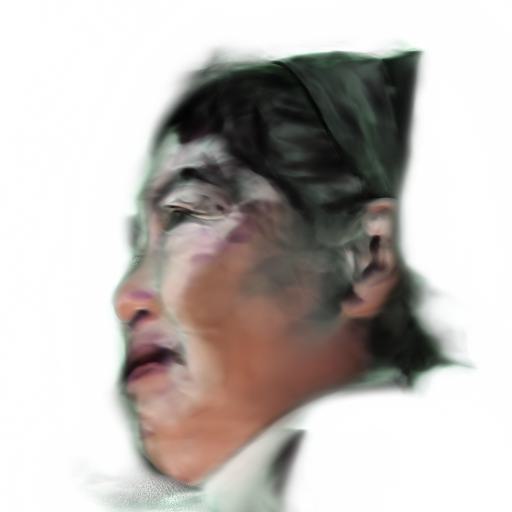}
    \includegraphics[width=0.09\textwidth]{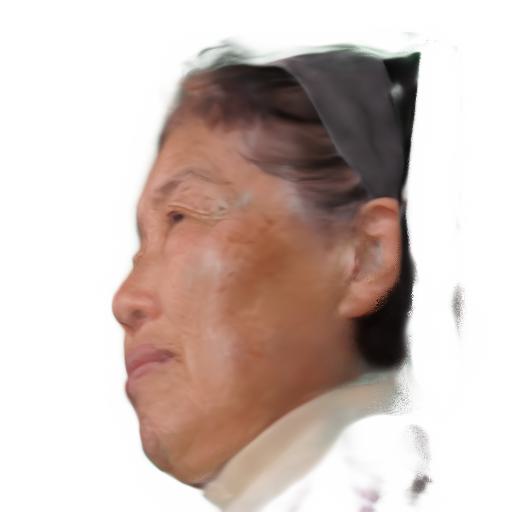}\\
    % \rotatebox{90}{\tiny}
    \includegraphics[width=0.09\textwidth]{./results/template_effects/gt_571_blank.jpg}
    \includegraphics[width=0.09\textwidth]{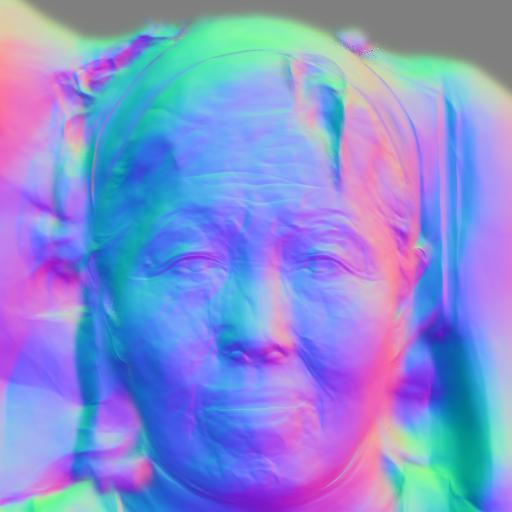}
    \includegraphics[width=0.09\textwidth]{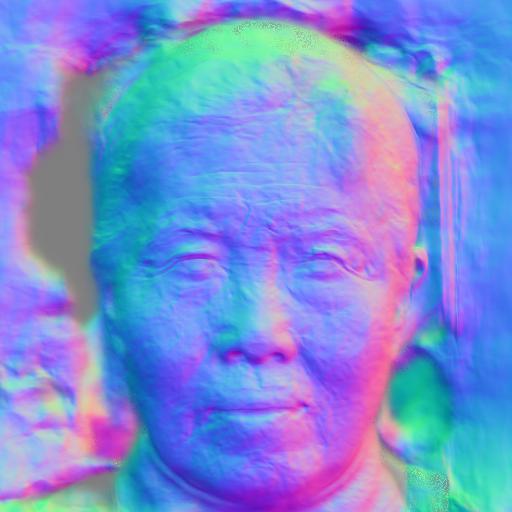}
    \includegraphics[width=0.09\textwidth]{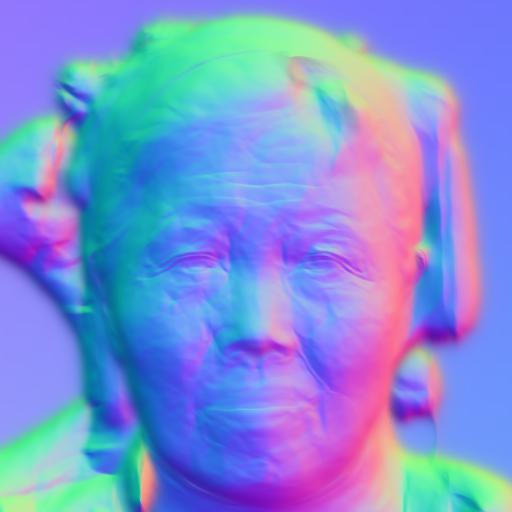}
    \includegraphics[width=0.09\textwidth]{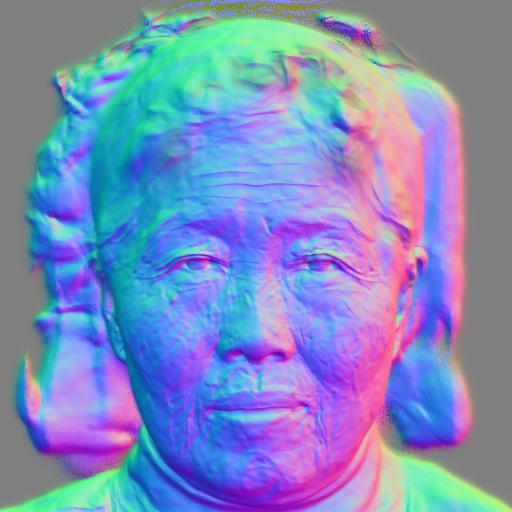}
    \includegraphics[width=0.09\textwidth]{./results/template_effects/gt_571_blank.jpg}
    \includegraphics[width=0.09\textwidth]{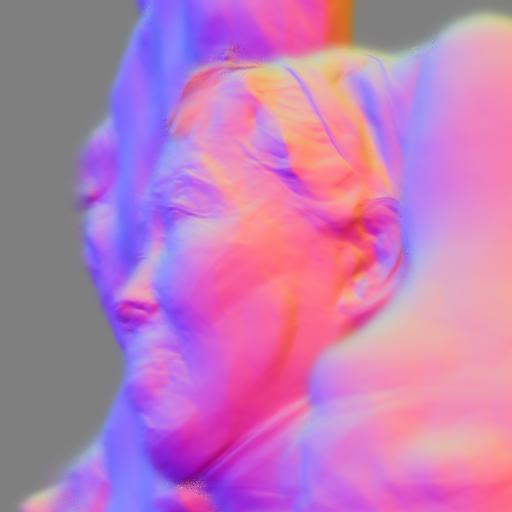}
    \includegraphics[width=0.09\textwidth]{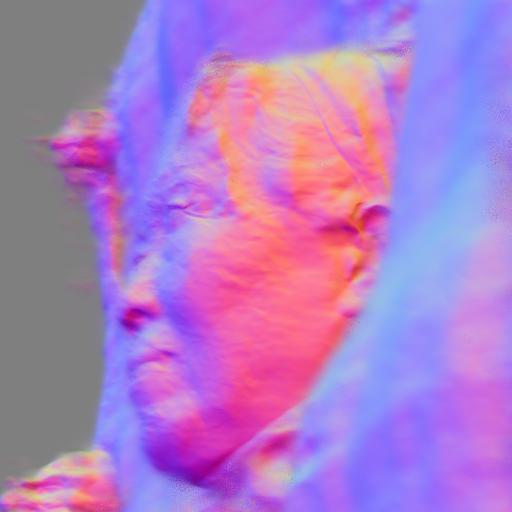}
    \includegraphics[width=0.09\textwidth]{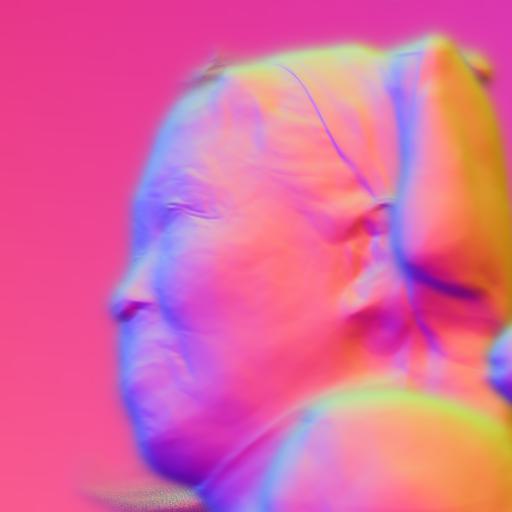}
    \includegraphics[width=0.09\textwidth]{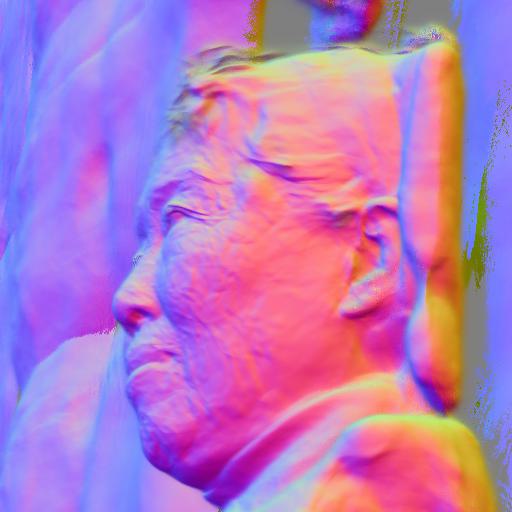}\\
    \rotatebox{90}{\textbf{376}}
    \includegraphics[width=0.09\textwidth]{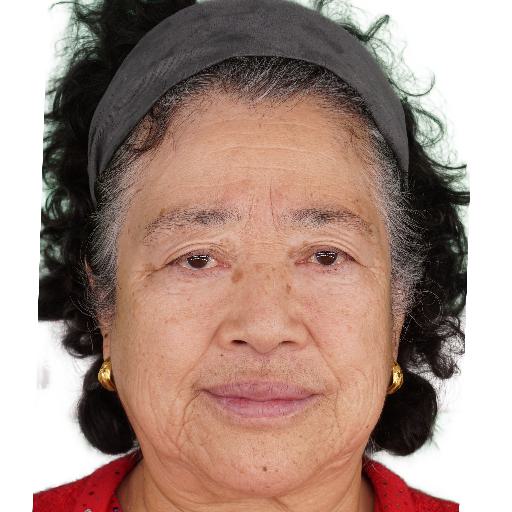}
    \includegraphics[width=0.09\textwidth]{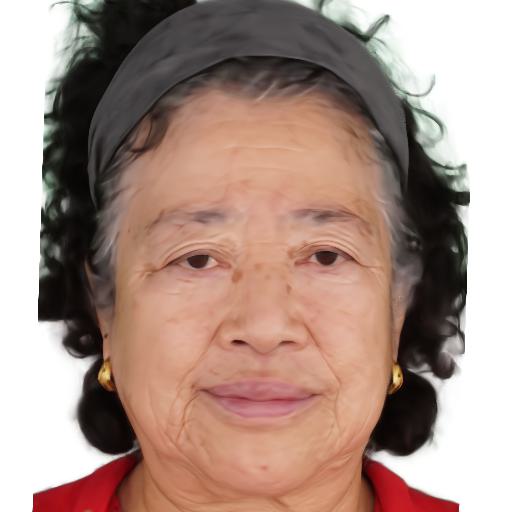}
    \includegraphics[width=0.09\textwidth]{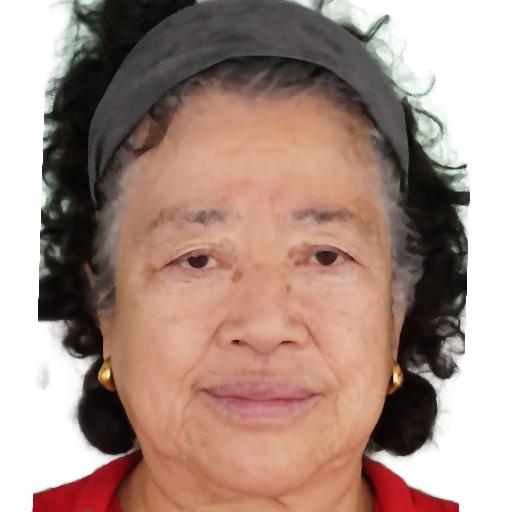}
    \includegraphics[width=0.09\textwidth]{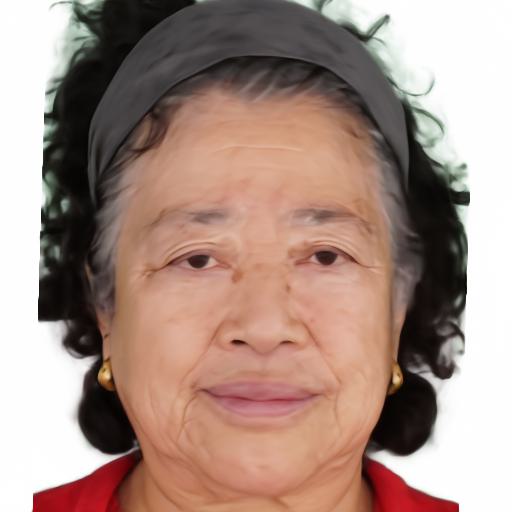}
    \includegraphics[width=0.09\textwidth]{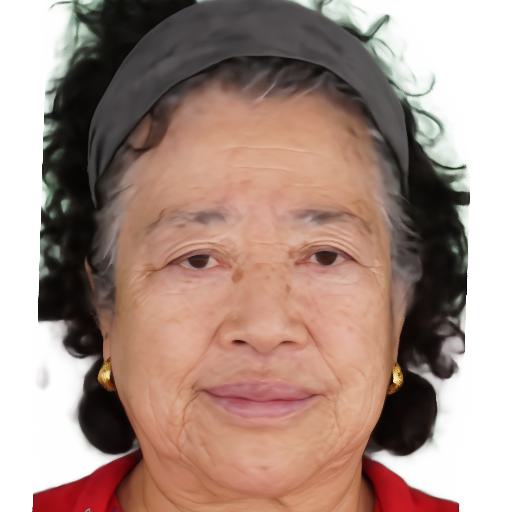}
    \includegraphics[width=0.09\textwidth]{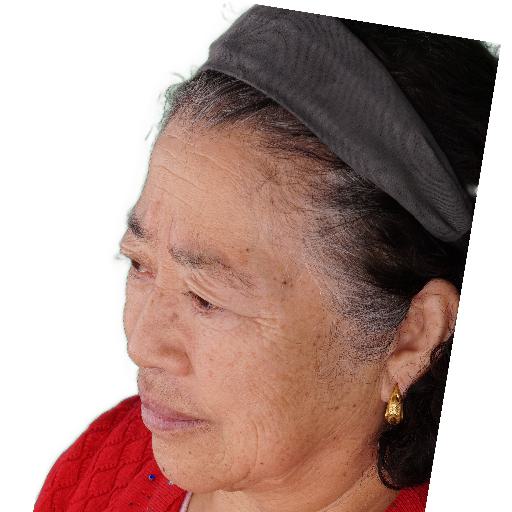}
    \includegraphics[width=0.09\textwidth]{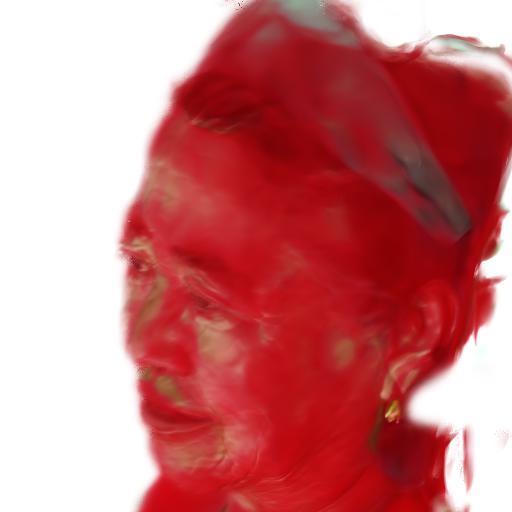}
    \includegraphics[width=0.09\textwidth]{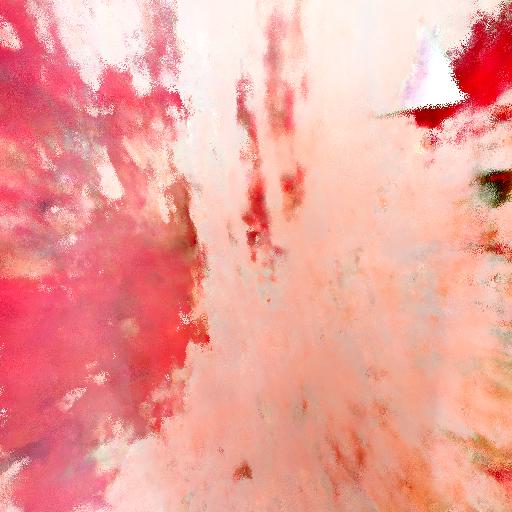}
    \includegraphics[width=0.09\textwidth]{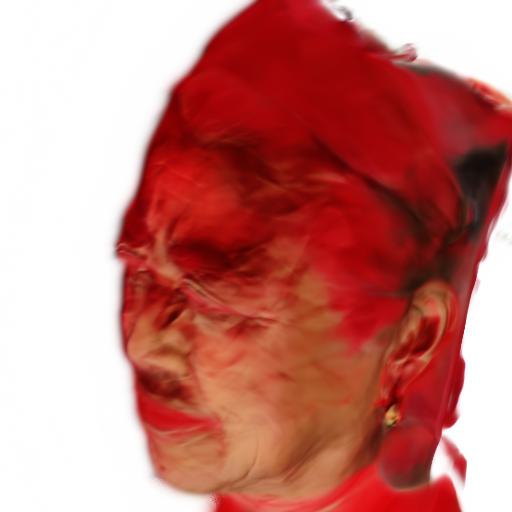}
    \includegraphics[width=0.09\textwidth]{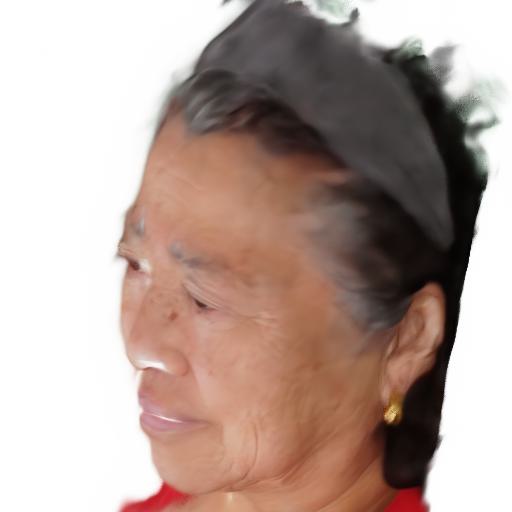}\\
    % \rotatebox{90}{\tiny}
    \includegraphics[width=0.09\textwidth]{./results/template_effects/gt_571_blank.jpg}
    \includegraphics[width=0.09\textwidth]{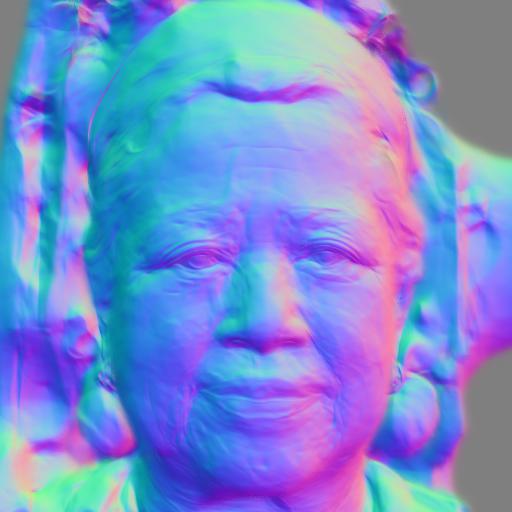}
    \includegraphics[width=0.09\textwidth]{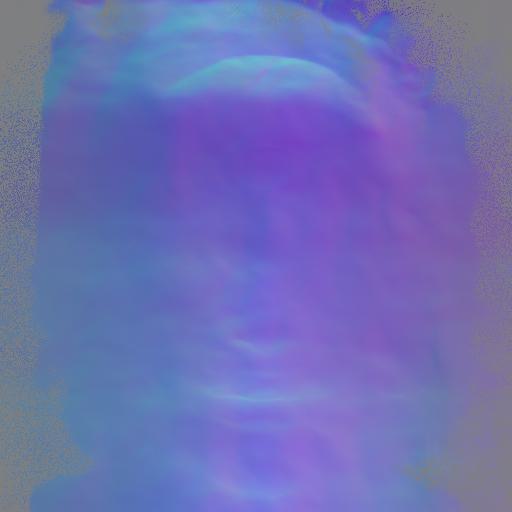}
    \includegraphics[width=0.09\textwidth]{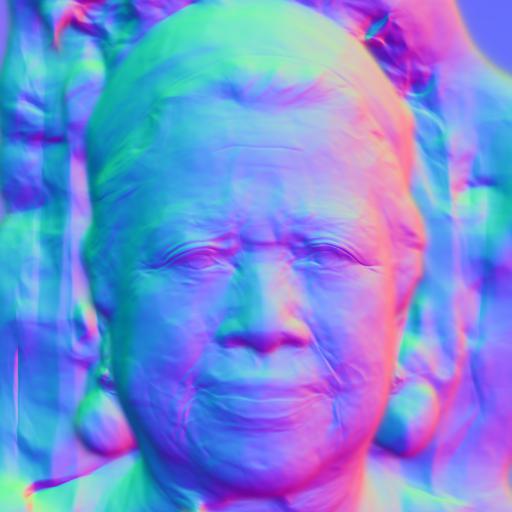}
    \includegraphics[width=0.09\textwidth]{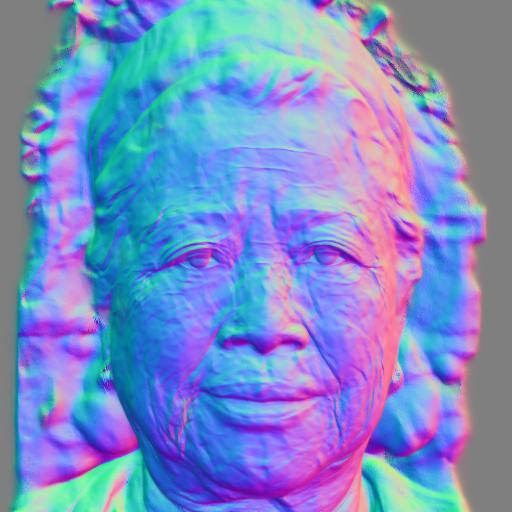}
    \includegraphics[width=0.09\textwidth]{./results/template_effects/gt_571_blank.jpg}
    \includegraphics[width=0.09\textwidth]{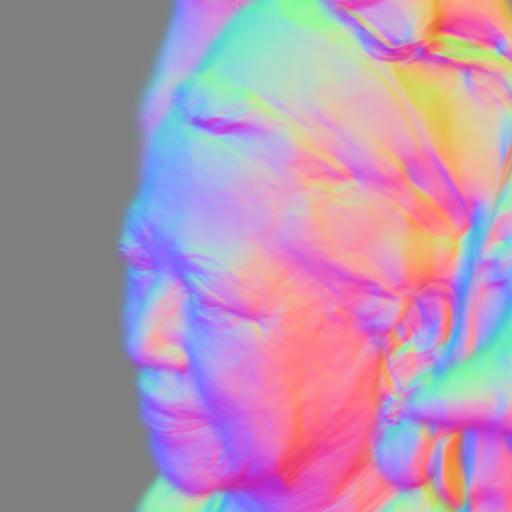}
    \includegraphics[width=0.09\textwidth]{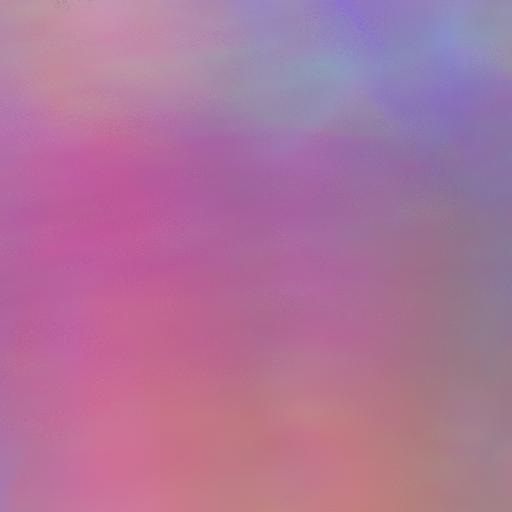}
    \includegraphics[width=0.09\textwidth]{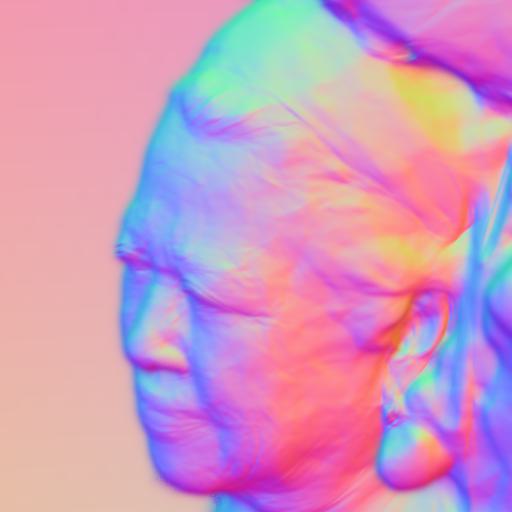}
    \includegraphics[width=0.09\textwidth]{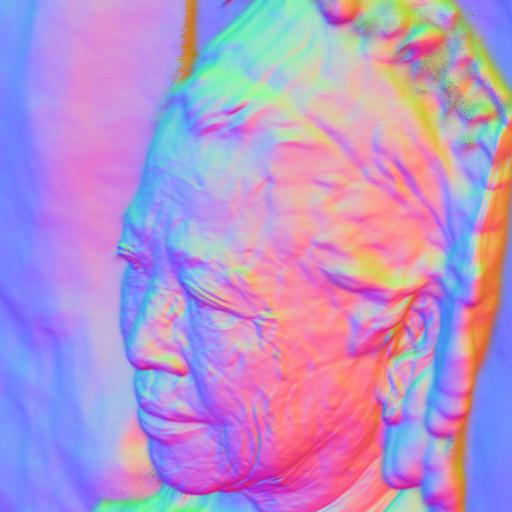}\\
    \rotatebox{90}{\textbf{377}}
    \includegraphics[width=0.09\textwidth]{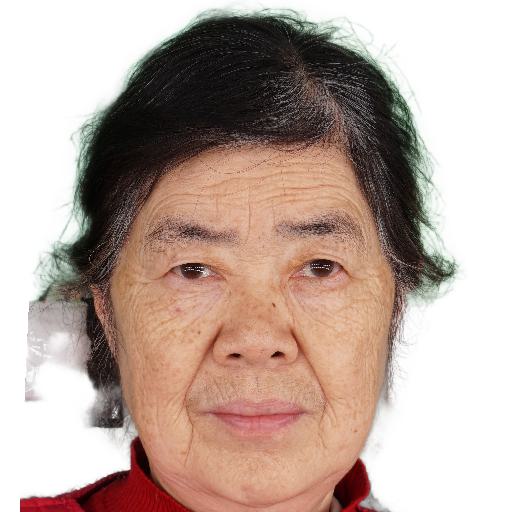}
    \includegraphics[width=0.09\textwidth]{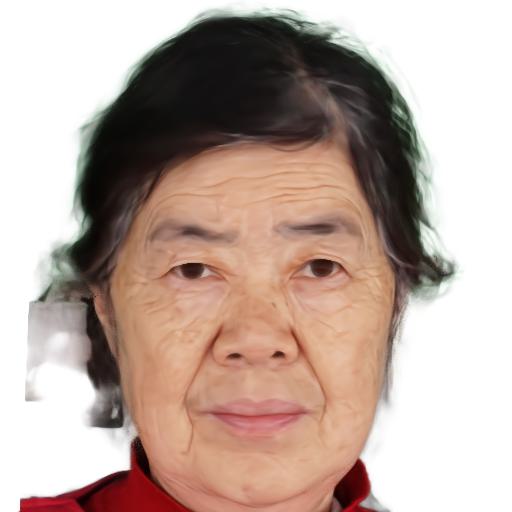}
    \includegraphics[width=0.09\textwidth]{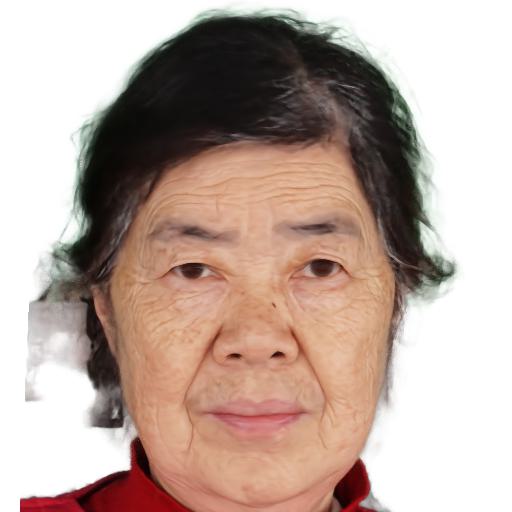}
    \includegraphics[width=0.09\textwidth]{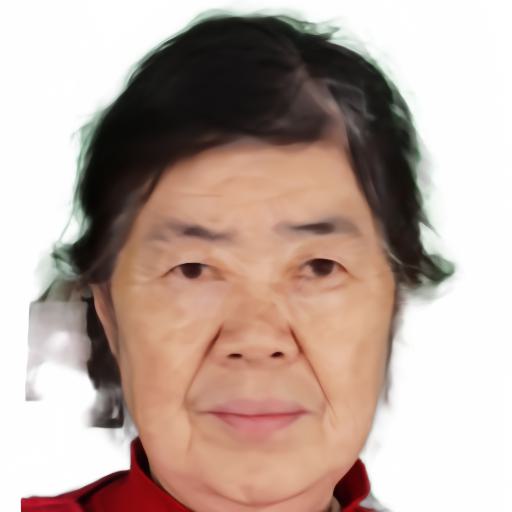}
    \includegraphics[width=0.09\textwidth]{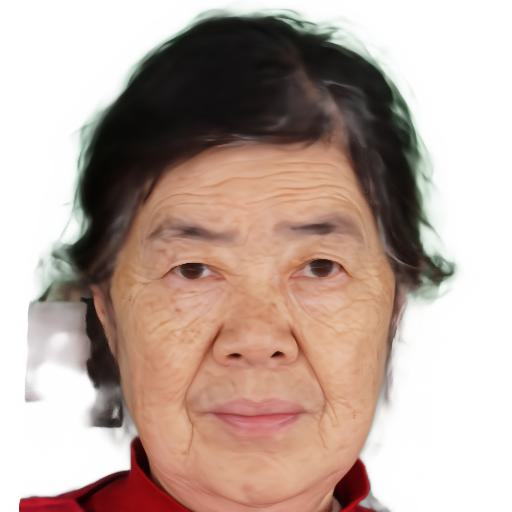}
    \includegraphics[width=0.09\textwidth]{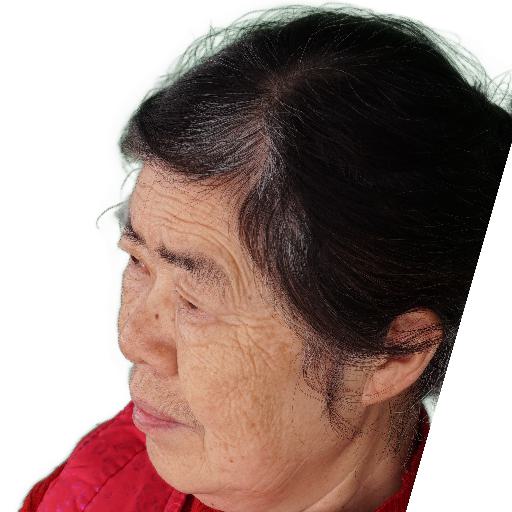}
    \includegraphics[width=0.09\textwidth]{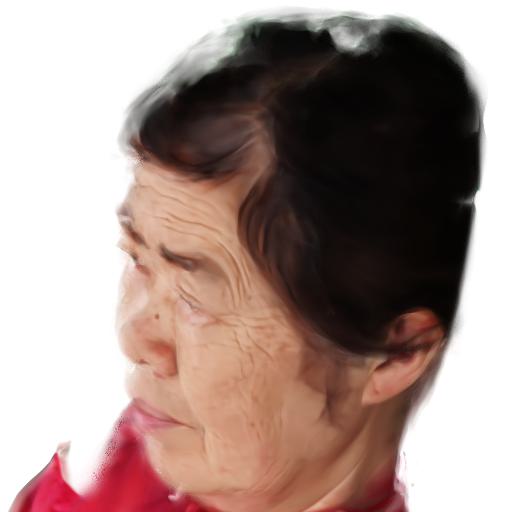}
    \includegraphics[width=0.09\textwidth]{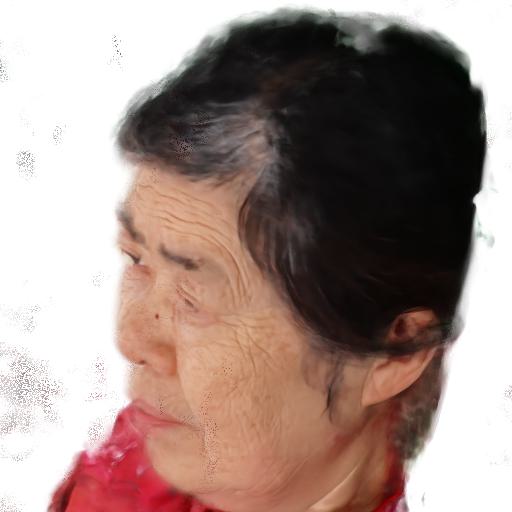}
    \includegraphics[width=0.09\textwidth]{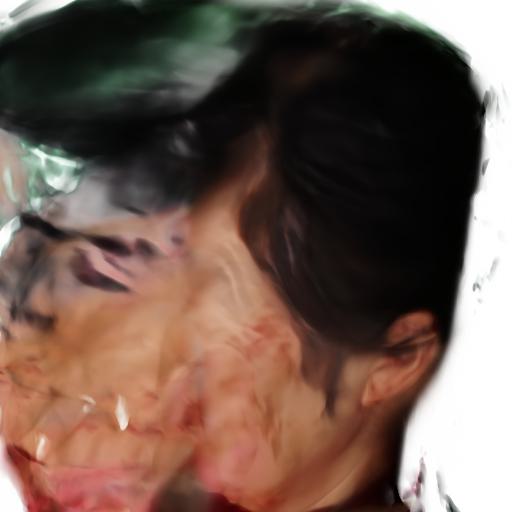}
    \includegraphics[width=0.09\textwidth]{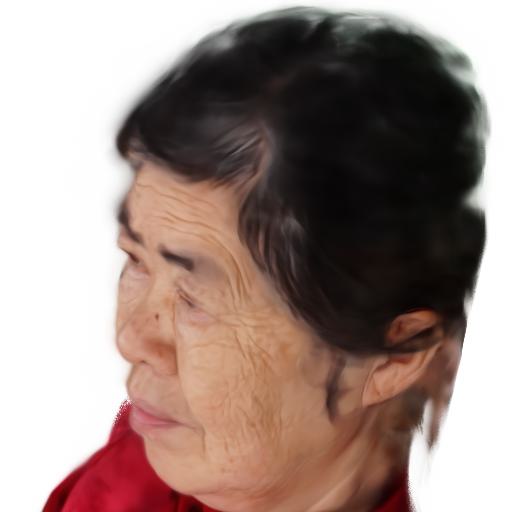}\\
    % \rotatebox{90}{\tiny}
    \includegraphics[width=0.09\textwidth]{./results/template_effects/gt_571_blank.jpg}
    \includegraphics[width=0.09\textwidth]{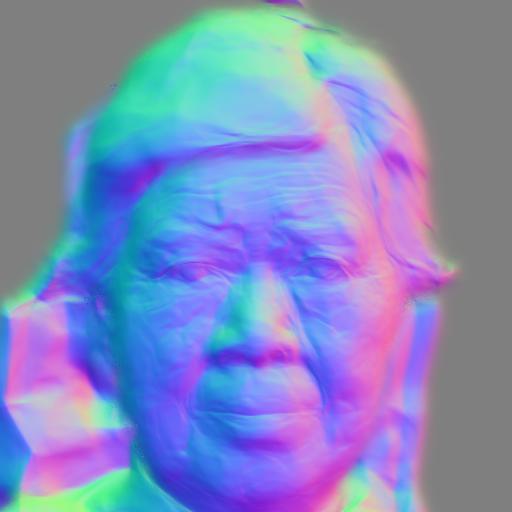}
    \includegraphics[width=0.09\textwidth]{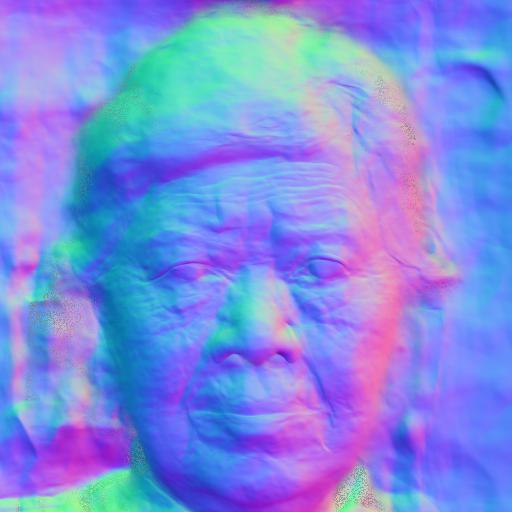}
    \includegraphics[width=0.09\textwidth]{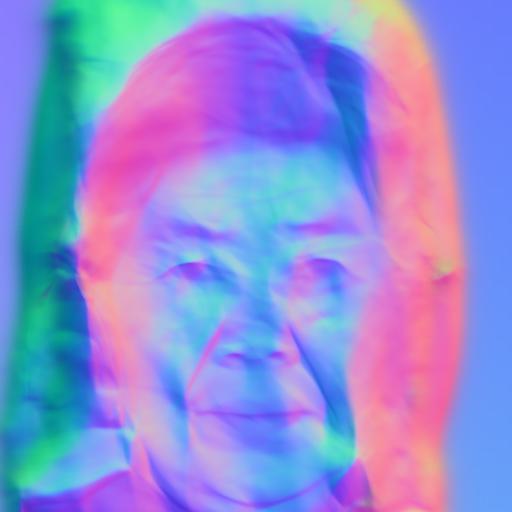}
    \includegraphics[width=0.09\textwidth]{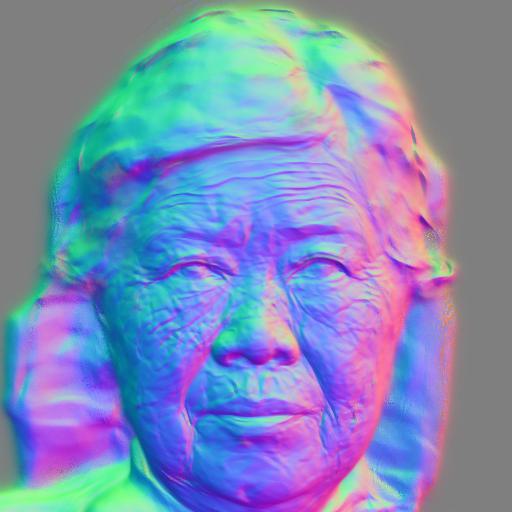}
    \includegraphics[width=0.09\textwidth]{./results/template_effects/gt_571_blank.jpg}
    \includegraphics[width=0.09\textwidth]{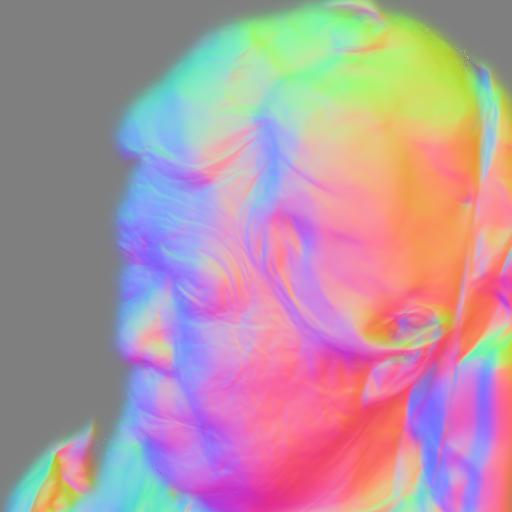}
    \includegraphics[width=0.09\textwidth]{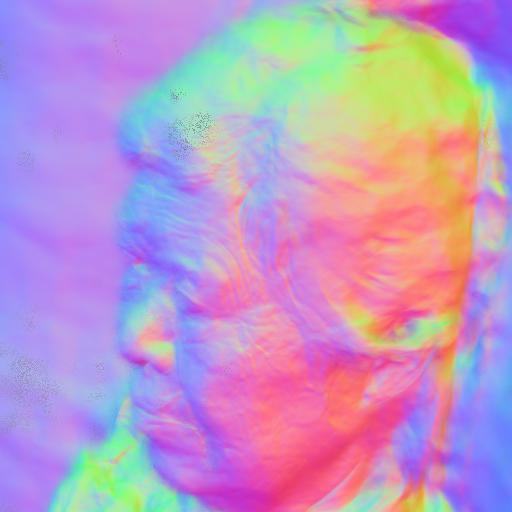}
    \includegraphics[width=0.09\textwidth]{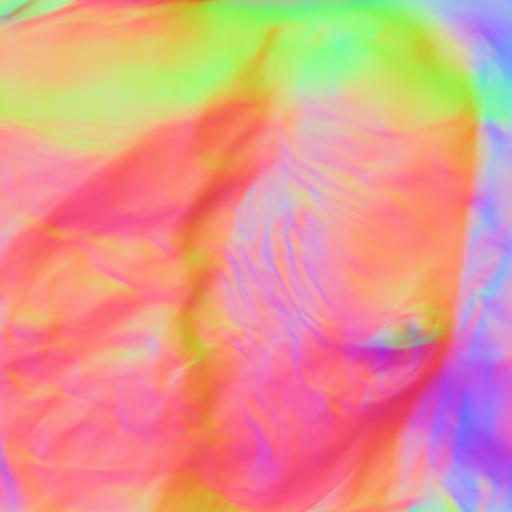}
    \includegraphics[width=0.09\textwidth]{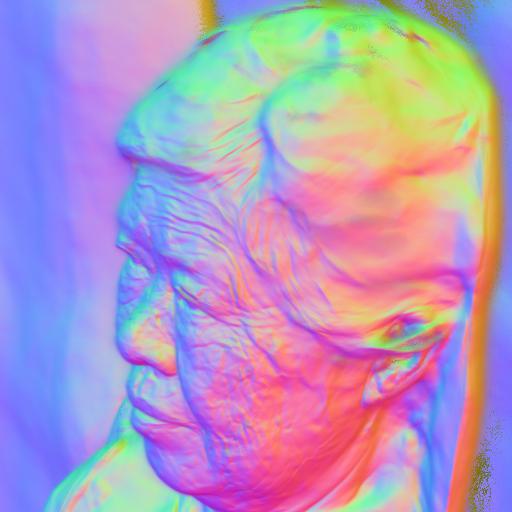}\\
    \rotatebox{90}{\textbf{383}}
    \includegraphics[width=0.09\textwidth]{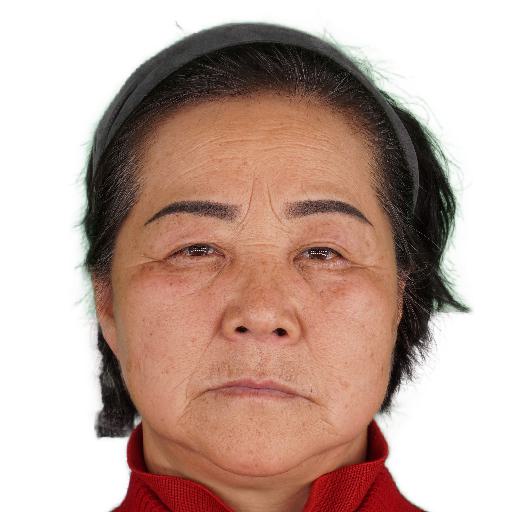}
    \includegraphics[width=0.09\textwidth]{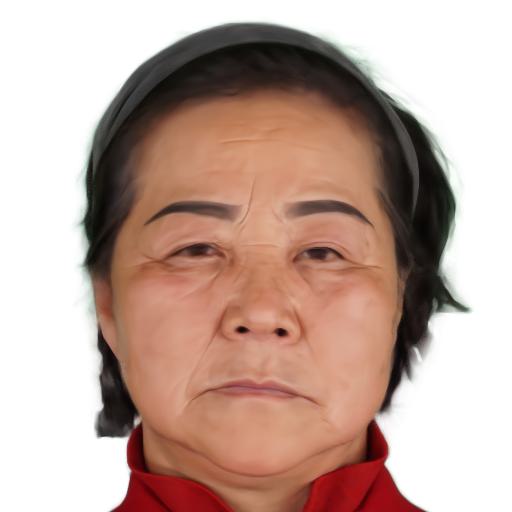}
    \includegraphics[width=0.09\textwidth]{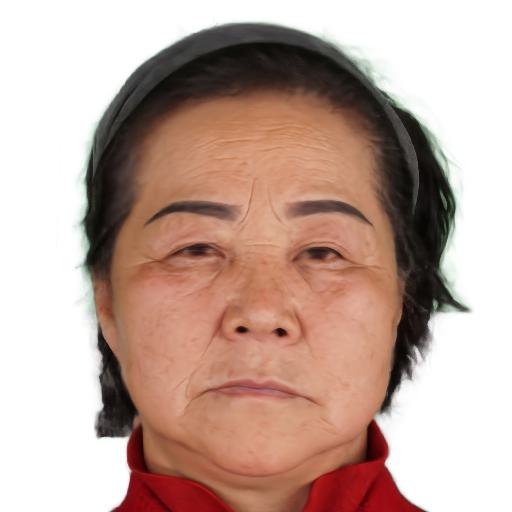}
    \includegraphics[width=0.09\textwidth]{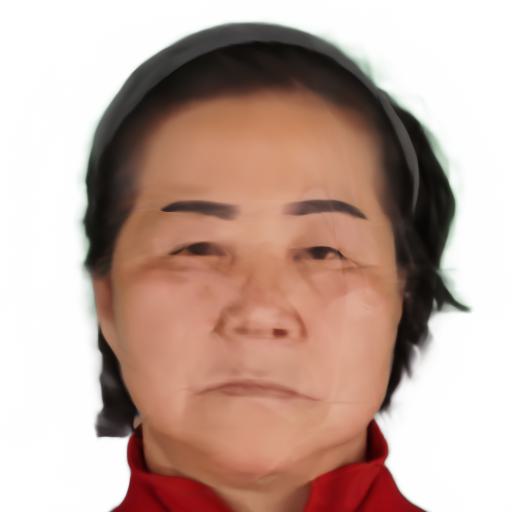}
    \includegraphics[width=0.09\textwidth]{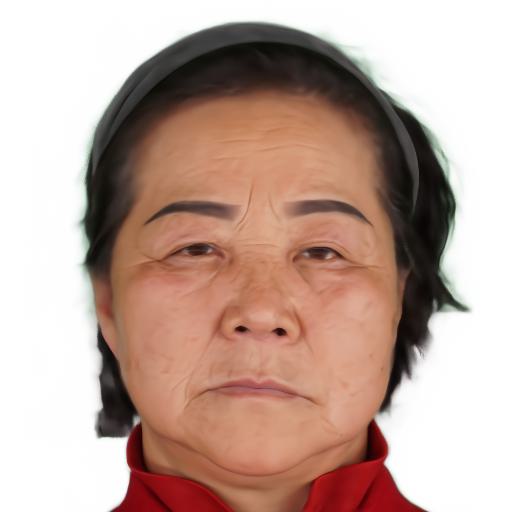}
    \includegraphics[width=0.09\textwidth]{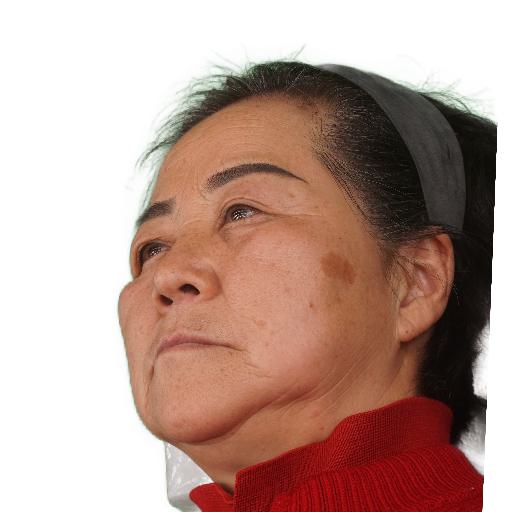}
    \includegraphics[width=0.09\textwidth]{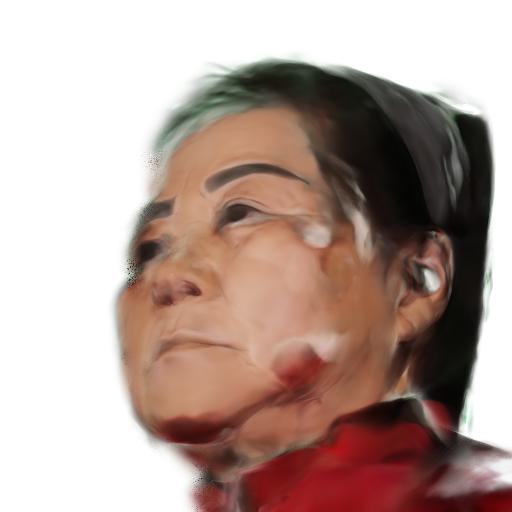}
    \includegraphics[width=0.09\textwidth]{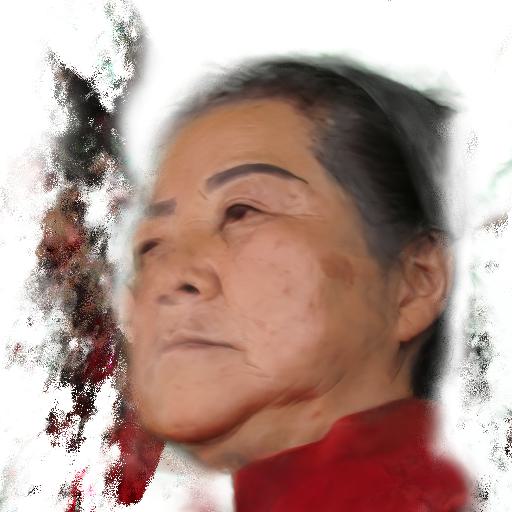}
    \includegraphics[width=0.09\textwidth]{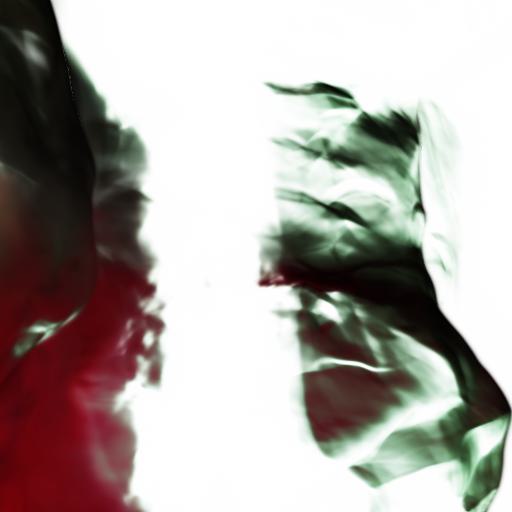}
    \includegraphics[width=0.09\textwidth]{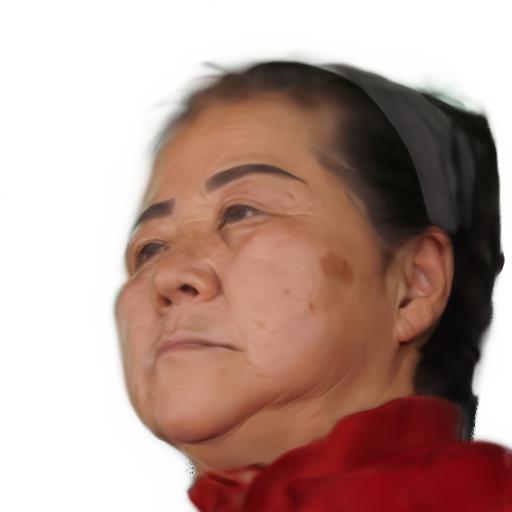}\\
    % \rotatebox{90}{\tiny}
    \includegraphics[width=0.09\textwidth]{./results/template_effects/gt_571_blank.jpg}
    \includegraphics[width=0.09\textwidth]{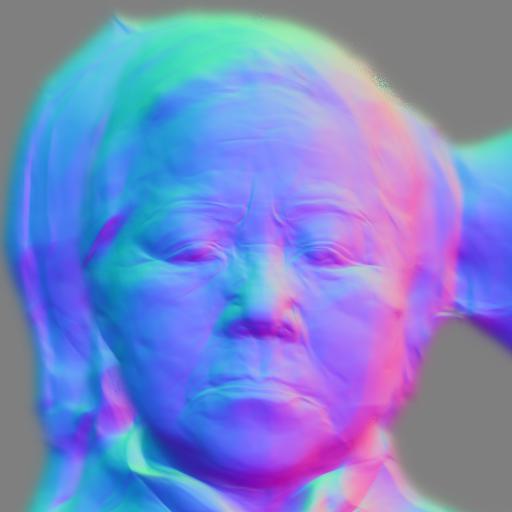}
    \includegraphics[width=0.09\textwidth]{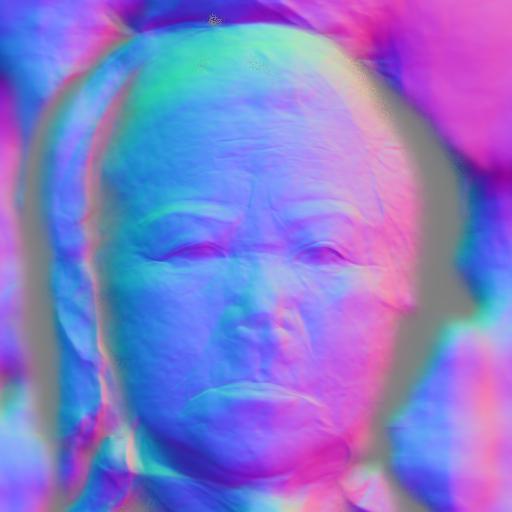}
    \includegraphics[width=0.09\textwidth]{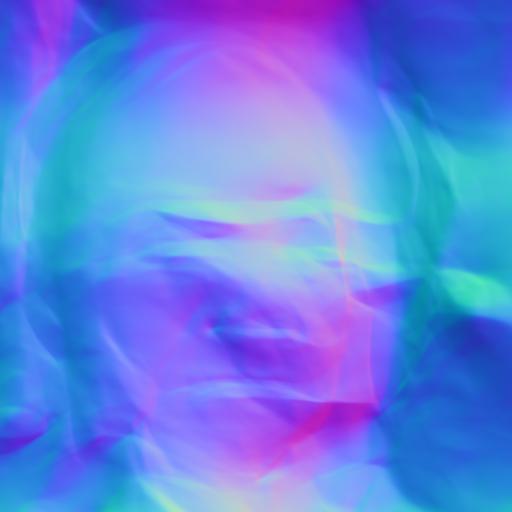}
    \includegraphics[width=0.09\textwidth]{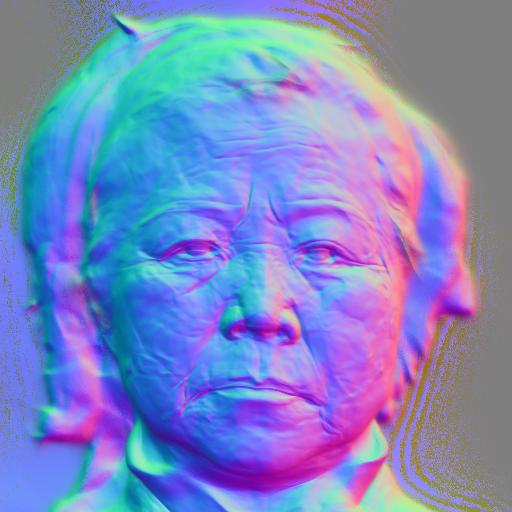}
    \includegraphics[width=0.09\textwidth]{./results/template_effects/gt_571_blank.jpg}
    \includegraphics[width=0.09\textwidth]{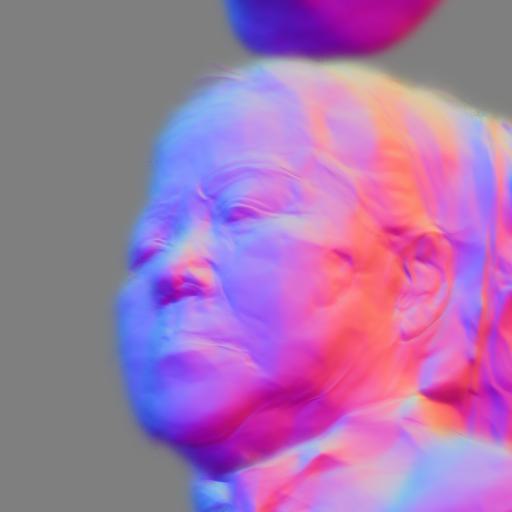}
    \includegraphics[width=0.09\textwidth]{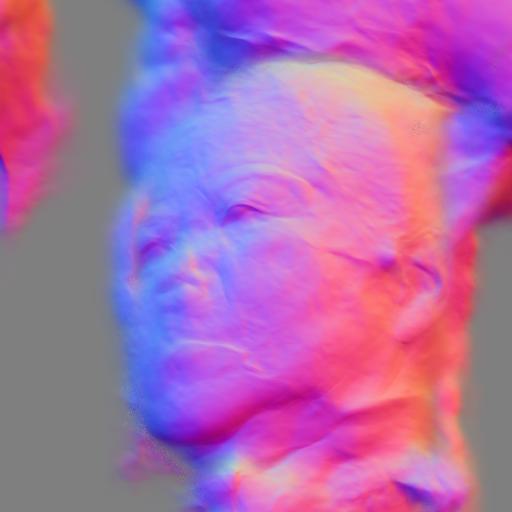}
    \includegraphics[width=0.09\textwidth]{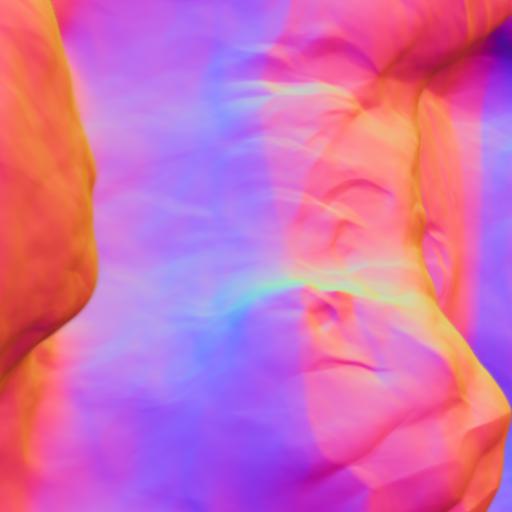}
    \includegraphics[width=0.09\textwidth]{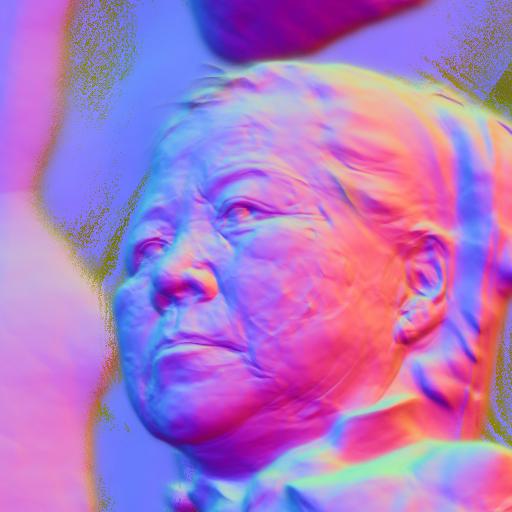}\\
    \rotatebox{90}{\textbf{389}}
    \includegraphics[width=0.09\textwidth]{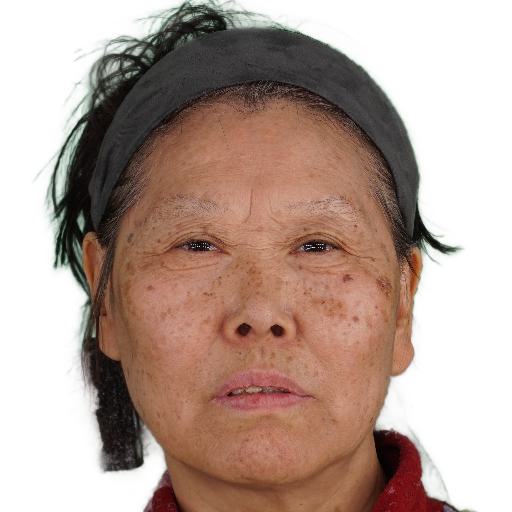}
    \includegraphics[width=0.09\textwidth]{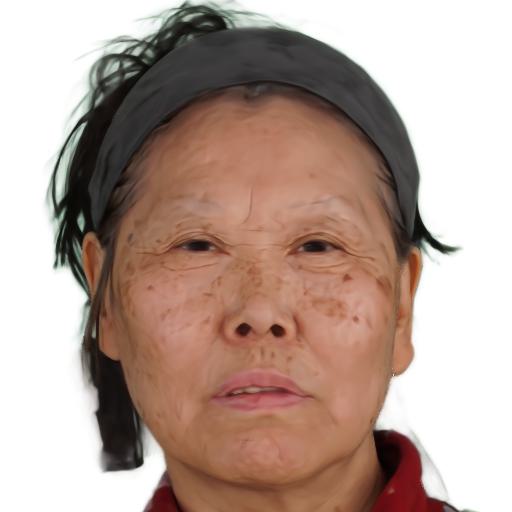}
    \includegraphics[width=0.09\textwidth]{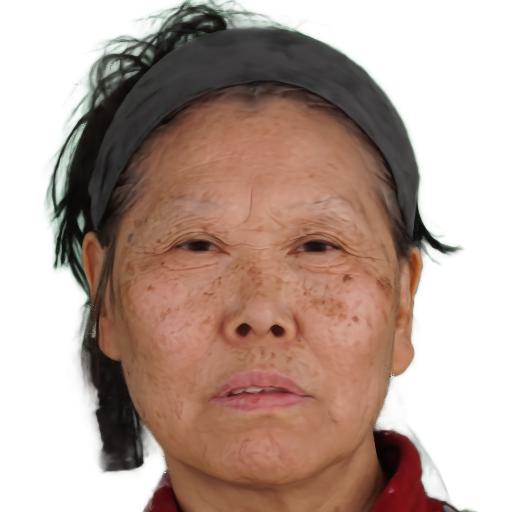}
    \includegraphics[width=0.09\textwidth]{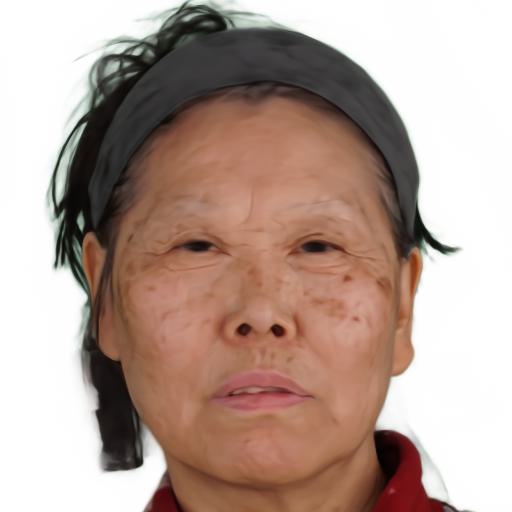}
    \includegraphics[width=0.09\textwidth]{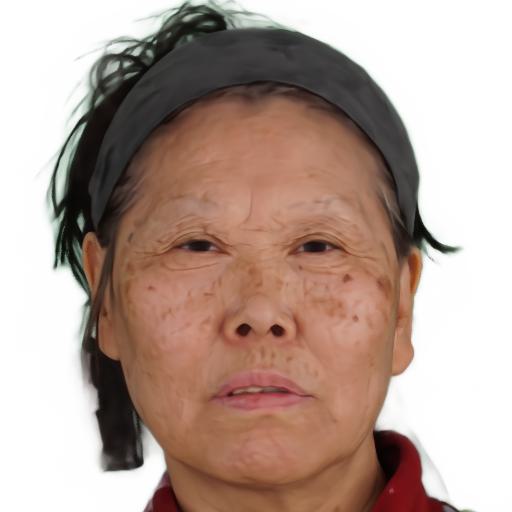}
    \includegraphics[width=0.09\textwidth]{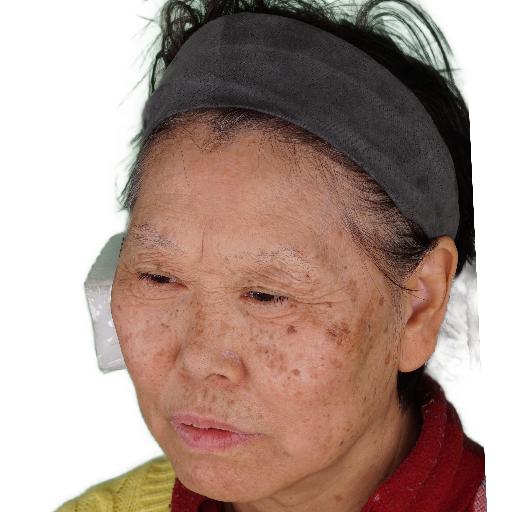}
    \includegraphics[width=0.09\textwidth]{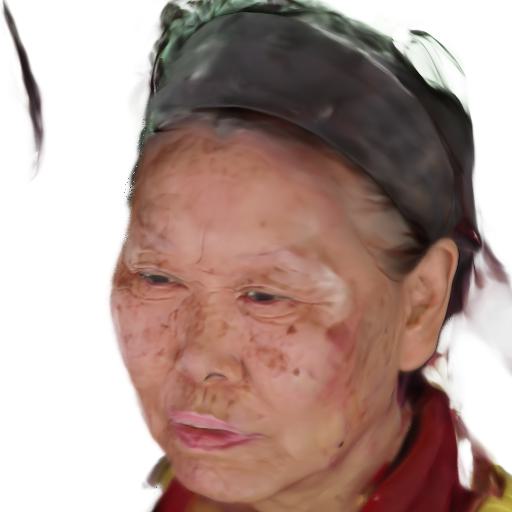}
    \includegraphics[width=0.09\textwidth]{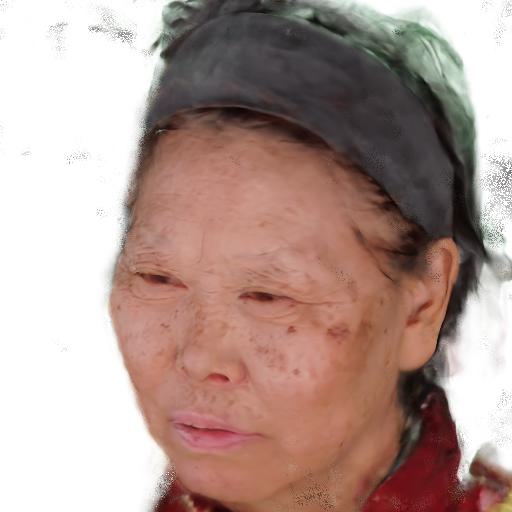}
    \includegraphics[width=0.09\textwidth]{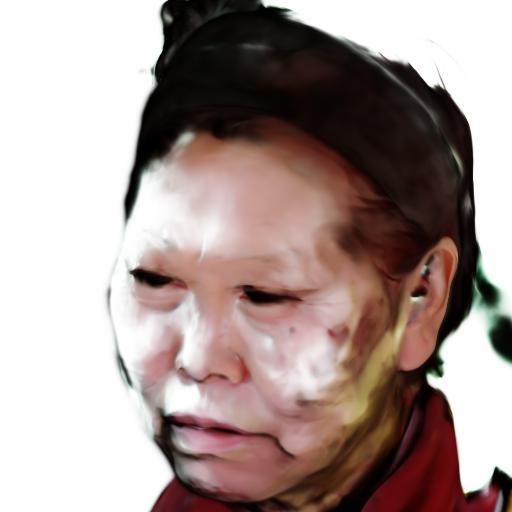}
    \includegraphics[width=0.09\textwidth]{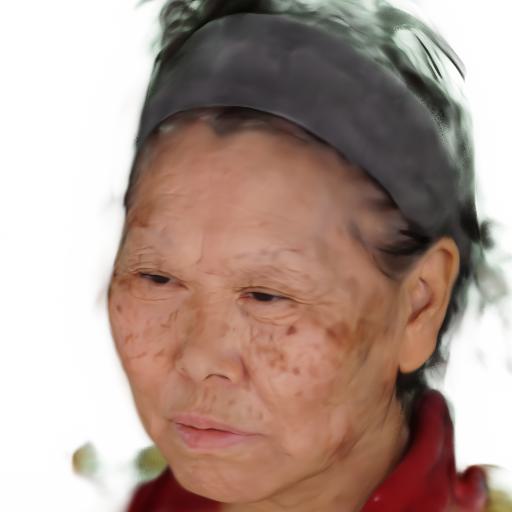}\\
    % \rotatebox{90}{\tiny}
    \includegraphics[width=0.09\textwidth]{./results/template_effects/gt_571_blank.jpg}
    \includegraphics[width=0.09\textwidth]{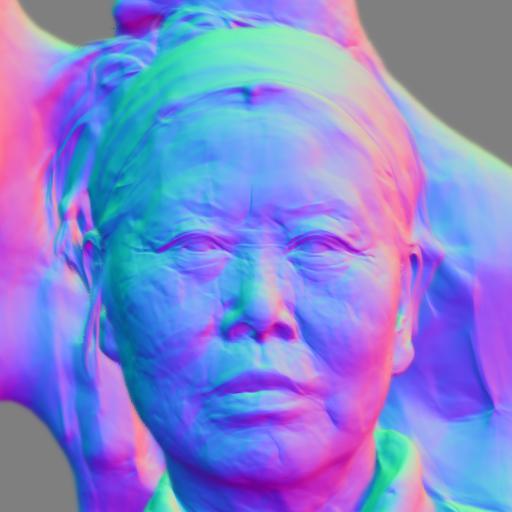}
    \includegraphics[width=0.09\textwidth]{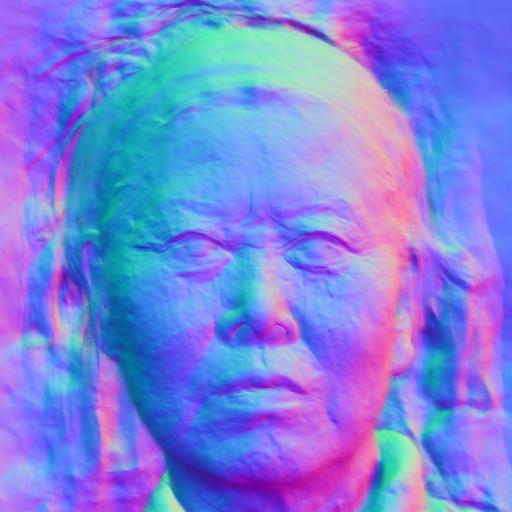}
    \includegraphics[width=0.09\textwidth]{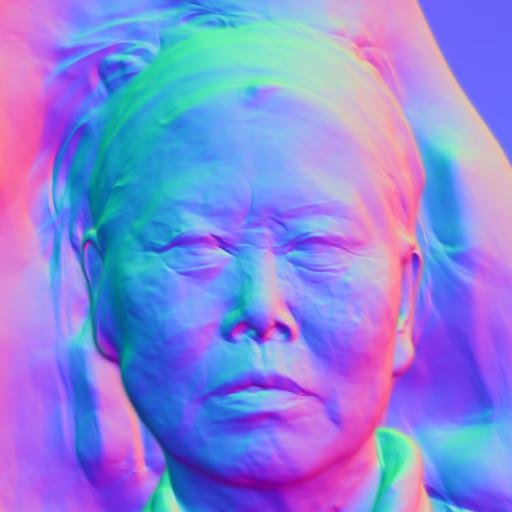}
    \includegraphics[width=0.09\textwidth]{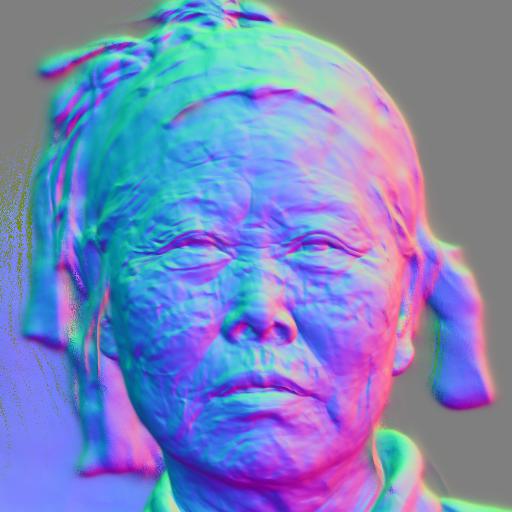}
    \includegraphics[width=0.09\textwidth]{./results/template_effects/gt_571_blank.jpg}
    \includegraphics[width=0.09\textwidth]{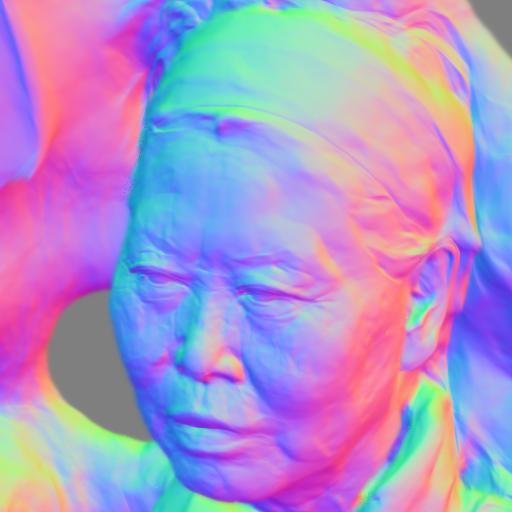}
    \includegraphics[width=0.09\textwidth]{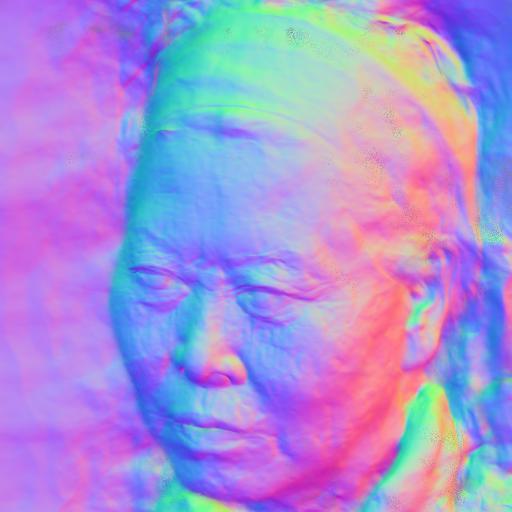}
    \includegraphics[width=0.09\textwidth]{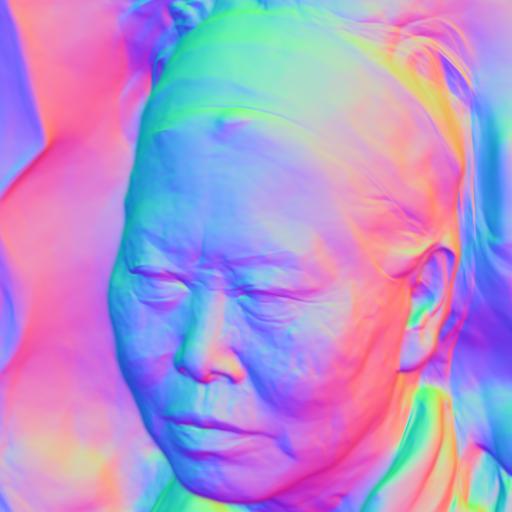}
    \includegraphics[width=0.09\textwidth]{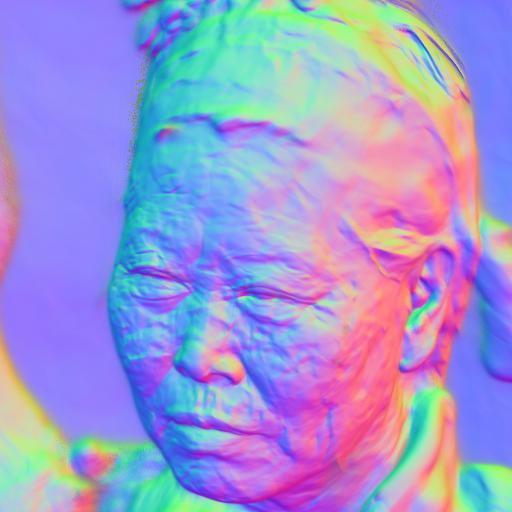}\\
    \makebox[0.09\textwidth]{GT}
    \makebox[0.09\textwidth]{NeuS}
    \makebox[0.09\textwidth]{HF-NeuS}
    \makebox[0.09\textwidth]{VolSDF}
    \makebox[0.09\textwidth]{Ours}
    \makebox[0.09\textwidth]{GT}
    \makebox[0.09\textwidth]{NeuS}
    \makebox[0.09\textwidth]{HF-NeuS}
    \makebox[0.09\textwidth]{VolSDF}
    \makebox[0.09\textwidth]{Ours}\\
    \caption{Comparison of various approaches under a 10-view setting (from Model 371 to Model 389).
    For each model, we show the results on one training view (left) and one novel view (right).}
    \label{fig:all_result}
\end{figure*}

In the 10-view setting, VolSDF generates erroneous geometry for Models 377, 383, and 401 due to the insufficient number of views. As illustrated in Figure~\ref{fig:all_result}, VolSDF only learns a partial geometry for the training views, resulting in poor novel view synthesis results.

Although the reconstruction quality improves with 15 views, VolSDF still fails to reconstruct Models 558 and 608. It is possible that VolSDF was successful on these two models with 10 views, but failed on them with 15 views because we chose the input views \textbf{randomly} from the original PR dataset in order to test the robustness of various approaches. The redundant information in the given views may not be helpful for improving the reconstruction quality of VolSDF.

\begin{figure*}[htbp]
    \centering
    \rotatebox{90}{\textbf{558}}
    \includegraphics[width=0.09\textwidth]{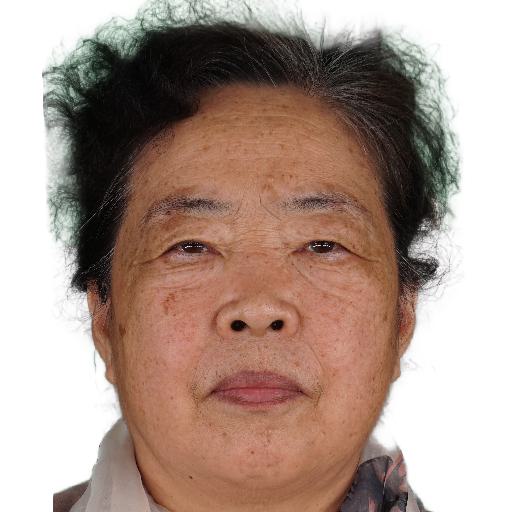}
    \includegraphics[width=0.09\textwidth]{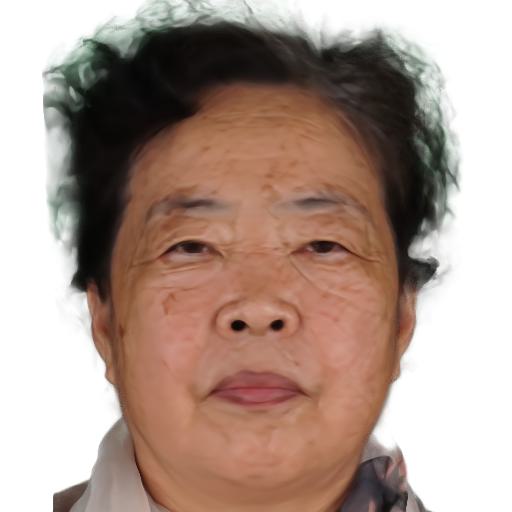}
    \includegraphics[width=0.09\textwidth]{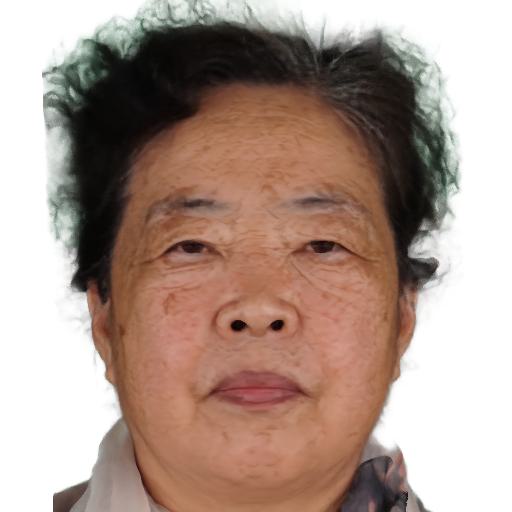}
    \includegraphics[width=0.09\textwidth]{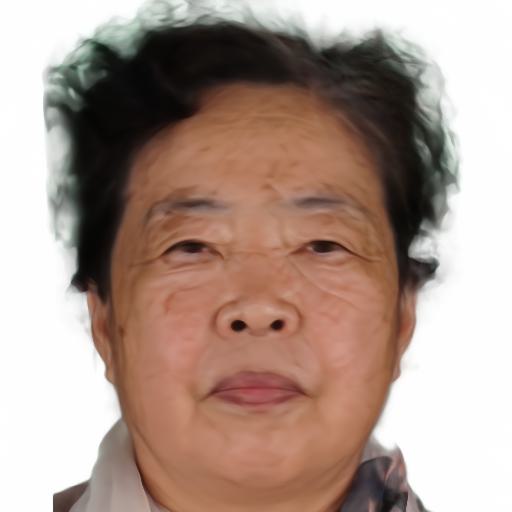}
    \includegraphics[width=0.09\textwidth]{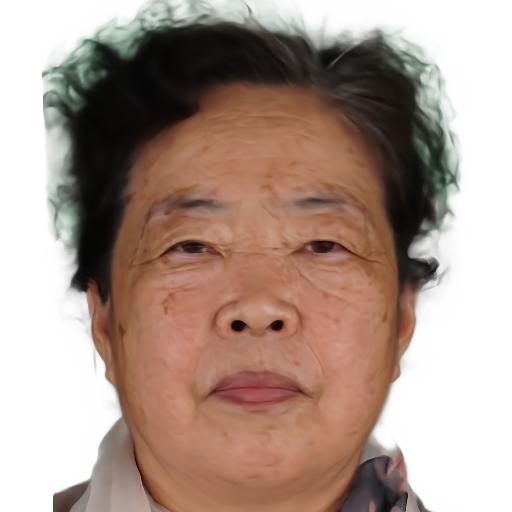}
    \includegraphics[width=0.09\textwidth]{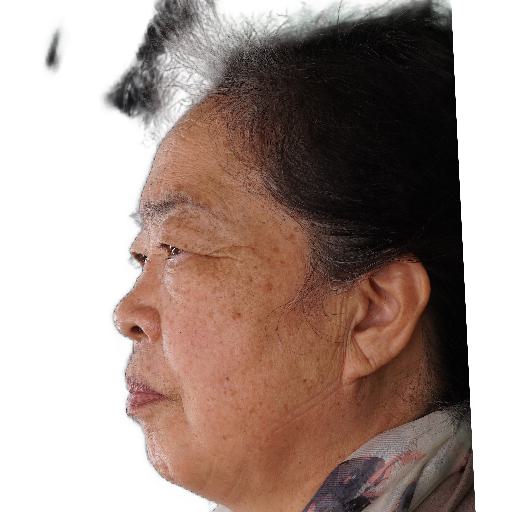}
    \includegraphics[width=0.09\textwidth]{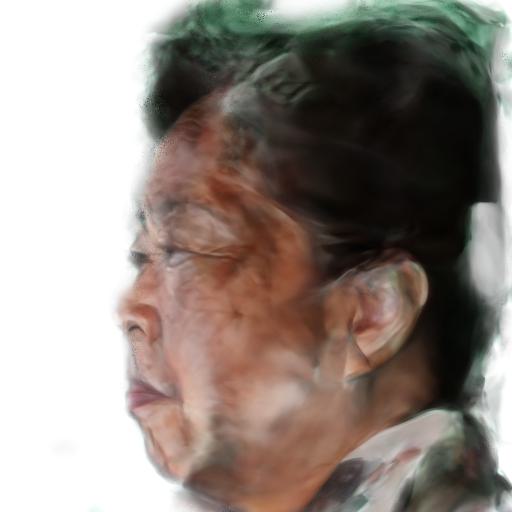}
    \includegraphics[width=0.09\textwidth]{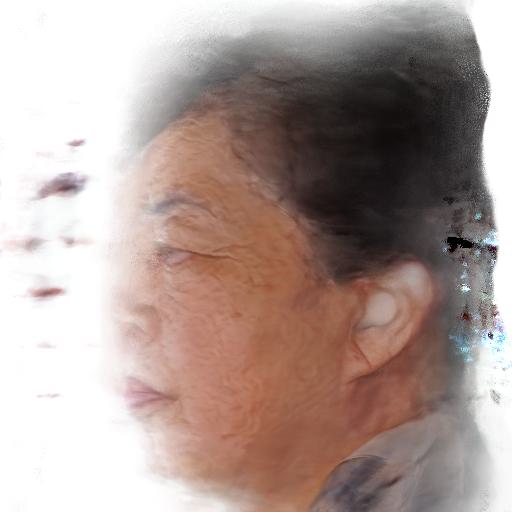}
    \includegraphics[width=0.09\textwidth]{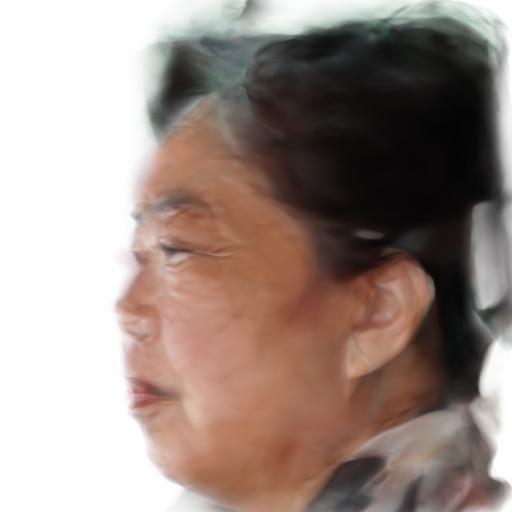}
    \includegraphics[width=0.09\textwidth]{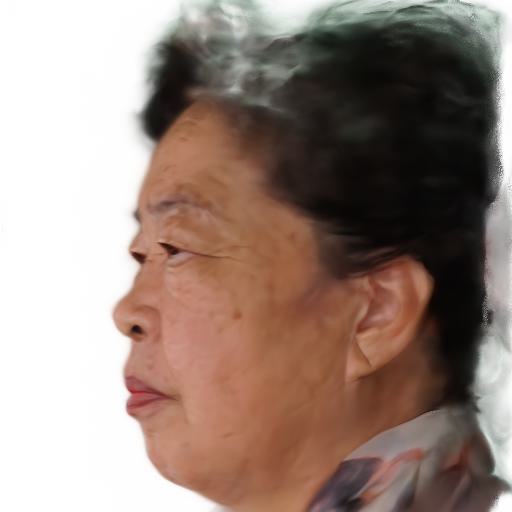}
    \rotatebox{90}{\tiny}
    \includegraphics[width=0.09\textwidth]{./results/template_effects/gt_571_blank.jpg}
    \includegraphics[width=0.09\textwidth]{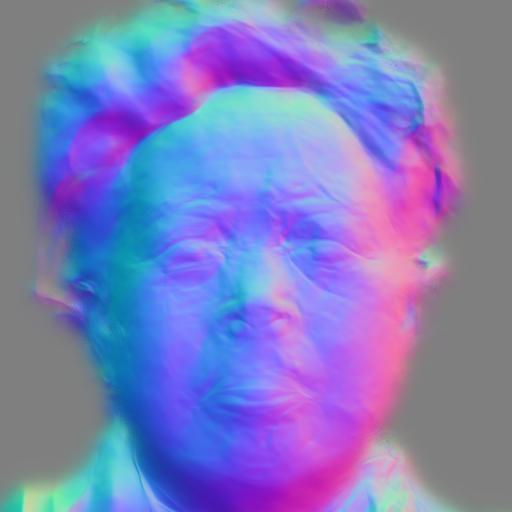}
    \includegraphics[width=0.09\textwidth]{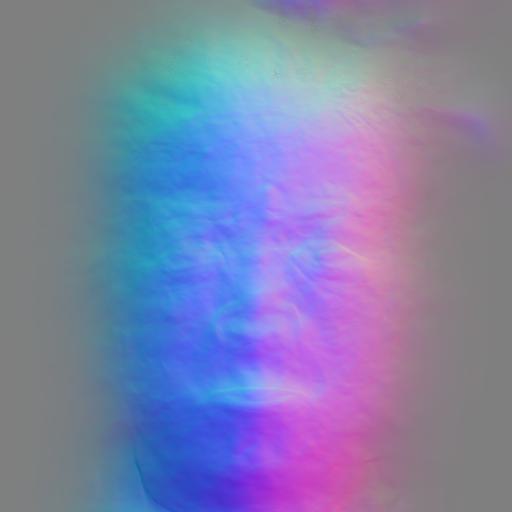}
    \includegraphics[width=0.09\textwidth]{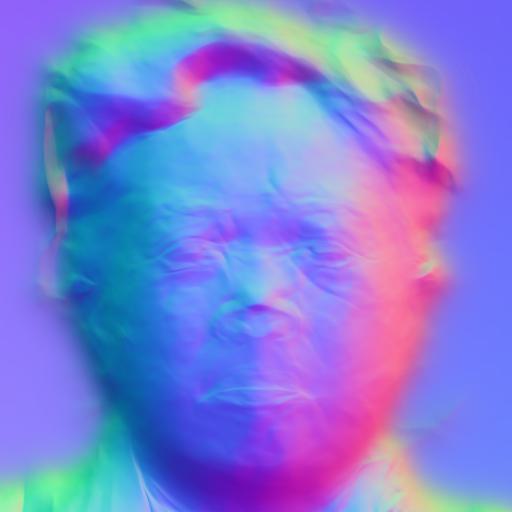}
    \includegraphics[width=0.09\textwidth]{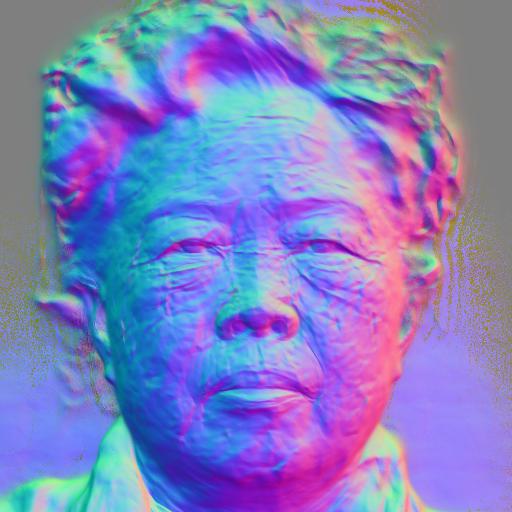}
    \includegraphics[width=0.09\textwidth]{./results/template_effects/gt_571_blank.jpg}
    \includegraphics[width=0.09\textwidth]{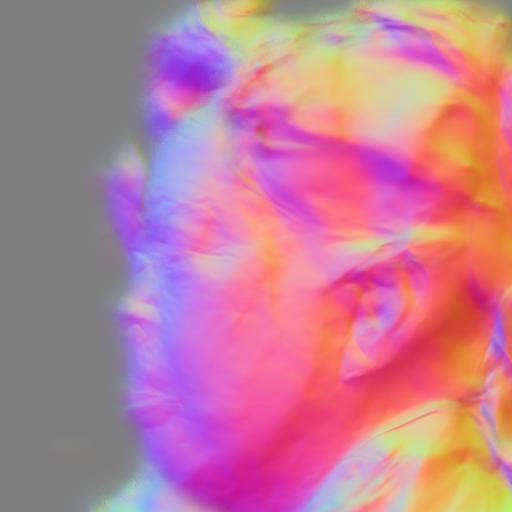}
    \includegraphics[width=0.09\textwidth]{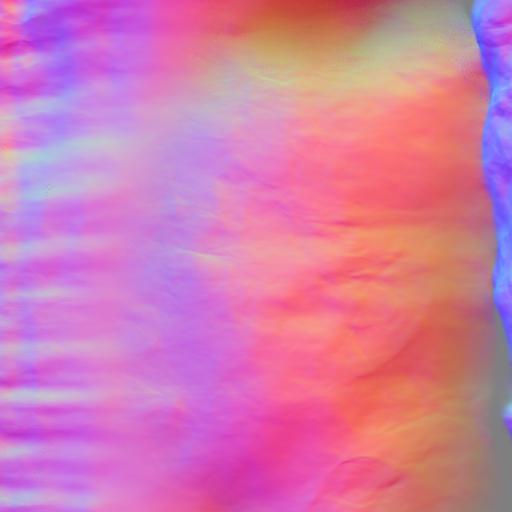}
    \includegraphics[width=0.09\textwidth]{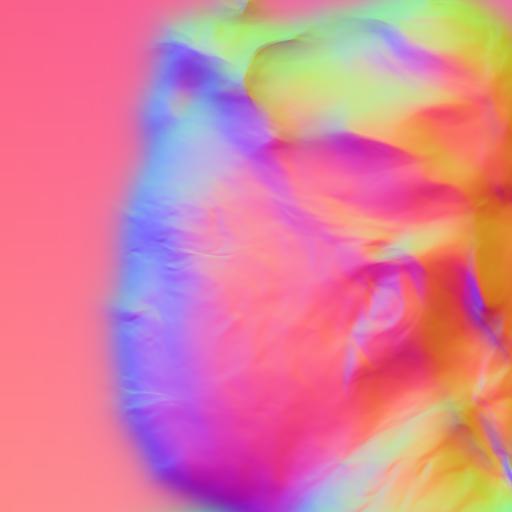}
    \includegraphics[width=0.09\textwidth]{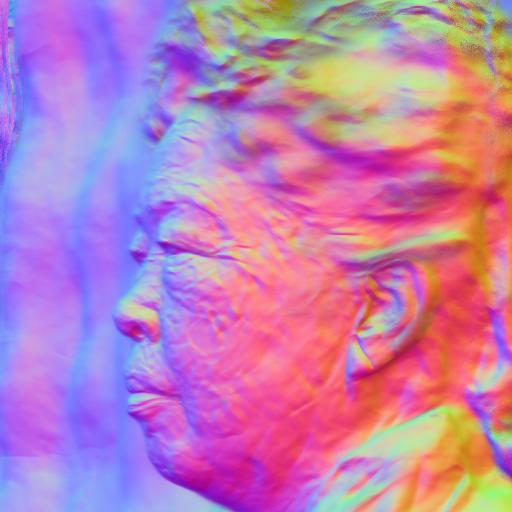}\\
    \rotatebox{90}{\textbf{566}}
    \includegraphics[width=0.09\textwidth]{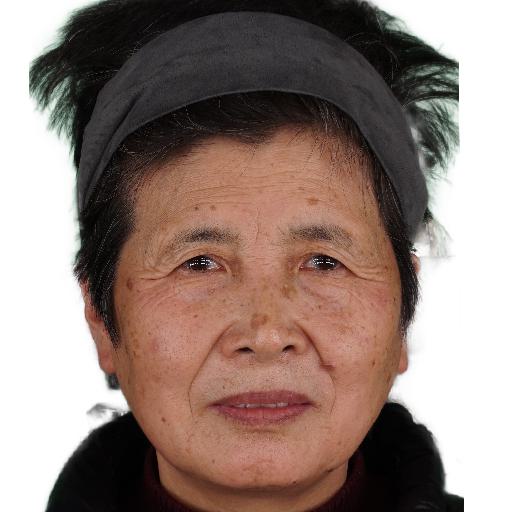}
    \includegraphics[width=0.09\textwidth]{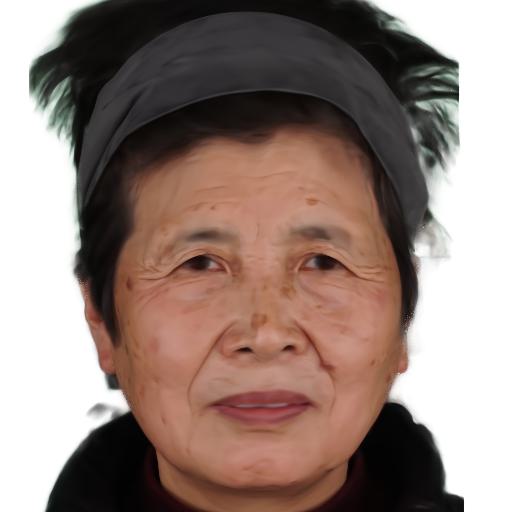}
    \includegraphics[width=0.09\textwidth]{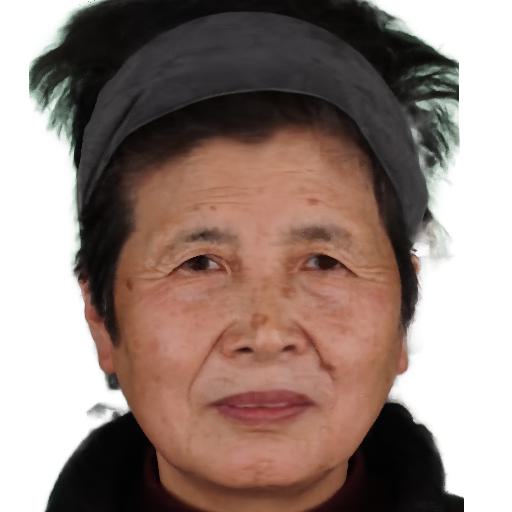}
    \includegraphics[width=0.09\textwidth]{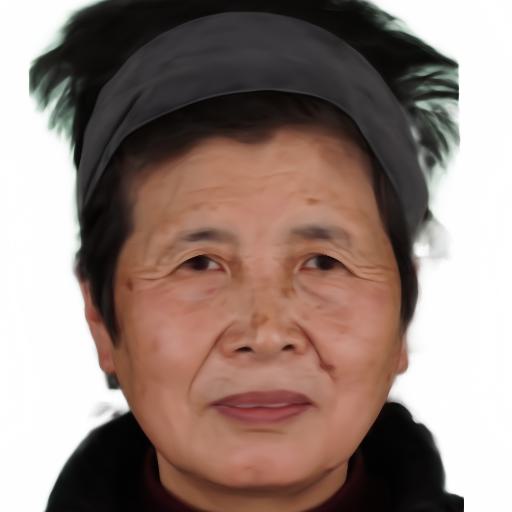}
    \includegraphics[width=0.09\textwidth]{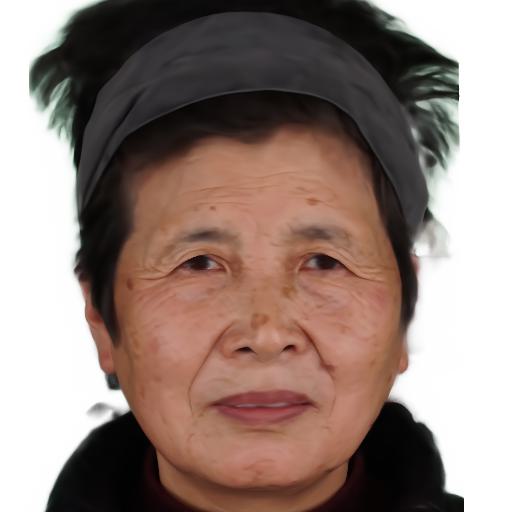}
    \includegraphics[width=0.09\textwidth]{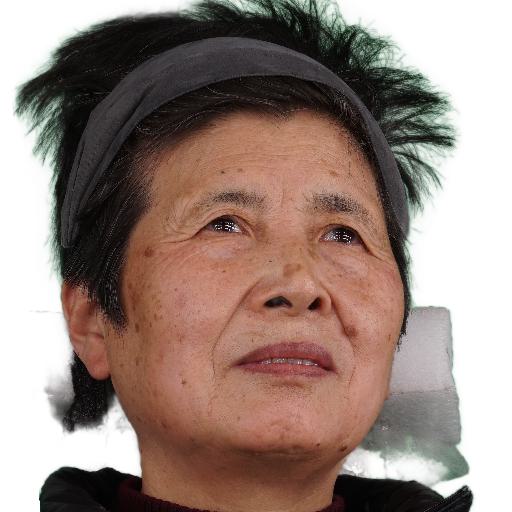}
    \includegraphics[width=0.09\textwidth]{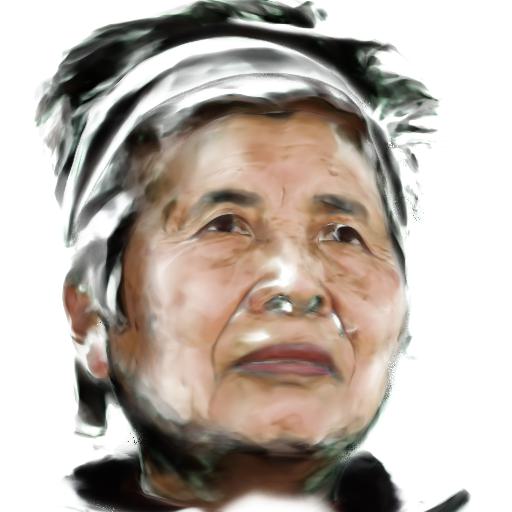}
    \includegraphics[width=0.09\textwidth]{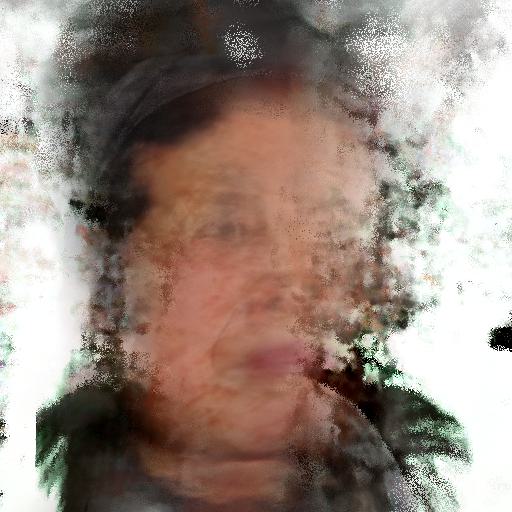}
    \includegraphics[width=0.09\textwidth]{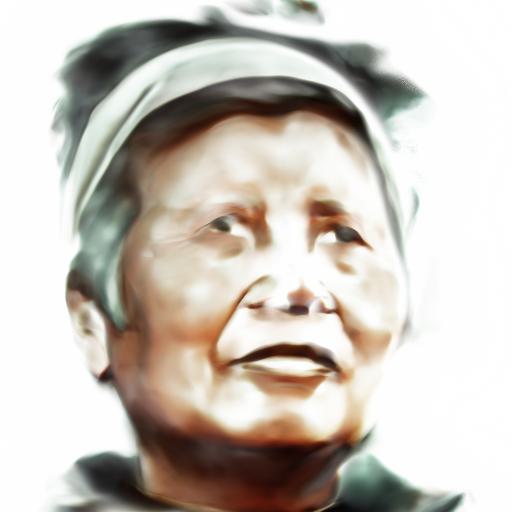}
    \includegraphics[width=0.09\textwidth]{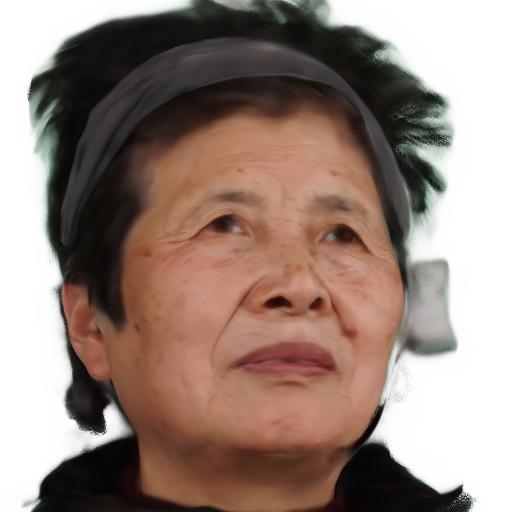}
    \rotatebox{90}{\tiny}
    \includegraphics[width=0.09\textwidth]{./results/template_effects/gt_571_blank.jpg}
    \includegraphics[width=0.09\textwidth]{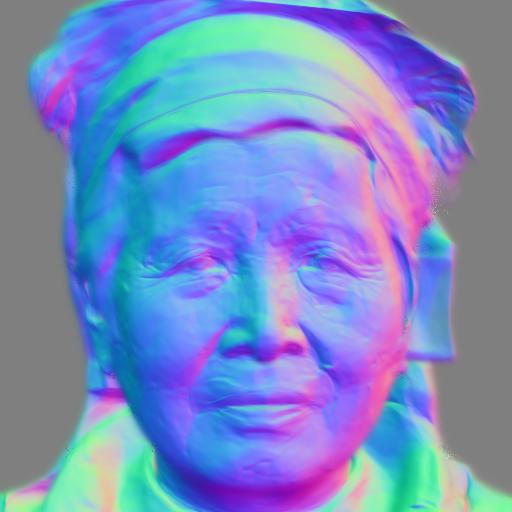}
    \includegraphics[width=0.09\textwidth]{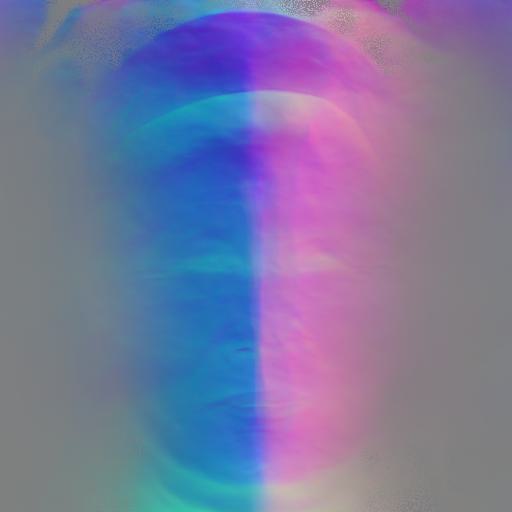}
    \includegraphics[width=0.09\textwidth]{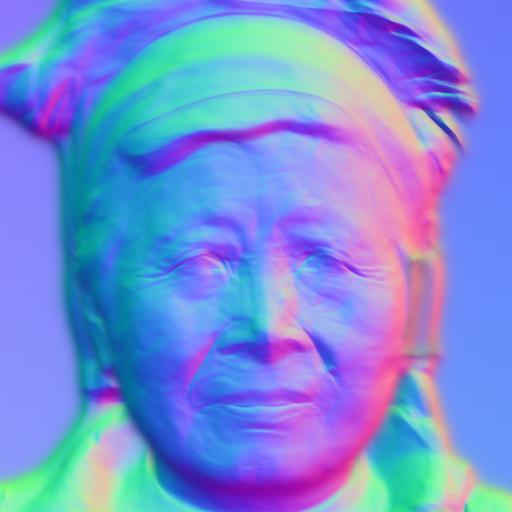}
    \includegraphics[width=0.09\textwidth]{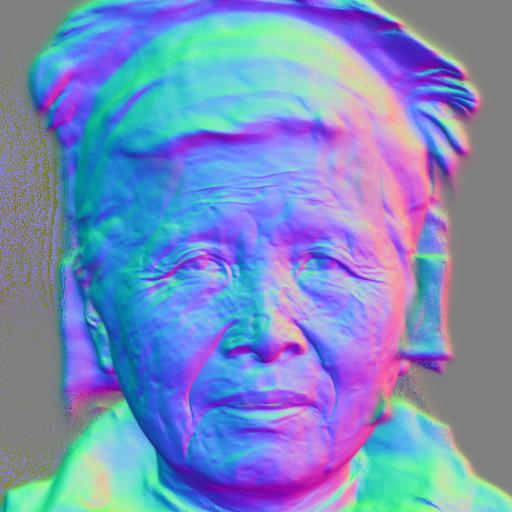}
    \includegraphics[width=0.09\textwidth]{./results/template_effects/gt_571_blank.jpg}
    \includegraphics[width=0.09\textwidth]{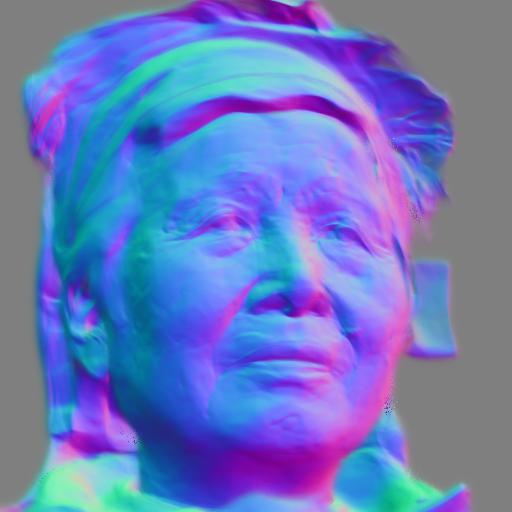}
    \includegraphics[width=0.09\textwidth]{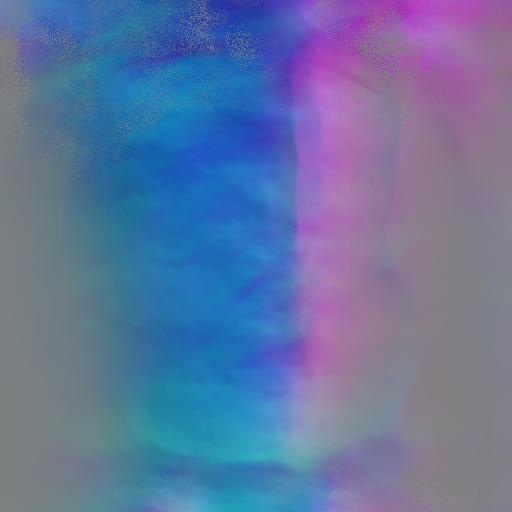}
    \includegraphics[width=0.09\textwidth]{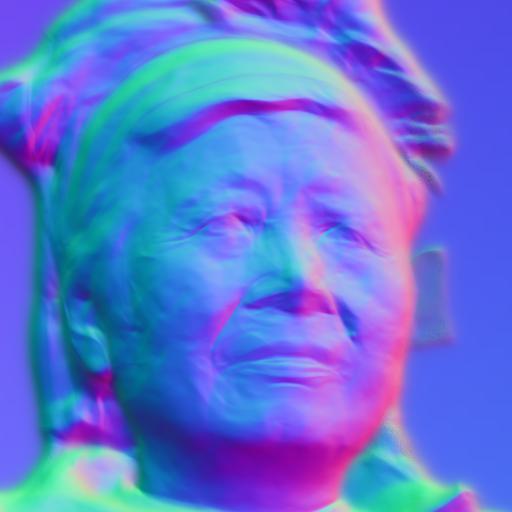}
    \includegraphics[width=0.09\textwidth]{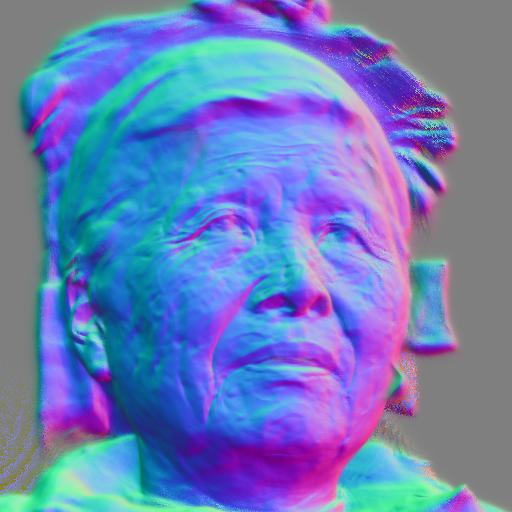}\\
    \rotatebox{90}{\textbf{571}}
    \includegraphics[width=0.09\textwidth]{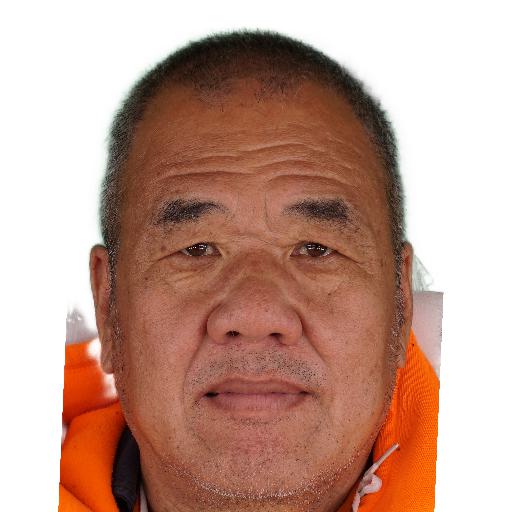}
    \includegraphics[width=0.09\textwidth]{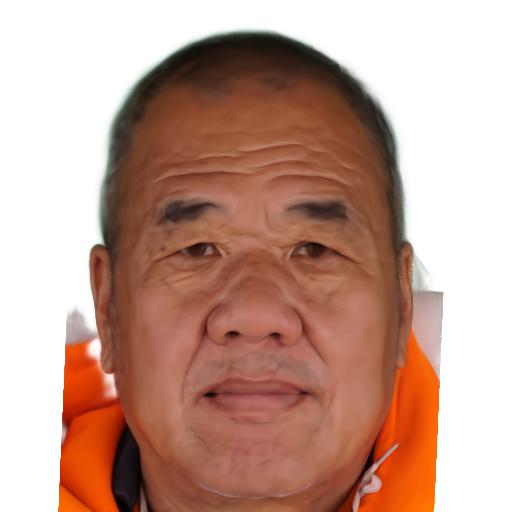}
    \includegraphics[width=0.09\textwidth]{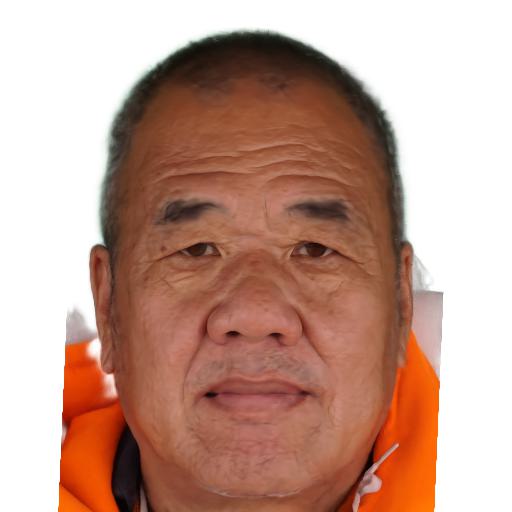}
    \includegraphics[width=0.09\textwidth]{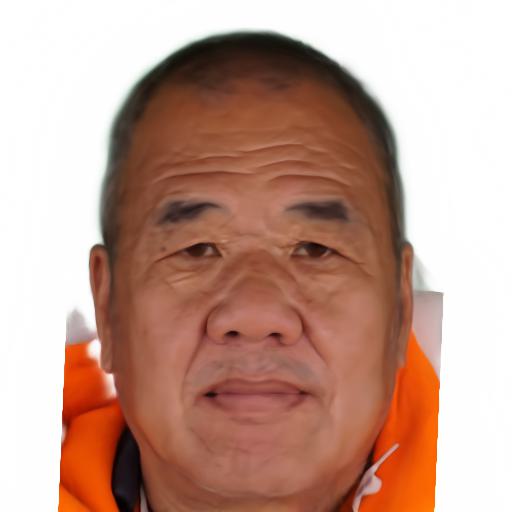}
    \includegraphics[width=0.09\textwidth]{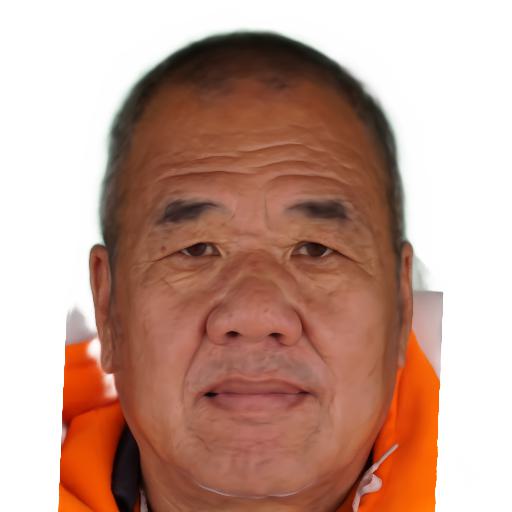}
    \includegraphics[width=0.09\textwidth]{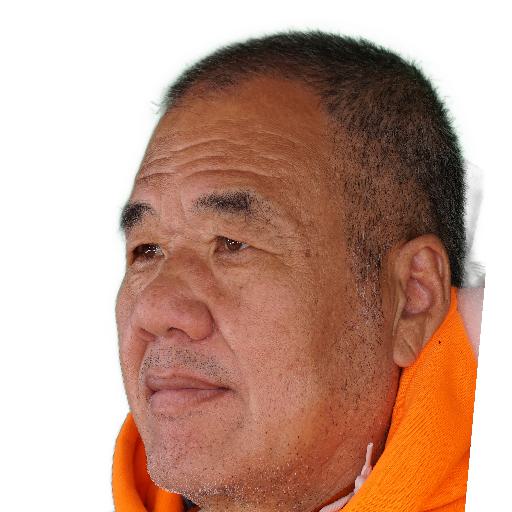}
    \includegraphics[width=0.09\textwidth]{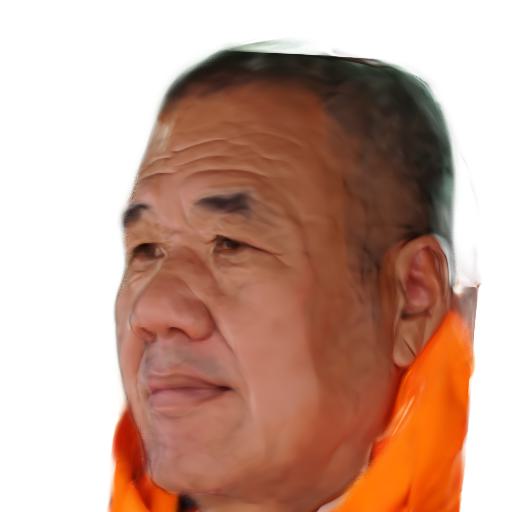}
    \includegraphics[width=0.09\textwidth]{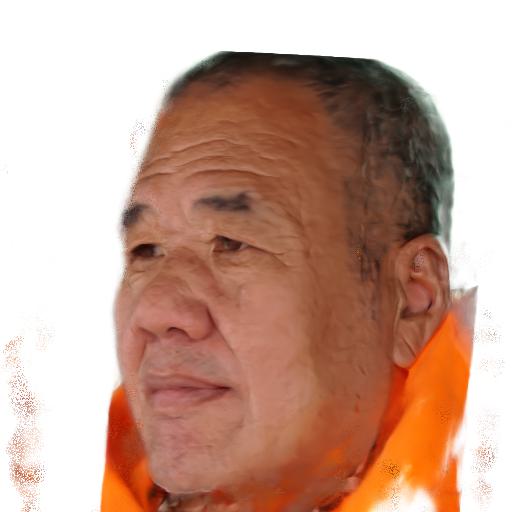}
    \includegraphics[width=0.09\textwidth]{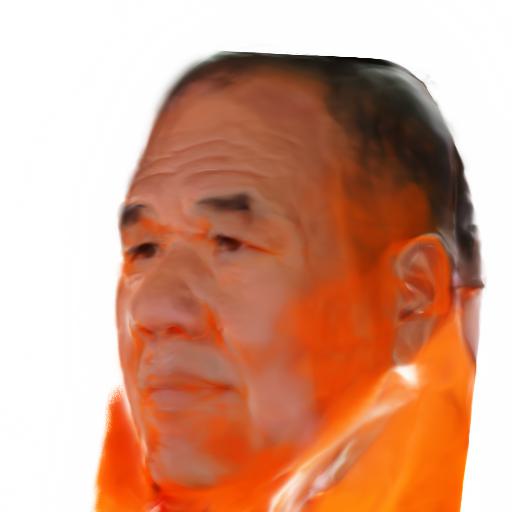}
    \includegraphics[width=0.09\textwidth]{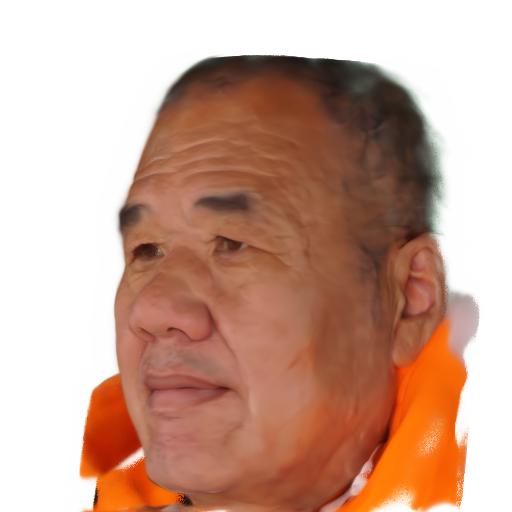}
    \rotatebox{90}{\tiny}
    \includegraphics[width=0.09\textwidth]{./results/template_effects/gt_571_blank.jpg}
    \includegraphics[width=0.09\textwidth]{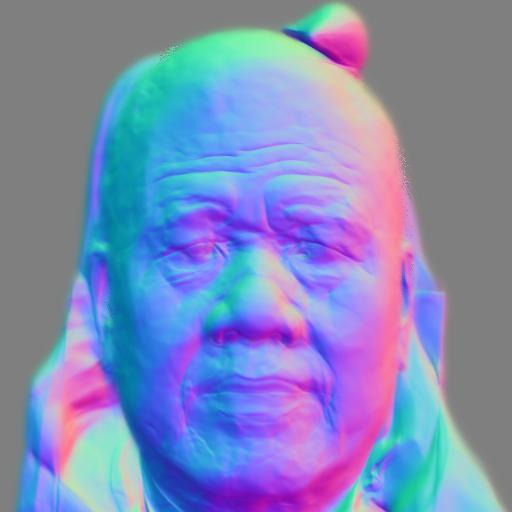}
    \includegraphics[width=0.09\textwidth]{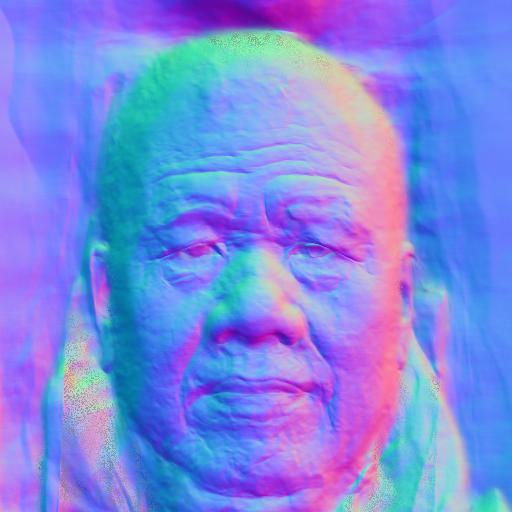}
    \includegraphics[width=0.09\textwidth]{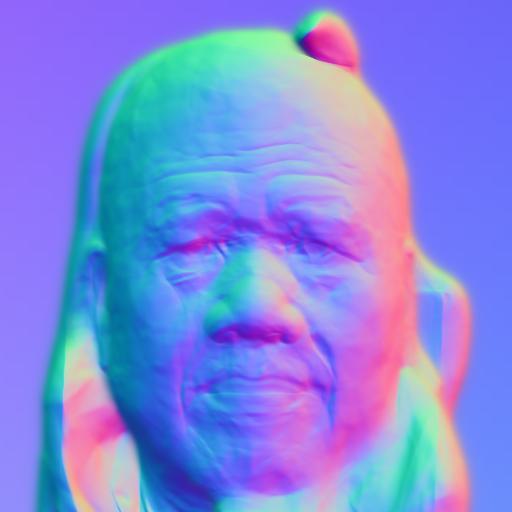}
    \includegraphics[width=0.09\textwidth]{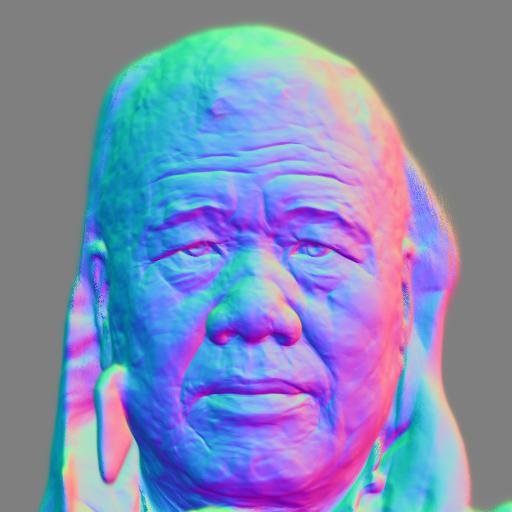}
    \includegraphics[width=0.09\textwidth]{./results/template_effects/gt_571_blank.jpg}
    \includegraphics[width=0.09\textwidth]{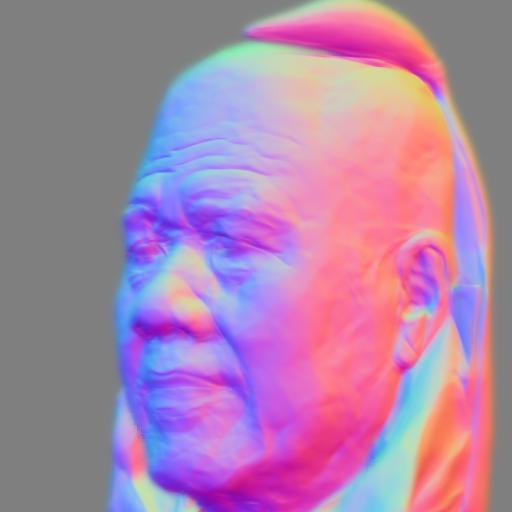}
    \includegraphics[width=0.09\textwidth]{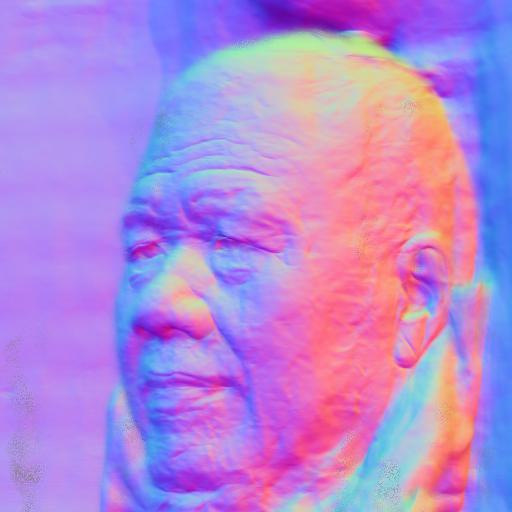}
    \includegraphics[width=0.09\textwidth]{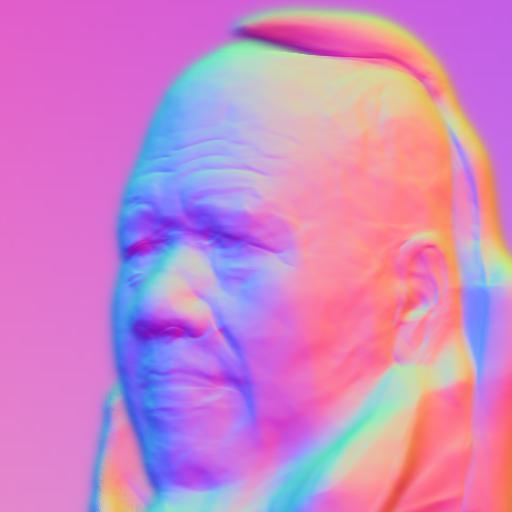}
    \includegraphics[width=0.09\textwidth]{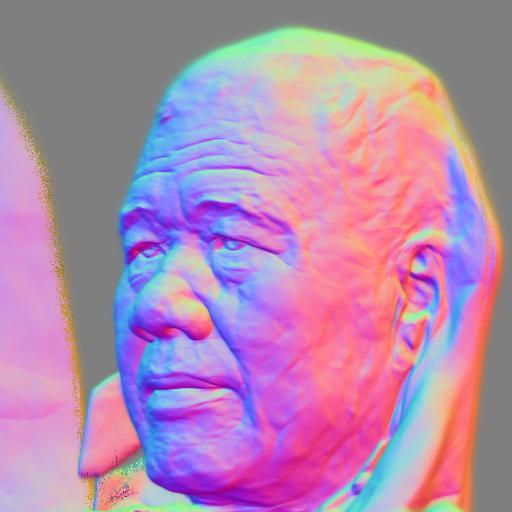}\\
    \rotatebox{90}{\textbf{608}}
    \includegraphics[width=0.09\textwidth]{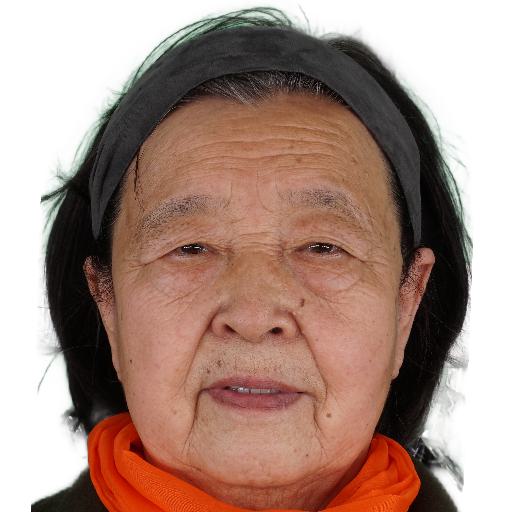}
    \includegraphics[width=0.09\textwidth]{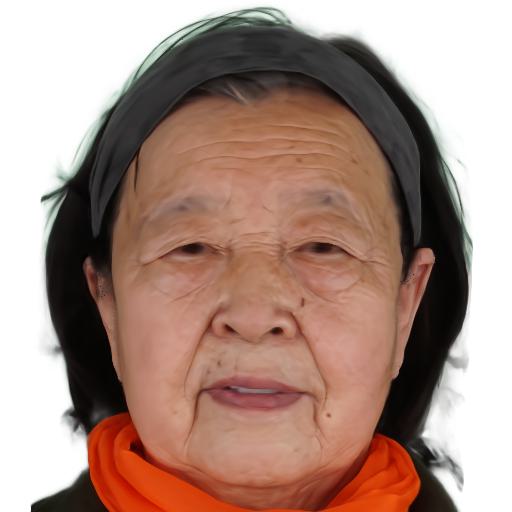}
    \includegraphics[width=0.09\textwidth]{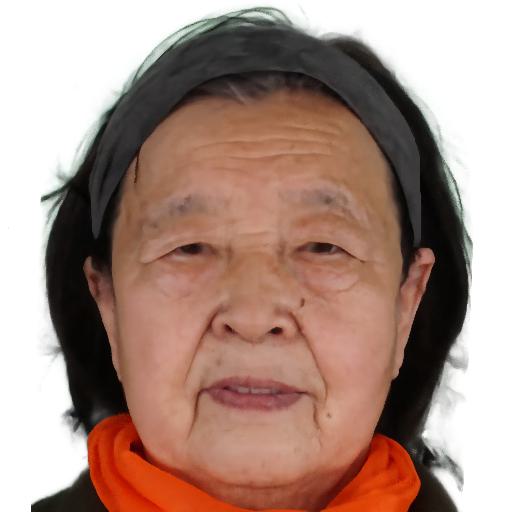}
    \includegraphics[width=0.09\textwidth]{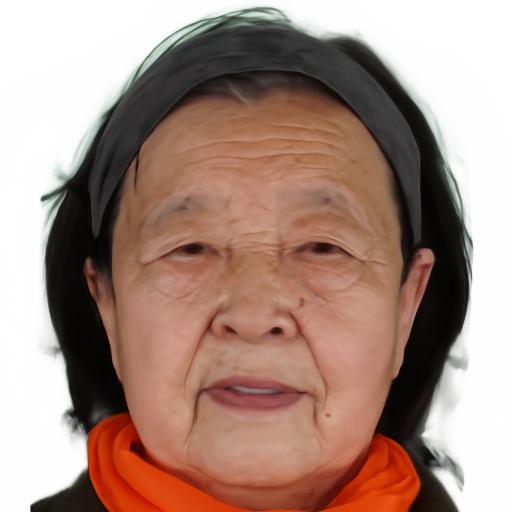}
    \includegraphics[width=0.09\textwidth]{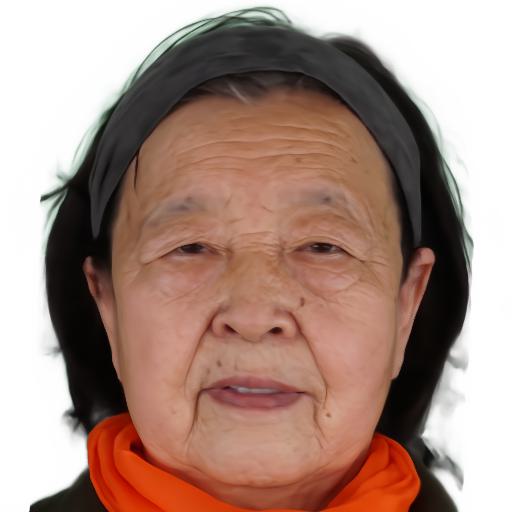}
    \includegraphics[width=0.09\textwidth]{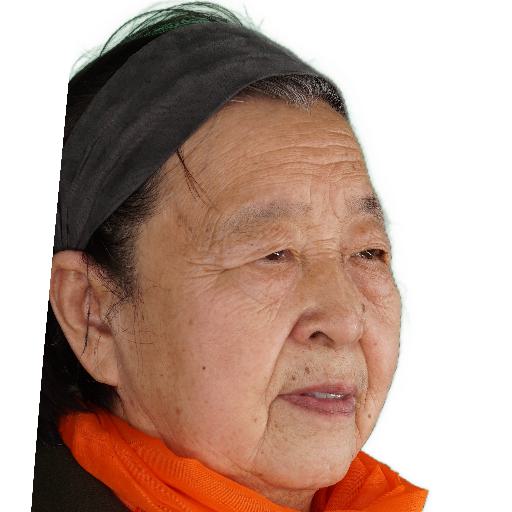}
    \includegraphics[width=0.09\textwidth]{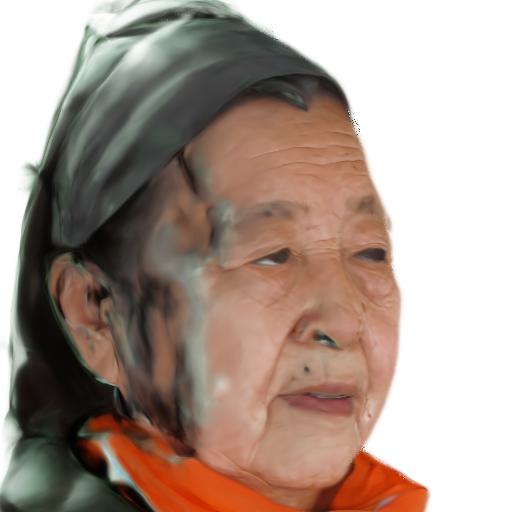}
    \includegraphics[width=0.09\textwidth]{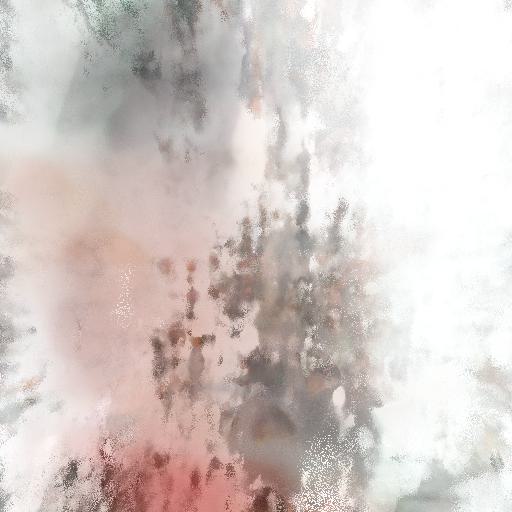}
    \includegraphics[width=0.09\textwidth]{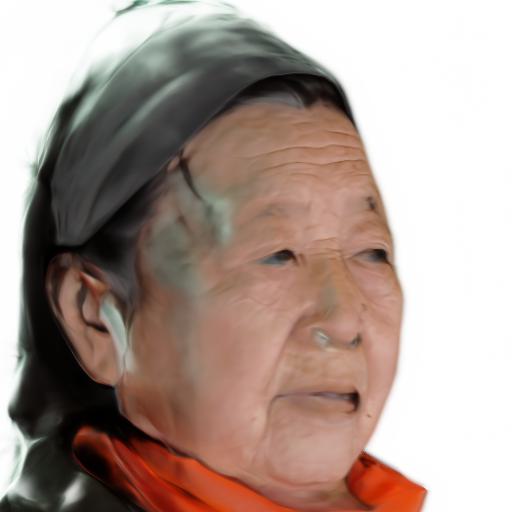}
    \includegraphics[width=0.09\textwidth]{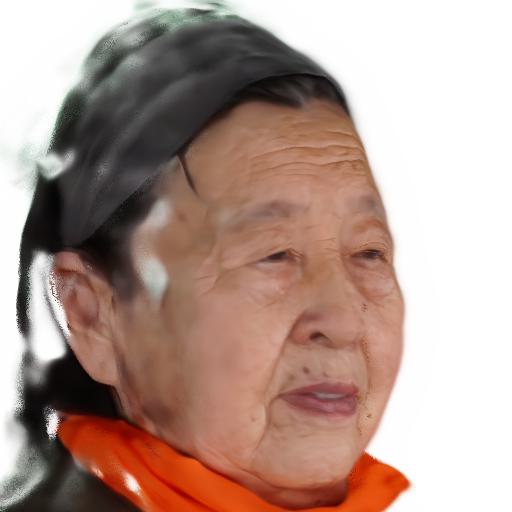}
    \rotatebox{90}{\tiny}
    \includegraphics[width=0.09\textwidth]{./results/template_effects/gt_571_blank.jpg}
    \includegraphics[width=0.09\textwidth]{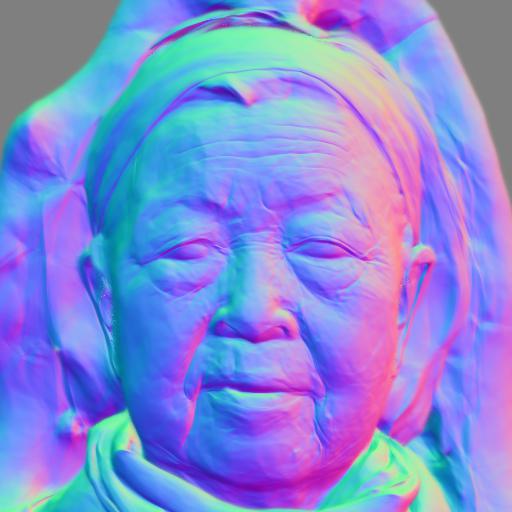}
    \includegraphics[width=0.09\textwidth]{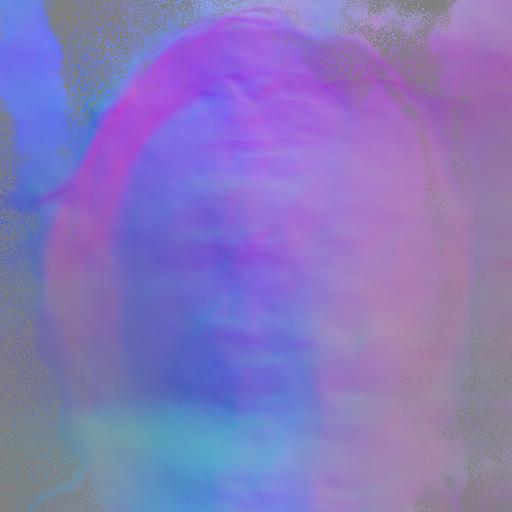}
    \includegraphics[width=0.09\textwidth]{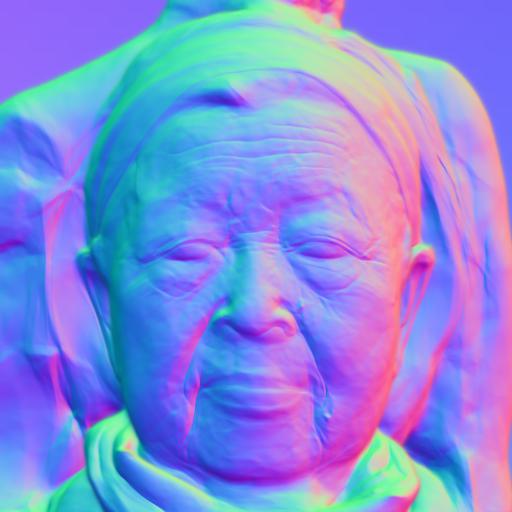}
    \includegraphics[width=0.09\textwidth]{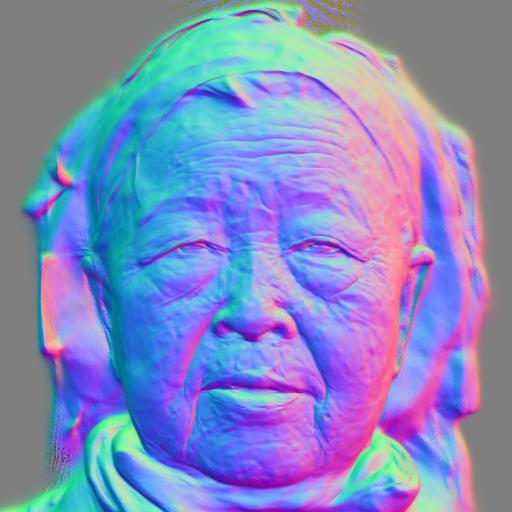}
    \includegraphics[width=0.09\textwidth]{./results/template_effects/gt_571_blank.jpg}
    \includegraphics[width=0.09\textwidth]{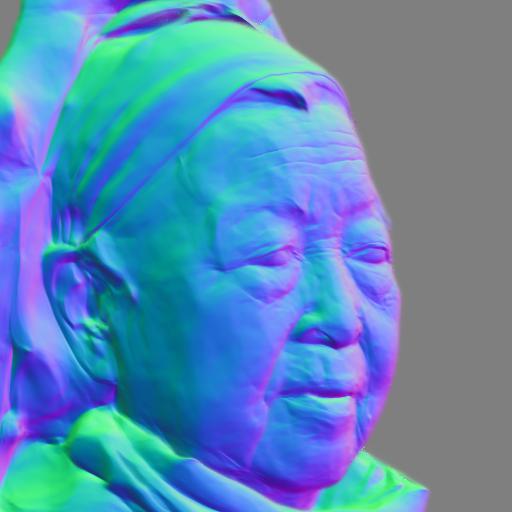}
    \includegraphics[width=0.09\textwidth]{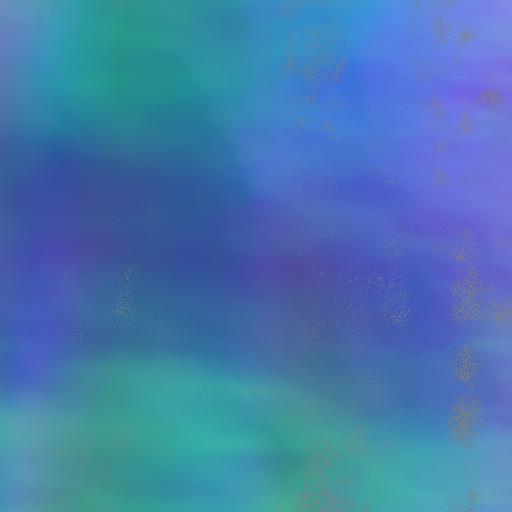}
    \includegraphics[width=0.09\textwidth]{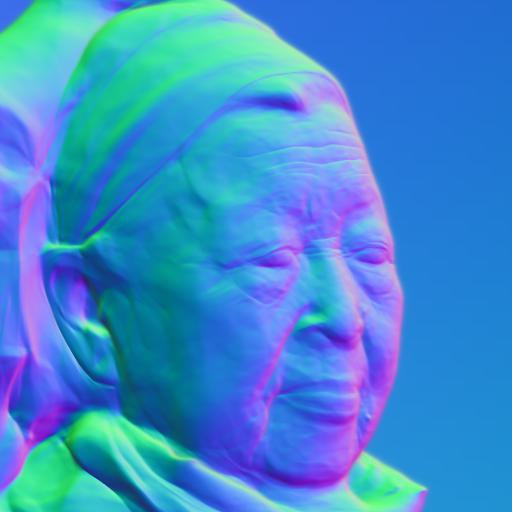}
    \includegraphics[width=0.09\textwidth]{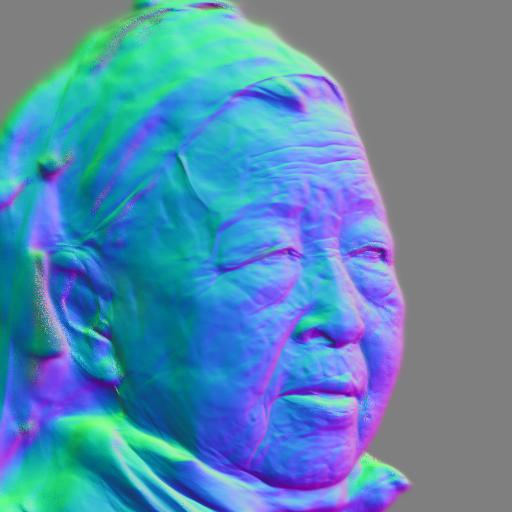}\\
    \rotatebox{90}{\textbf{619}}
    \includegraphics[width=0.09\textwidth]{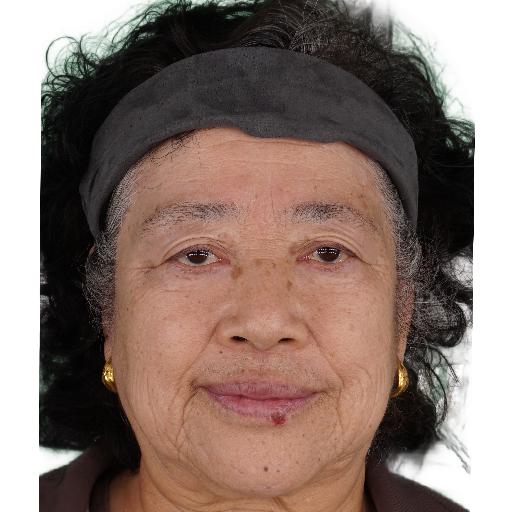}
    \includegraphics[width=0.09\textwidth]{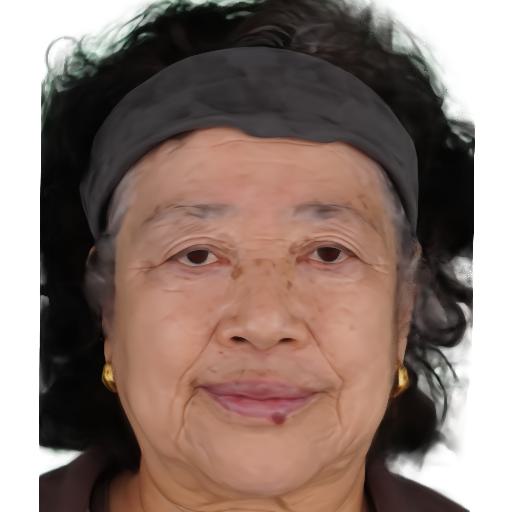}
    \includegraphics[width=0.09\textwidth]{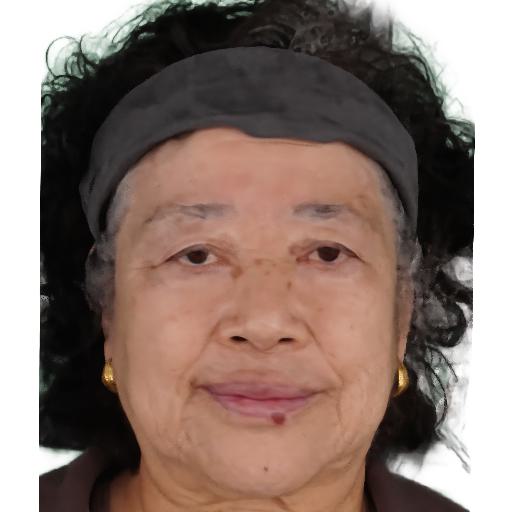}
    \includegraphics[width=0.09\textwidth]{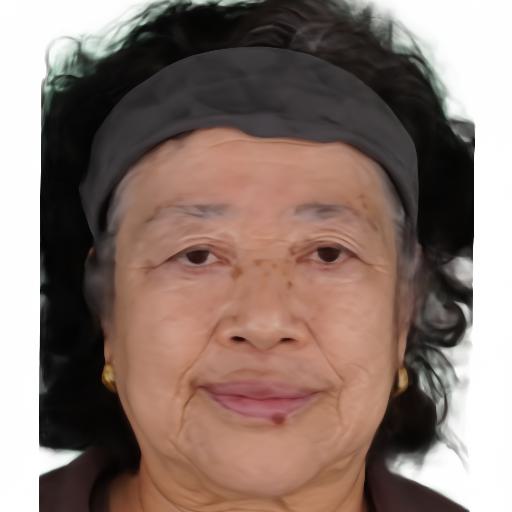}
    \includegraphics[width=0.09\textwidth]{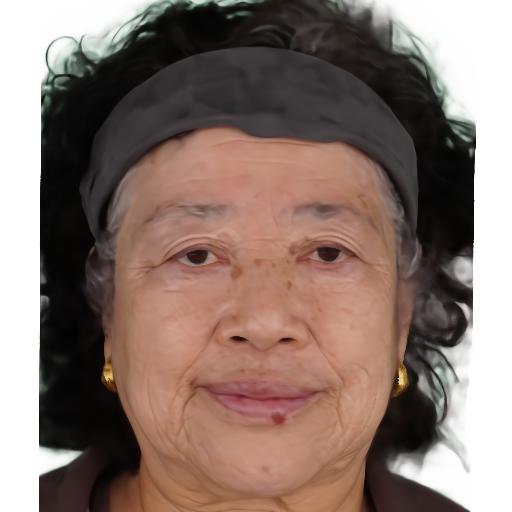}
    \includegraphics[width=0.09\textwidth]{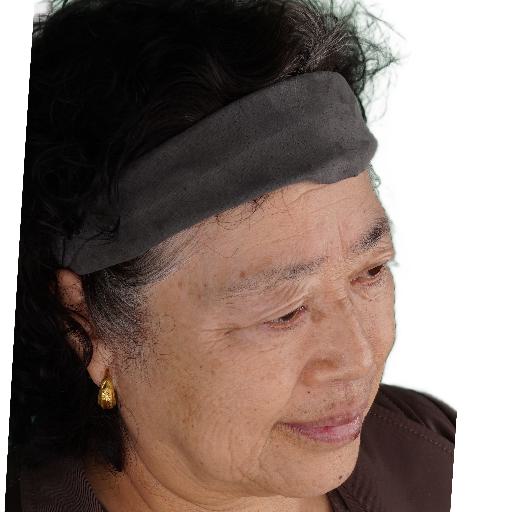}
    \includegraphics[width=0.09\textwidth]{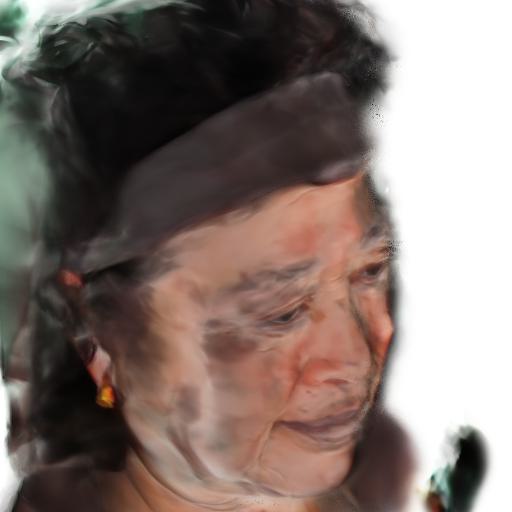}
    \includegraphics[width=0.09\textwidth]{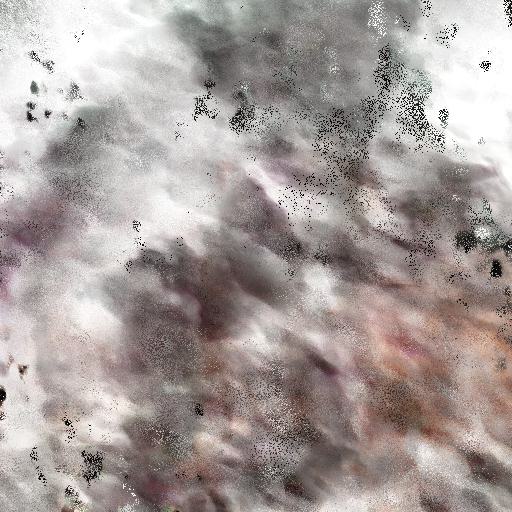}
    \includegraphics[width=0.09\textwidth]{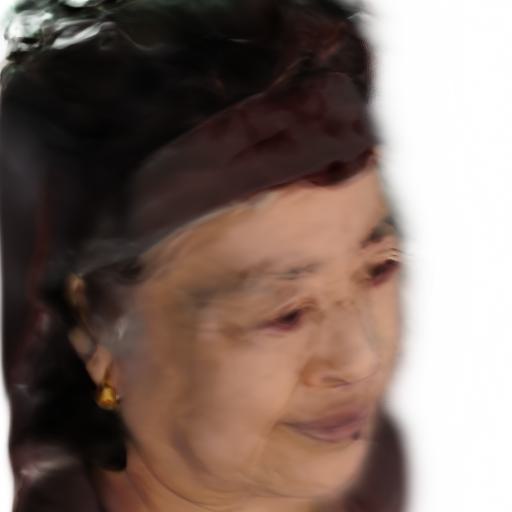}
    \includegraphics[width=0.09\textwidth]{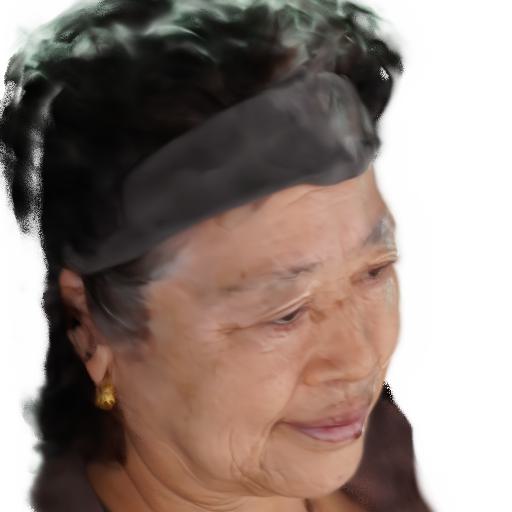}
    \rotatebox{90}{\tiny}
    \includegraphics[width=0.09\textwidth]{./results/template_effects/gt_571_blank.jpg}
    \includegraphics[width=0.09\textwidth]{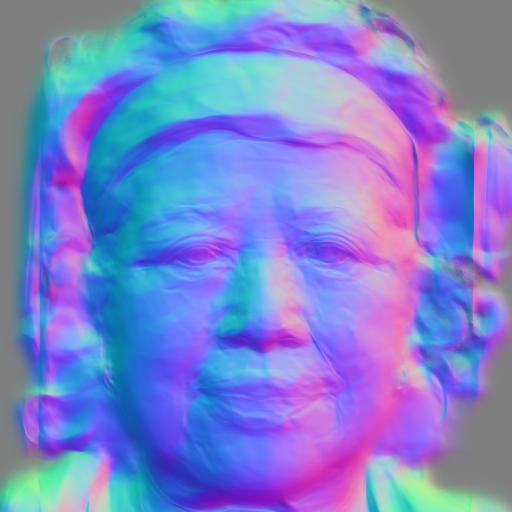}
    \includegraphics[width=0.09\textwidth]{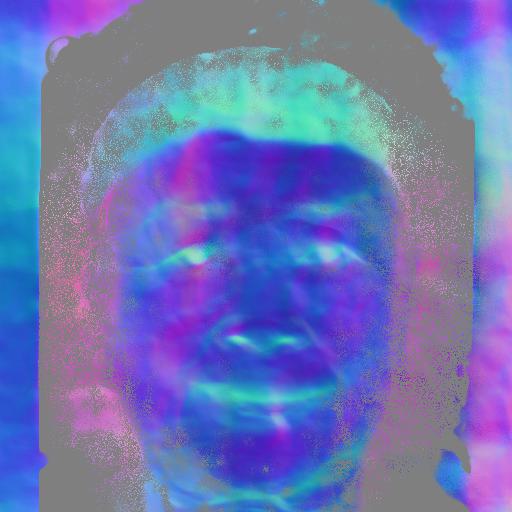}
    \includegraphics[width=0.09\textwidth]{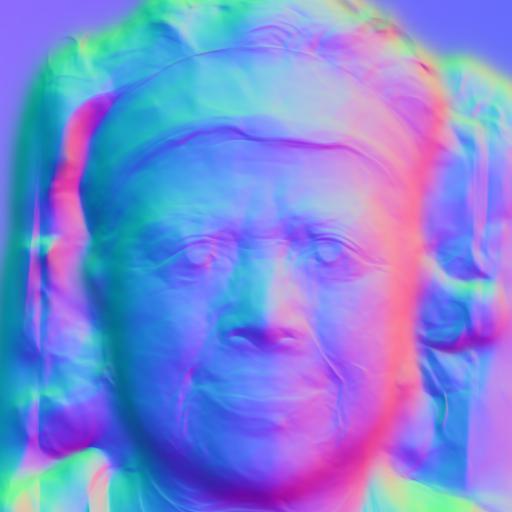}
    \includegraphics[width=0.09\textwidth]{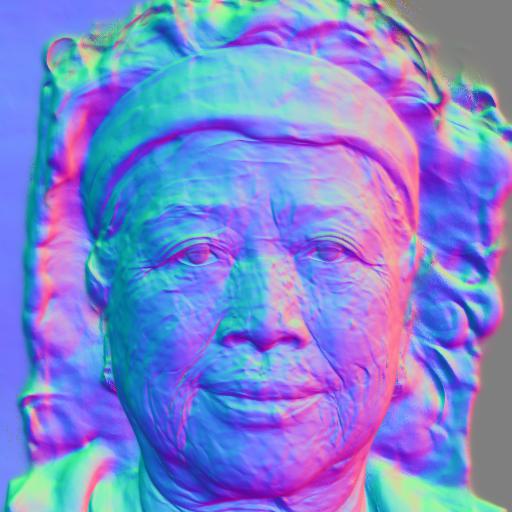}
    \includegraphics[width=0.09\textwidth]{./results/template_effects/gt_571_blank.jpg}
    \includegraphics[width=0.09\textwidth]{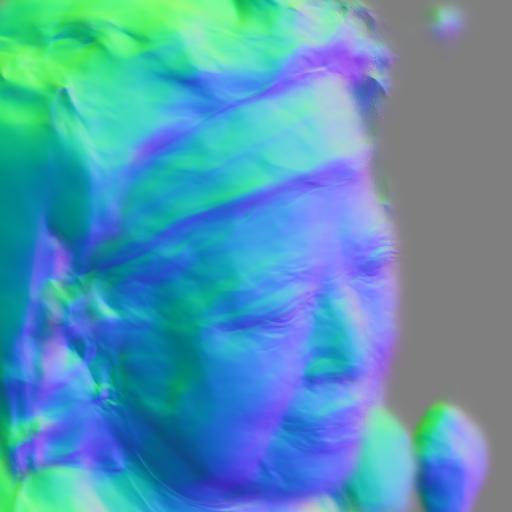}
    \includegraphics[width=0.09\textwidth]{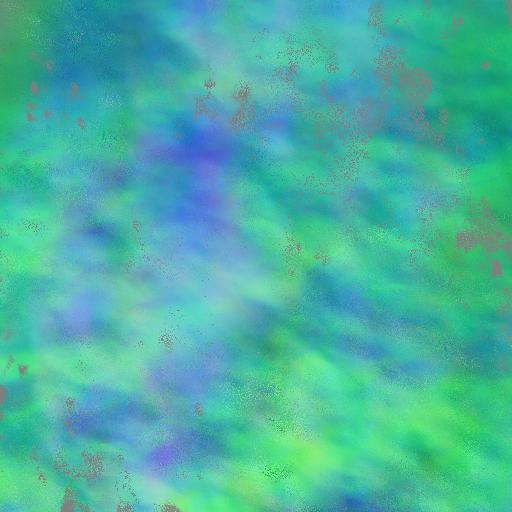}
    \includegraphics[width=0.09\textwidth]{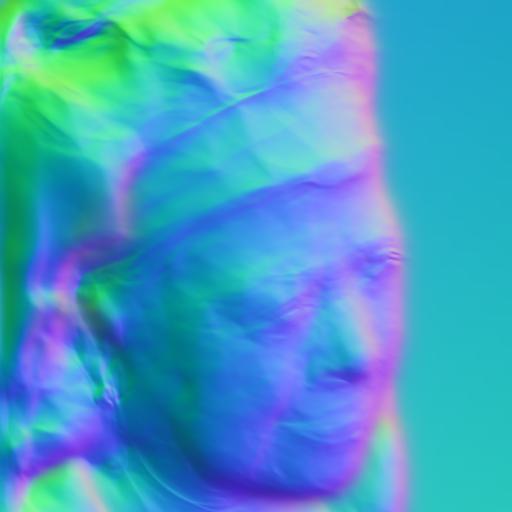}
    \includegraphics[width=0.09\textwidth]{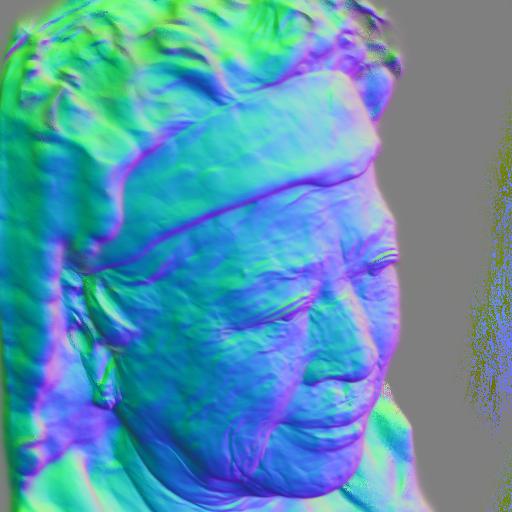}\\
    \rotatebox{90}{\textbf{635}}
    \includegraphics[width=0.09\textwidth]{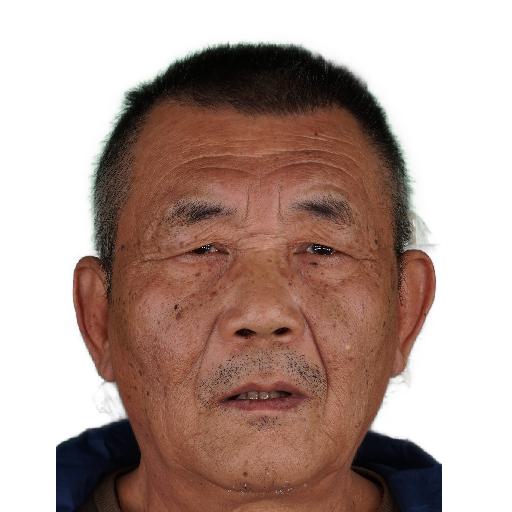}
    \includegraphics[width=0.09\textwidth]{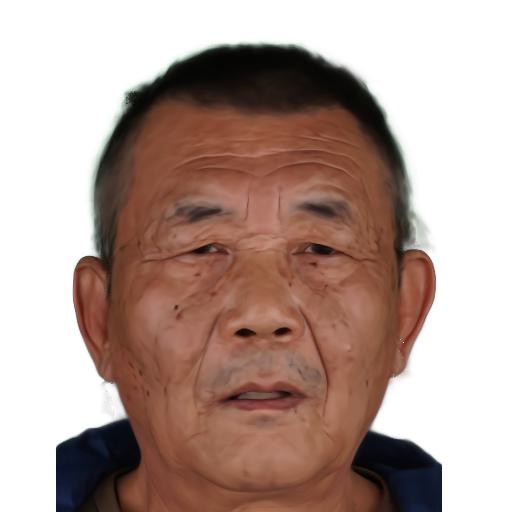}
    \includegraphics[width=0.09\textwidth]{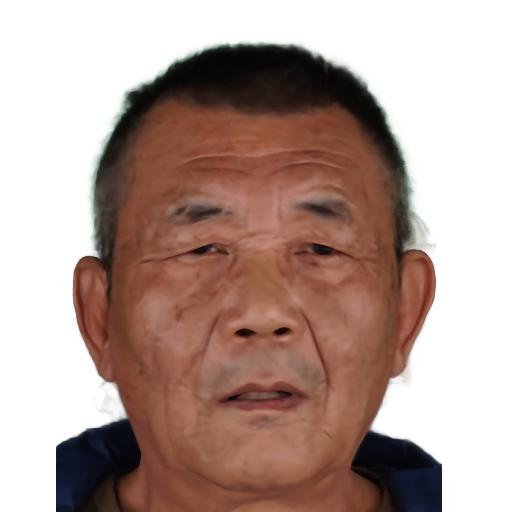}
    \includegraphics[width=0.09\textwidth]{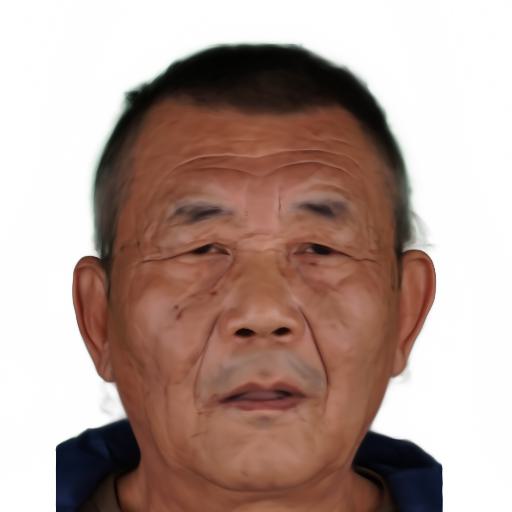}
    \includegraphics[width=0.09\textwidth]{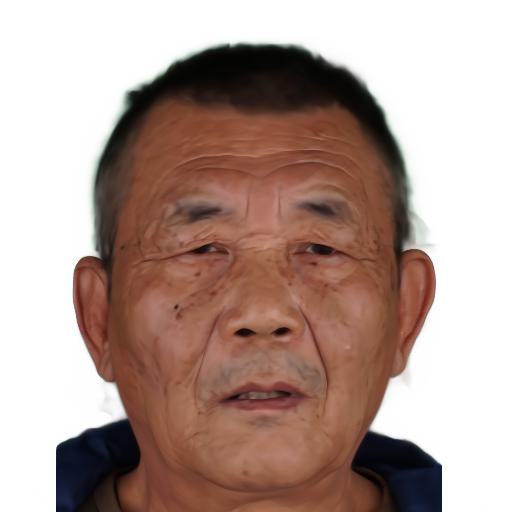}
    \includegraphics[width=0.09\textwidth]{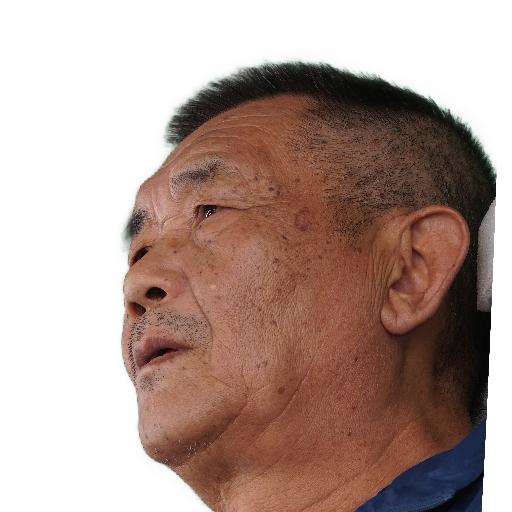}
    \includegraphics[width=0.09\textwidth]{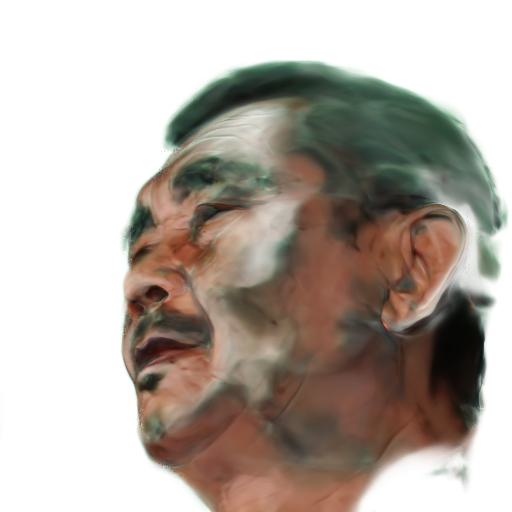}
    \includegraphics[width=0.09\textwidth]{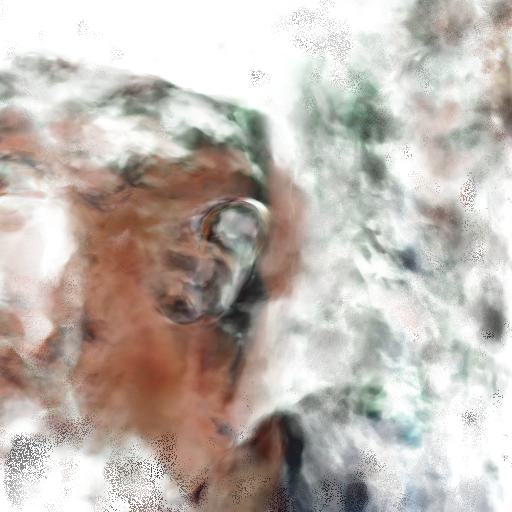}
    \includegraphics[width=0.09\textwidth]{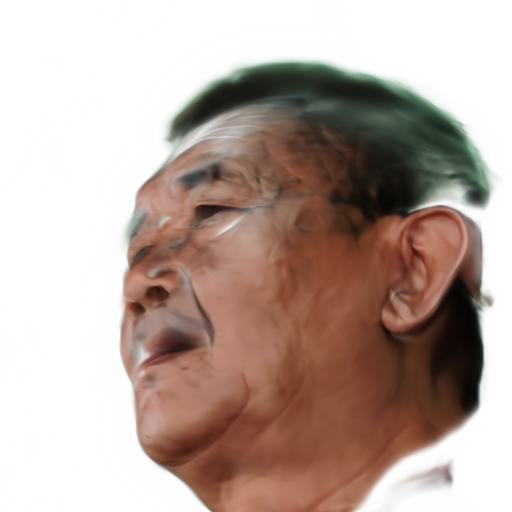}
    \includegraphics[width=0.09\textwidth]{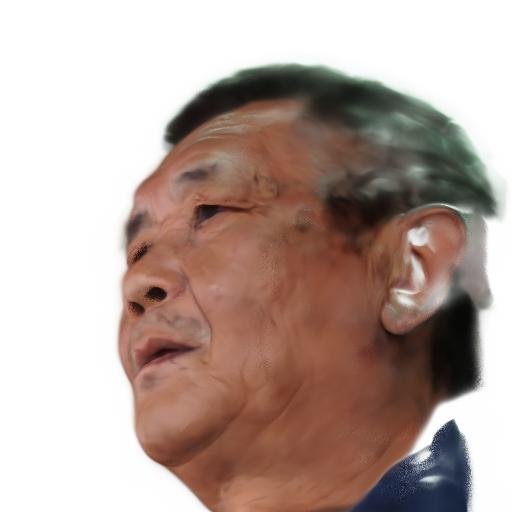}
    \rotatebox{90}{\tiny}
    \includegraphics[width=0.09\textwidth]{./results/template_effects/gt_571_blank.jpg}
    \includegraphics[width=0.09\textwidth]{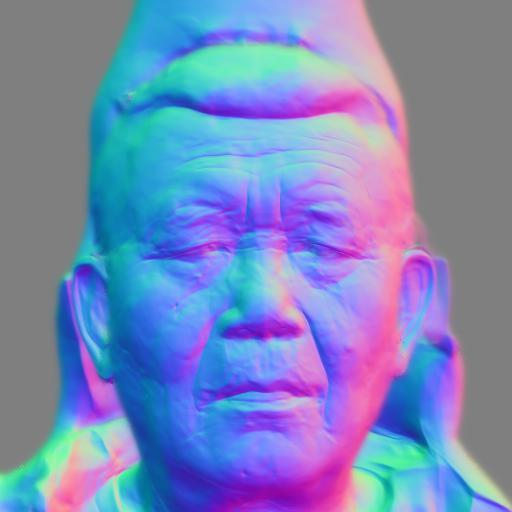}
    \includegraphics[width=0.09\textwidth]{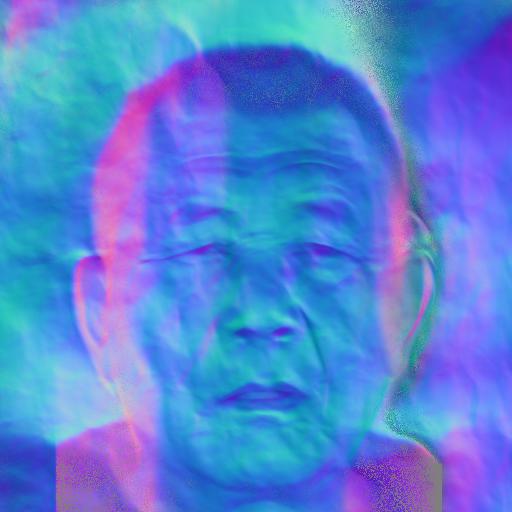}
    \includegraphics[width=0.09\textwidth]{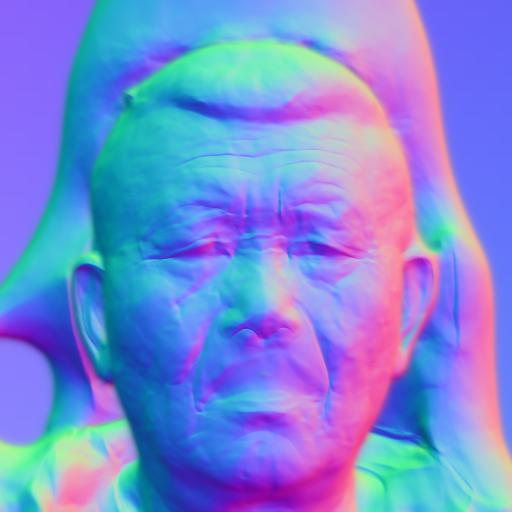}
    \includegraphics[width=0.09\textwidth]{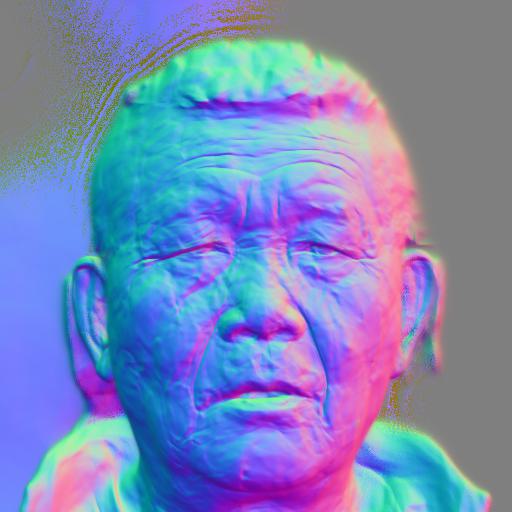}
    \includegraphics[width=0.09\textwidth]{./results/template_effects/gt_571_blank.jpg}
    \includegraphics[width=0.09\textwidth]{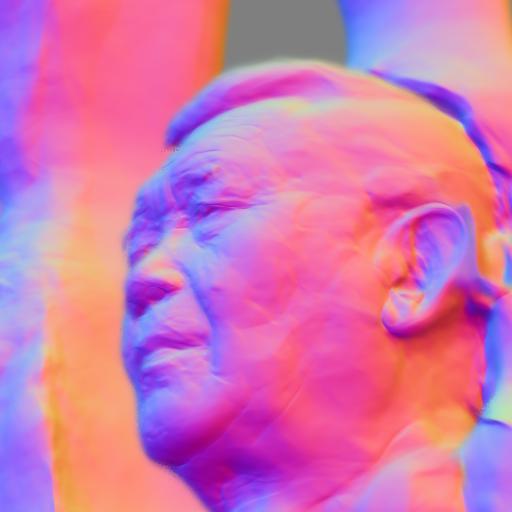}
    \includegraphics[width=0.09\textwidth]{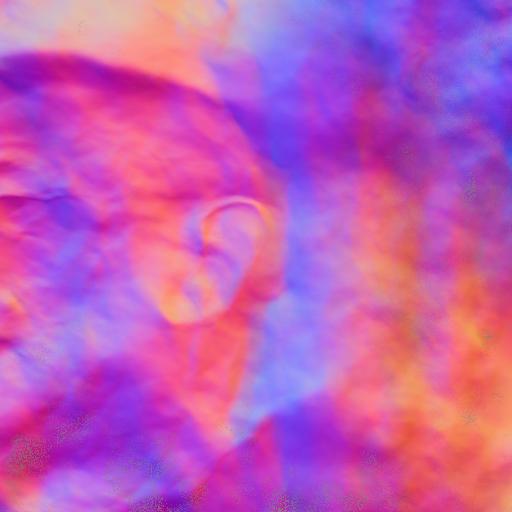}
    \includegraphics[width=0.09\textwidth]{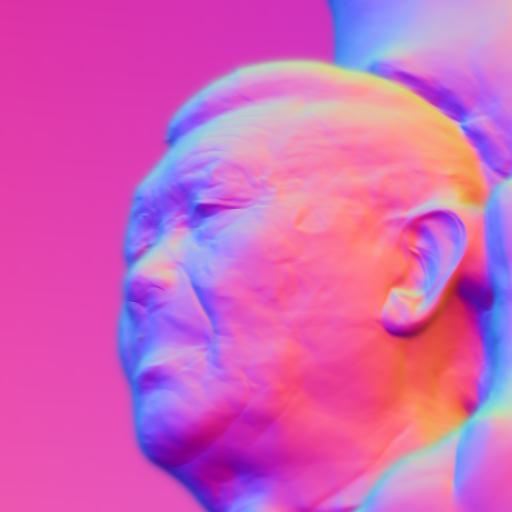}
    \includegraphics[width=0.09\textwidth]{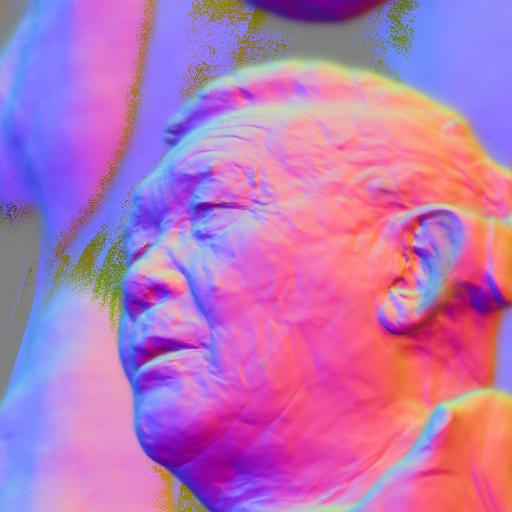}\\
    \makebox[0.09\textwidth]{GT}
    \makebox[0.09\textwidth]{NeuS}
    \makebox[0.09\textwidth]{HF-NeuS}
    \makebox[0.09\textwidth]{VolSDF}
    \makebox[0.09\textwidth]{Ours}
    \makebox[0.09\textwidth]{GT}
    \makebox[0.09\textwidth]{NeuS}
    \makebox[0.09\textwidth]{HF-NeuS}
    \makebox[0.09\textwidth]{VolSDF}
    \makebox[0.09\textwidth]{Ours}\\
    \caption{Comparison of various approaches under a 10-view setting (from Model 558 to Model 635).
    For each model, we show the results on one training view (left) and one novel view (right).}
    \label{fig:all_result5}
\end{figure*}

In the 20-view setting, VolSDF still failed on Model 571. We noticed that this model is affected by the failure reconstruction of the neck shown in Figure~\ref{fig:all_result5}, which leads to inaccurate cropping of the face.

\subsubsection{Comparison to NeuS \& HF-NeuS} 

We found that NeuS was successful in reconstructing all 3D human heads in our experiments. However, due to the lack of modeling high-frequency signals, it cannot recover fine details, resulting in Chamfer distances in their results that are 1 times larger than ours. In contrast, our method can reconstruct fine details such as wrinkles, scarves, and hair, thanks to the additional degree of freedom provided by the displacement field.

HF-NeuS extends NeuS by learning a displacement field for representing high-frequency details. It typically achieves the best performance with a sufficient number of views. However, as the number of views decreases, the 3D reconstruction quality often degrades significantly. The reason is that HF-NeuS learns both the base surface and the high-frequency details at the same time. Such a learning process is unstable under a low-view setting. With only 10 views, HF-NeuS failed to reconstruct geometry for 19 out of 30 subjects, while with 15 views, it failed on Models 376, 377, 383, 396, 435, 469, 487, 491, 548, and 566. Even with 20 views, HF-NeuS still failed to reconstruct six models, which are 487, 548, 608, 399, 413, and 397. These results confirm that learning high-frequency details from low-view inputs is a challenging task.

In contrast, our method tackles this challenge by adopting a geometry-decomposition and a two-stage training framework. The template is trained on multiple persons with randomly chosen views. Although the number of views for each person is still low, the randomly selected views complement each other and provide a complete head. This template provides a good initialization for training the displacement in Stage 2. 

Comparing to NeuS and HF-NeuS, our method performs consistently well in terms of geometry measure under the same low-view settings, thanks to the use of a pre-trained template and the displacement field. 

While both NeuS and HF-NeuS produce high-quality RGB images for training views, which are 0.5-1.5 dB higher than ours, their novel view synthesis results are consistently worse than ours with a 2-4 dB lower score. This is attributed to the fact that these methods have less accurate geometry reconstruction and do not incorporate multiple views from various identities.

\subsubsection{Summary}

Our approach is specifically designed to enhance the performance of 3D reconstruction in low-view settings and complements the existing methods, such as VolSDF, NeuS and HF-NeuS, by utilizing a pre-trained template and a two-stage training framework. We do not intend to replace these methods, but rather to \textbf{improve their performance in such scenarios}.

\begin{figure*}[htbp]
    \centering
    \rotatebox{90}{\textbf{395}}
    \includegraphics[width=0.09\textwidth]{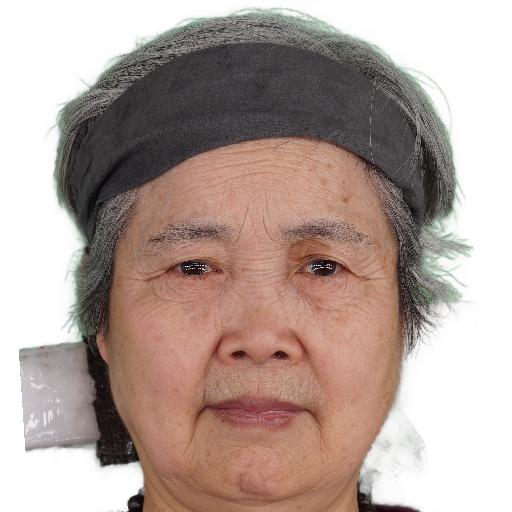}
    \includegraphics[width=0.09\textwidth]{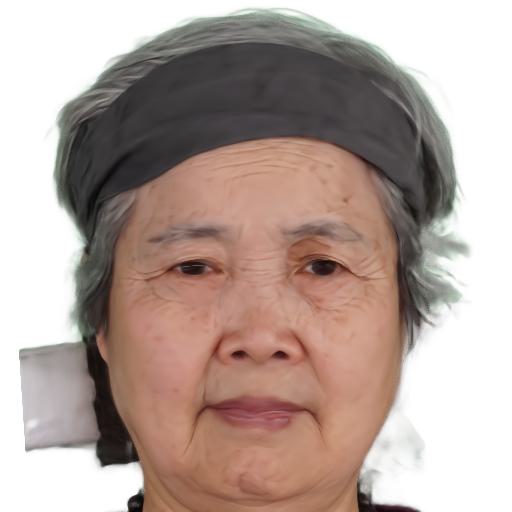}
    \includegraphics[width=0.09\textwidth]{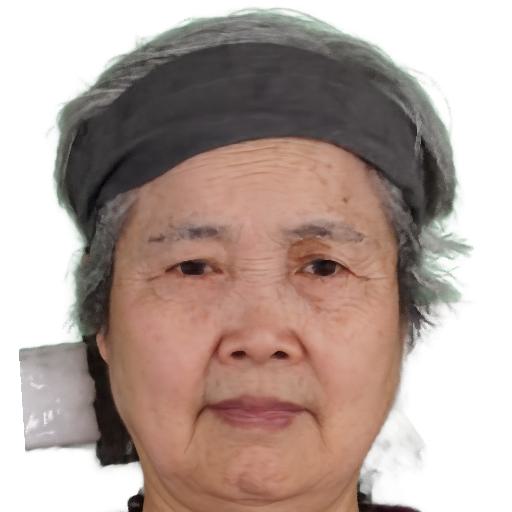}
    \includegraphics[width=0.09\textwidth]{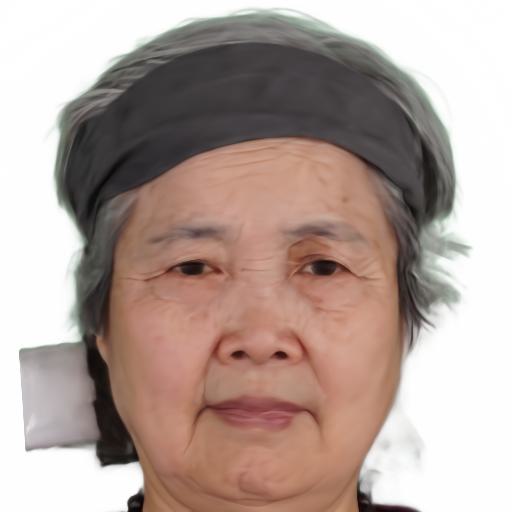}
    \includegraphics[width=0.09\textwidth]{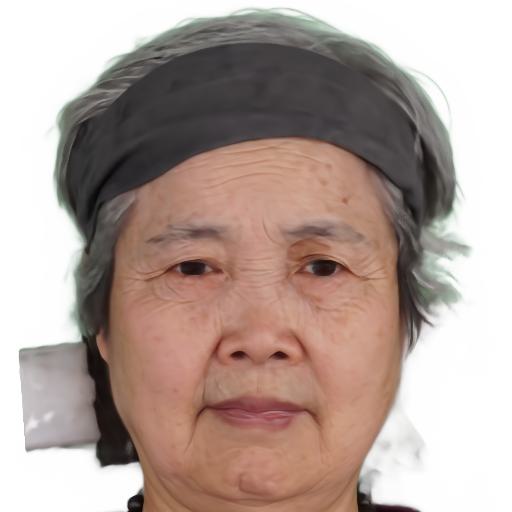}
    \includegraphics[width=0.09\textwidth]{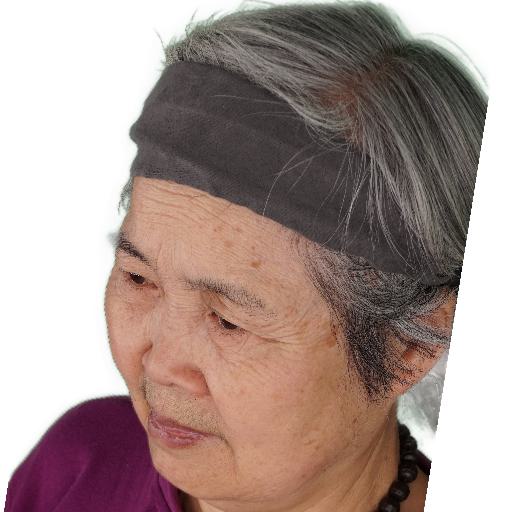}
    \includegraphics[width=0.09\textwidth]{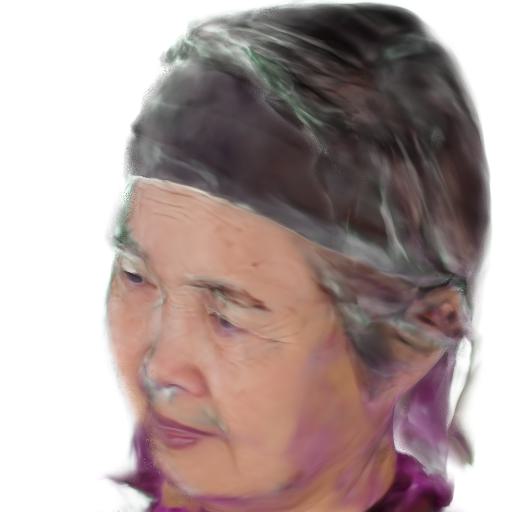}
    \includegraphics[width=0.09\textwidth]{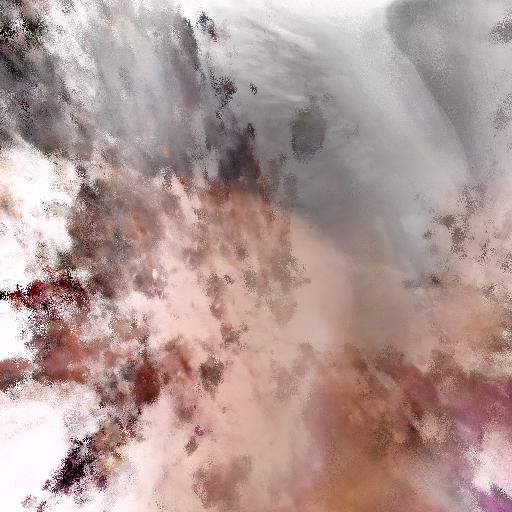}
    \includegraphics[width=0.09\textwidth]{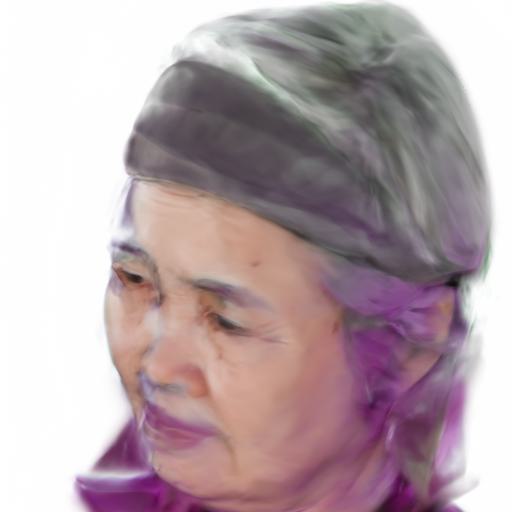}
    \includegraphics[width=0.09\textwidth]{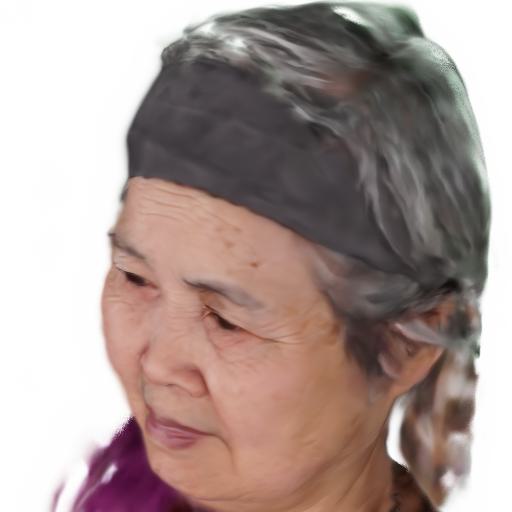}\\
    % \rotatebox{90}{\tiny}
    \includegraphics[width=0.09\textwidth]{./results/template_effects/gt_571_blank.jpg}
    \includegraphics[width=0.09\textwidth]{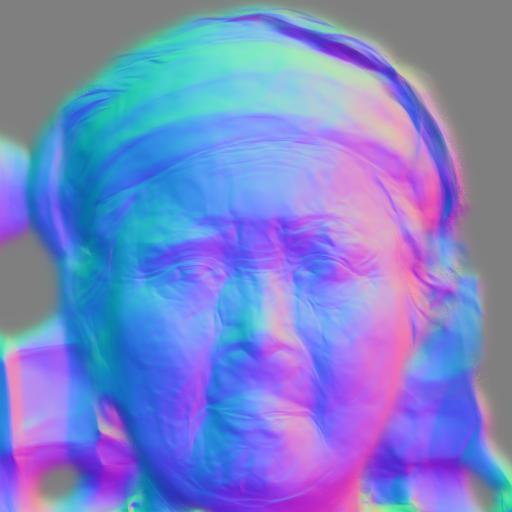}
    \includegraphics[width=0.09\textwidth]{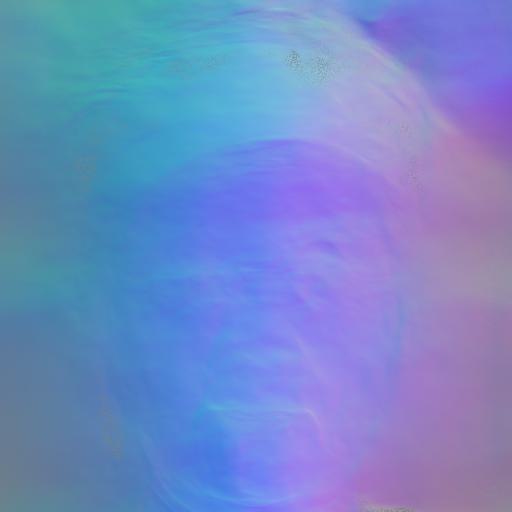}
    \includegraphics[width=0.09\textwidth]{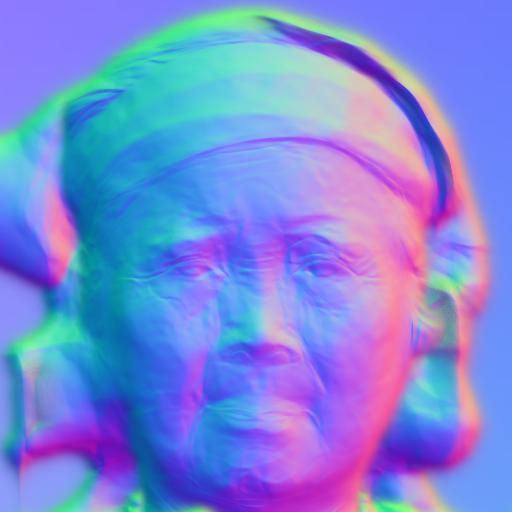}
    \includegraphics[width=0.09\textwidth]{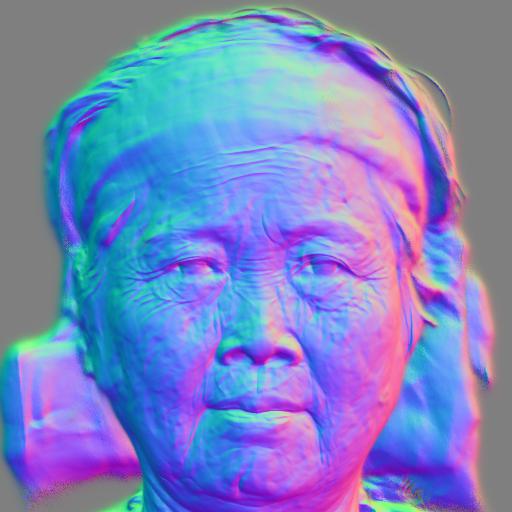}
    \includegraphics[width=0.09\textwidth]{./results/template_effects/gt_571_blank.jpg}
    \includegraphics[width=0.09\textwidth]{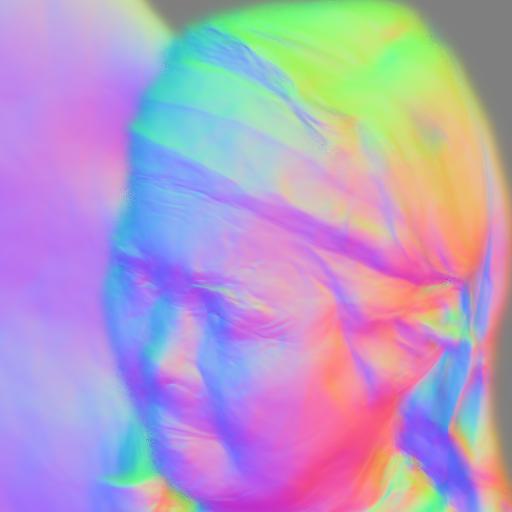}
    \includegraphics[width=0.09\textwidth]{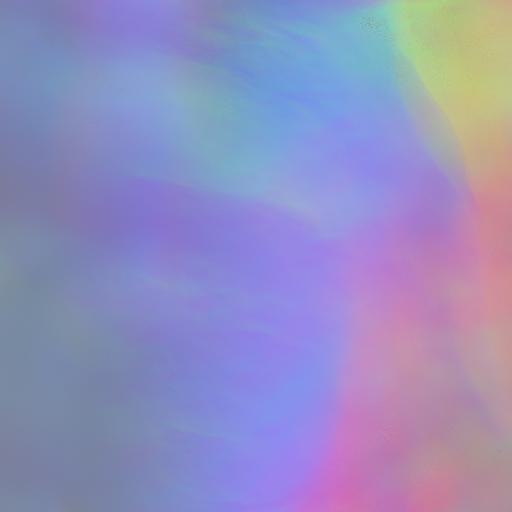}
    \includegraphics[width=0.09\textwidth]{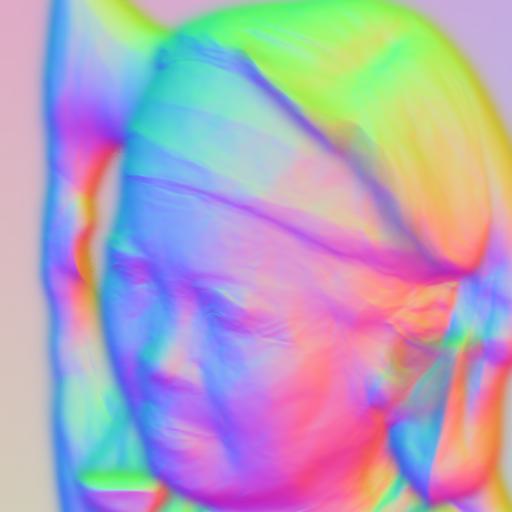}
    \includegraphics[width=0.09\textwidth]{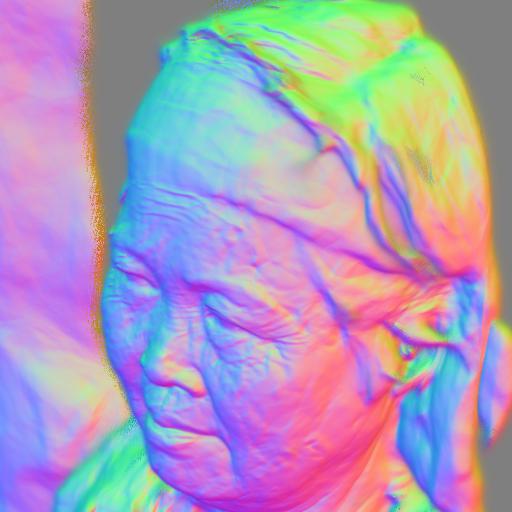}\\
    \rotatebox{90}{\textbf{396}}
    \includegraphics[width=0.09\textwidth]{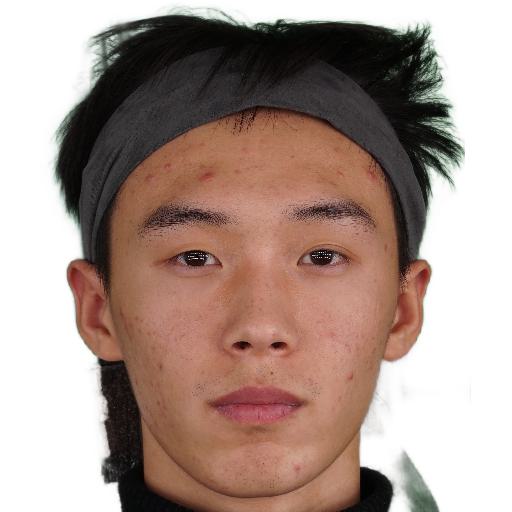}
    \includegraphics[width=0.09\textwidth]{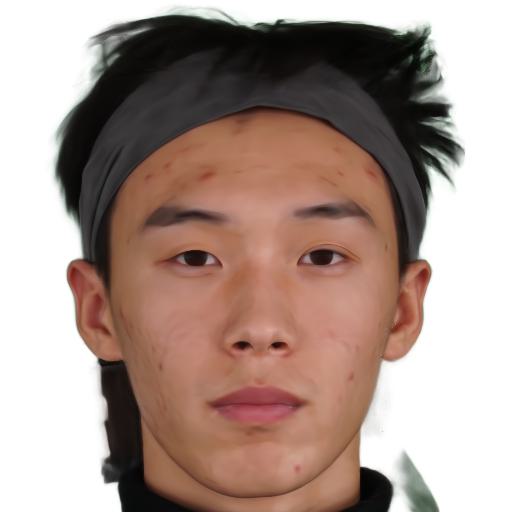}
    \includegraphics[width=0.09\textwidth]{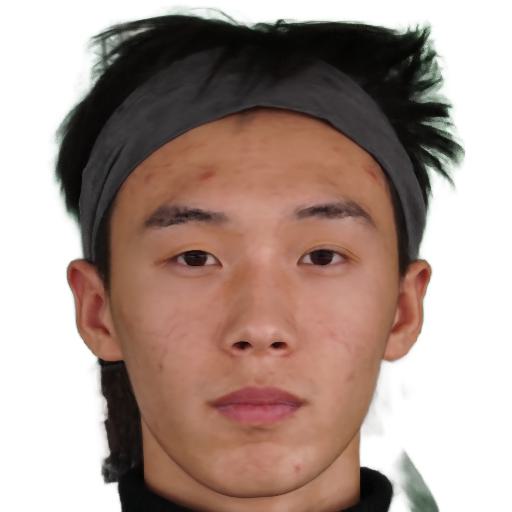}
    \includegraphics[width=0.09\textwidth]{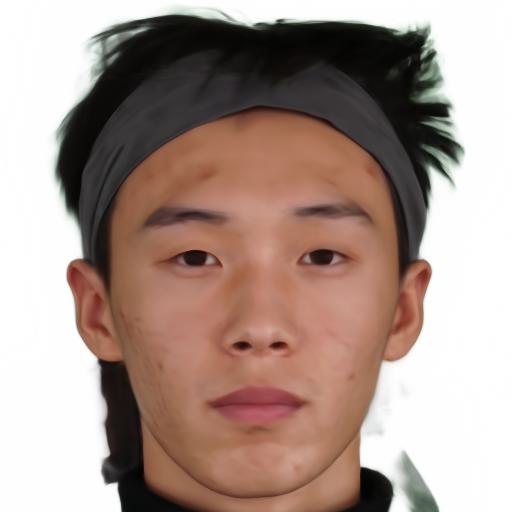}
    \includegraphics[width=0.09\textwidth]{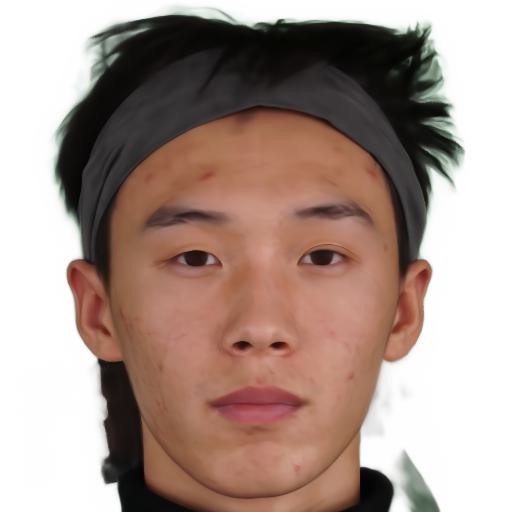}
    \includegraphics[width=0.09\textwidth]{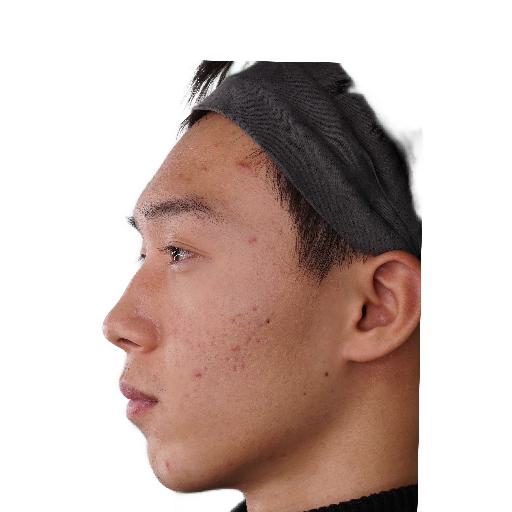}
    \includegraphics[width=0.09\textwidth]{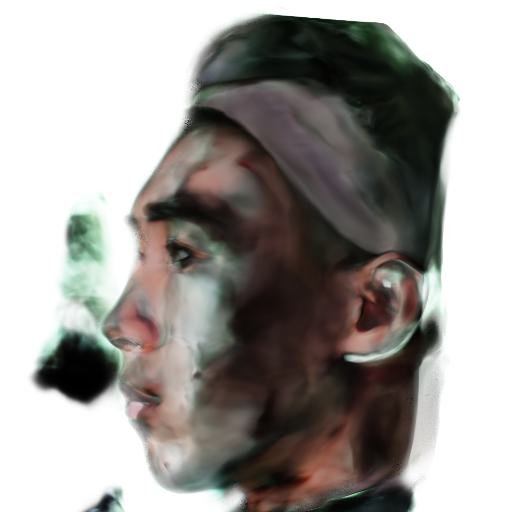}
    \includegraphics[width=0.09\textwidth]{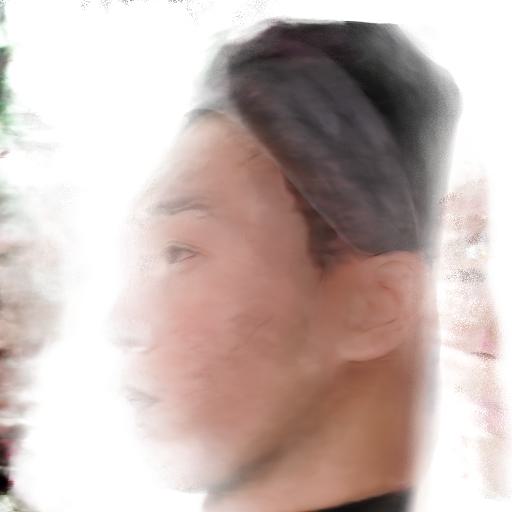}
    \includegraphics[width=0.09\textwidth]{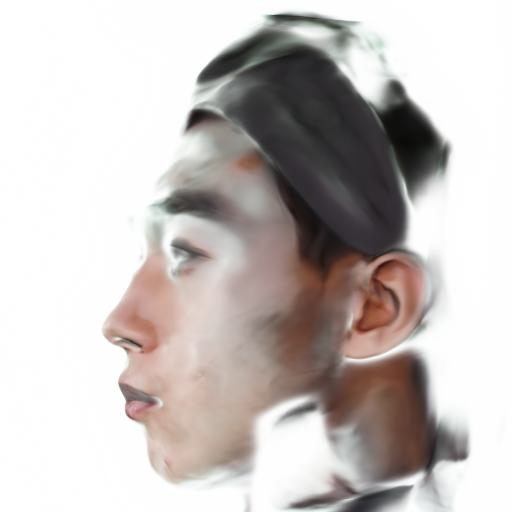}
    \includegraphics[width=0.09\textwidth]{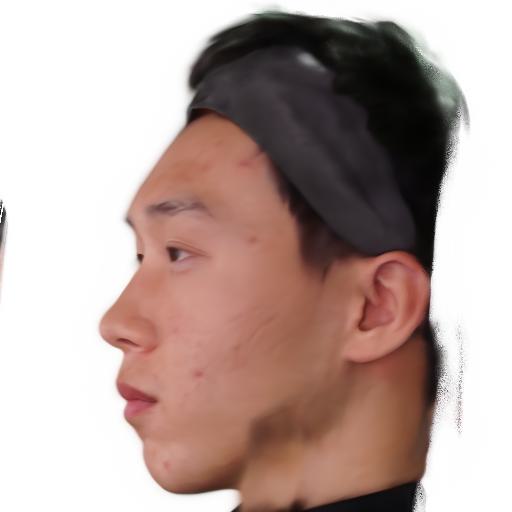}
    \rotatebox{90}{\tiny}
    \includegraphics[width=0.09\textwidth]{./results/template_effects/gt_571_blank.jpg}
    \includegraphics[width=0.09\textwidth]{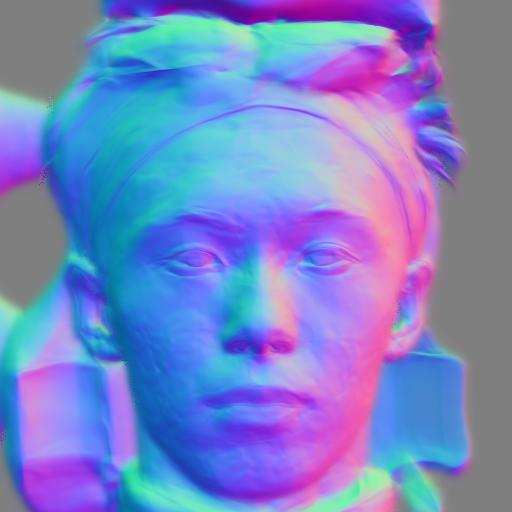}
    \includegraphics[width=0.09\textwidth]{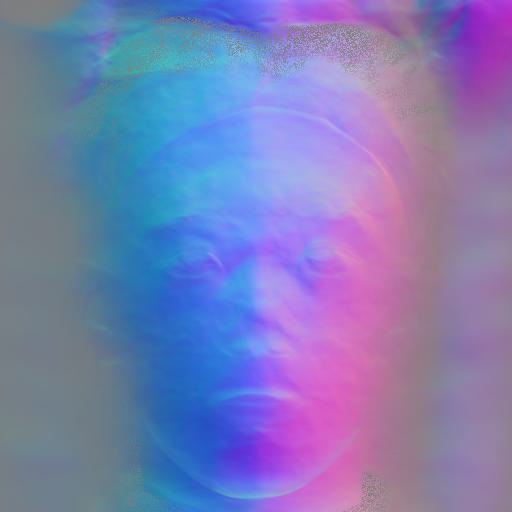}
    \includegraphics[width=0.09\textwidth]{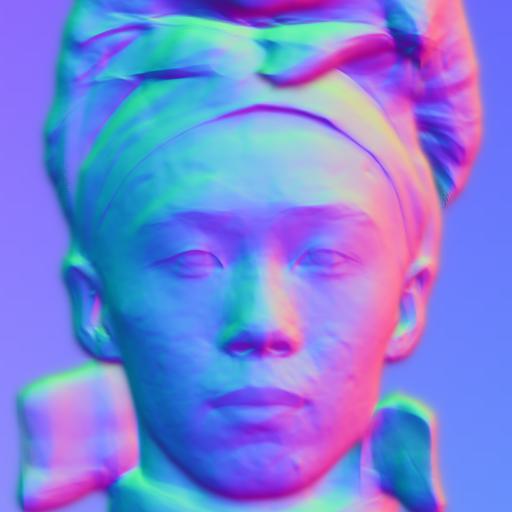}
    \includegraphics[width=0.09\textwidth]{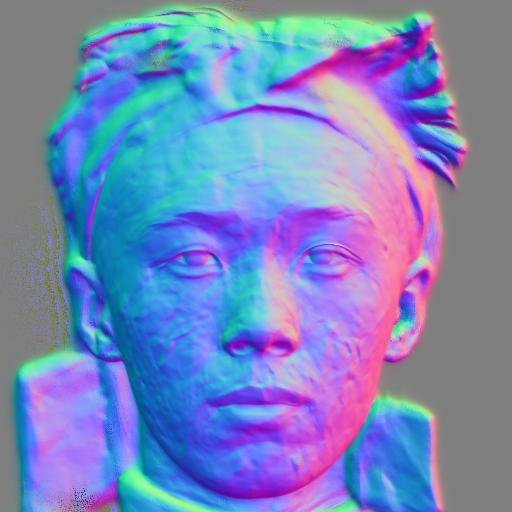}
    \includegraphics[width=0.09\textwidth]{./results/template_effects/gt_571_blank.jpg}
    \includegraphics[width=0.09\textwidth]{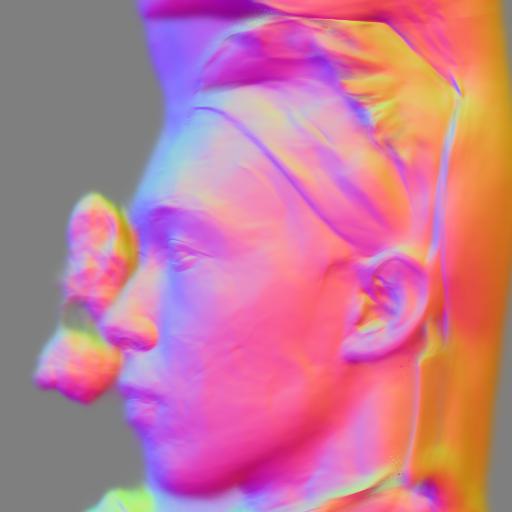}
    \includegraphics[width=0.09\textwidth]{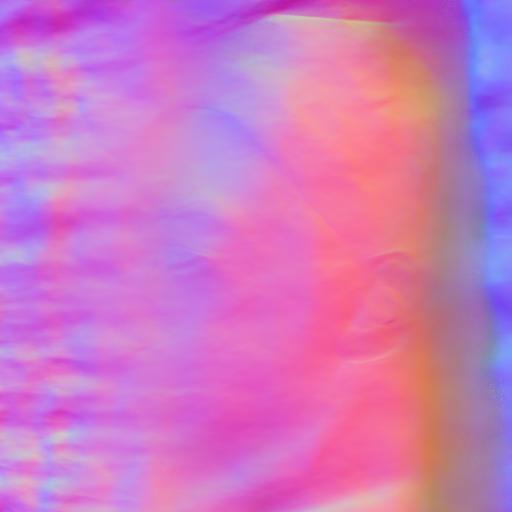}
    \includegraphics[width=0.09\textwidth]{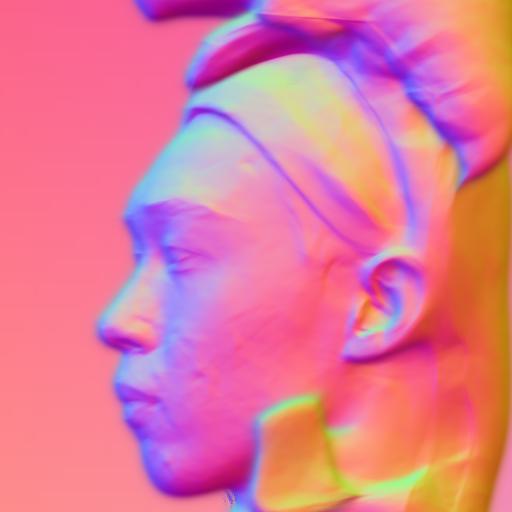}
    \includegraphics[width=0.09\textwidth]{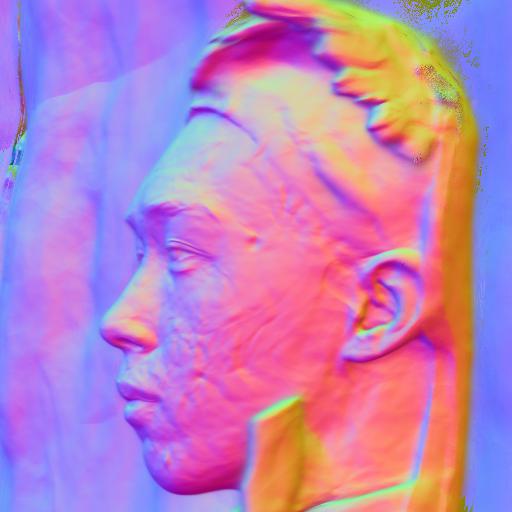}\\
    \rotatebox{90}{\textbf{397}}
    \includegraphics[width=0.09\textwidth]{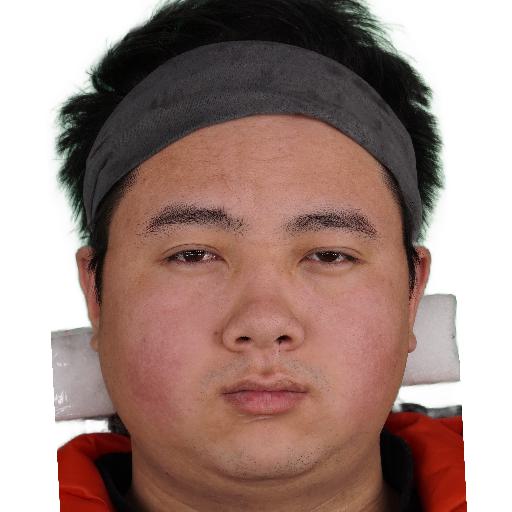}
    \includegraphics[width=0.09\textwidth]{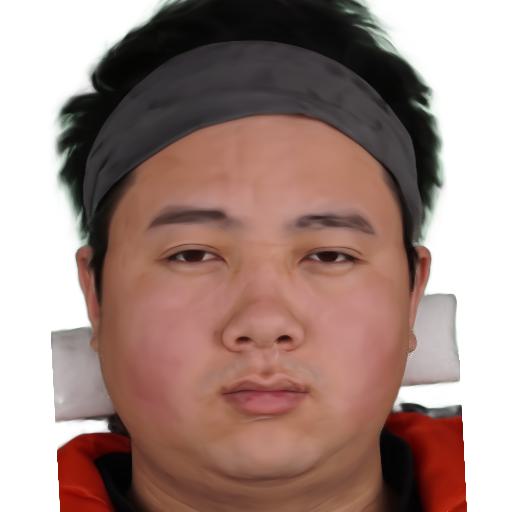}
    \includegraphics[width=0.09\textwidth]{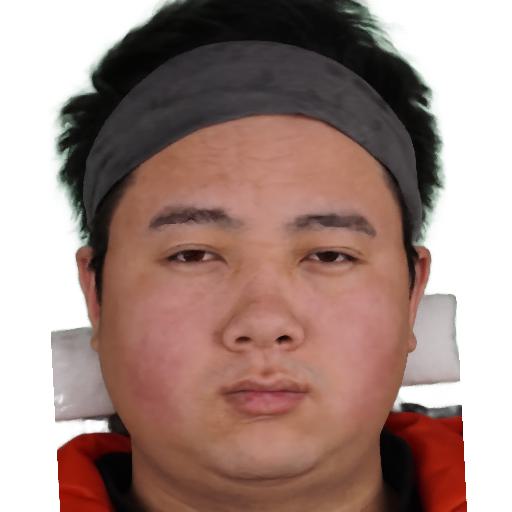}
    \includegraphics[width=0.09\textwidth]{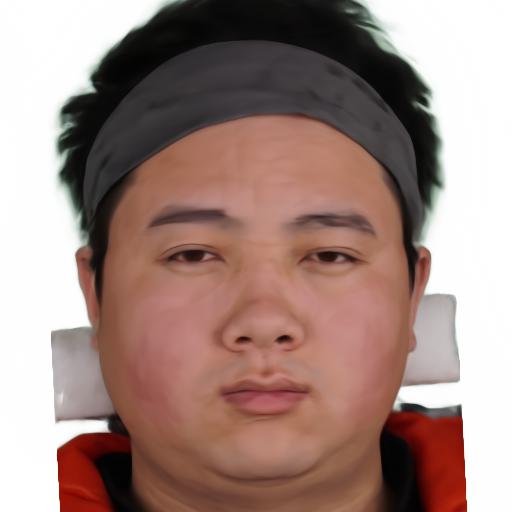}
    \includegraphics[width=0.09\textwidth]{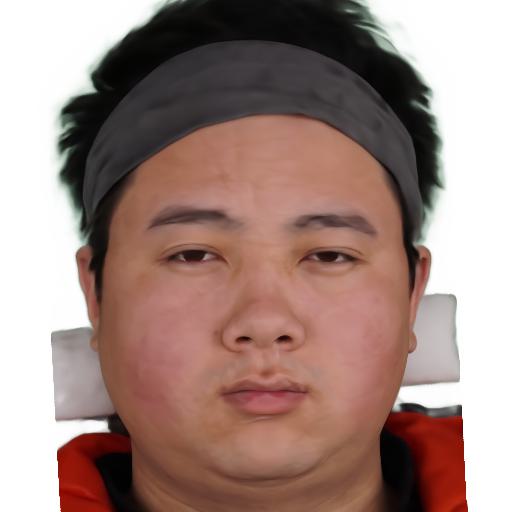}
    \includegraphics[width=0.09\textwidth]{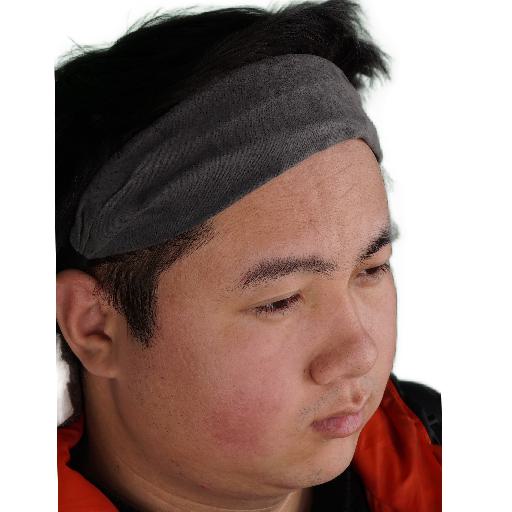}
    \includegraphics[width=0.09\textwidth]{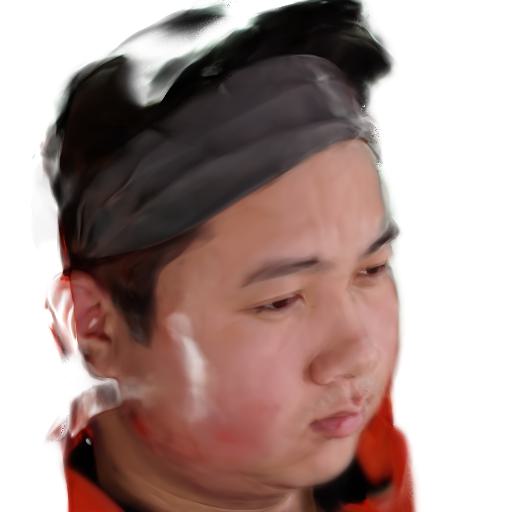}
    \includegraphics[width=0.09\textwidth]{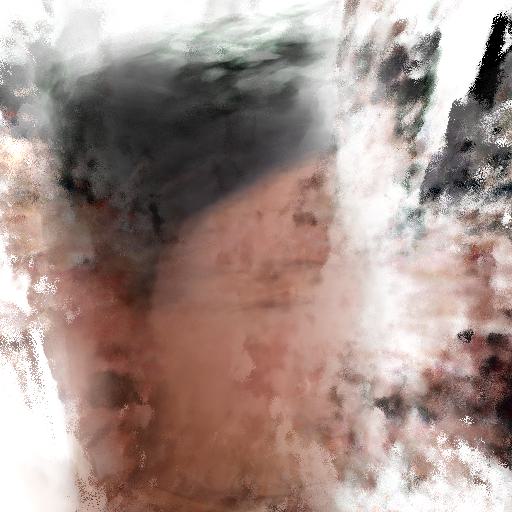}
    \includegraphics[width=0.09\textwidth]{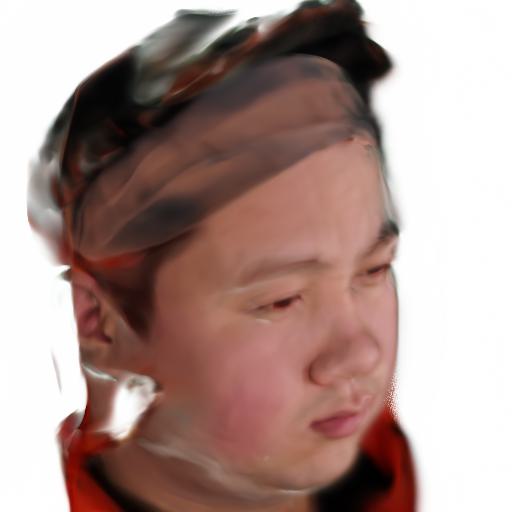}
    \includegraphics[width=0.09\textwidth]{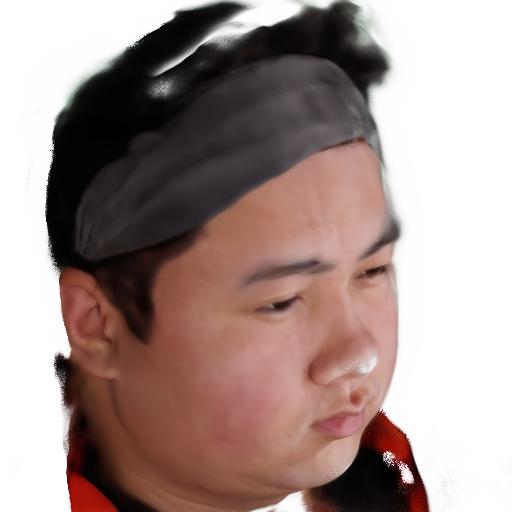}
    \rotatebox{90}{\tiny}
    \includegraphics[width=0.09\textwidth]{./results/template_effects/gt_571_blank.jpg}
    \includegraphics[width=0.09\textwidth]{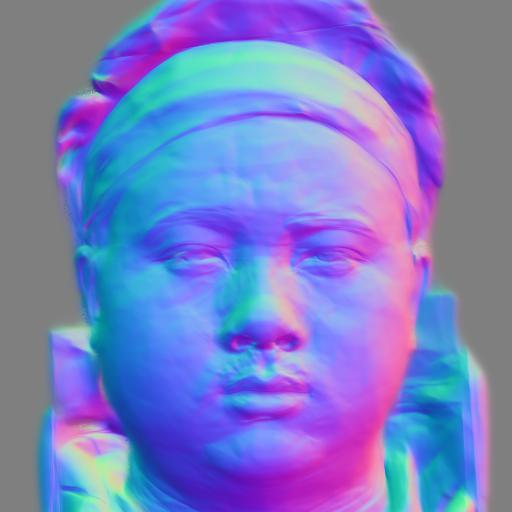}
    \includegraphics[width=0.09\textwidth]{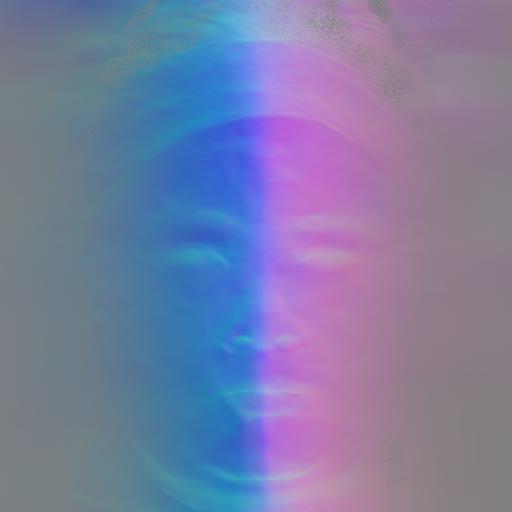}
    \includegraphics[width=0.09\textwidth]{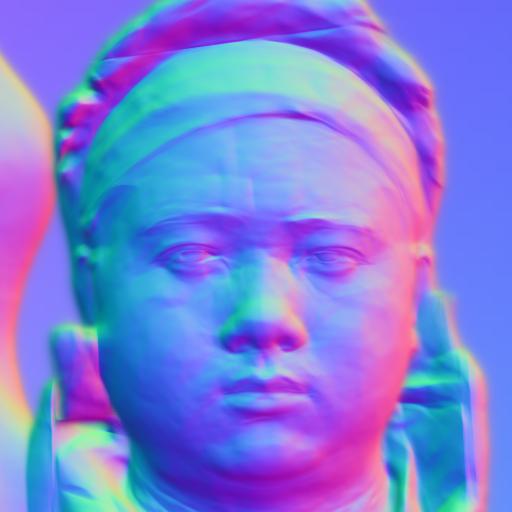}
    \includegraphics[width=0.09\textwidth]{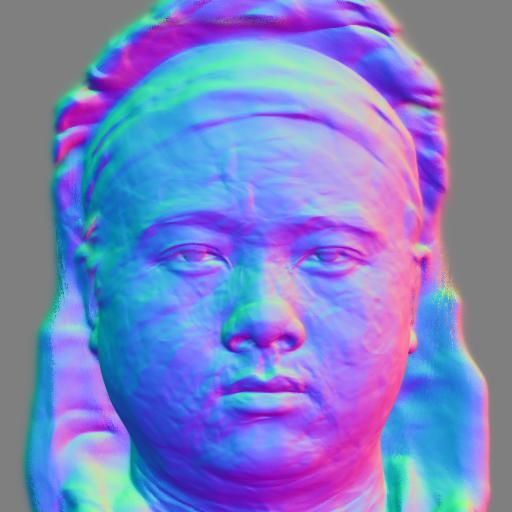}
    \includegraphics[width=0.09\textwidth]{./results/template_effects/gt_571_blank.jpg}
    \includegraphics[width=0.09\textwidth]{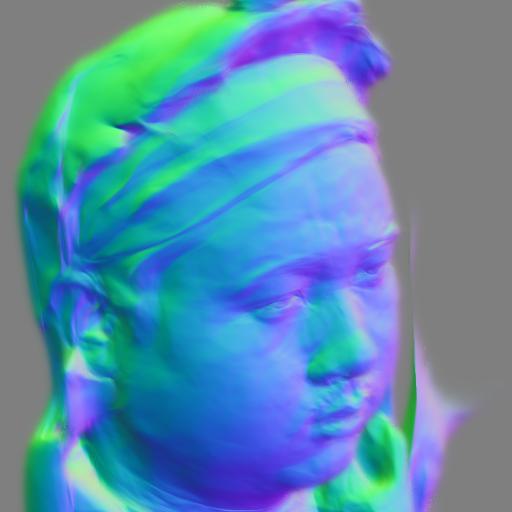}
    \includegraphics[width=0.09\textwidth]{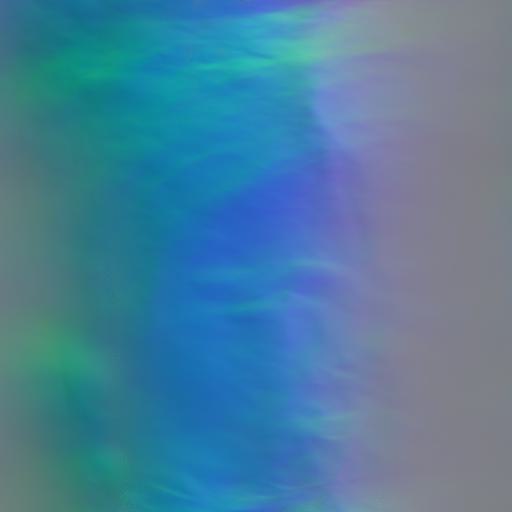}
    \includegraphics[width=0.09\textwidth]{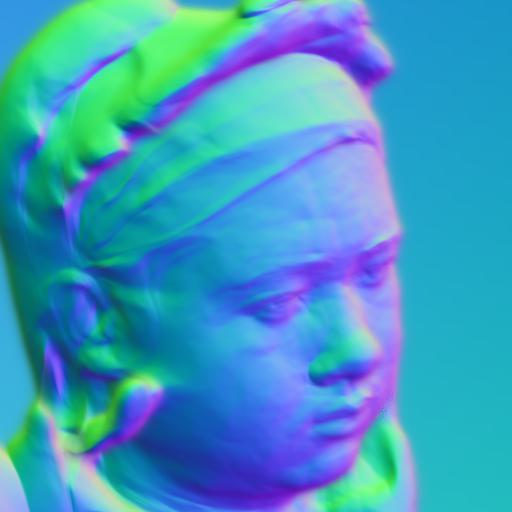}
    \includegraphics[width=0.09\textwidth]{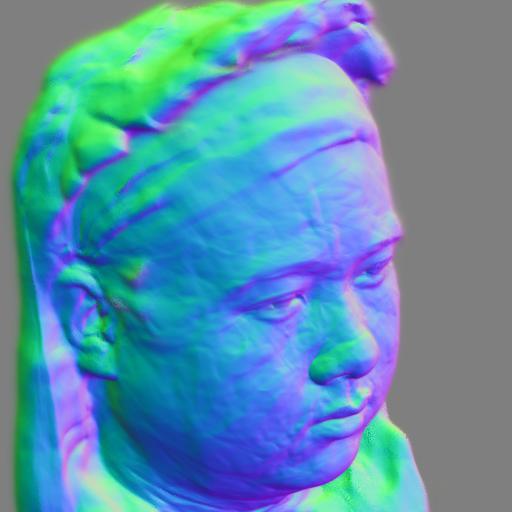}\\
    \rotatebox{90}{\textbf{399}}
    \includegraphics[width=0.09\textwidth]{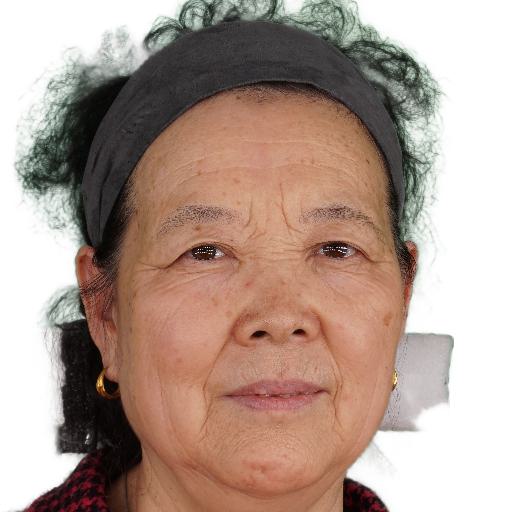}
    \includegraphics[width=0.09\textwidth]{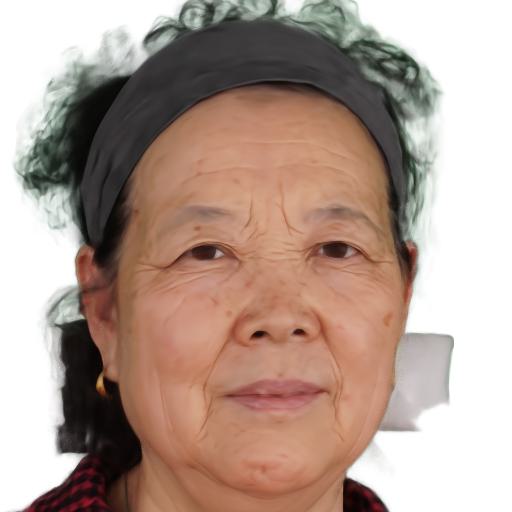}
    \includegraphics[width=0.09\textwidth]{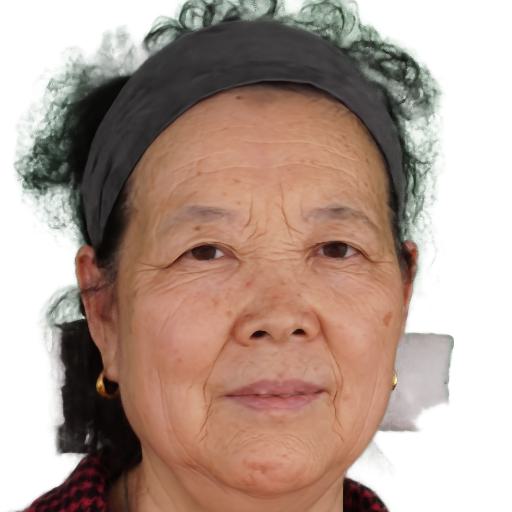}
    \includegraphics[width=0.09\textwidth]{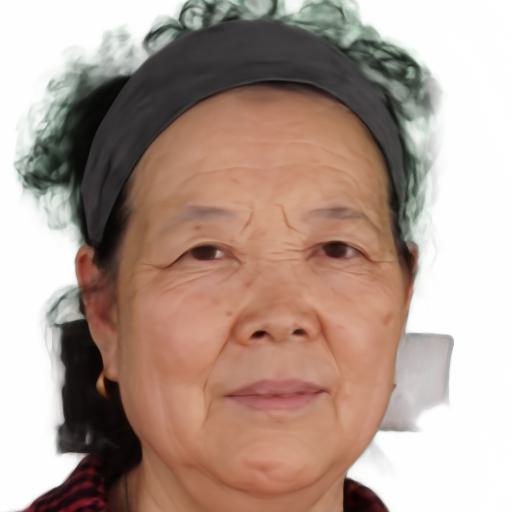}
    \includegraphics[width=0.09\textwidth]{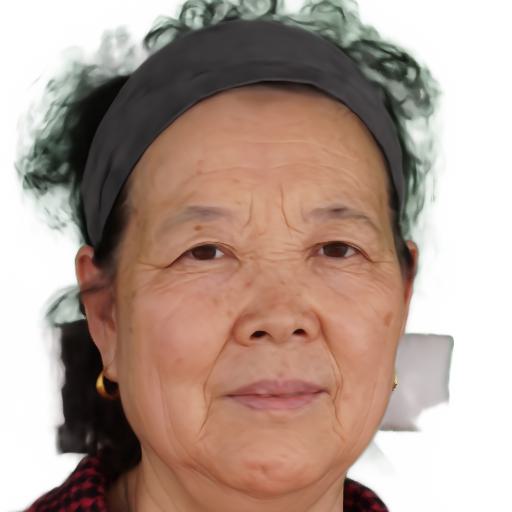}
    \includegraphics[width=0.09\textwidth]{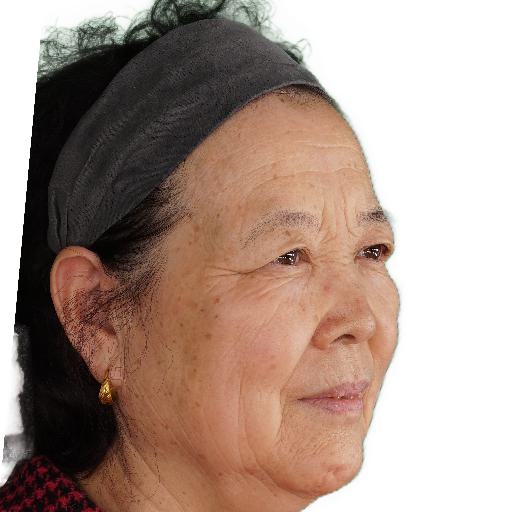}
    \includegraphics[width=0.09\textwidth]{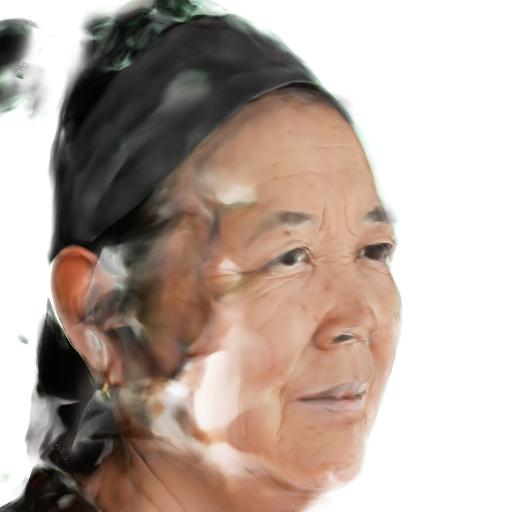}
    \includegraphics[width=0.09\textwidth]{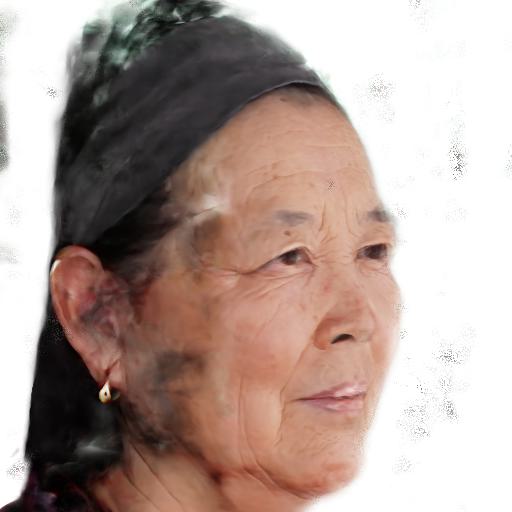}
    \includegraphics[width=0.09\textwidth]{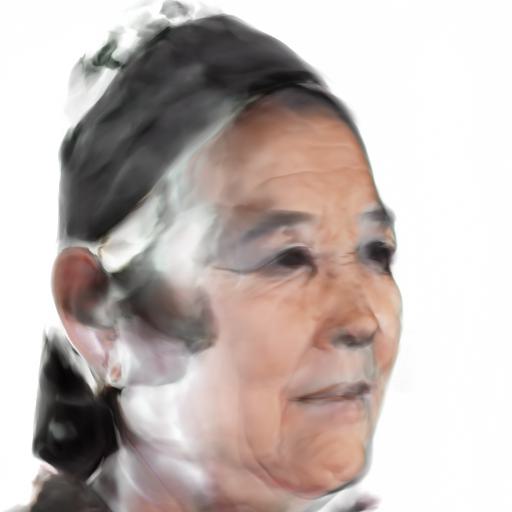}
    \includegraphics[width=0.09\textwidth]{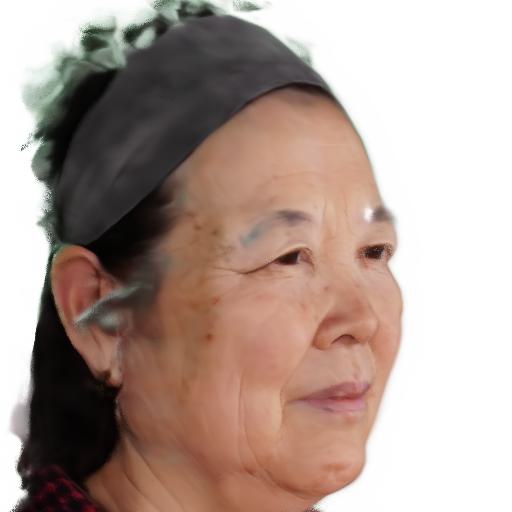}
    \rotatebox{90}{\tiny}
    \includegraphics[width=0.09\textwidth]{./results/template_effects/gt_571_blank.jpg}
    \includegraphics[width=0.09\textwidth]{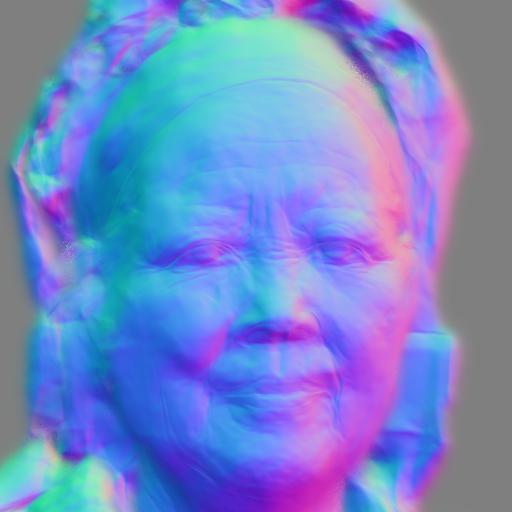}
    \includegraphics[width=0.09\textwidth]{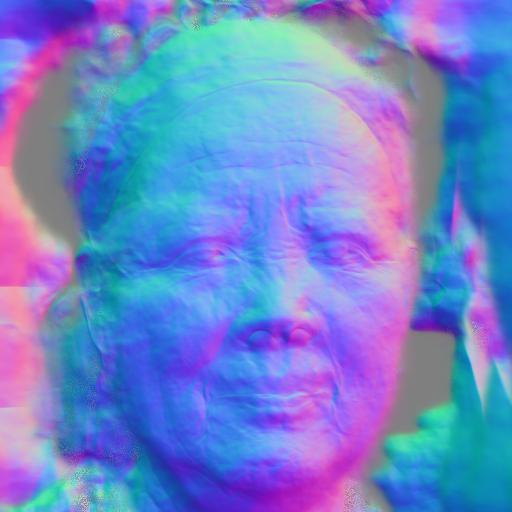}
    \includegraphics[width=0.09\textwidth]{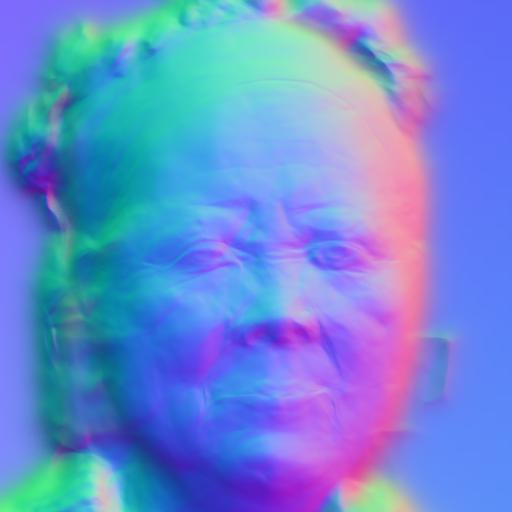}
    \includegraphics[width=0.09\textwidth]{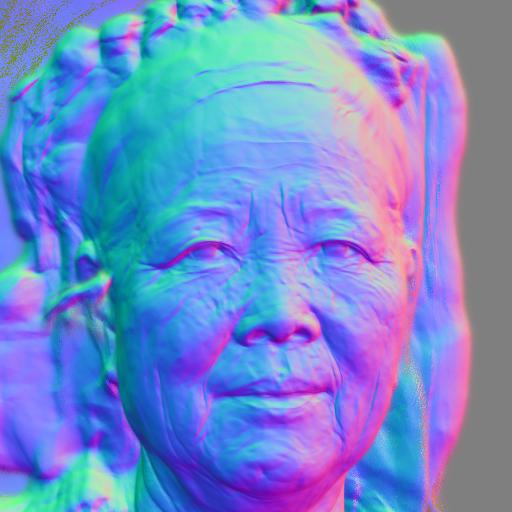}
    \includegraphics[width=0.09\textwidth]{./results/template_effects/gt_571_blank.jpg}
    \includegraphics[width=0.09\textwidth]{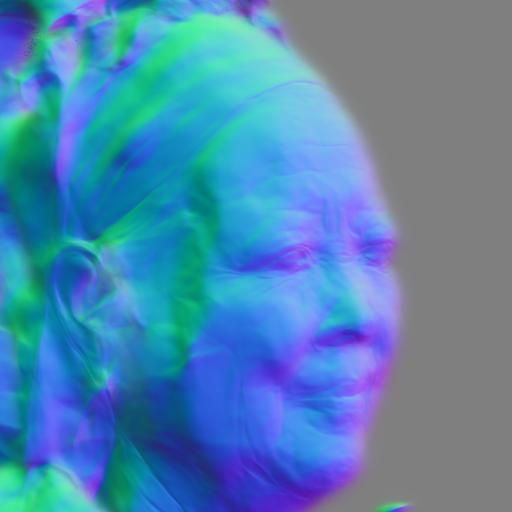}
    \includegraphics[width=0.09\textwidth]{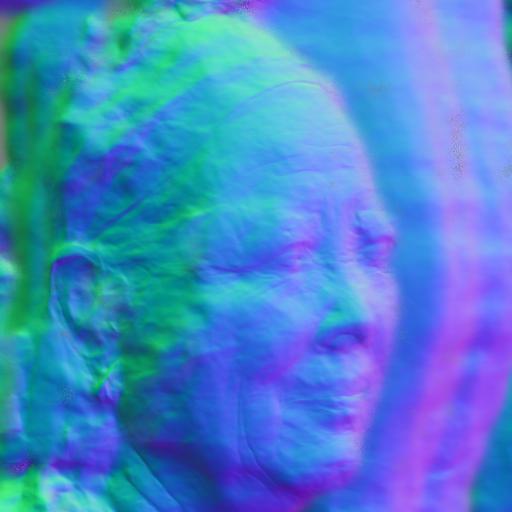}
    \includegraphics[width=0.09\textwidth]{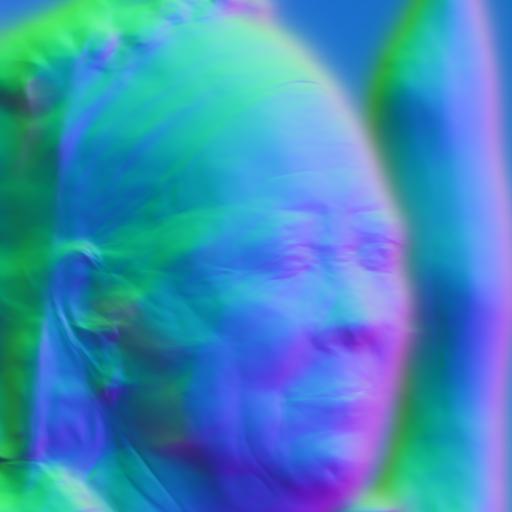}
    \includegraphics[width=0.09\textwidth]{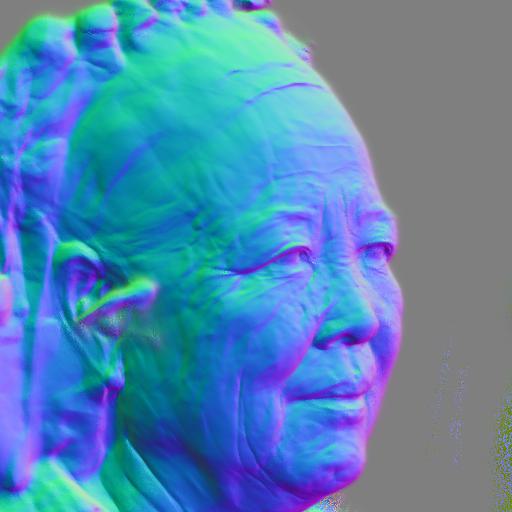}\\
    \rotatebox{90}{\textbf{401}}
    \includegraphics[width=0.09\textwidth]{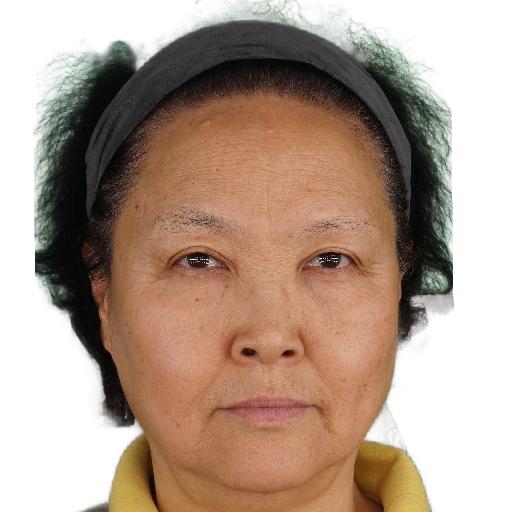}
    \includegraphics[width=0.09\textwidth]{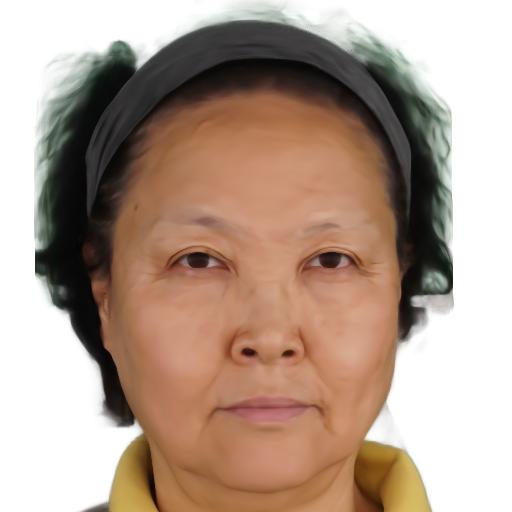}
    \includegraphics[width=0.09\textwidth]{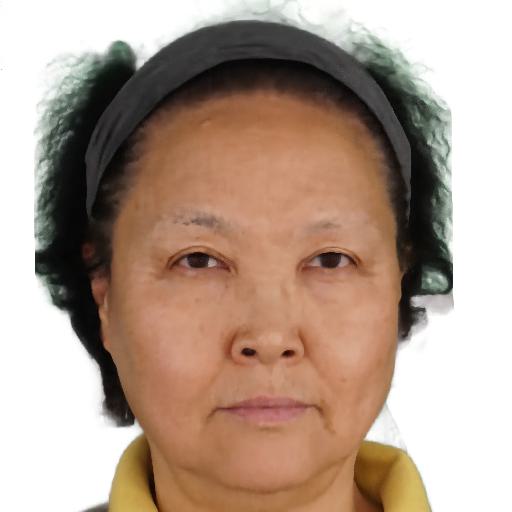}
    \includegraphics[width=0.09\textwidth]{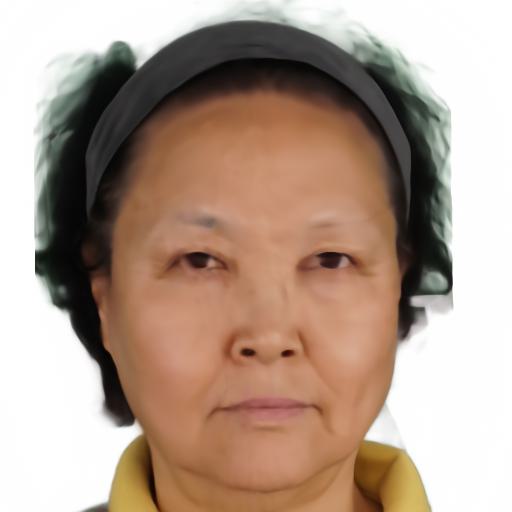}
    \includegraphics[width=0.09\textwidth]{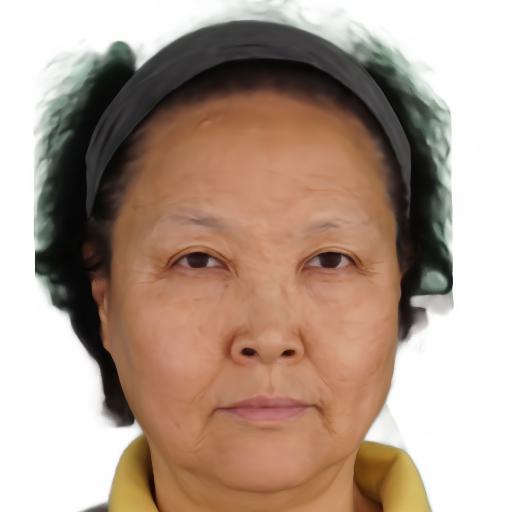}
    \includegraphics[width=0.09\textwidth]{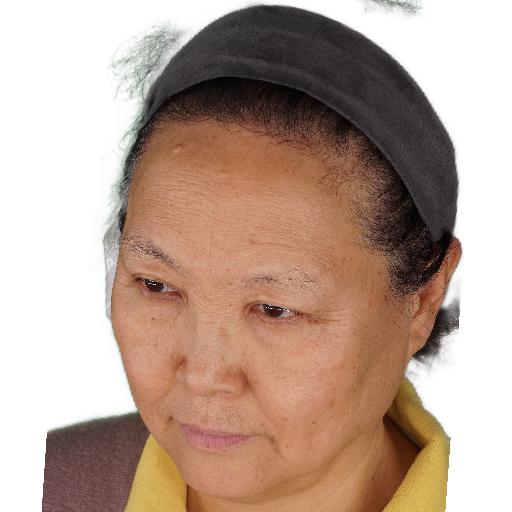}
    \includegraphics[width=0.09\textwidth]{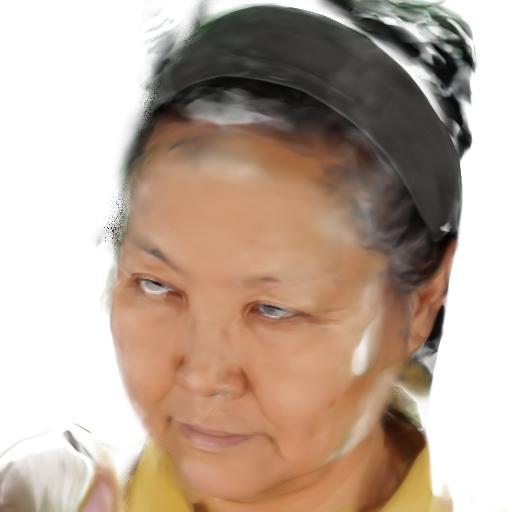}
    \includegraphics[width=0.09\textwidth]{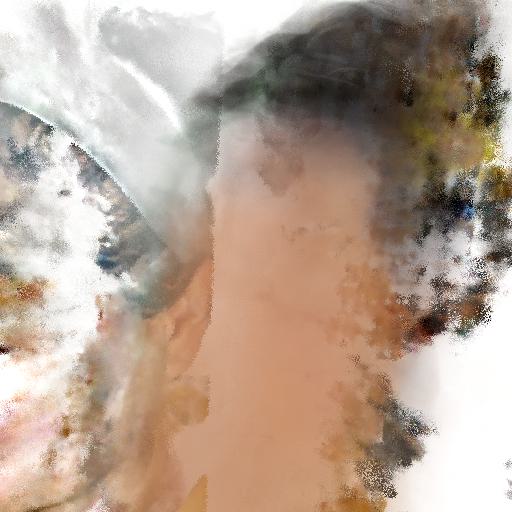}
    \includegraphics[width=0.09\textwidth]{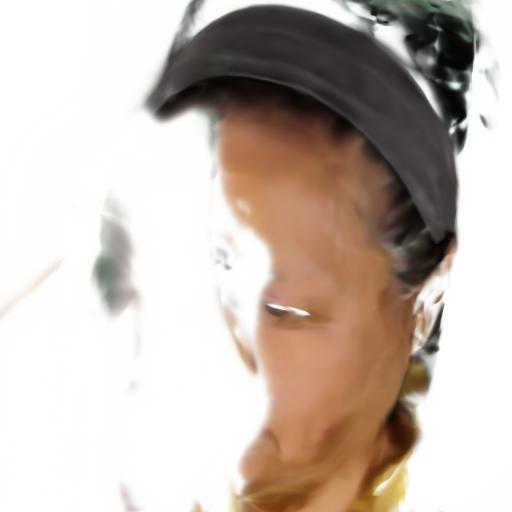}
    \includegraphics[width=0.09\textwidth]{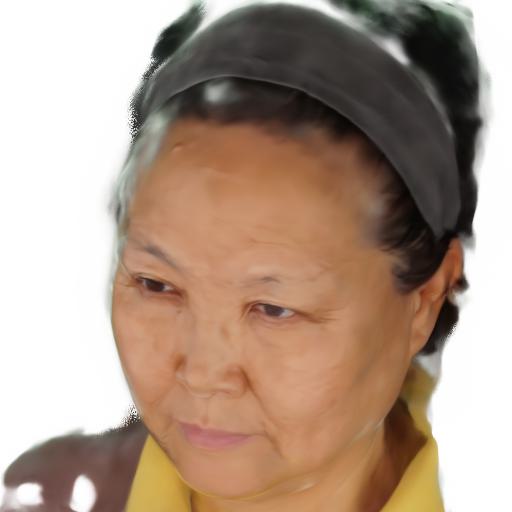}
    \rotatebox{90}{\tiny}
    \includegraphics[width=0.09\textwidth]{./results/template_effects/gt_571_blank.jpg}
    \includegraphics[width=0.09\textwidth]{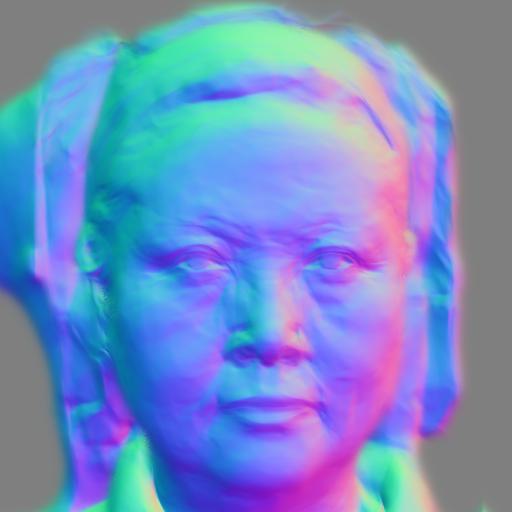}
    \includegraphics[width=0.09\textwidth]{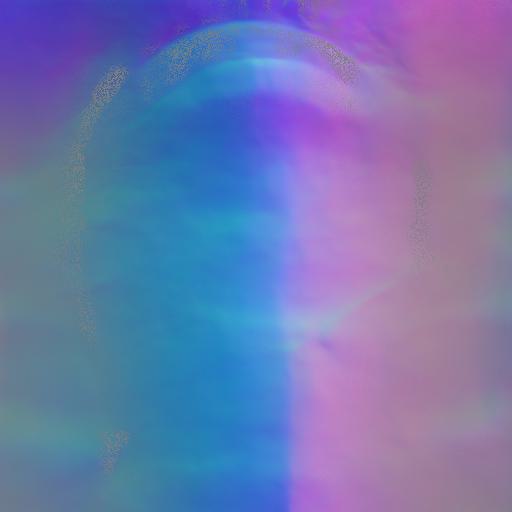}
    \includegraphics[width=0.09\textwidth]{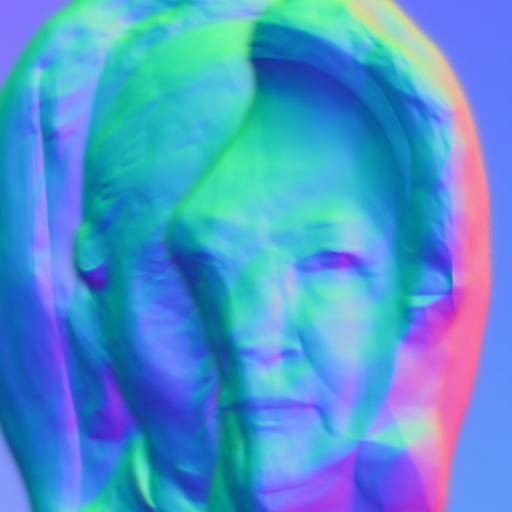}
    \includegraphics[width=0.09\textwidth]{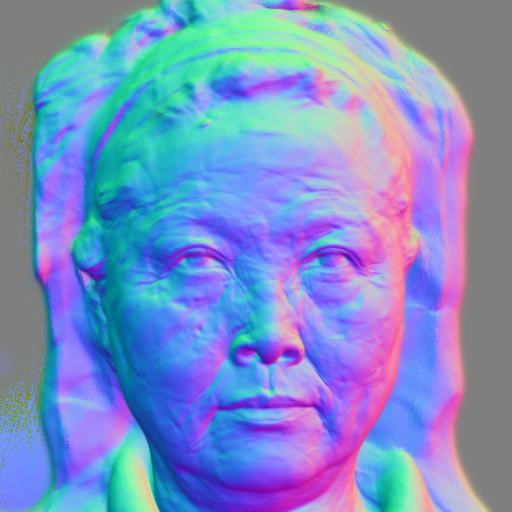}
    \includegraphics[width=0.09\textwidth]{./results/template_effects/gt_571_blank.jpg}
    \includegraphics[width=0.09\textwidth]{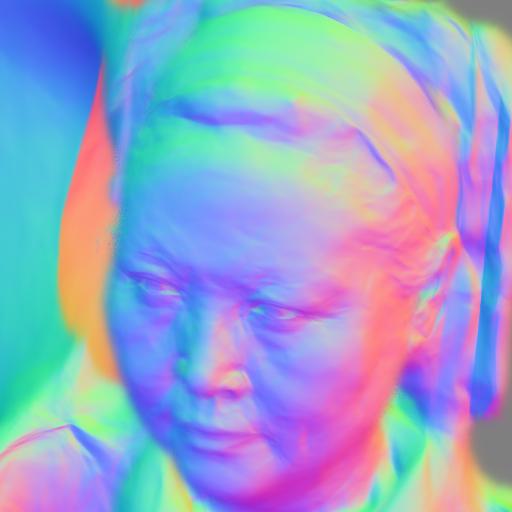}
    \includegraphics[width=0.09\textwidth]{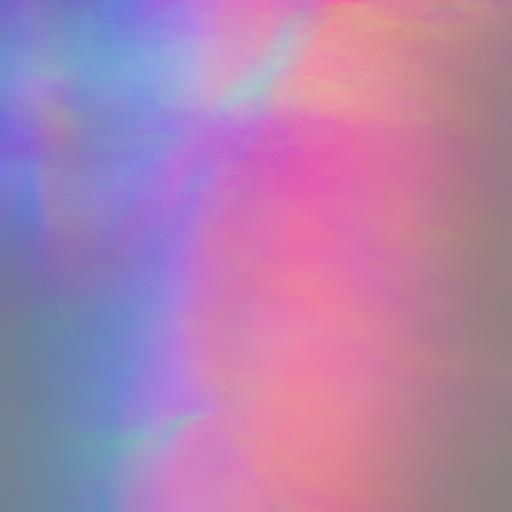}
    \includegraphics[width=0.09\textwidth]{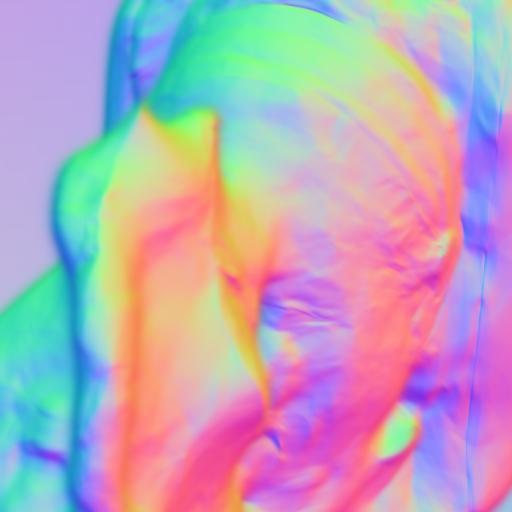}
    \includegraphics[width=0.09\textwidth]{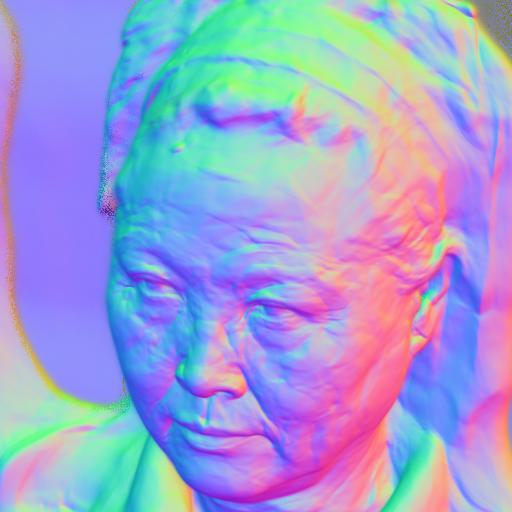}\\
    \rotatebox{90}{\textbf{413}}
    \includegraphics[width=0.09\textwidth]{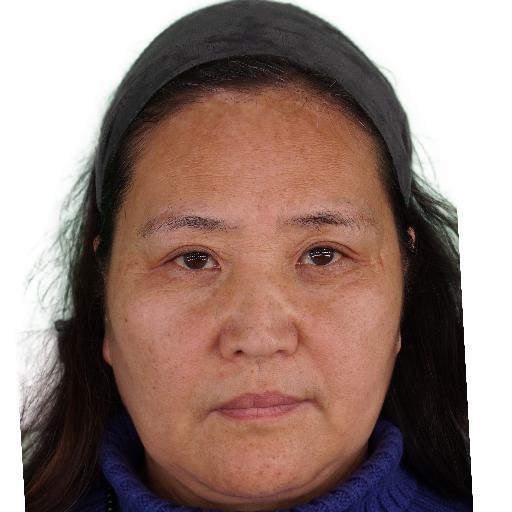}
    \includegraphics[width=0.09\textwidth]{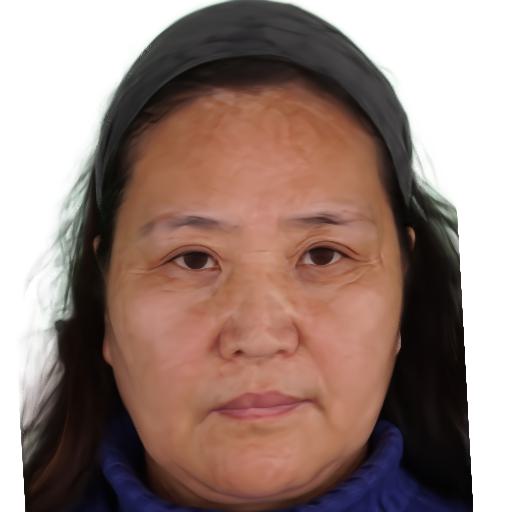}
    \includegraphics[width=0.09\textwidth]{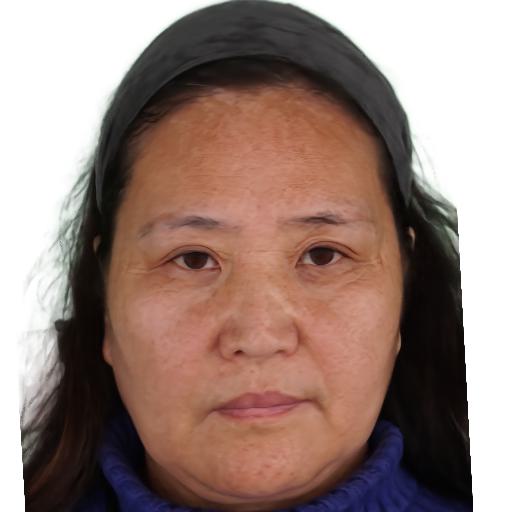}
    \includegraphics[width=0.09\textwidth]{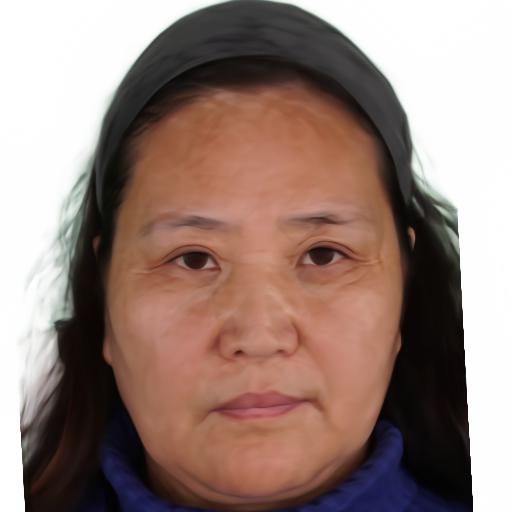}
    \includegraphics[width=0.09\textwidth]{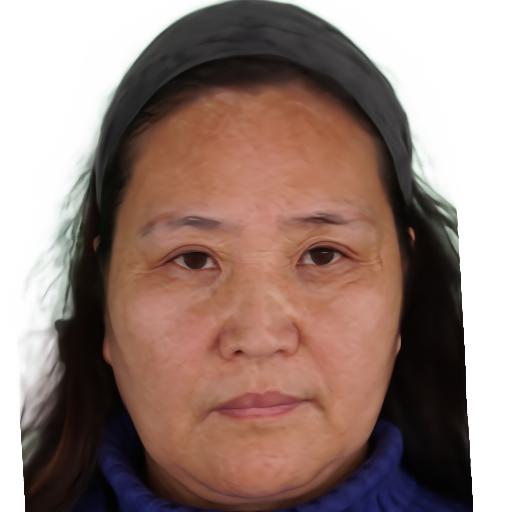}
    \includegraphics[width=0.09\textwidth]{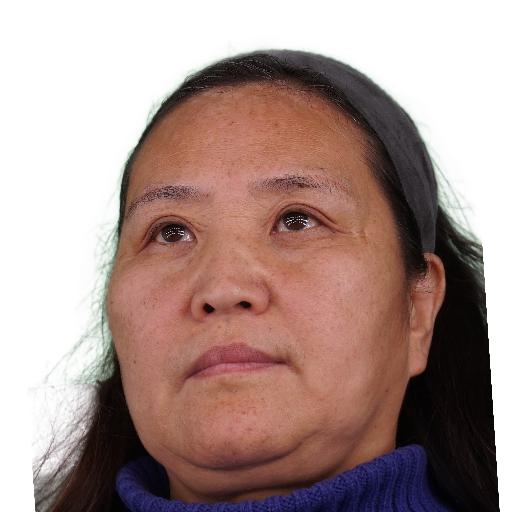}
    \includegraphics[width=0.09\textwidth]{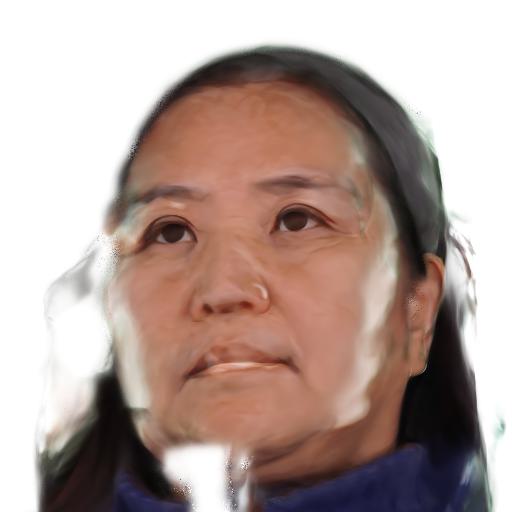}
    \includegraphics[width=0.09\textwidth]{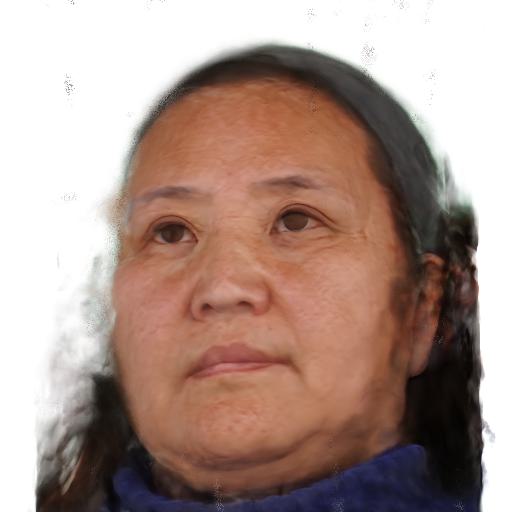}
    \includegraphics[width=0.09\textwidth]{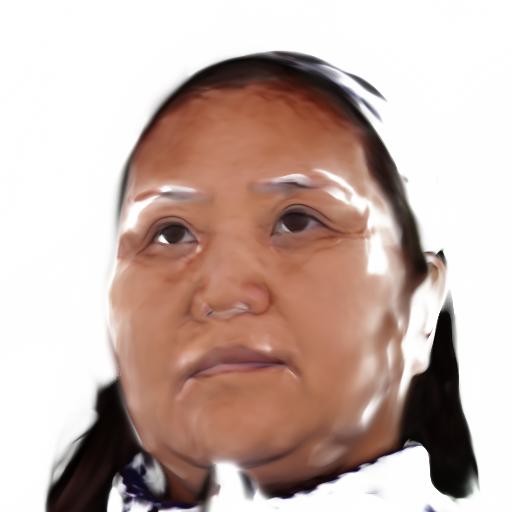}
    \includegraphics[width=0.09\textwidth]{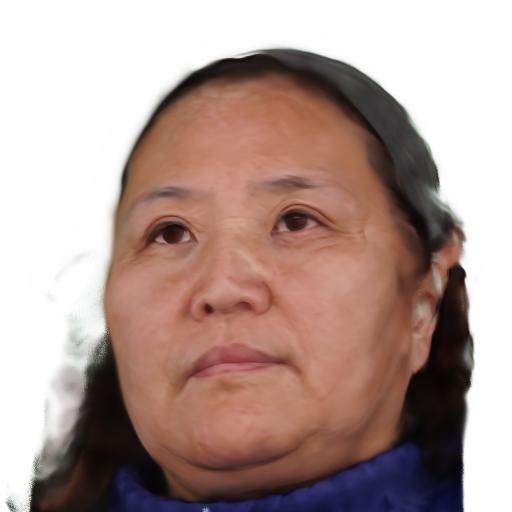}
    \rotatebox{90}{\tiny}
    \includegraphics[width=0.09\textwidth]{./results/template_effects/gt_571_blank.jpg}
    \includegraphics[width=0.09\textwidth]{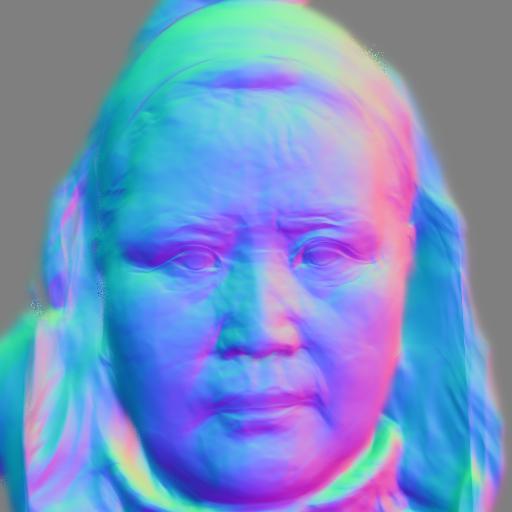}
    \includegraphics[width=0.09\textwidth]{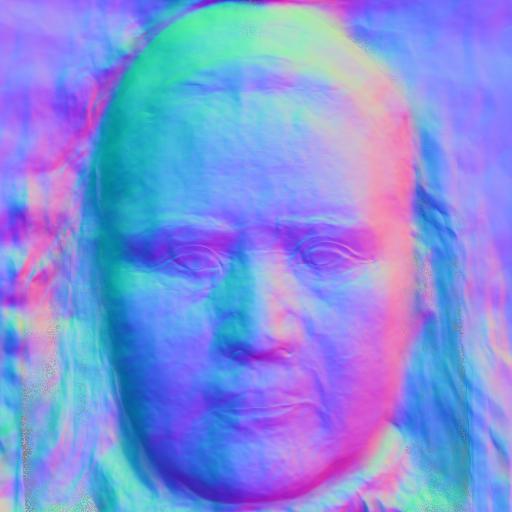}
    \includegraphics[width=0.09\textwidth]{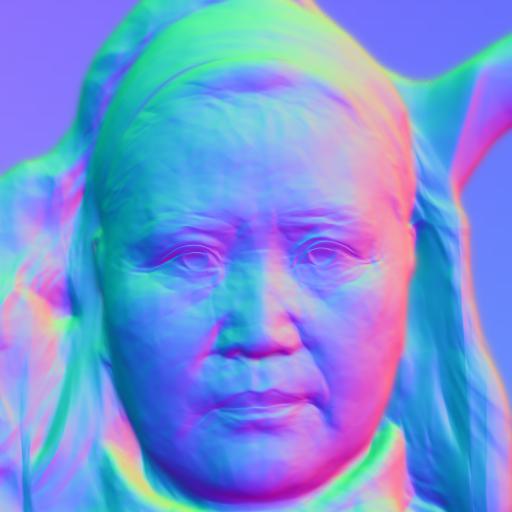}
    \includegraphics[width=0.09\textwidth]{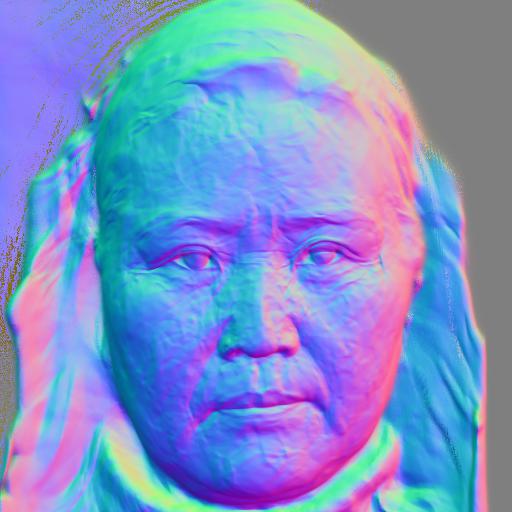}
    \includegraphics[width=0.09\textwidth]{./results/template_effects/gt_571_blank.jpg}
    \includegraphics[width=0.09\textwidth]{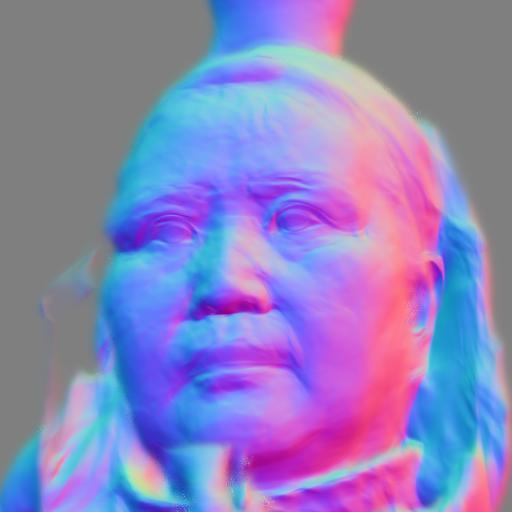}
    \includegraphics[width=0.09\textwidth]{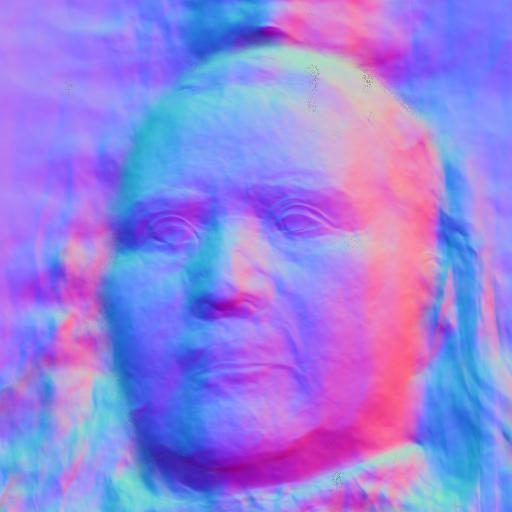}
    \includegraphics[width=0.09\textwidth]{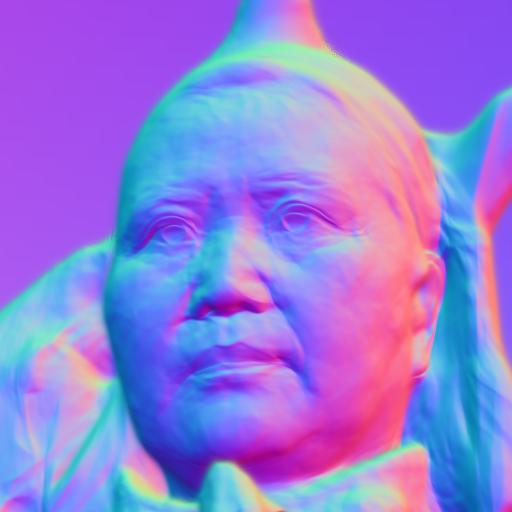}
    \includegraphics[width=0.09\textwidth]{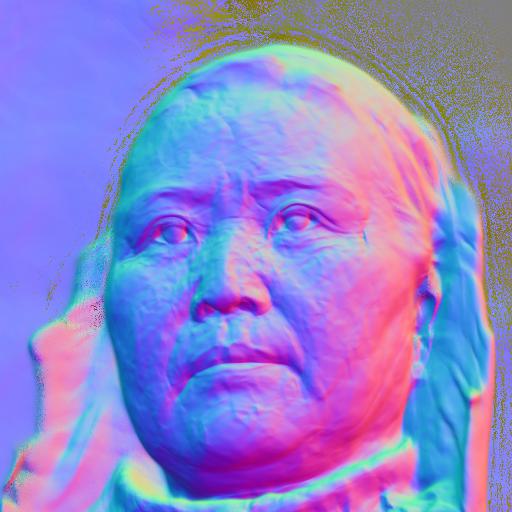}\\
    \makebox[0.09\textwidth]{GT}
    \makebox[0.09\textwidth]{NeuS}
    \makebox[0.09\textwidth]{HF-NeuS}
    \makebox[0.09\textwidth]{VolSDF}
    \makebox[0.09\textwidth]{Ours}
    \makebox[0.09\textwidth]{GT}
    \makebox[0.09\textwidth]{NeuS}
    \makebox[0.09\textwidth]{HF-NeuS}
    \makebox[0.09\textwidth]{VolSDF}
    \makebox[0.09\textwidth]{Ours}\\
    \caption{Comparison of various approaches under a 10-view setting (from Model 395 to Model 413).
    For each model, we show the results on one training view (left) and one novel view (right).}
    \label{fig:all_result2}
\end{figure*}
\begin{figure*}[htbp]
    \centering
    \rotatebox{90}{\textbf{416}}
    \includegraphics[width=0.09\textwidth]{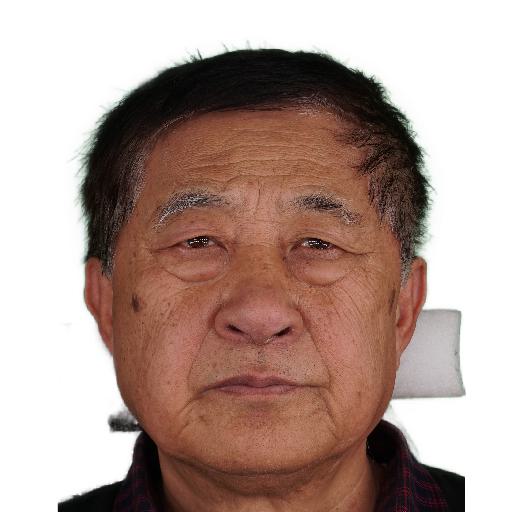}
    \includegraphics[width=0.09\textwidth]{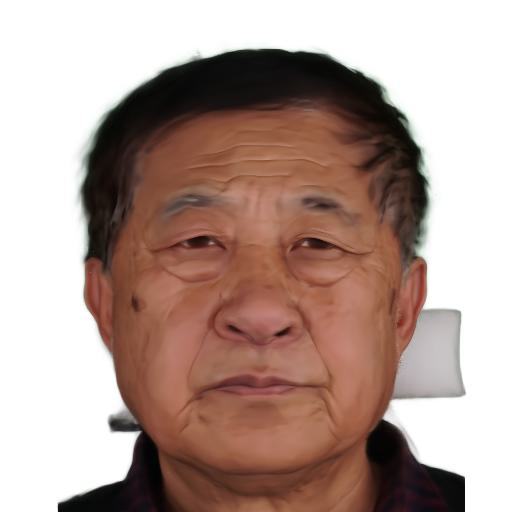}
    \includegraphics[width=0.09\textwidth]{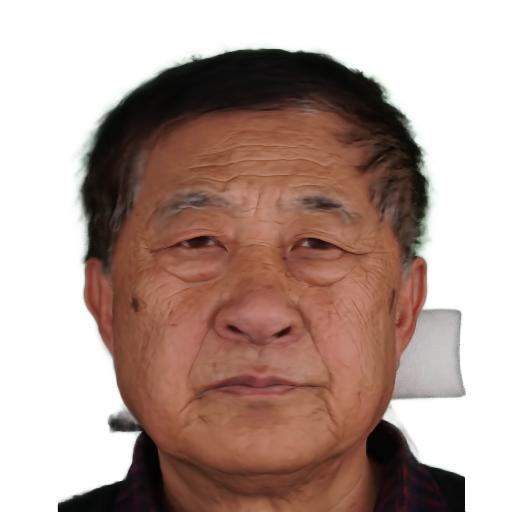}
    \includegraphics[width=0.09\textwidth]{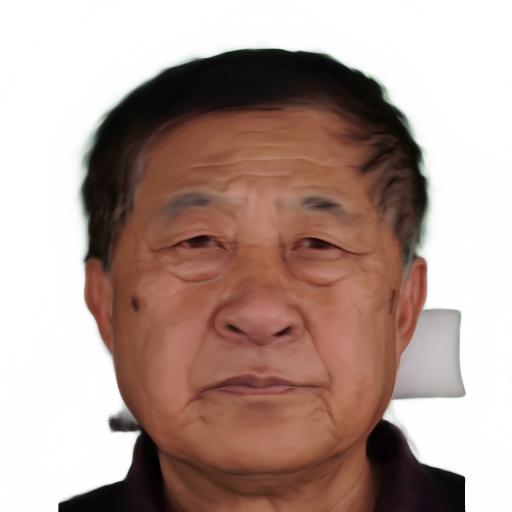}
    \includegraphics[width=0.09\textwidth]{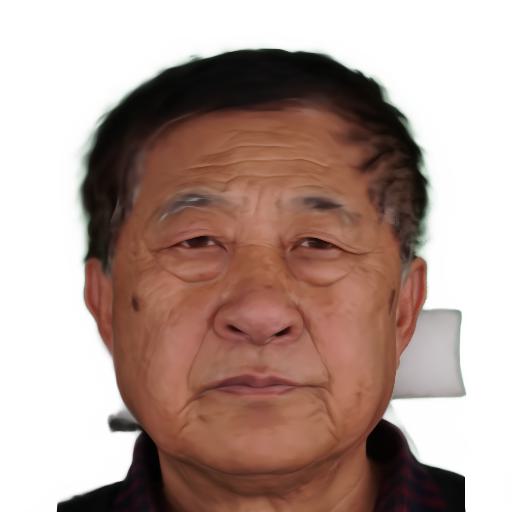}
    \includegraphics[width=0.09\textwidth]{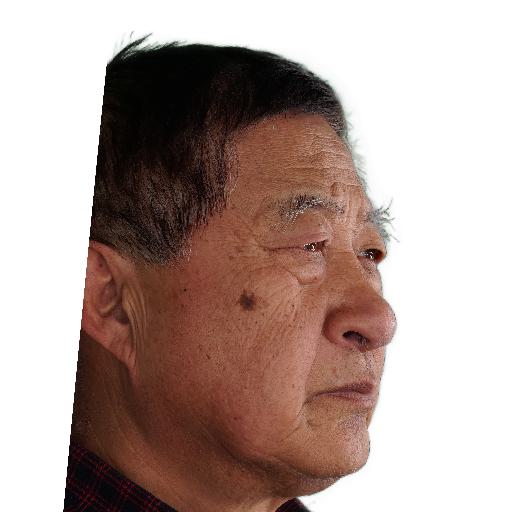}
    \includegraphics[width=0.09\textwidth]{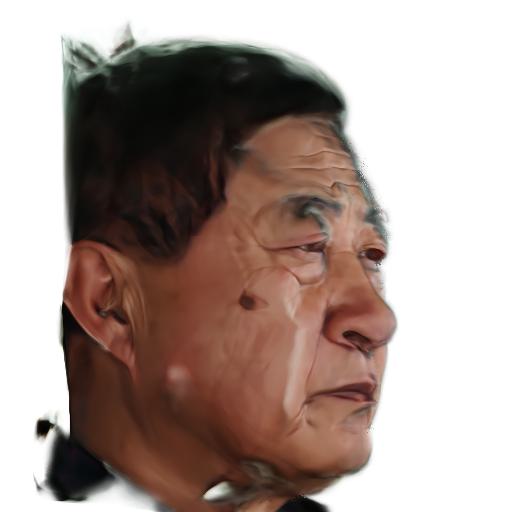}
    \includegraphics[width=0.09\textwidth]{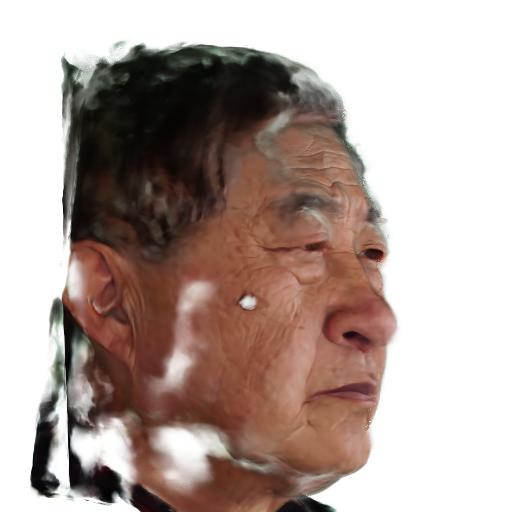}
    \includegraphics[width=0.09\textwidth]{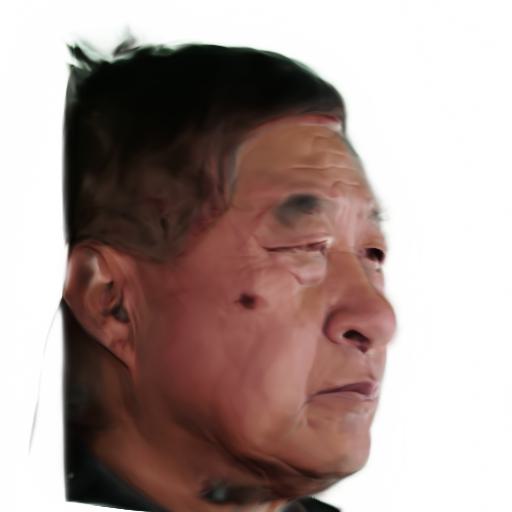}
    \includegraphics[width=0.09\textwidth]{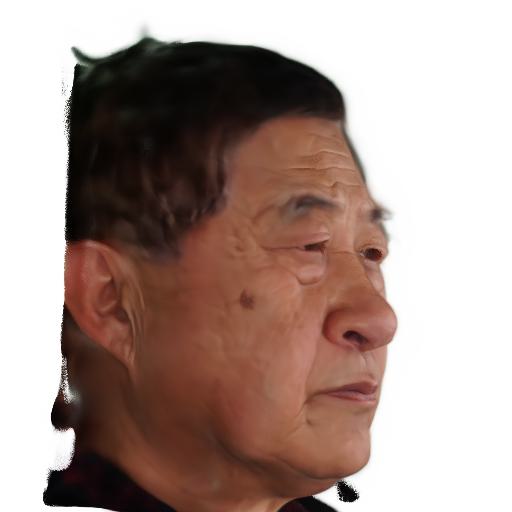}
    \rotatebox{90}{\tiny}
    \includegraphics[width=0.09\textwidth]{./results/template_effects/gt_571_blank.jpg}
    \includegraphics[width=0.09\textwidth]{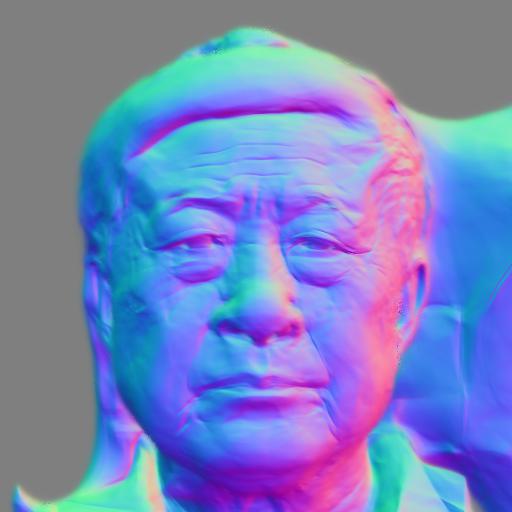}
    \includegraphics[width=0.09\textwidth]{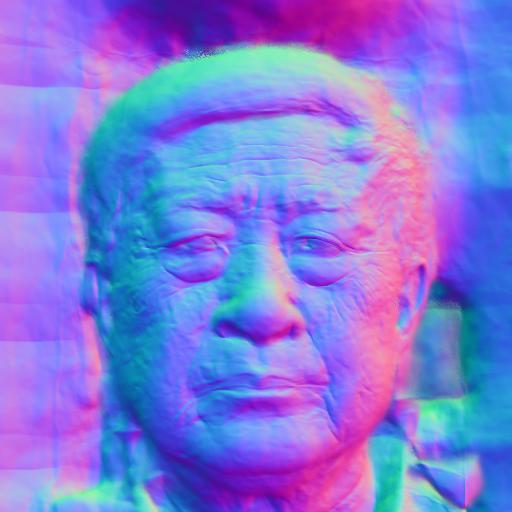}
    \includegraphics[width=0.09\textwidth]{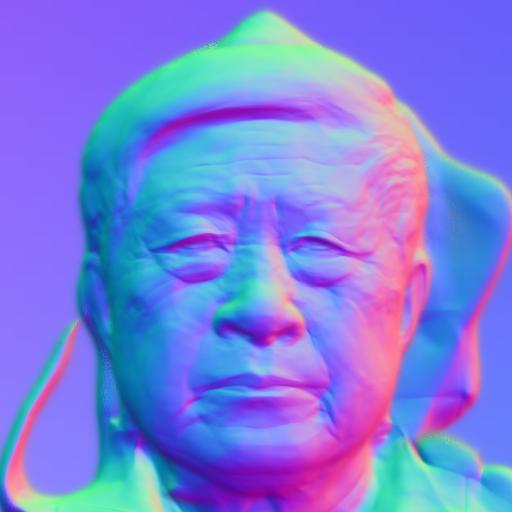}
    \includegraphics[width=0.09\textwidth]{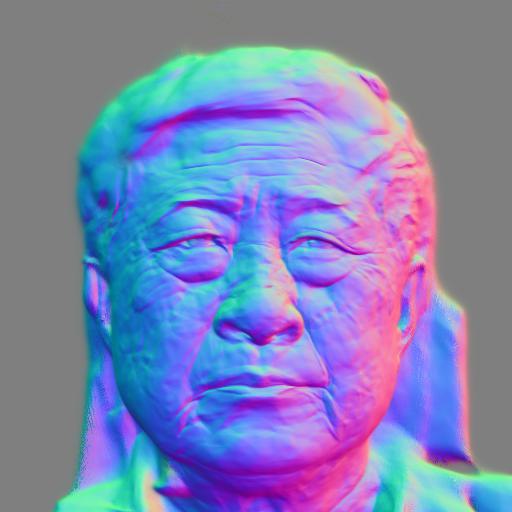}
    \includegraphics[width=0.09\textwidth]{./results/template_effects/gt_571_blank.jpg}
    \includegraphics[width=0.09\textwidth]{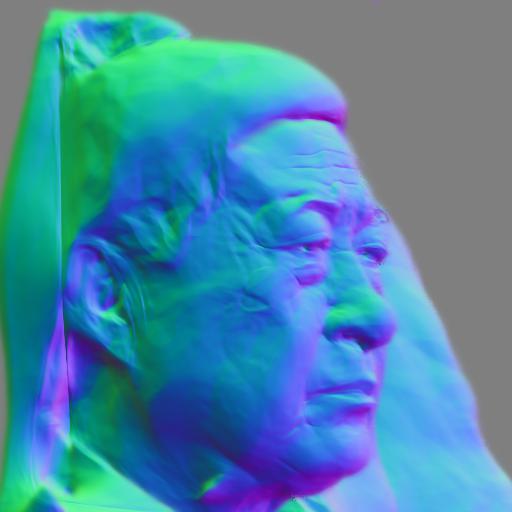}
    \includegraphics[width=0.09\textwidth]{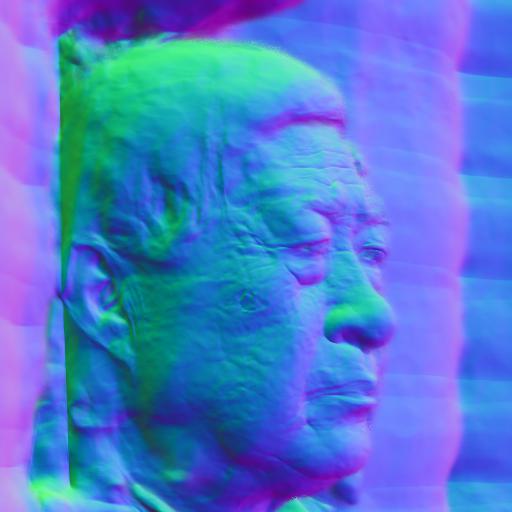}
    \includegraphics[width=0.09\textwidth]{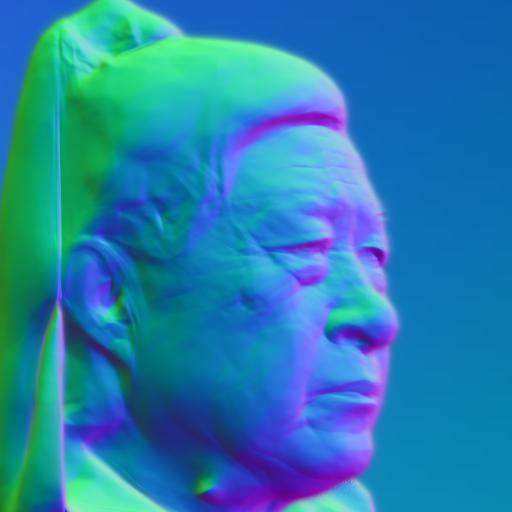}
    \includegraphics[width=0.09\textwidth]{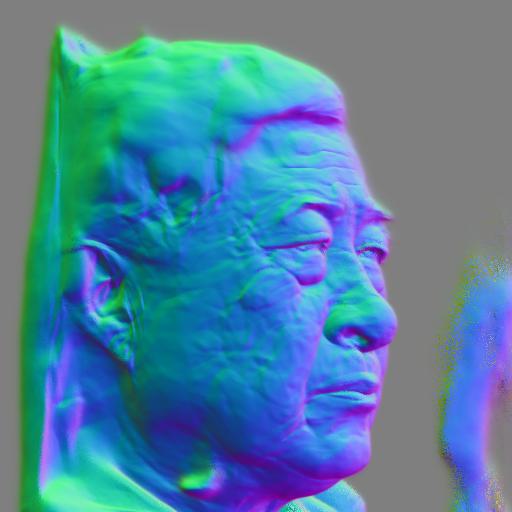}\\
    \rotatebox{90}{\textbf{421}}
    \includegraphics[width=0.09\textwidth]{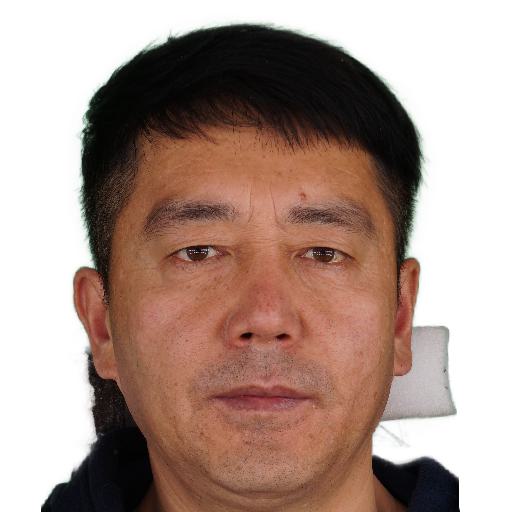}
    \includegraphics[width=0.09\textwidth]{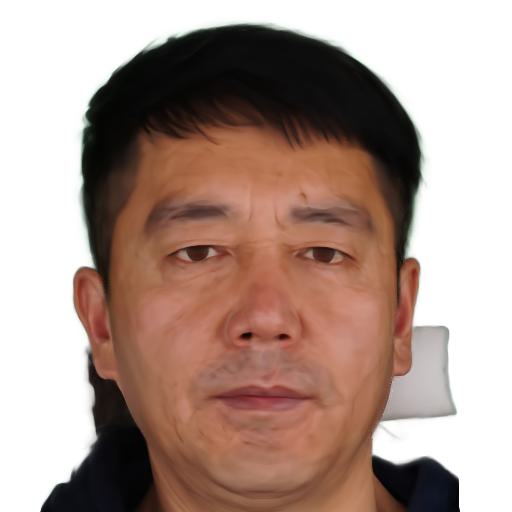}
    \includegraphics[width=0.09\textwidth]{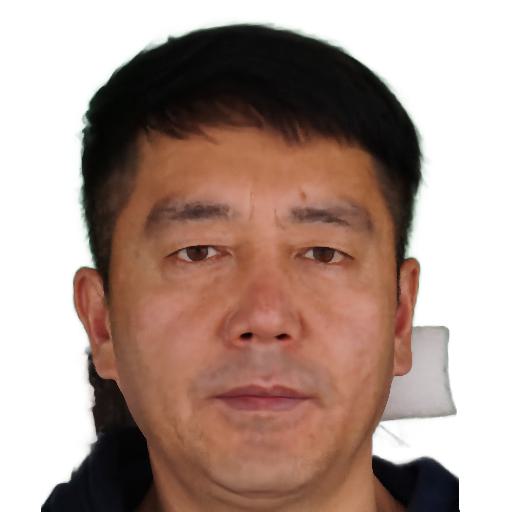}
    \includegraphics[width=0.09\textwidth]{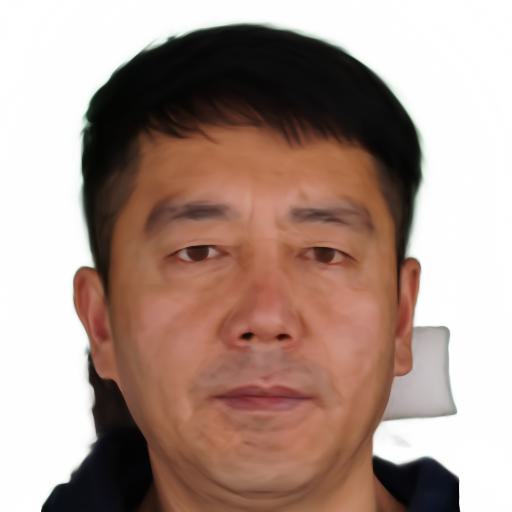}
    \includegraphics[width=0.09\textwidth]{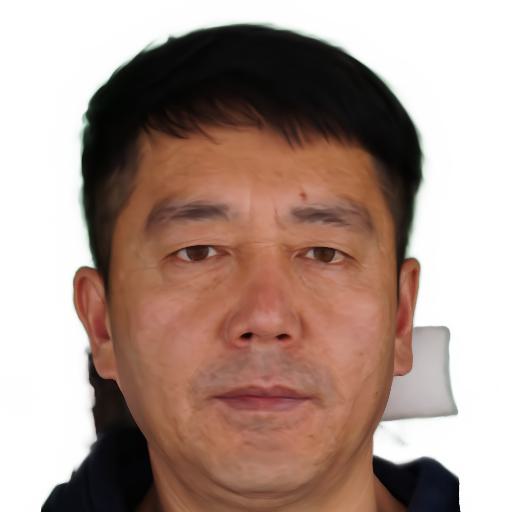}
    \includegraphics[width=0.09\textwidth]{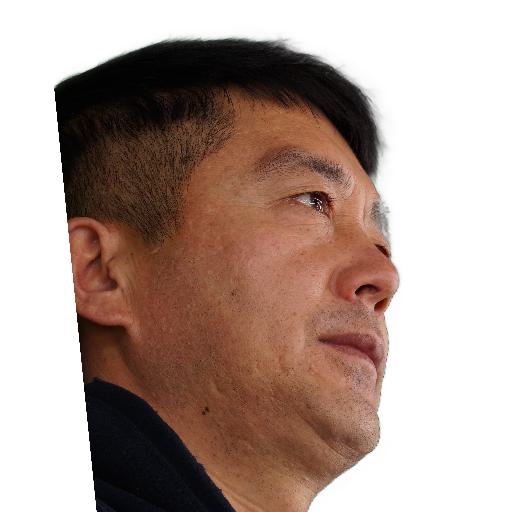}
    \includegraphics[width=0.09\textwidth]{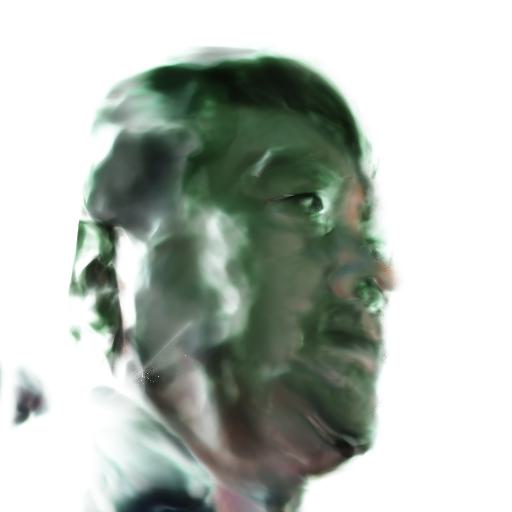}
    \includegraphics[width=0.09\textwidth]{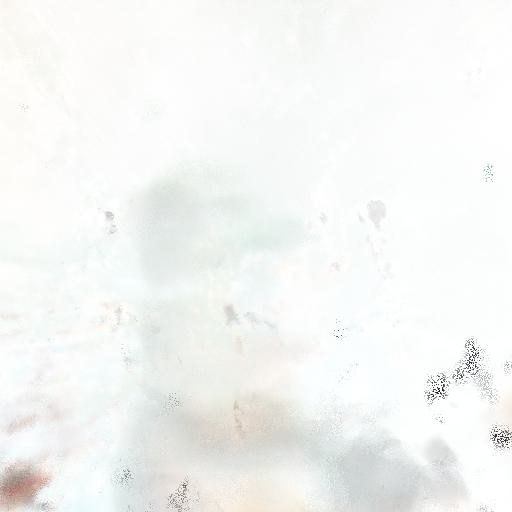}
    \includegraphics[width=0.09\textwidth]{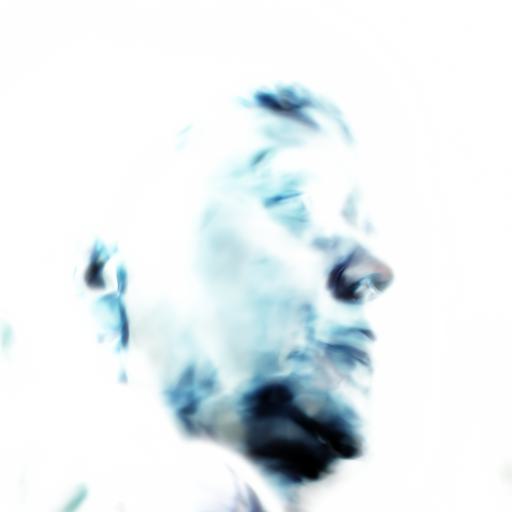}
    \includegraphics[width=0.09\textwidth]{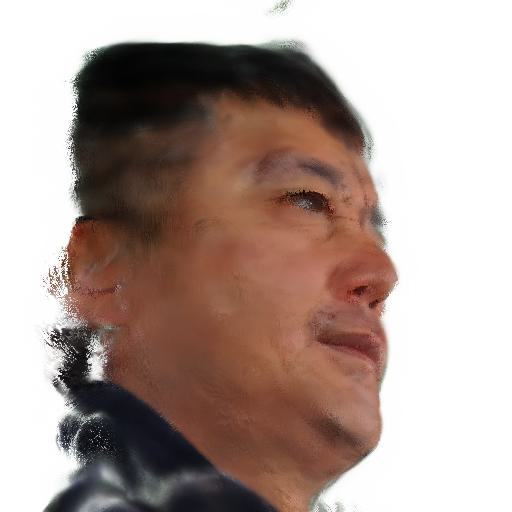}
    \rotatebox{90}{\tiny}
    \includegraphics[width=0.09\textwidth]{./results/template_effects/gt_571_blank.jpg}
    \includegraphics[width=0.09\textwidth]{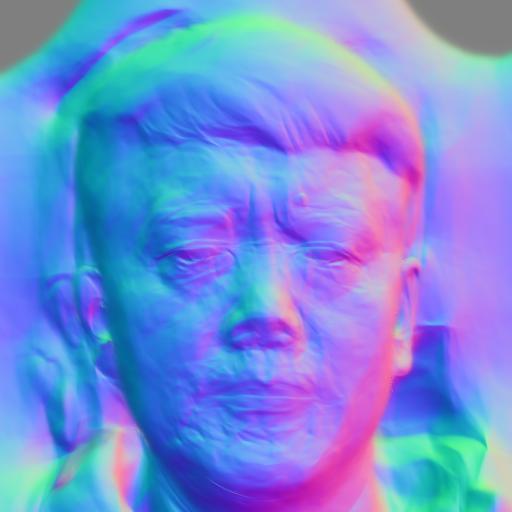}
    \includegraphics[width=0.09\textwidth]{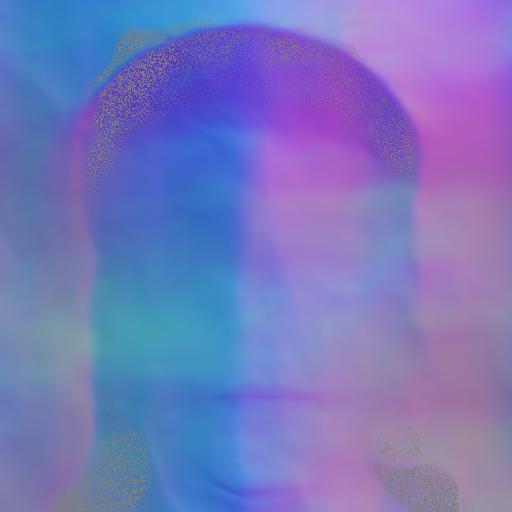}
    \includegraphics[width=0.09\textwidth]{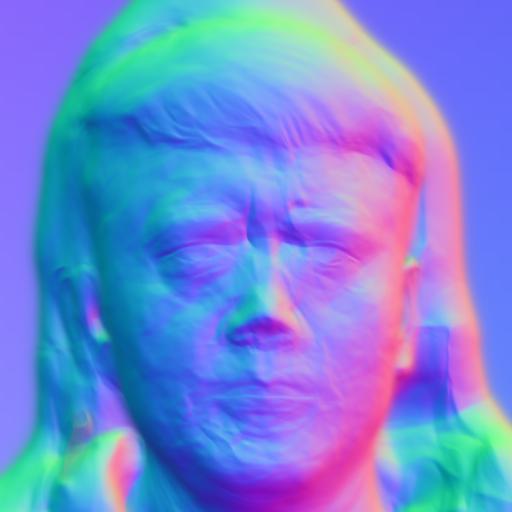}
    \includegraphics[width=0.09\textwidth]{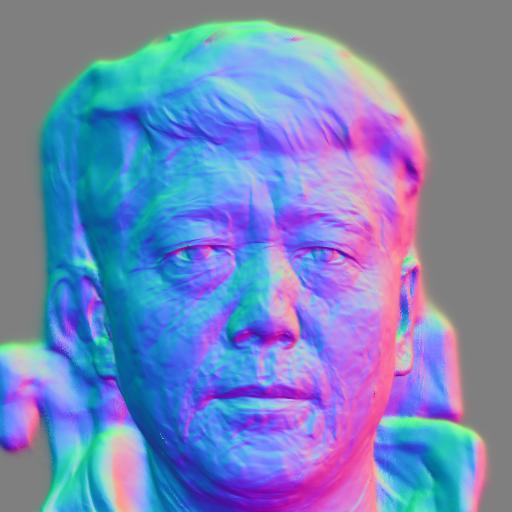}
    \includegraphics[width=0.09\textwidth]{./results/template_effects/gt_571_blank.jpg}
    \includegraphics[width=0.09\textwidth]{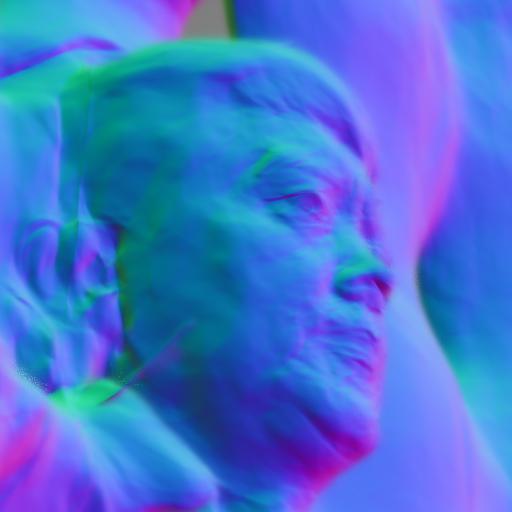}
    \includegraphics[width=0.09\textwidth]{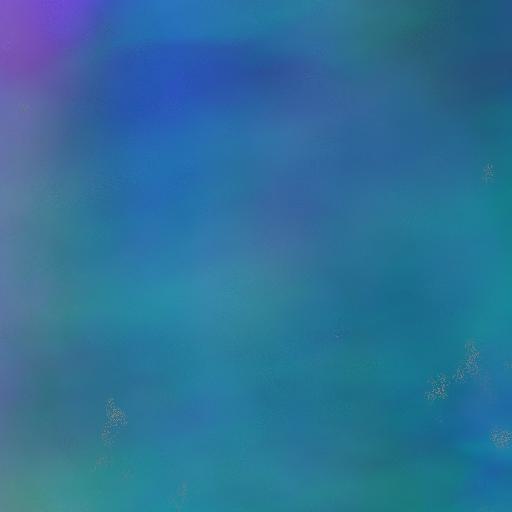}
    \includegraphics[width=0.09\textwidth]{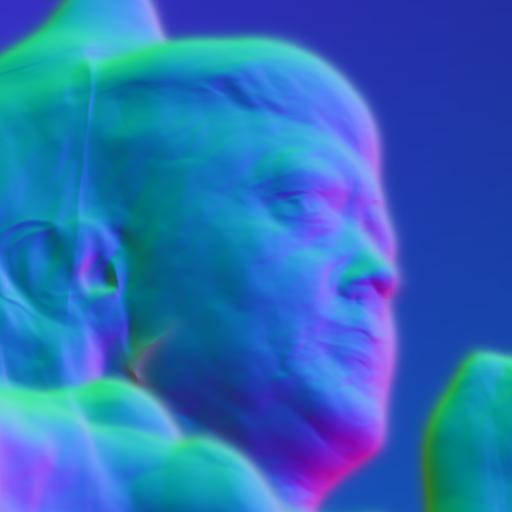}
    \includegraphics[width=0.09\textwidth]{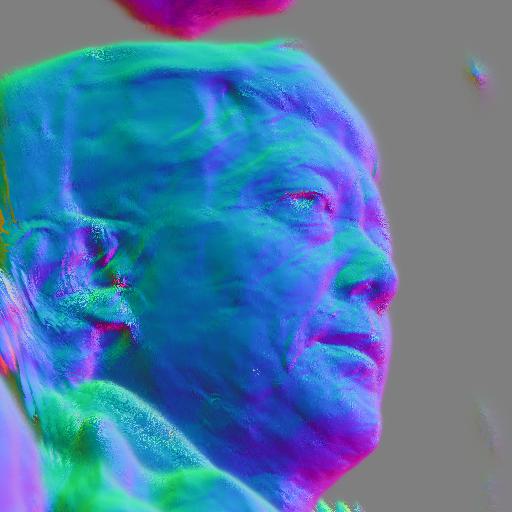}\\
    \rotatebox{90}{\textbf{427}}
    \includegraphics[width=0.09\textwidth]{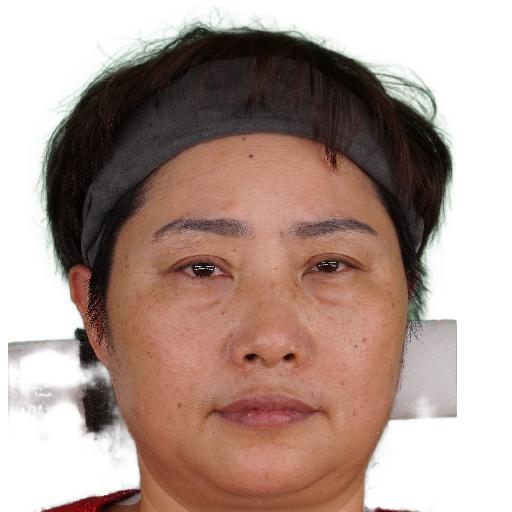}
    \includegraphics[width=0.09\textwidth]{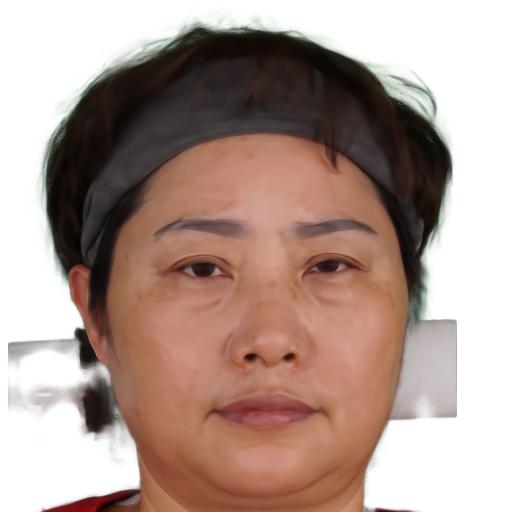}
    \includegraphics[width=0.09\textwidth]{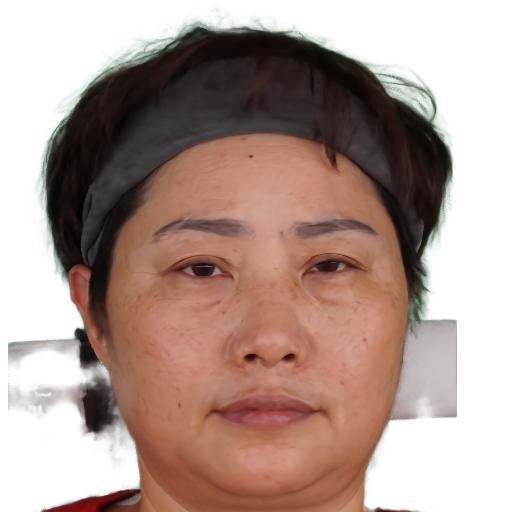}
    \includegraphics[width=0.09\textwidth]{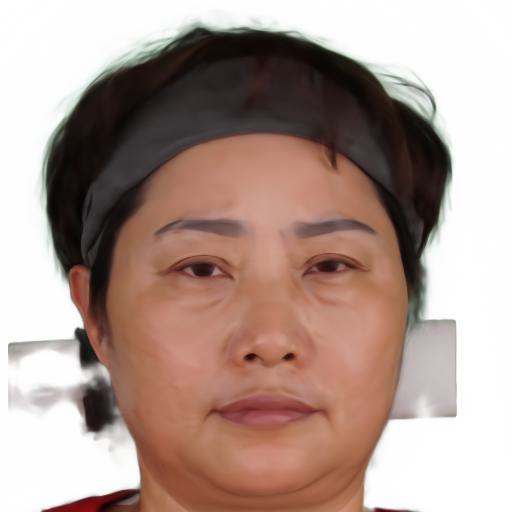}
    \includegraphics[width=0.09\textwidth]{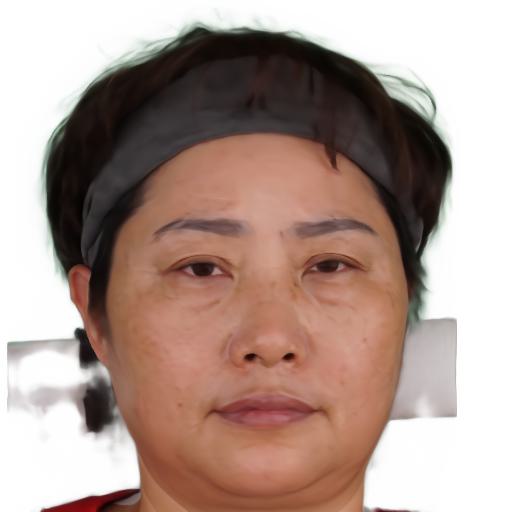}
    \includegraphics[width=0.09\textwidth]{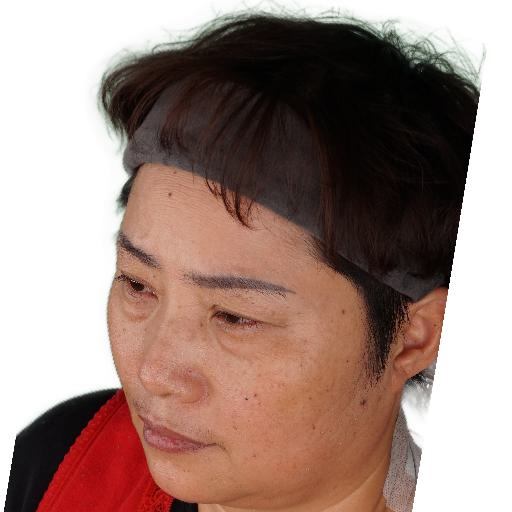}
    \includegraphics[width=0.09\textwidth]{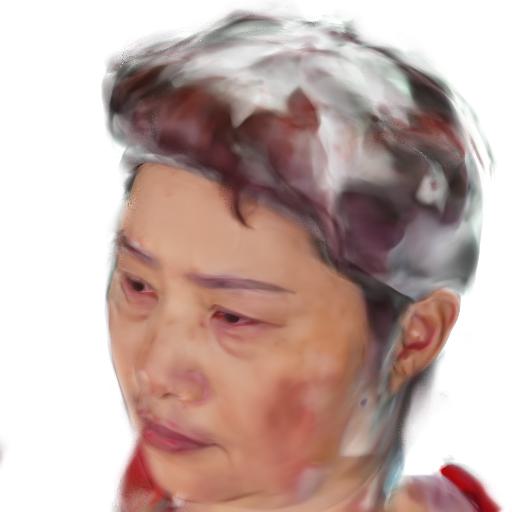}
    \includegraphics[width=0.09\textwidth]{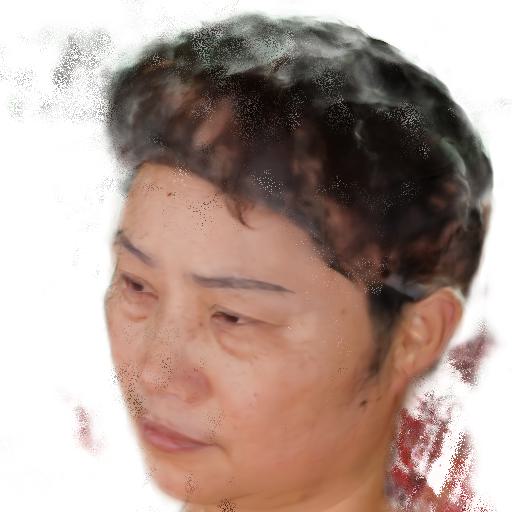}
    \includegraphics[width=0.09\textwidth]{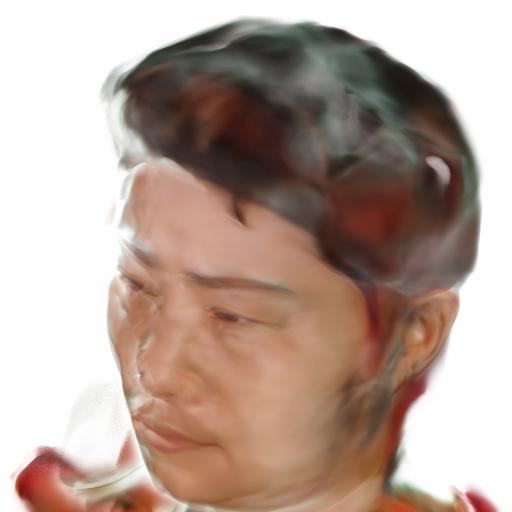}
    \includegraphics[width=0.09\textwidth]{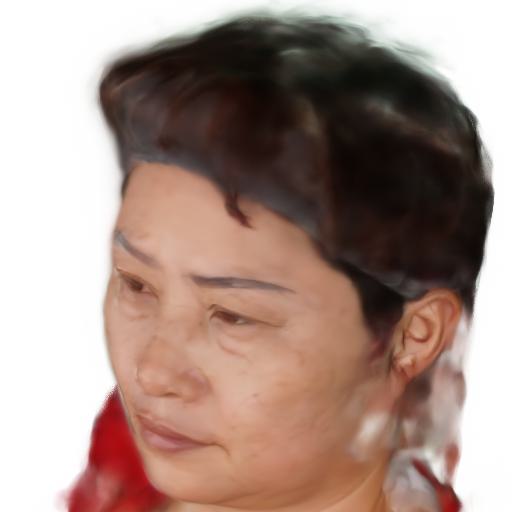}
    \rotatebox{90}{\tiny}
    \includegraphics[width=0.09\textwidth]{./results/template_effects/gt_571_blank.jpg}
    \includegraphics[width=0.09\textwidth]{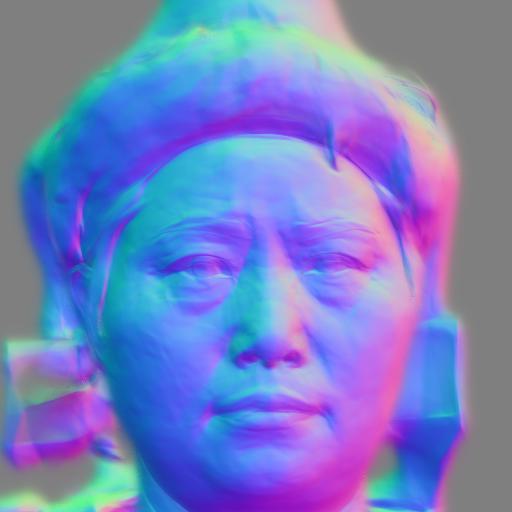}
    \includegraphics[width=0.09\textwidth]{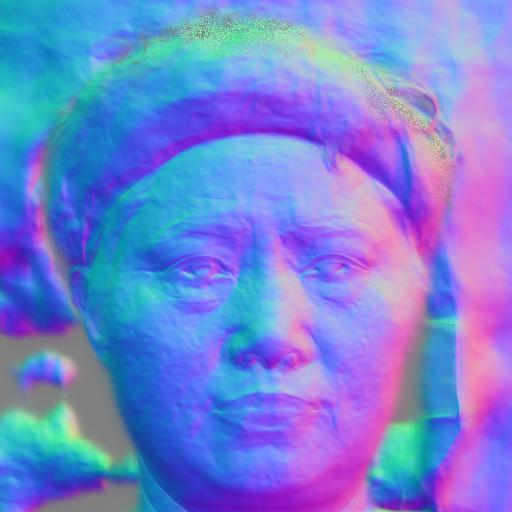}
    \includegraphics[width=0.09\textwidth]{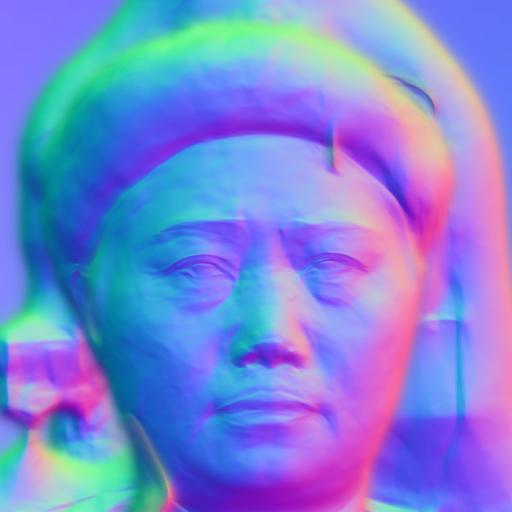}
    \includegraphics[width=0.09\textwidth]{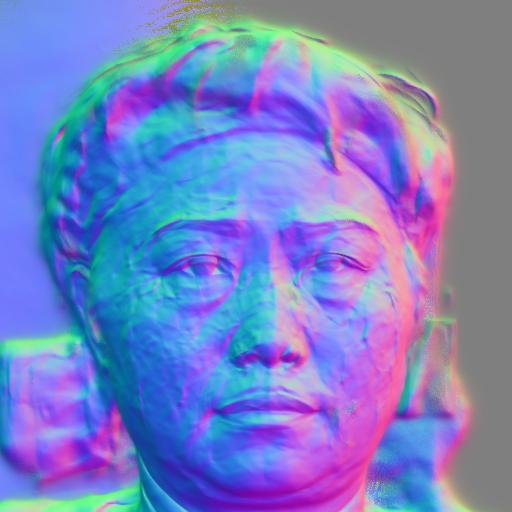}
    \includegraphics[width=0.09\textwidth]{./results/template_effects/gt_571_blank.jpg}
    \includegraphics[width=0.09\textwidth]{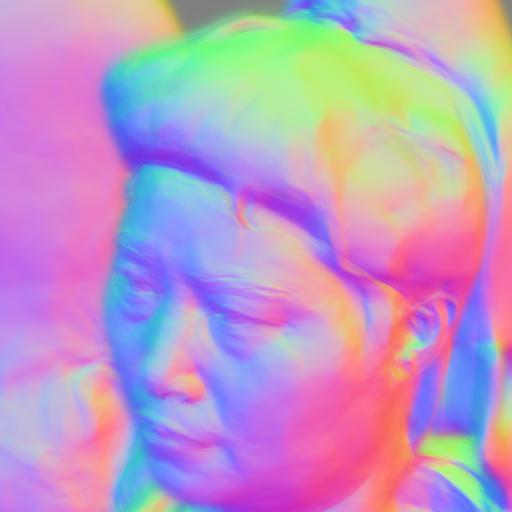}
    \includegraphics[width=0.09\textwidth]{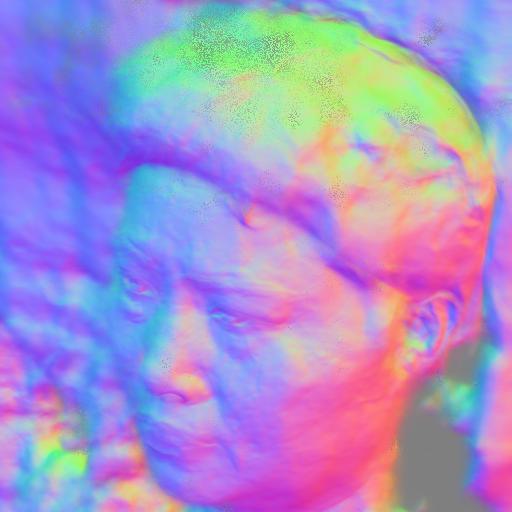}
    \includegraphics[width=0.09\textwidth]{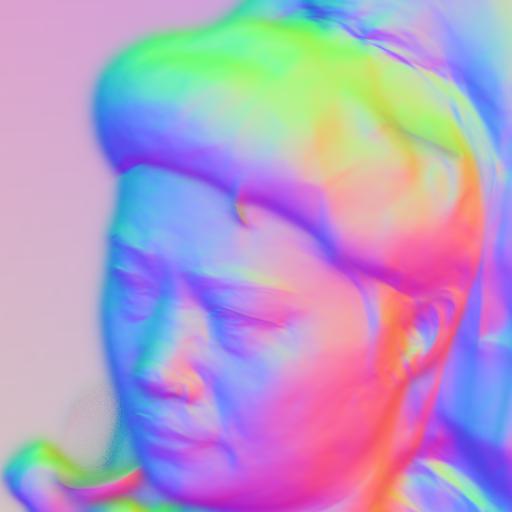}
    \includegraphics[width=0.09\textwidth]{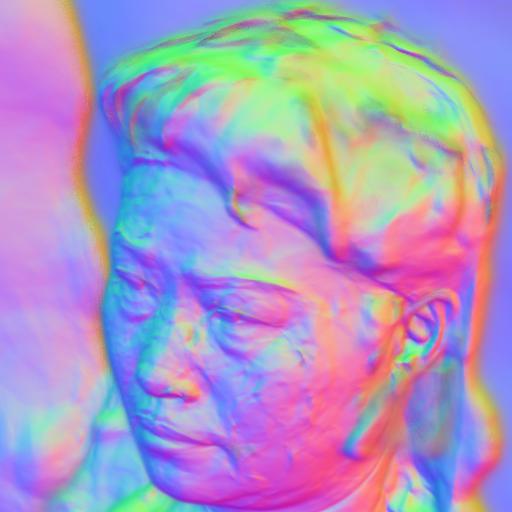}\\
    \rotatebox{90}{\textbf{429}}
    \includegraphics[width=0.09\textwidth]{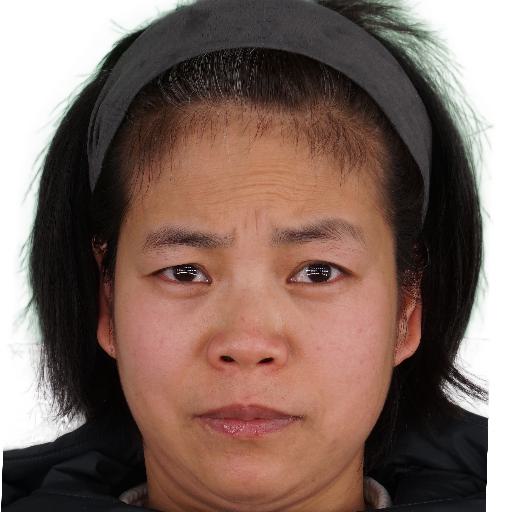}
    \includegraphics[width=0.09\textwidth]{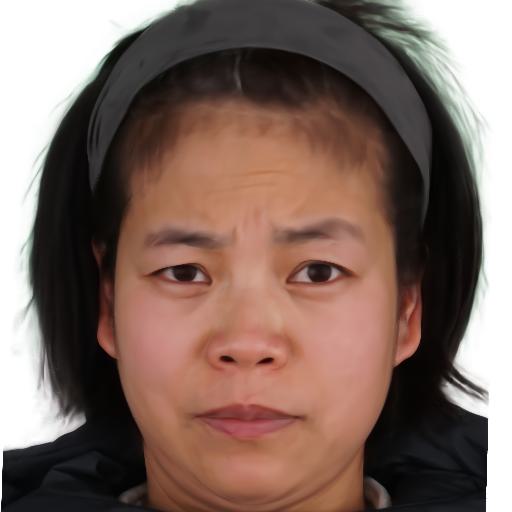}
    \includegraphics[width=0.09\textwidth]{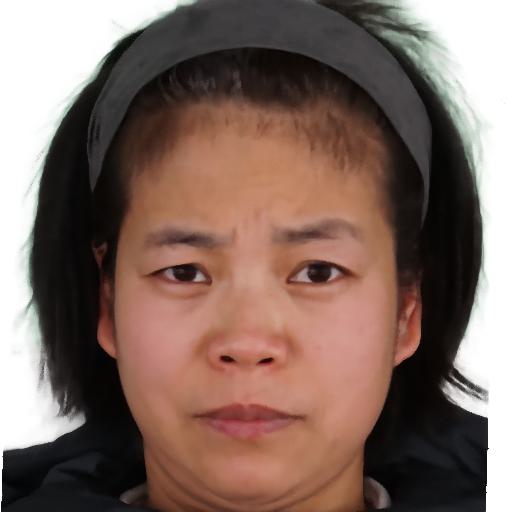}
    \includegraphics[width=0.09\textwidth]{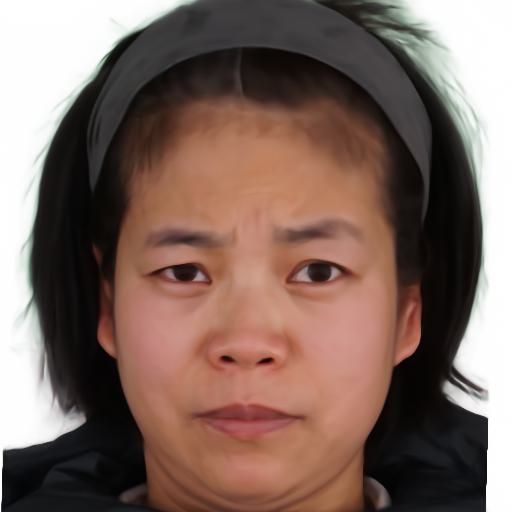}
    \includegraphics[width=0.09\textwidth]{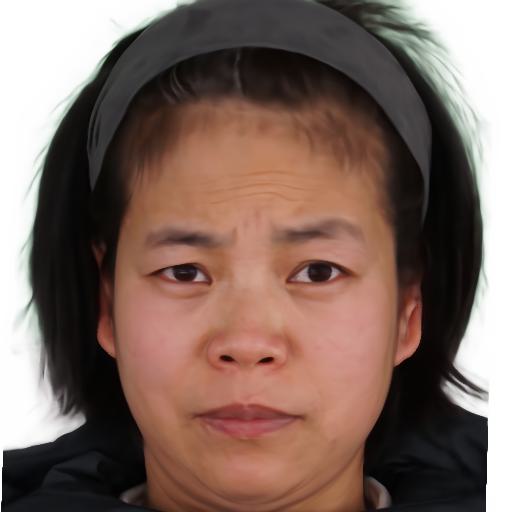}
    \includegraphics[width=0.09\textwidth]{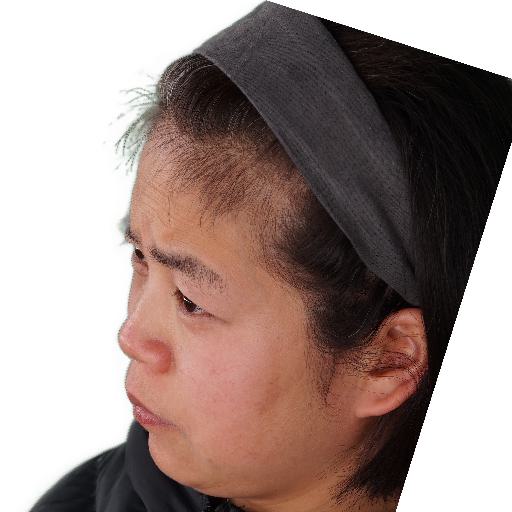}
    \includegraphics[width=0.09\textwidth]{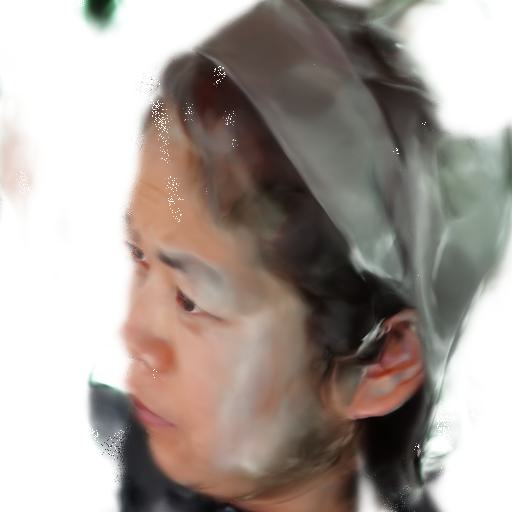}
    \includegraphics[width=0.09\textwidth]{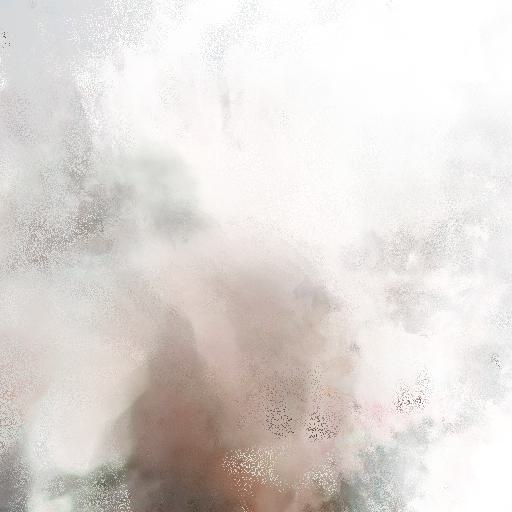}
    \includegraphics[width=0.09\textwidth]{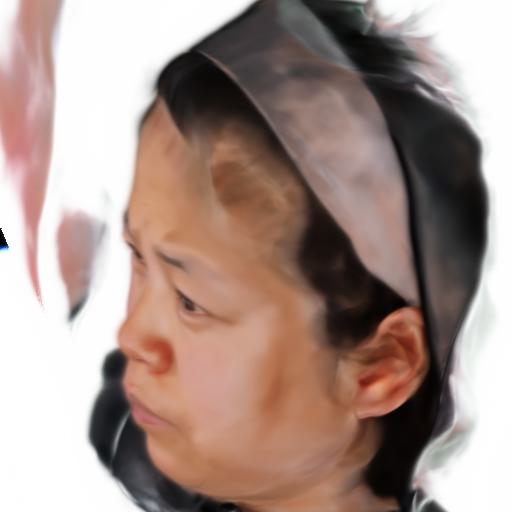}
    \includegraphics[width=0.09\textwidth]{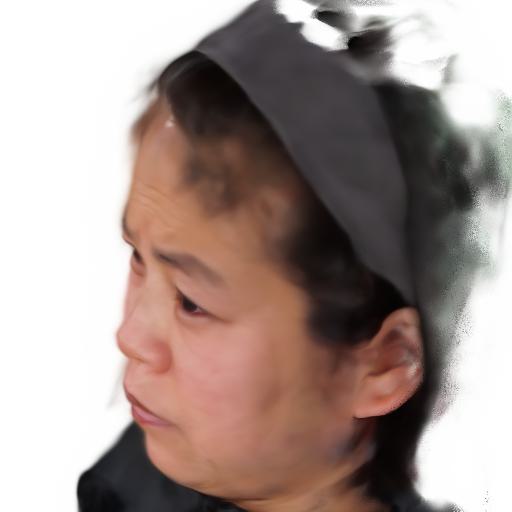}
    \rotatebox{90}{\tiny}
    \includegraphics[width=0.09\textwidth]{./results/template_effects/gt_571_blank.jpg}
    \includegraphics[width=0.09\textwidth]{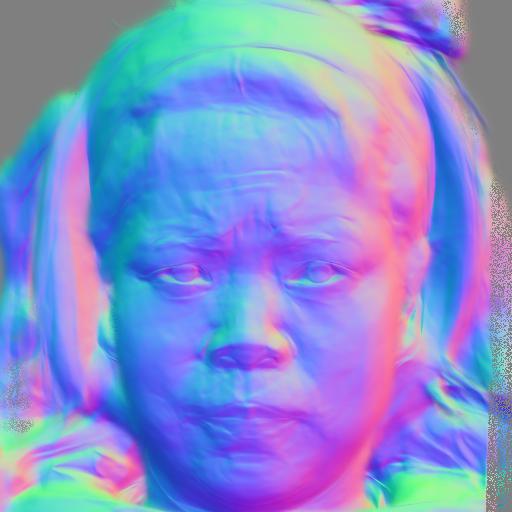}
    \includegraphics[width=0.09\textwidth]{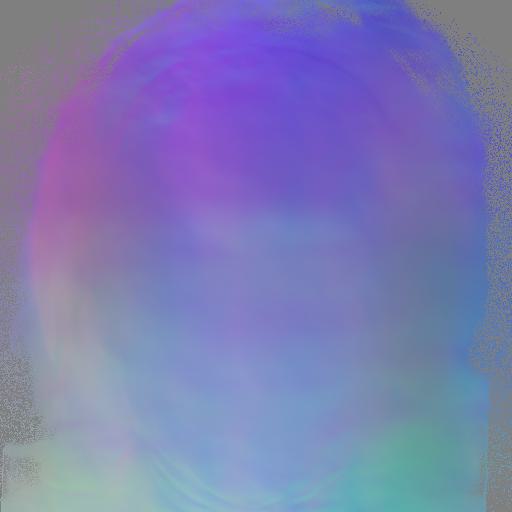}
    \includegraphics[width=0.09\textwidth]{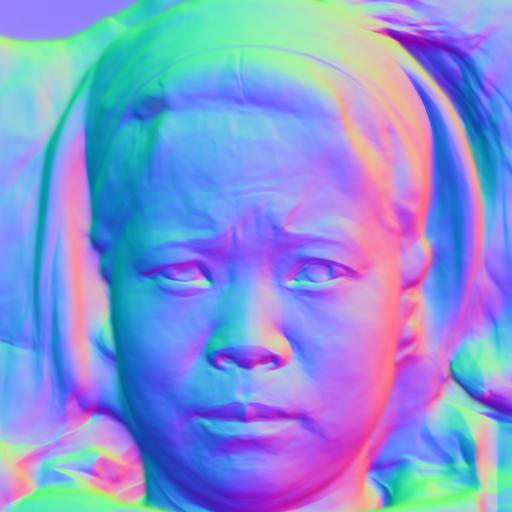}
    \includegraphics[width=0.09\textwidth]{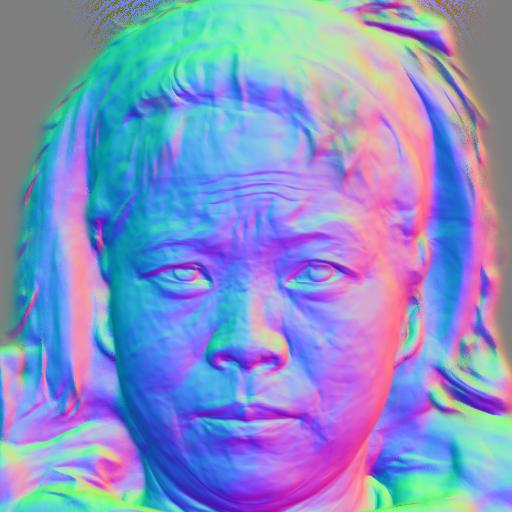}
    \includegraphics[width=0.09\textwidth]{./results/template_effects/gt_571_blank.jpg}
    \includegraphics[width=0.09\textwidth]{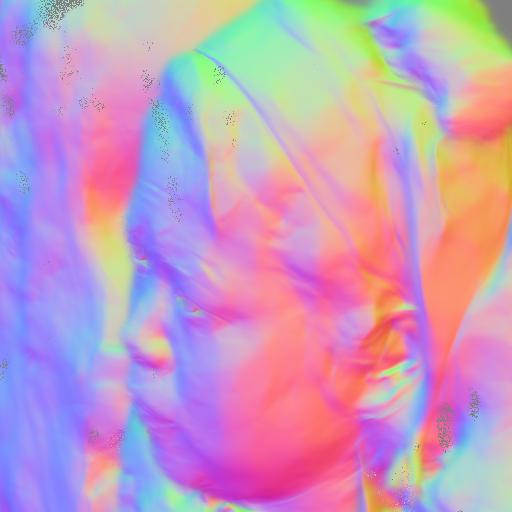}
    \includegraphics[width=0.09\textwidth]{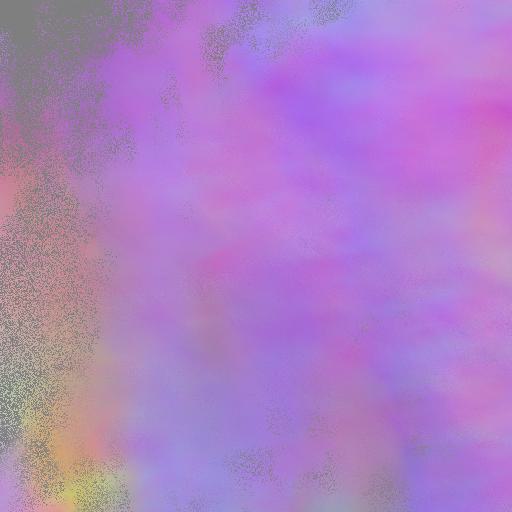}
    \includegraphics[width=0.09\textwidth]{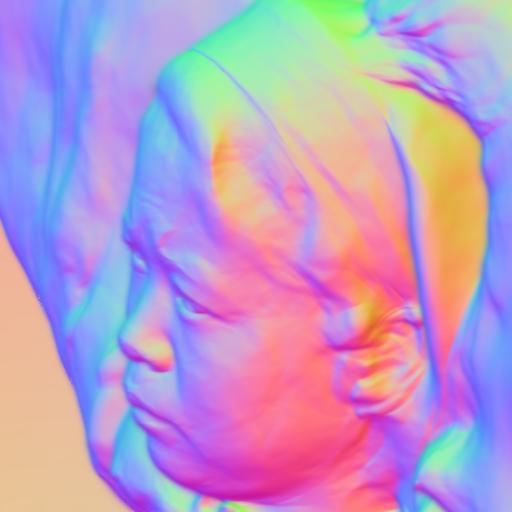}
    \includegraphics[width=0.09\textwidth]{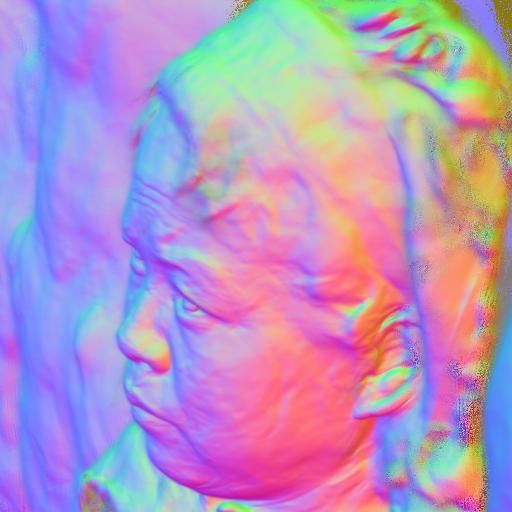}\\
    \rotatebox{90}{\textbf{435}}
    \includegraphics[width=0.09\textwidth]{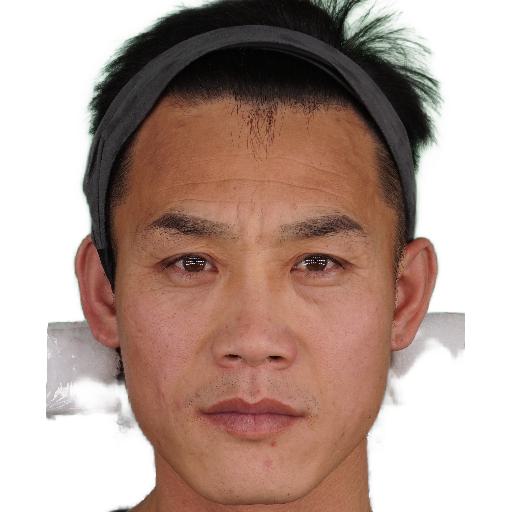}
    \includegraphics[width=0.09\textwidth]{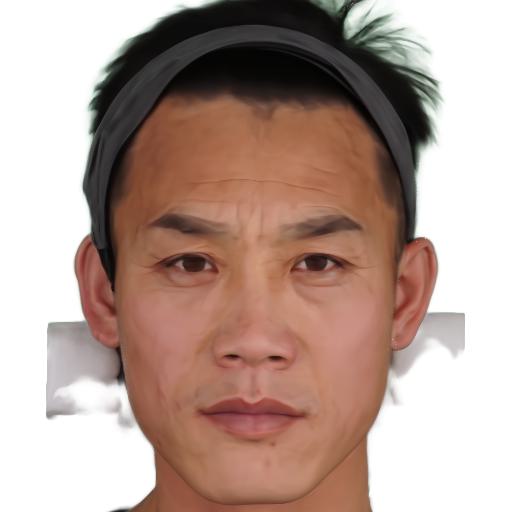}
    \includegraphics[width=0.09\textwidth]{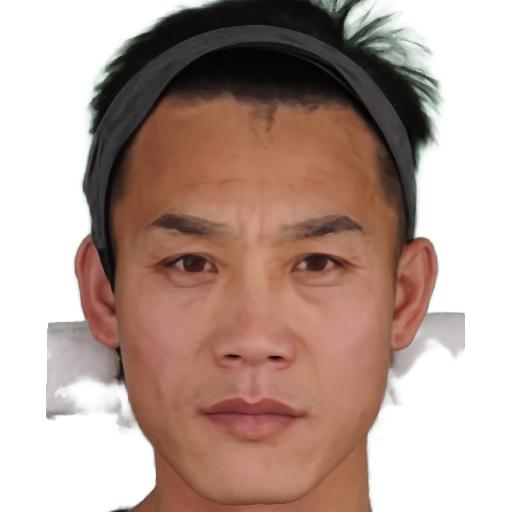}
    \includegraphics[width=0.09\textwidth]{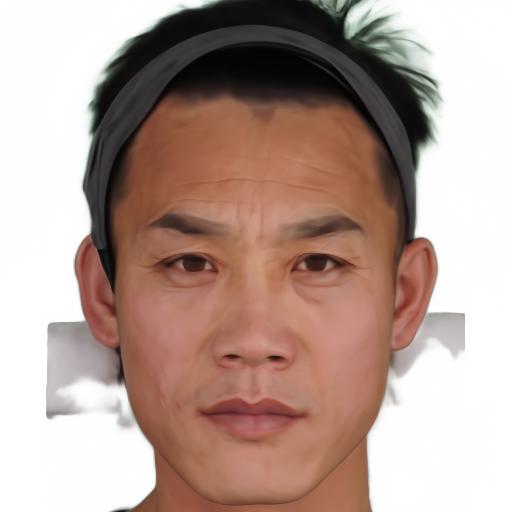}
    \includegraphics[width=0.09\textwidth]{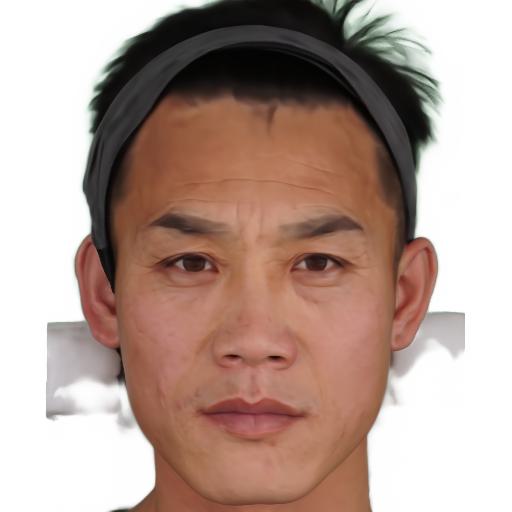}
    \includegraphics[width=0.09\textwidth]{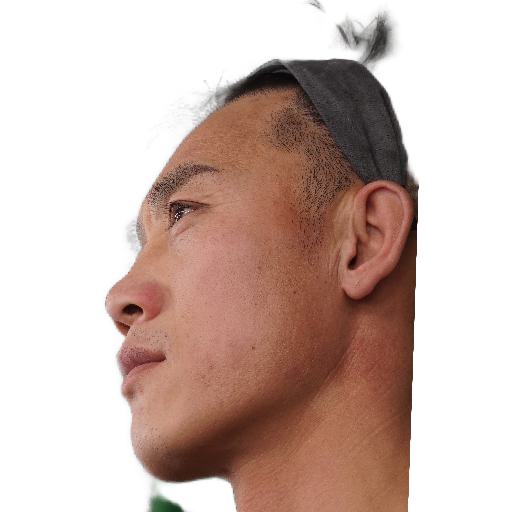}
    \includegraphics[width=0.09\textwidth]{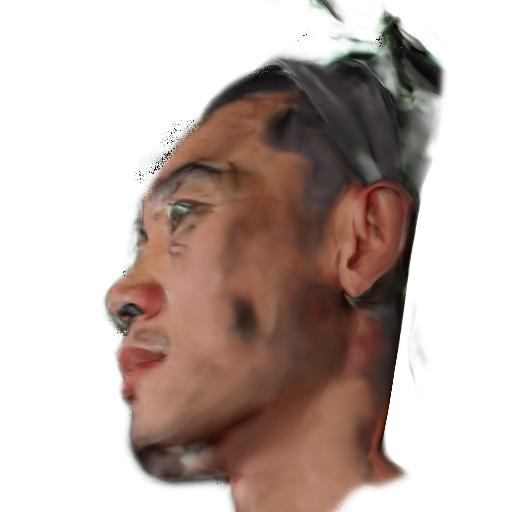}
    \includegraphics[width=0.09\textwidth]{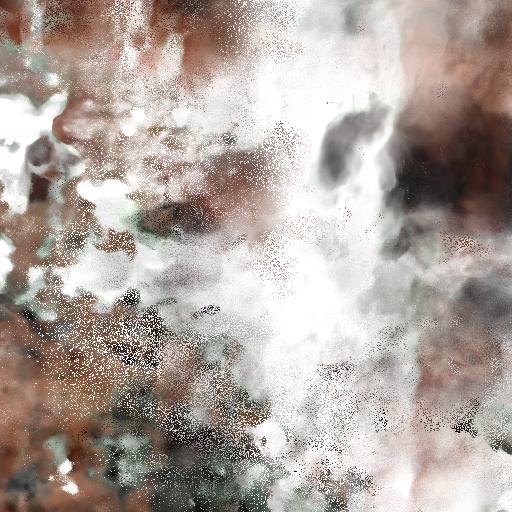}
    \includegraphics[width=0.09\textwidth]{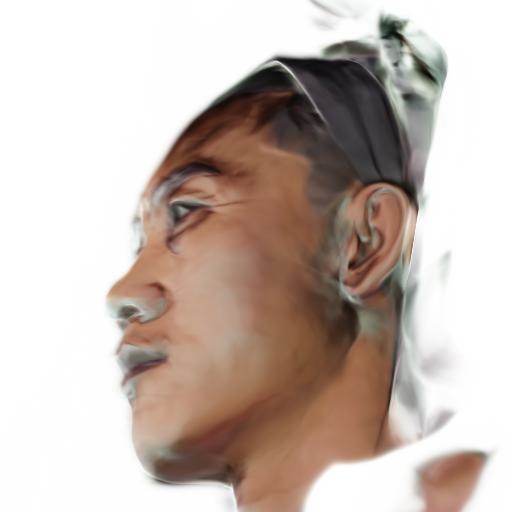}
    \includegraphics[width=0.09\textwidth]{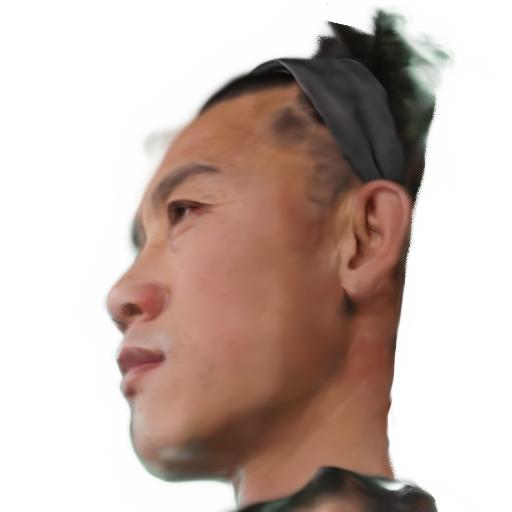}
    \rotatebox{90}{\tiny}
    \includegraphics[width=0.09\textwidth]{./results/template_effects/gt_571_blank.jpg}
    \includegraphics[width=0.09\textwidth]{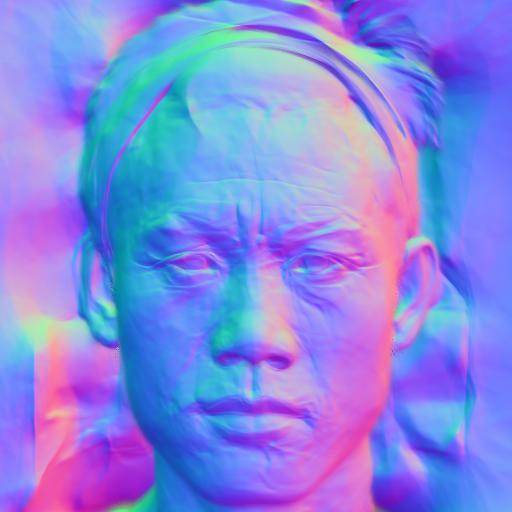}
    \includegraphics[width=0.09\textwidth]{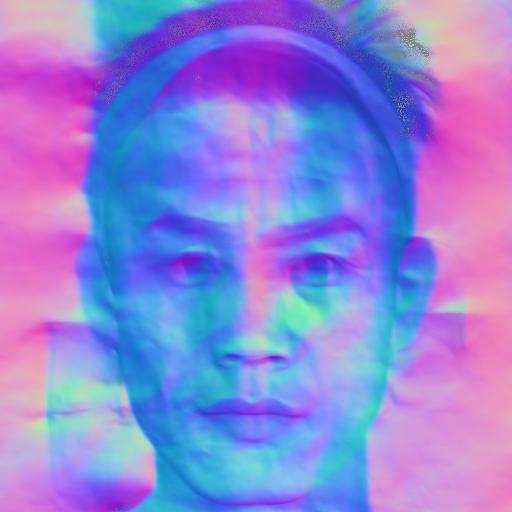}
    \includegraphics[width=0.09\textwidth]{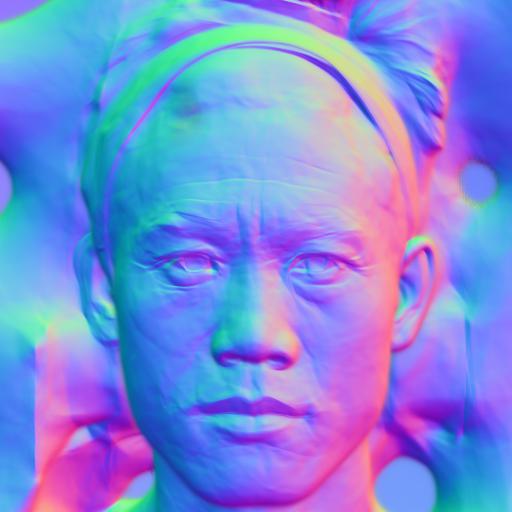}
    \includegraphics[width=0.09\textwidth]{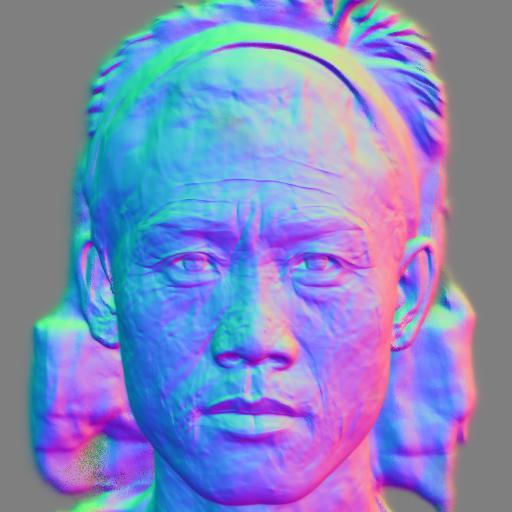}
    \includegraphics[width=0.09\textwidth]{./results/template_effects/gt_571_blank.jpg}
    \includegraphics[width=0.09\textwidth]{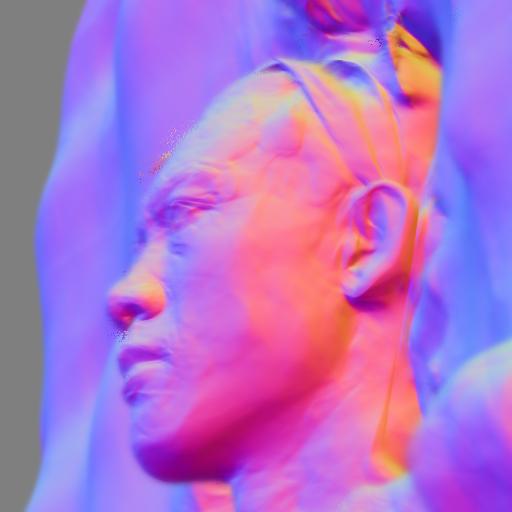}
    \includegraphics[width=0.09\textwidth]{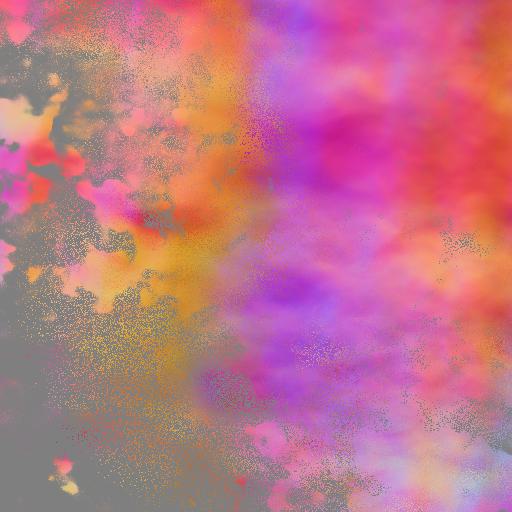}
    \includegraphics[width=0.09\textwidth]{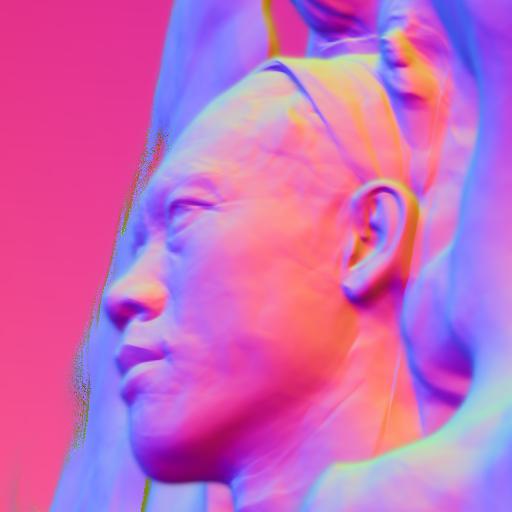}
    \includegraphics[width=0.09\textwidth]{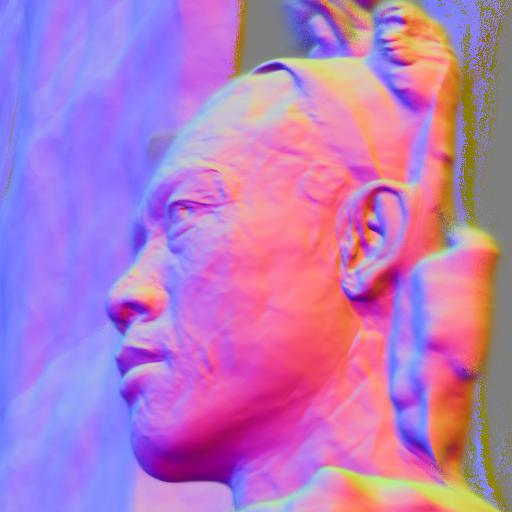}\\
    \rotatebox{90}{\textbf{451}}
    \includegraphics[width=0.09\textwidth]{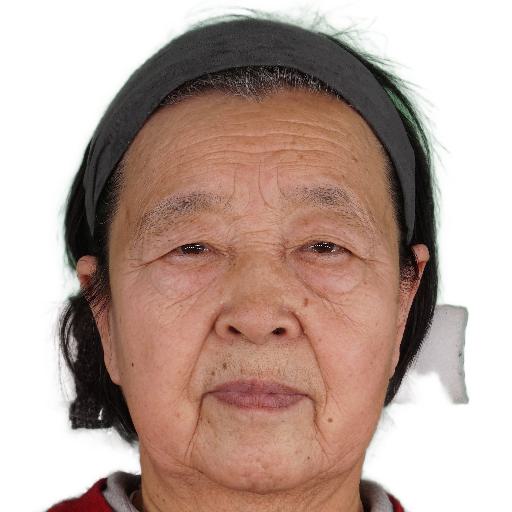}
    \includegraphics[width=0.09\textwidth]{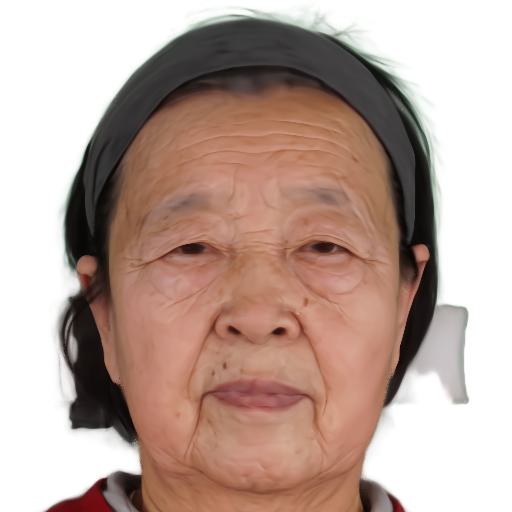}
    \includegraphics[width=0.09\textwidth]{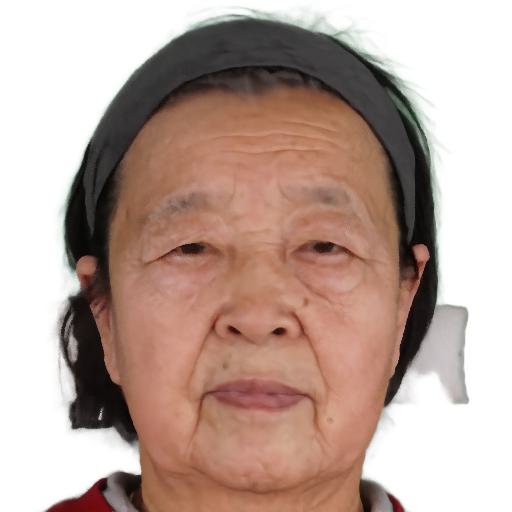}
    \includegraphics[width=0.09\textwidth]{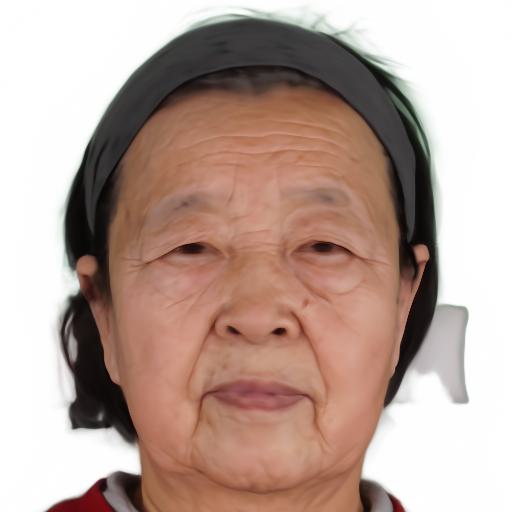}
    \includegraphics[width=0.09\textwidth]{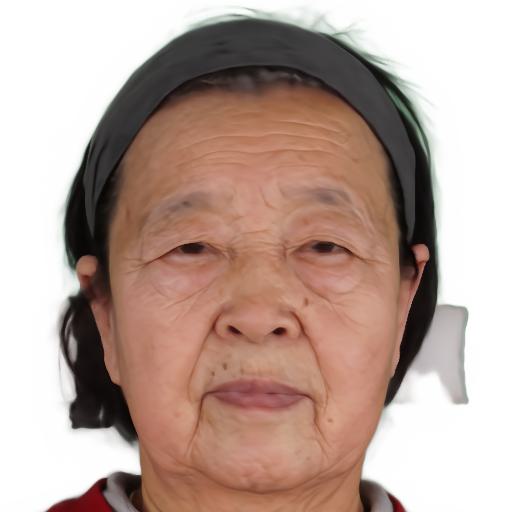}
    \includegraphics[width=0.09\textwidth]{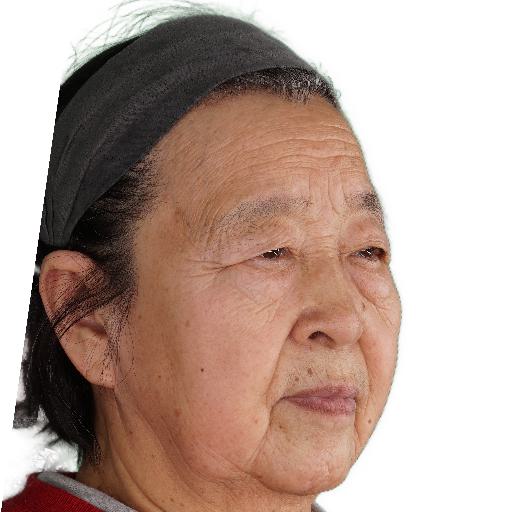}
    \includegraphics[width=0.09\textwidth]{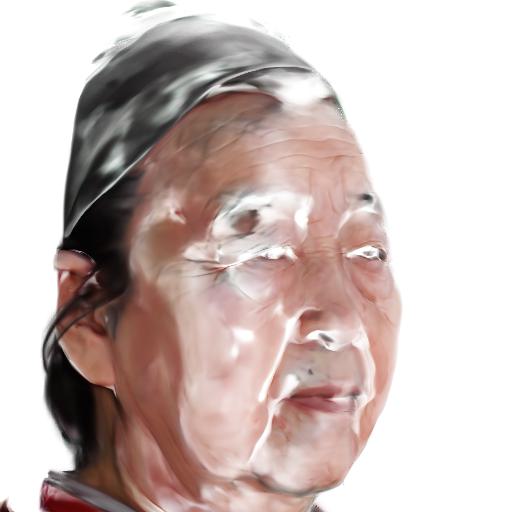}
    \includegraphics[width=0.09\textwidth]{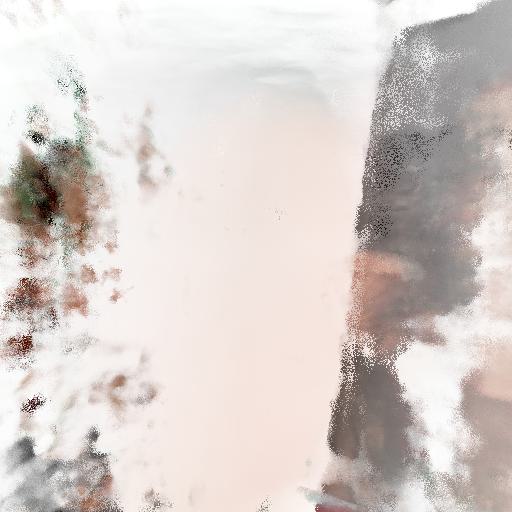}
    \includegraphics[width=0.09\textwidth]{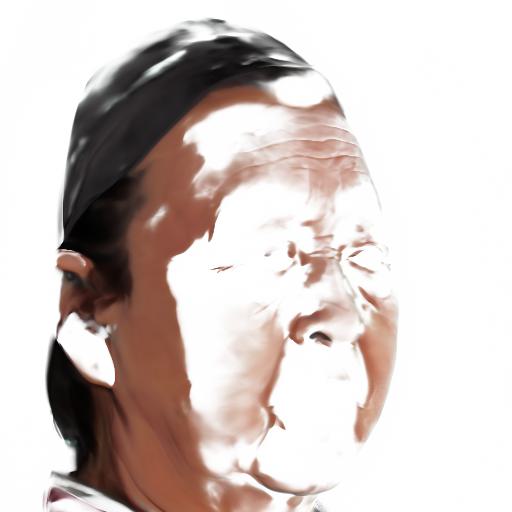}
    \includegraphics[width=0.09\textwidth]{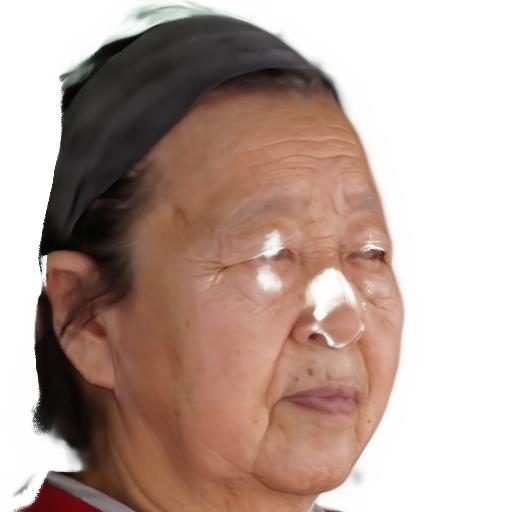}
    \rotatebox{90}{\tiny}
    \includegraphics[width=0.09\textwidth]{./results/template_effects/gt_571_blank.jpg}
    \includegraphics[width=0.09\textwidth]{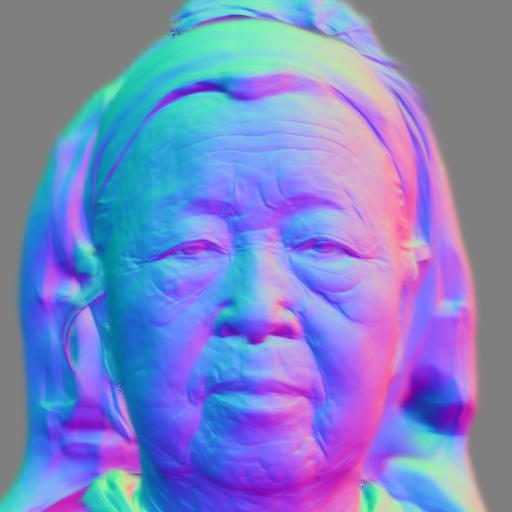}
    \includegraphics[width=0.09\textwidth]{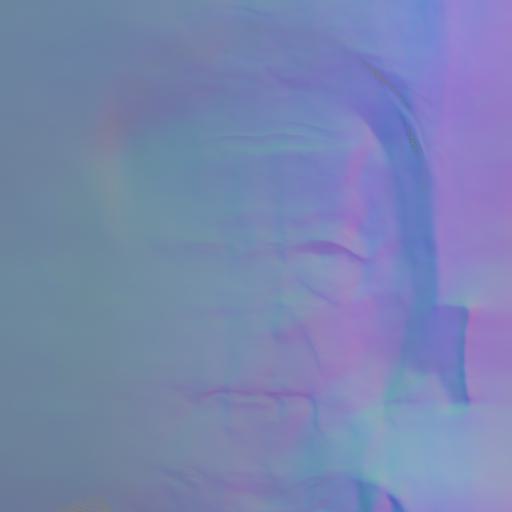}
    \includegraphics[width=0.09\textwidth]{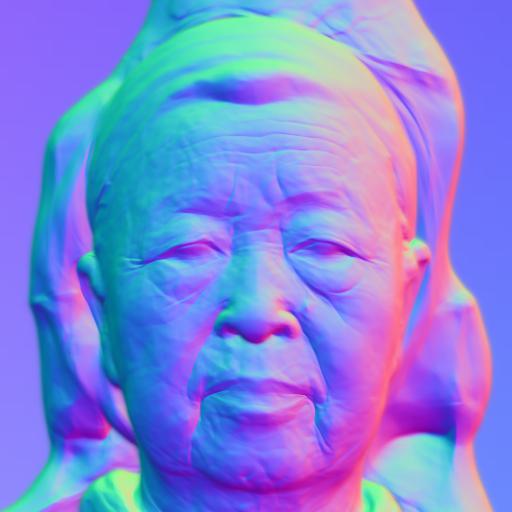}
    \includegraphics[width=0.09\textwidth]{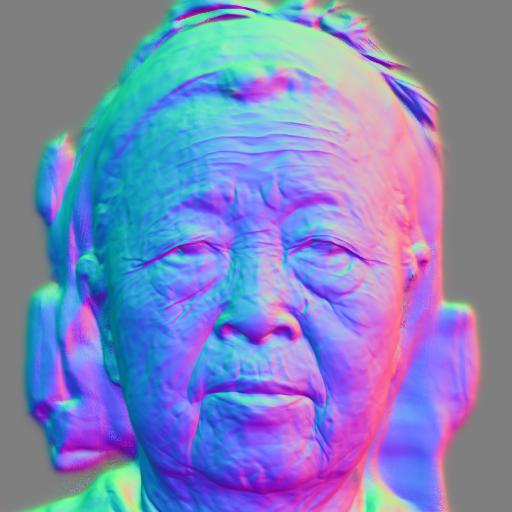}
    \includegraphics[width=0.09\textwidth]{./results/template_effects/gt_571_blank.jpg}
    \includegraphics[width=0.09\textwidth]{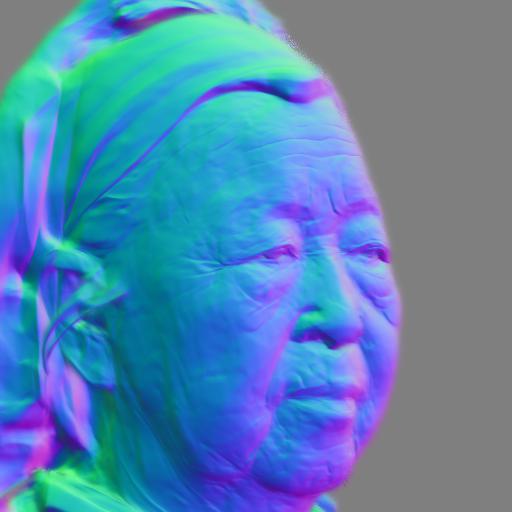}
    \includegraphics[width=0.09\textwidth]{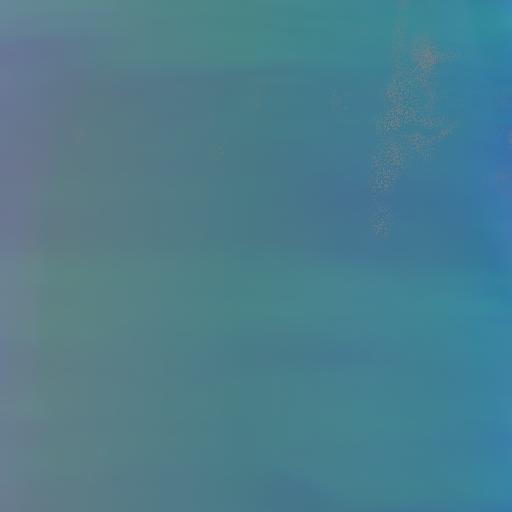}
    \includegraphics[width=0.09\textwidth]{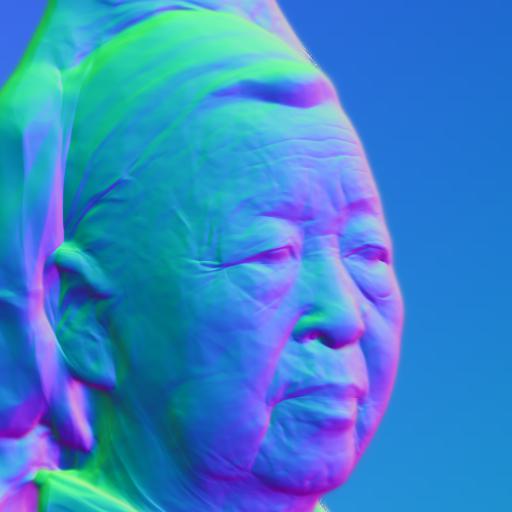}
    \includegraphics[width=0.09\textwidth]{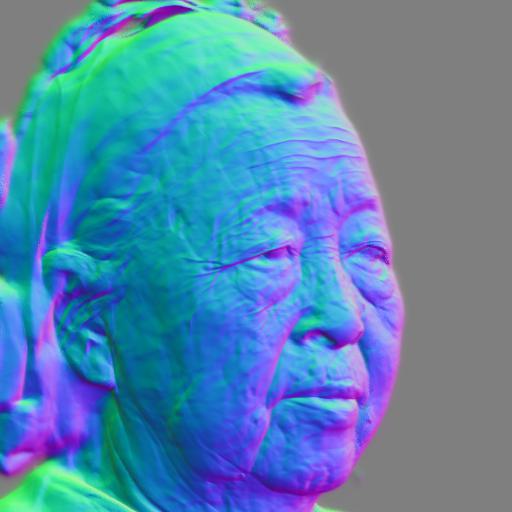}\\
    \makebox[0.09\textwidth]{GT}
    \makebox[0.09\textwidth]{NeuS}
    \makebox[0.09\textwidth]{HF-NeuS}
    \makebox[0.09\textwidth]{VolSDF}
    \makebox[0.09\textwidth]{Ours}
    \makebox[0.09\textwidth]{GT}
    \makebox[0.09\textwidth]{NeuS}
    \makebox[0.09\textwidth]{HF-NeuS}
    \makebox[0.09\textwidth]{VolSDF}
    \makebox[0.09\textwidth]{Ours}\\
    \caption{Comparison of various approaches under a 10-view setting (from Model 416 to Model 451).
    For each model, we show the results on one training view (left) and one novel view (right).}
    \label{fig:all_result3}
\end{figure*}
\begin{figure*}[htbp]
    \centering
    \rotatebox{90}{\textbf{454}}
    \includegraphics[width=0.09\textwidth]{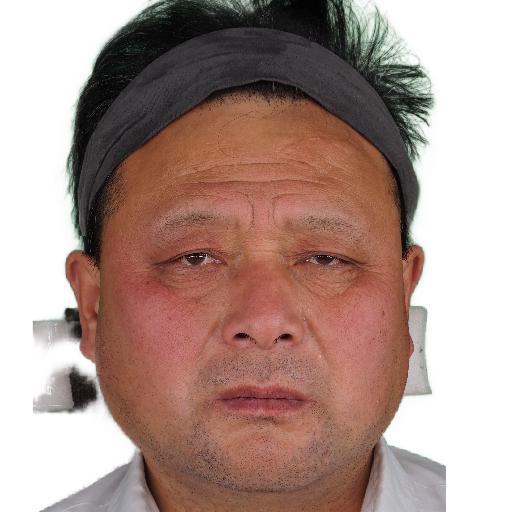}
    \includegraphics[width=0.09\textwidth]{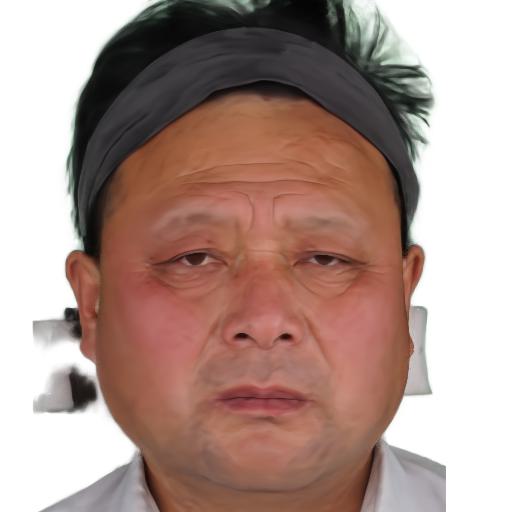}
    \includegraphics[width=0.09\textwidth]{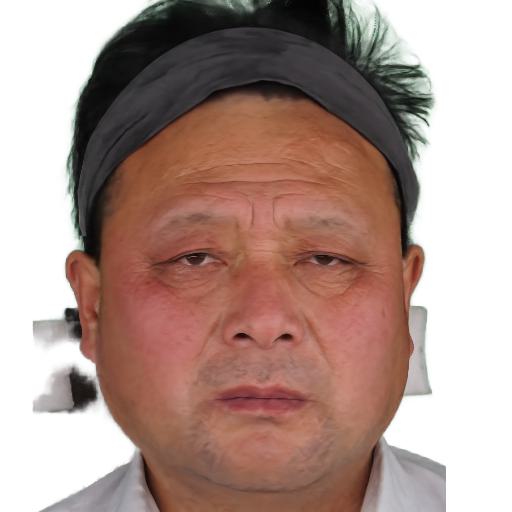}
    \includegraphics[width=0.09\textwidth]{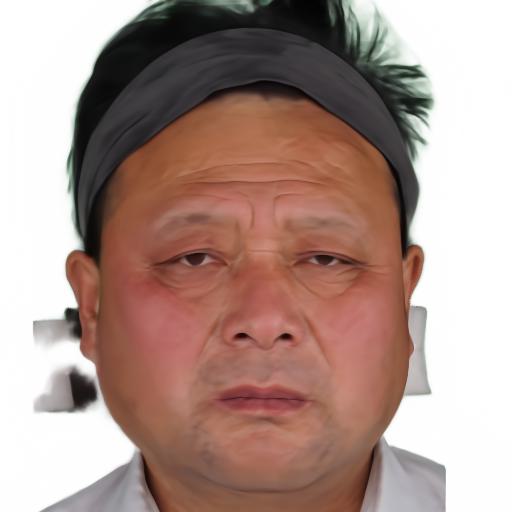}
    \includegraphics[width=0.09\textwidth]{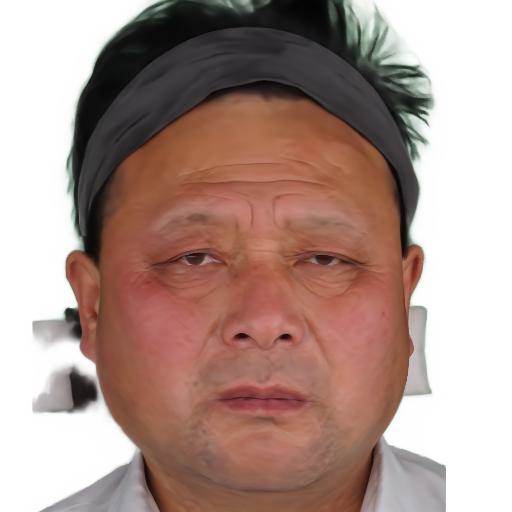}
    \includegraphics[width=0.09\textwidth]{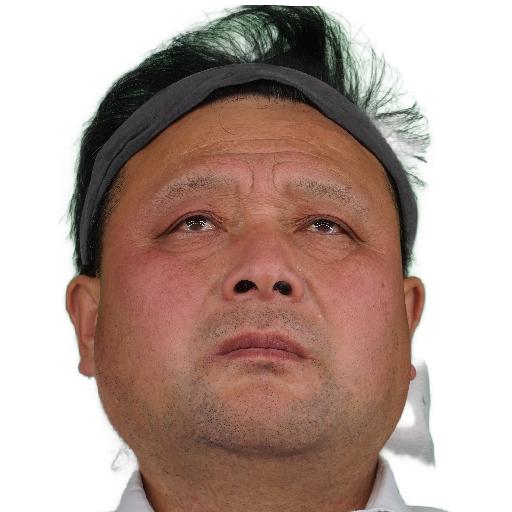}
    \includegraphics[width=0.09\textwidth]{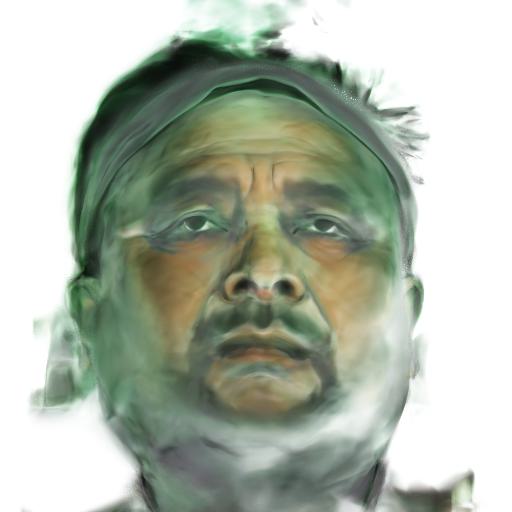}
    \includegraphics[width=0.09\textwidth]{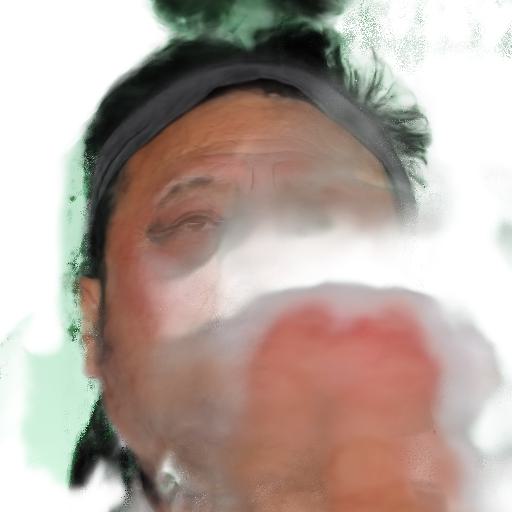}
    \includegraphics[width=0.09\textwidth]{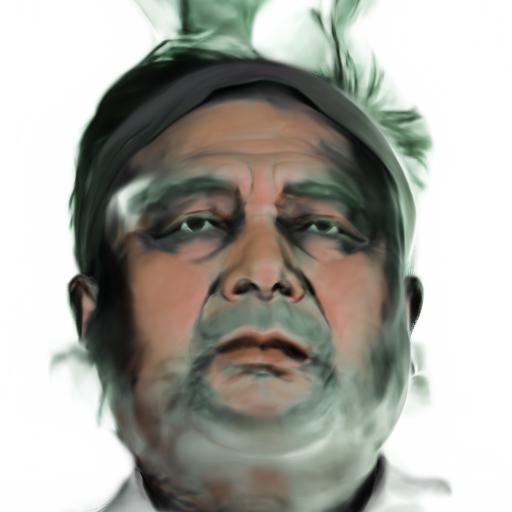}
    \includegraphics[width=0.09\textwidth]{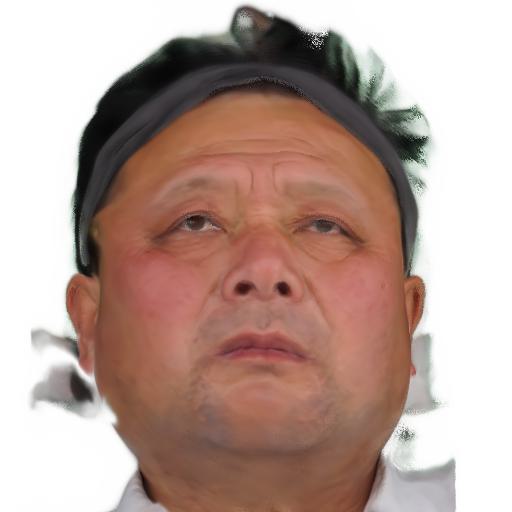}
    \rotatebox{90}{\tiny}
    \includegraphics[width=0.09\textwidth]{./results/template_effects/gt_571_blank.jpg}
    \includegraphics[width=0.09\textwidth]{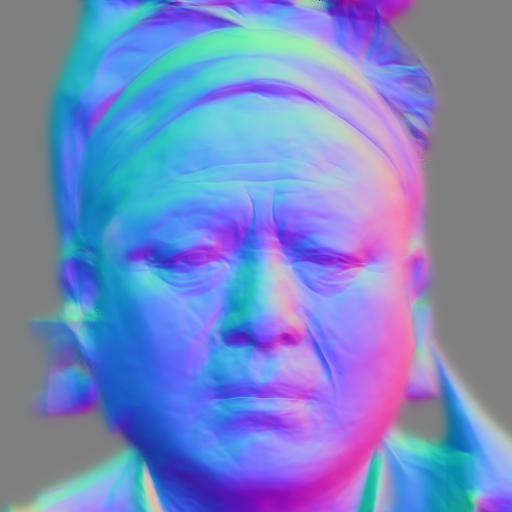}
    \includegraphics[width=0.09\textwidth]{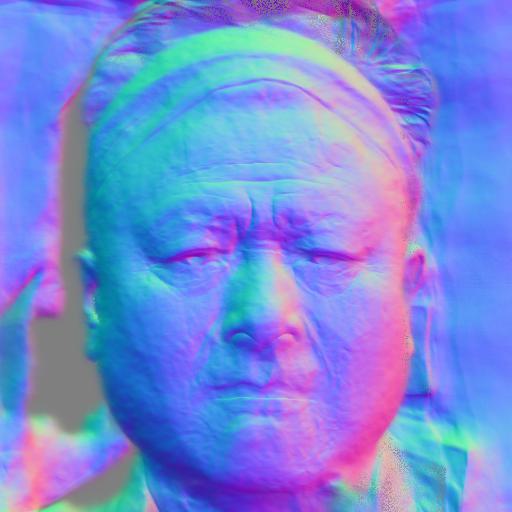}
    \includegraphics[width=0.09\textwidth]{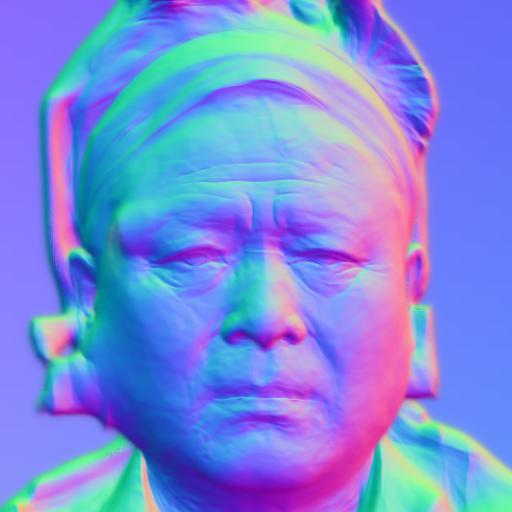}
    \includegraphics[width=0.09\textwidth]{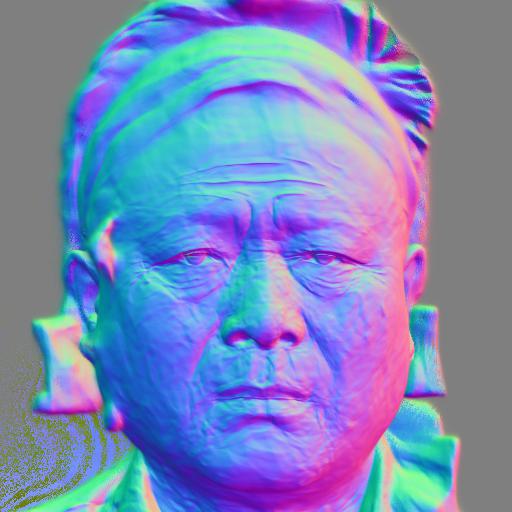}
    \includegraphics[width=0.09\textwidth]{./results/template_effects/gt_571_blank.jpg}
    \includegraphics[width=0.09\textwidth]{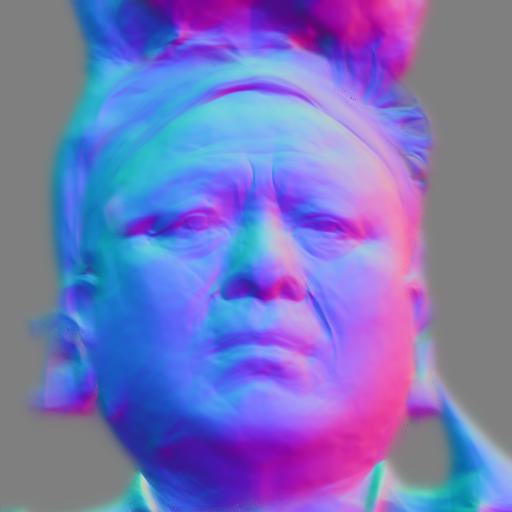}
    \includegraphics[width=0.09\textwidth]{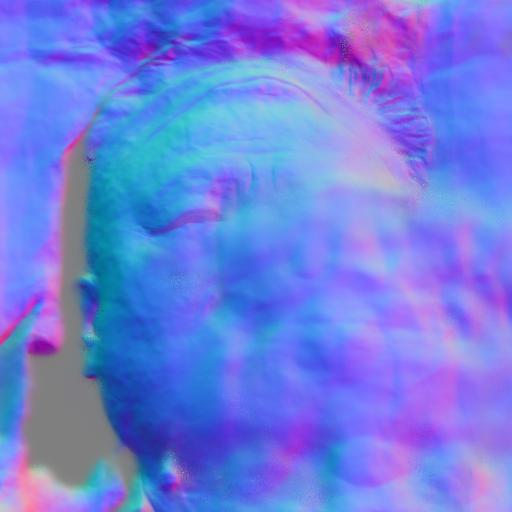}
    \includegraphics[width=0.09\textwidth]{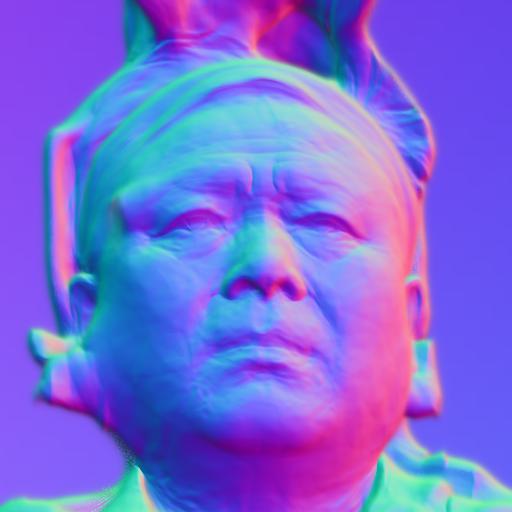}
    \includegraphics[width=0.09\textwidth]{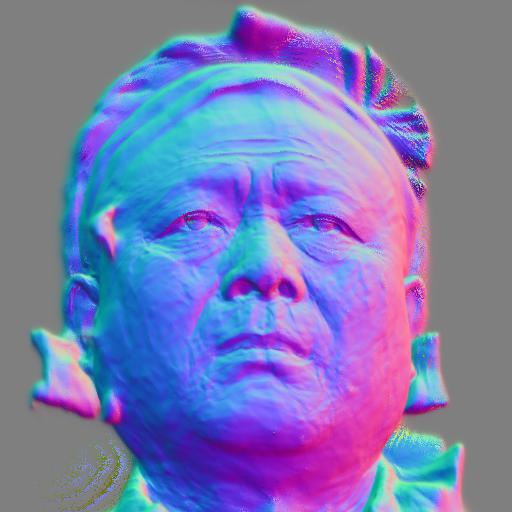}\\
    \rotatebox{90}{\textbf{469}}
    \includegraphics[width=0.09\textwidth]{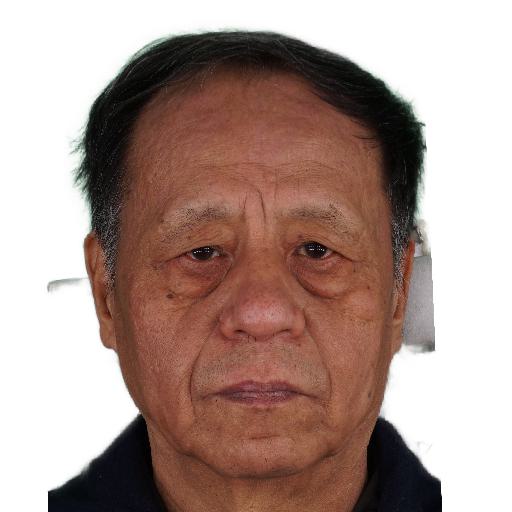}
    \includegraphics[width=0.09\textwidth]{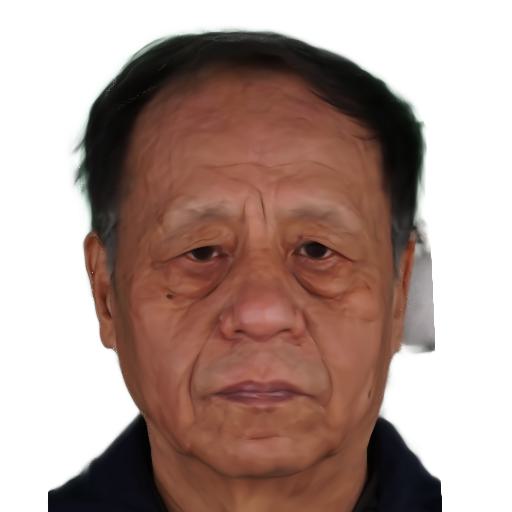}
    \includegraphics[width=0.09\textwidth]{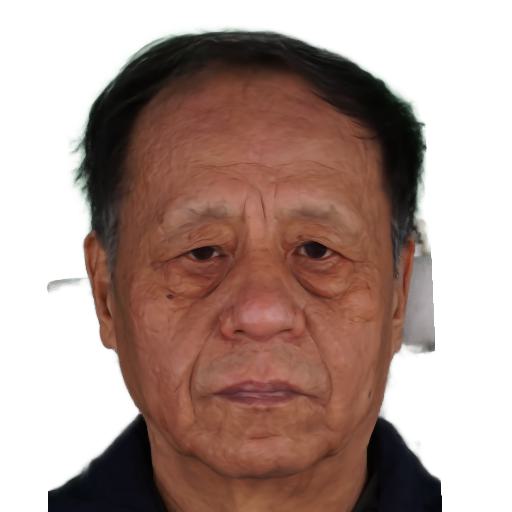}
    \includegraphics[width=0.09\textwidth]{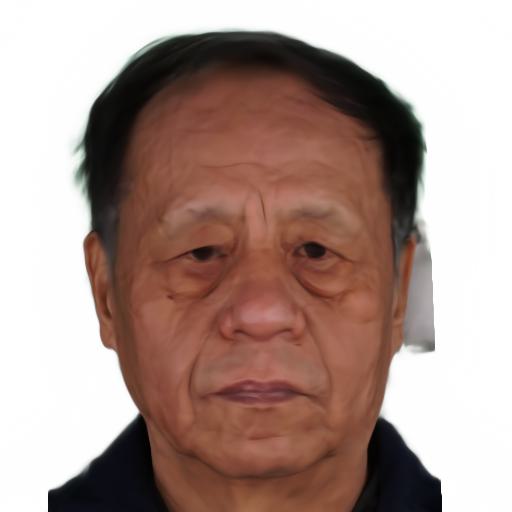}
    \includegraphics[width=0.09\textwidth]{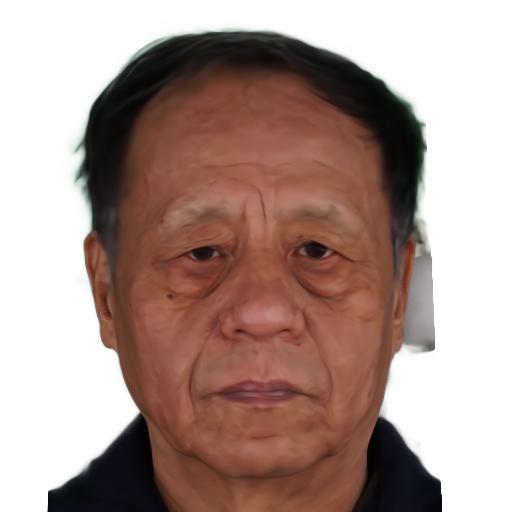}
    \includegraphics[width=0.09\textwidth]{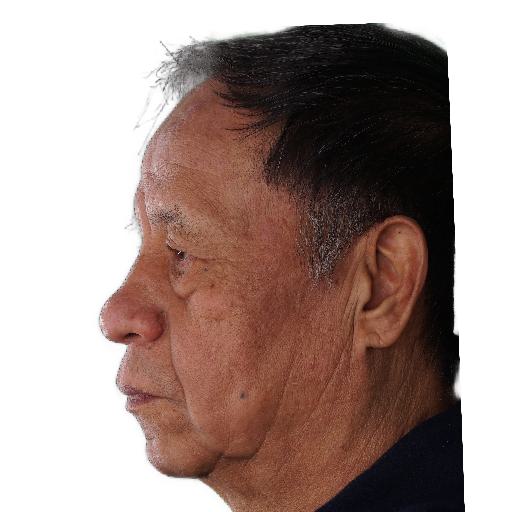}
    \includegraphics[width=0.09\textwidth]{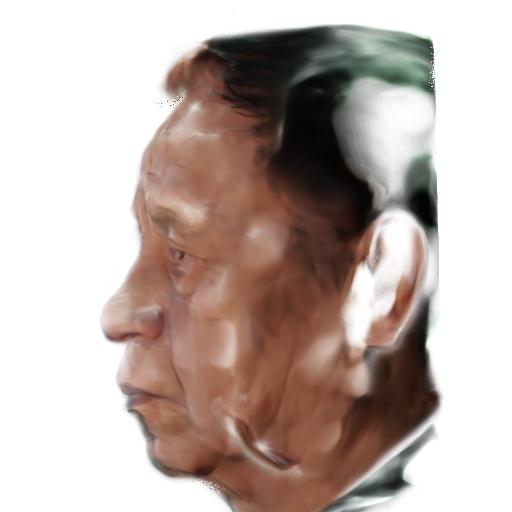}
    \includegraphics[width=0.09\textwidth]{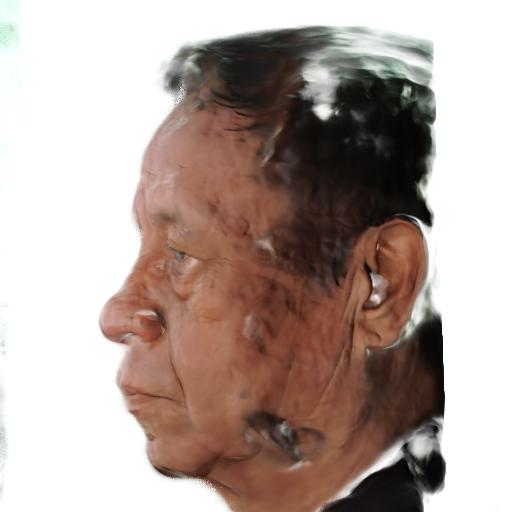}
    \includegraphics[width=0.09\textwidth]{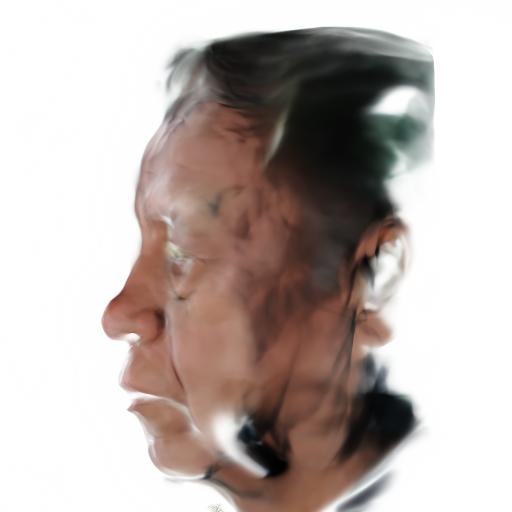}
    \includegraphics[width=0.09\textwidth]{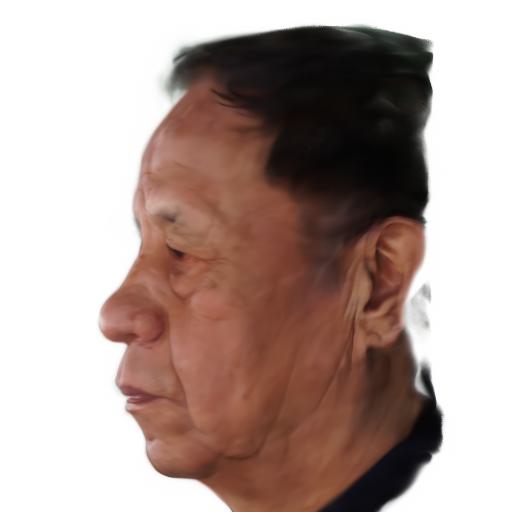}
    \rotatebox{90}{\tiny}
    \includegraphics[width=0.09\textwidth]{./results/template_effects/gt_571_blank.jpg}
    \includegraphics[width=0.09\textwidth]{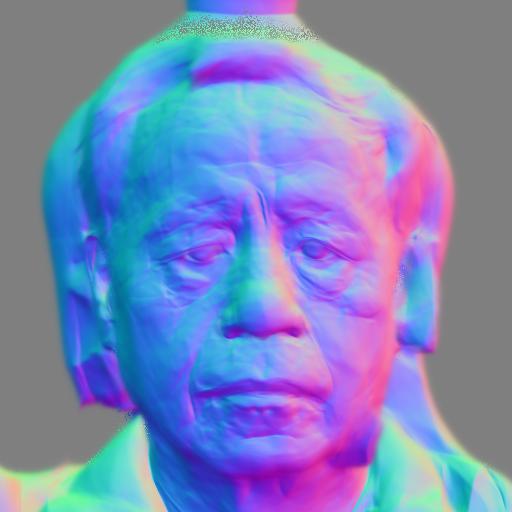}
    \includegraphics[width=0.09\textwidth]{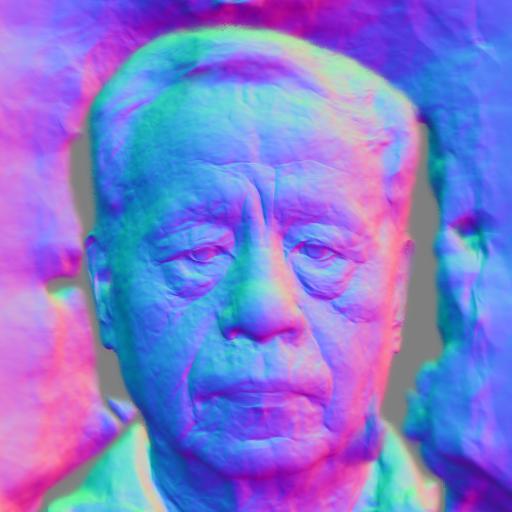}
    \includegraphics[width=0.09\textwidth]{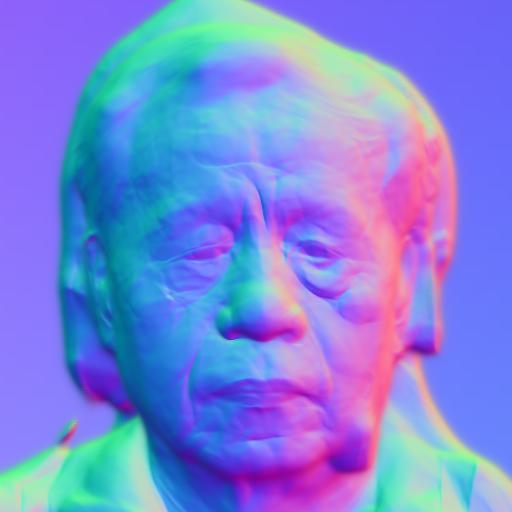}
    \includegraphics[width=0.09\textwidth]{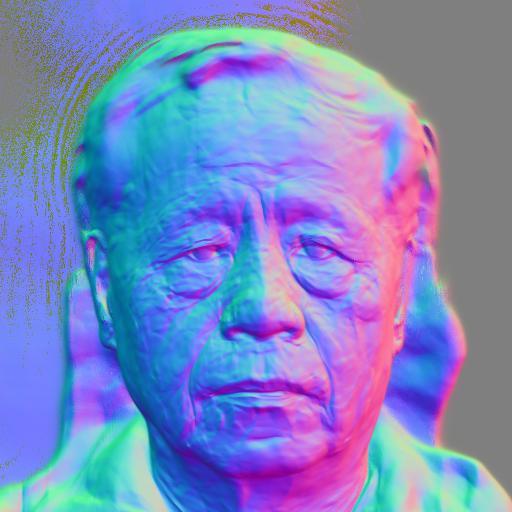}
    \includegraphics[width=0.09\textwidth]{./results/template_effects/gt_571_blank.jpg}
    \includegraphics[width=0.09\textwidth]{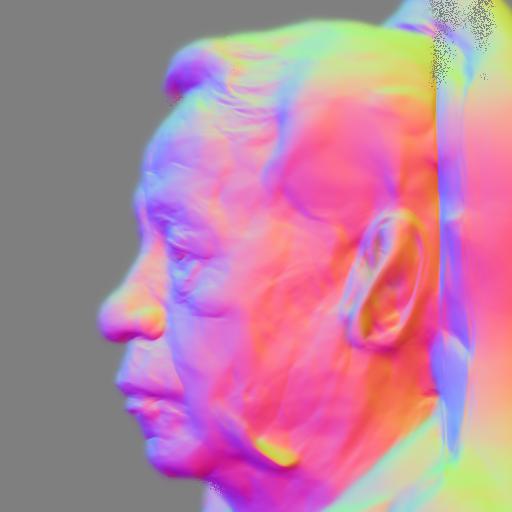}
    \includegraphics[width=0.09\textwidth]{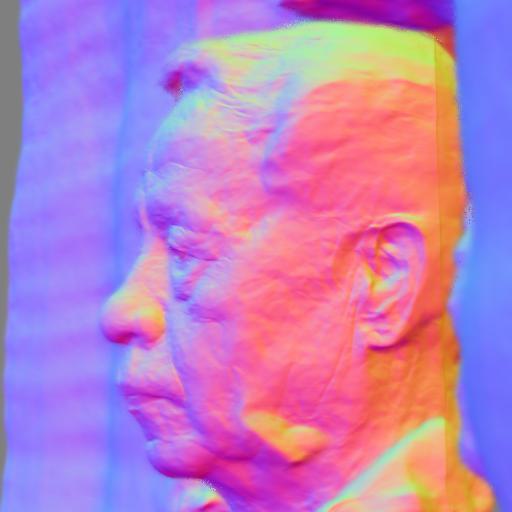}
    \includegraphics[width=0.09\textwidth]{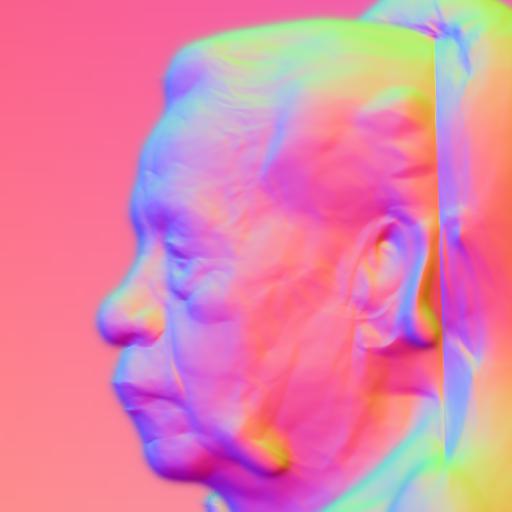}
    \includegraphics[width=0.09\textwidth]{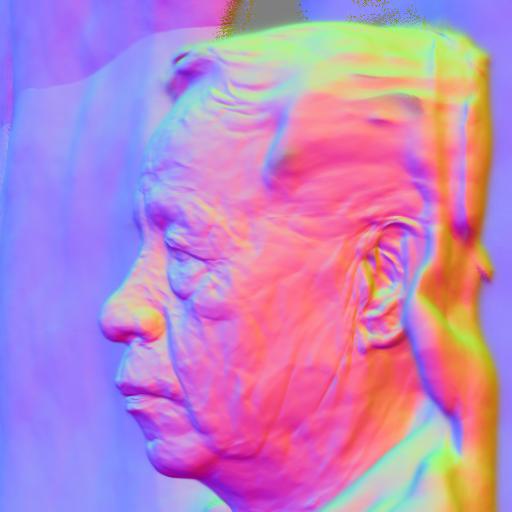}\\
    \rotatebox{90}{\textbf{470}}
    \includegraphics[width=0.09\textwidth]{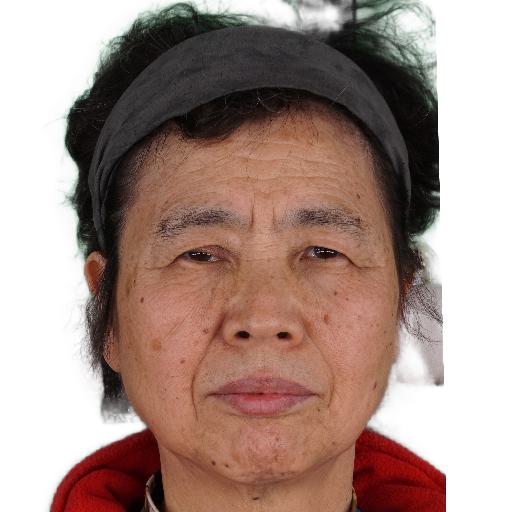}
    \includegraphics[width=0.09\textwidth]{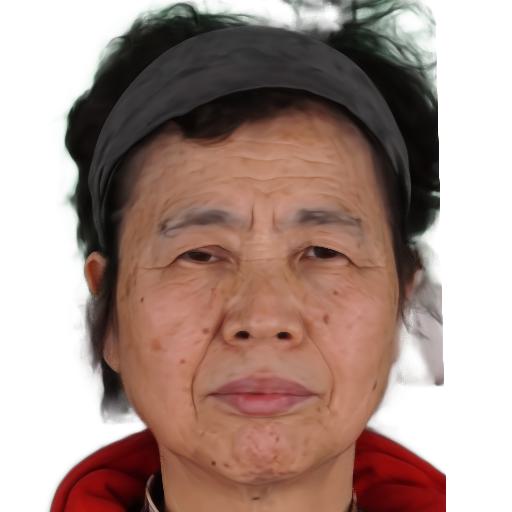}
    \includegraphics[width=0.09\textwidth]{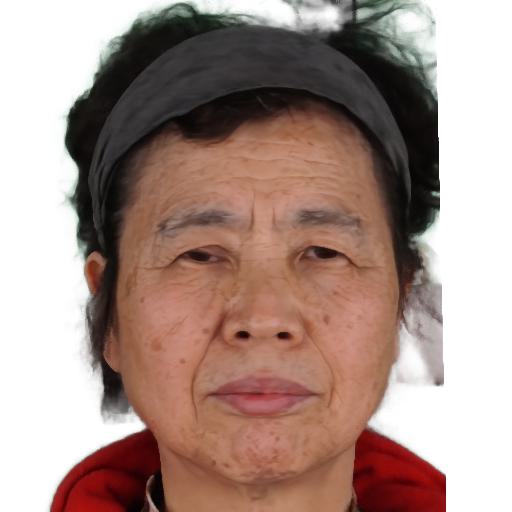}
    \includegraphics[width=0.09\textwidth]{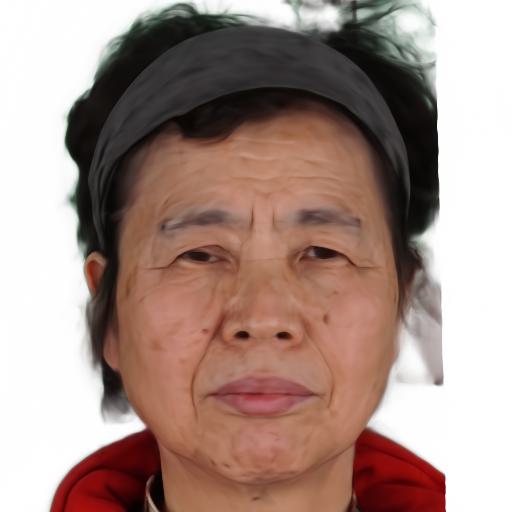}
    \includegraphics[width=0.09\textwidth]{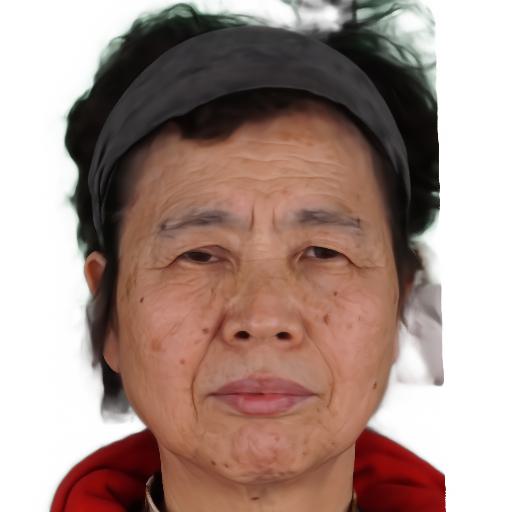}
    \includegraphics[width=0.09\textwidth]{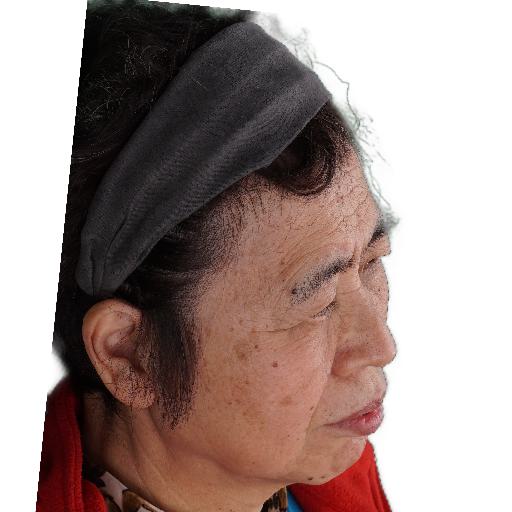}
    \includegraphics[width=0.09\textwidth]{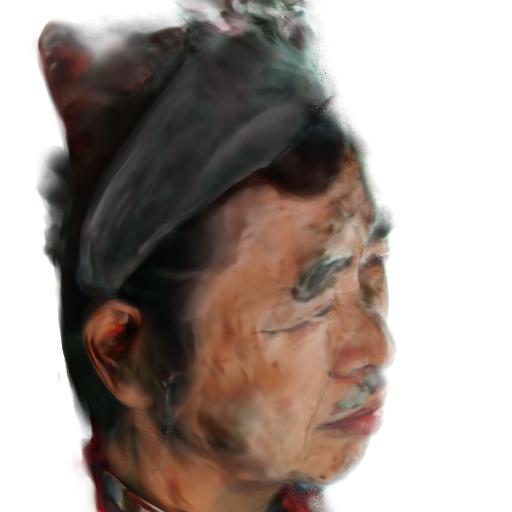}
    \includegraphics[width=0.09\textwidth]{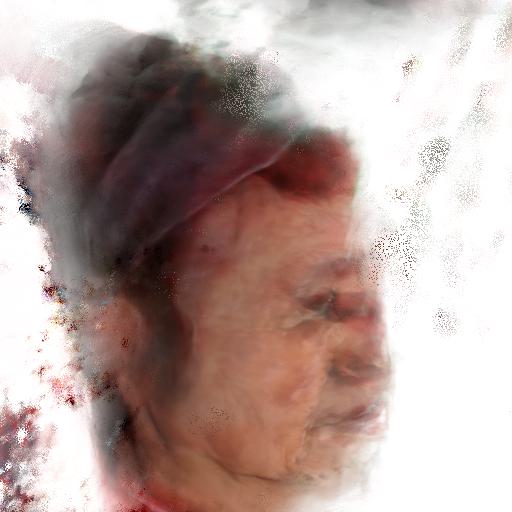}
    \includegraphics[width=0.09\textwidth]{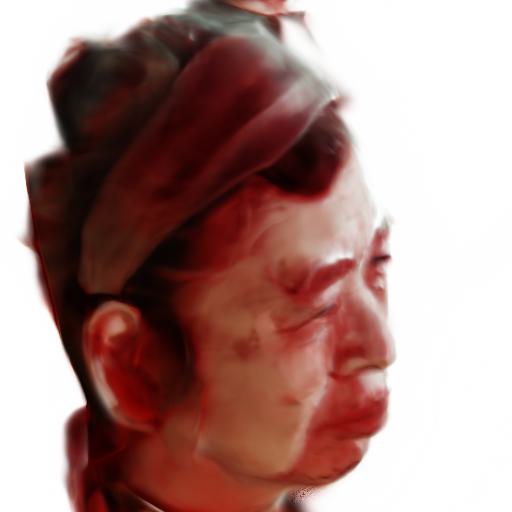}
    \includegraphics[width=0.09\textwidth]{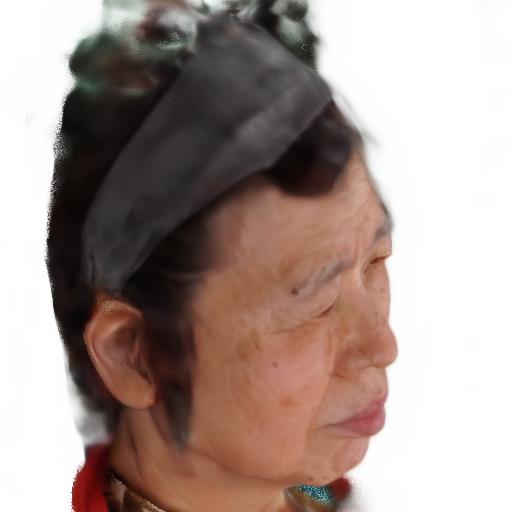}
    \rotatebox{90}{\tiny}
    \includegraphics[width=0.09\textwidth]{./results/template_effects/gt_571_blank.jpg}
    \includegraphics[width=0.09\textwidth]{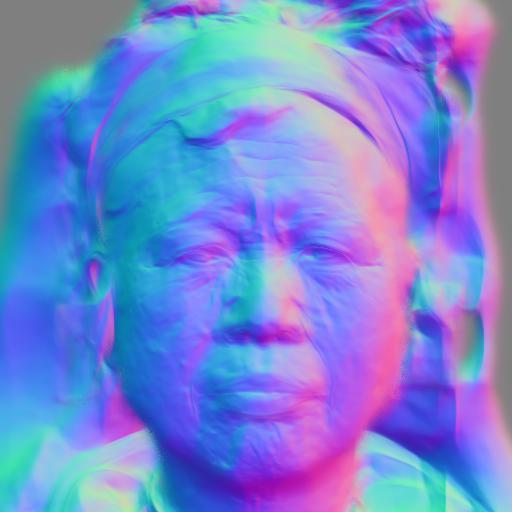}
    \includegraphics[width=0.09\textwidth]{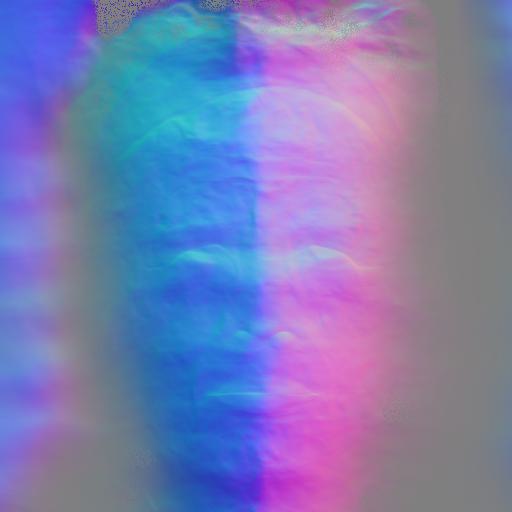}
    \includegraphics[width=0.09\textwidth]{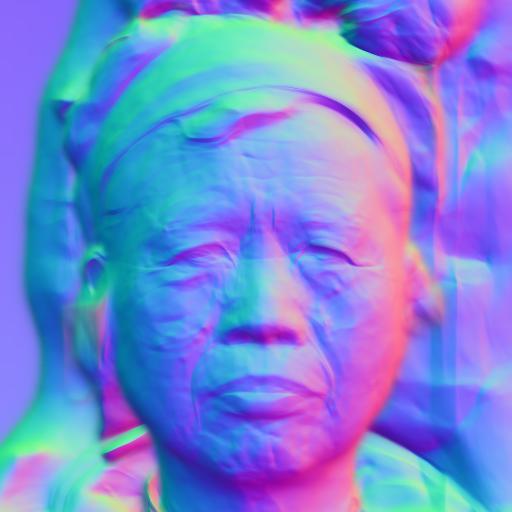}
    \includegraphics[width=0.09\textwidth]{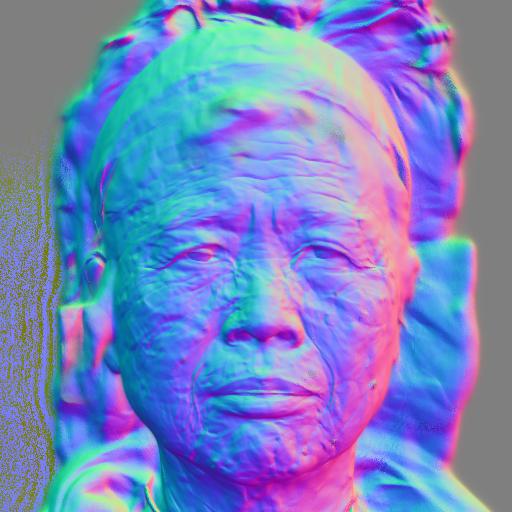}
    \includegraphics[width=0.09\textwidth]{./results/template_effects/gt_571_blank.jpg}
    \includegraphics[width=0.09\textwidth]{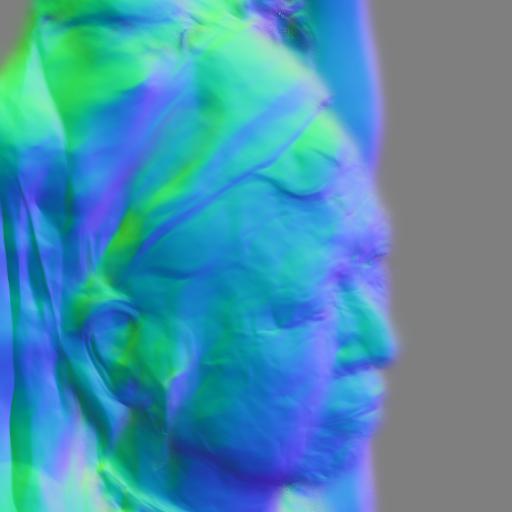}
    \includegraphics[width=0.09\textwidth]{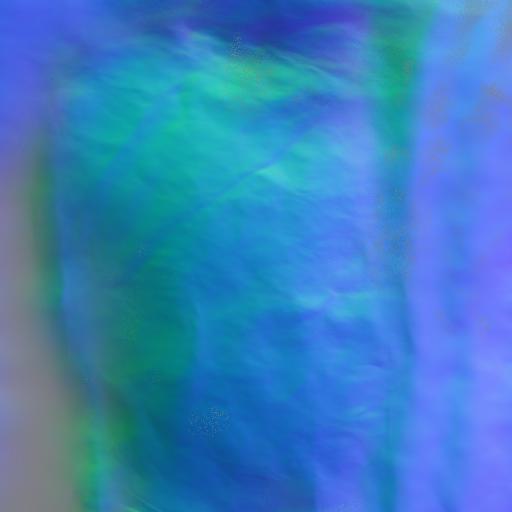}
    \includegraphics[width=0.09\textwidth]{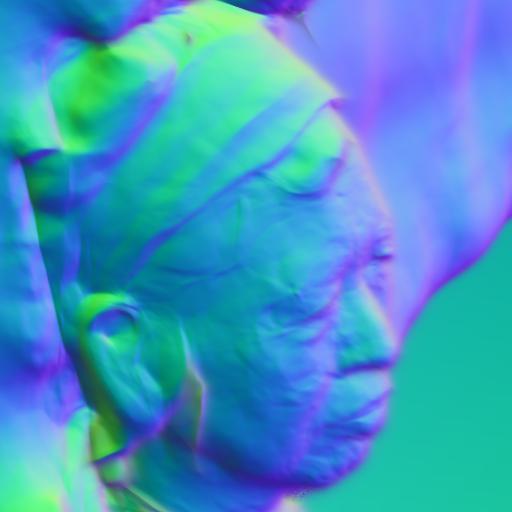}
    \includegraphics[width=0.09\textwidth]{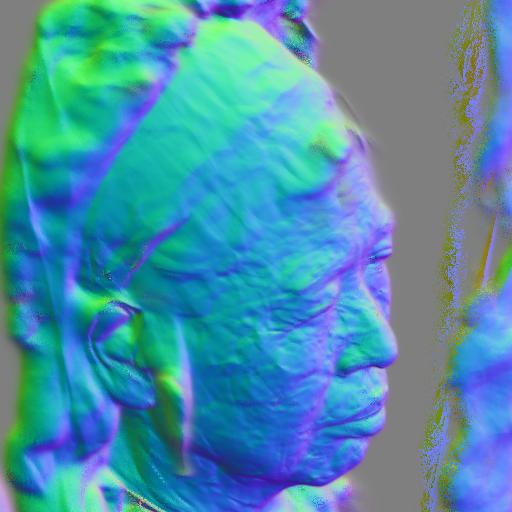}\\
    \rotatebox{90}{\textbf{487}}
    \includegraphics[width=0.09\textwidth]{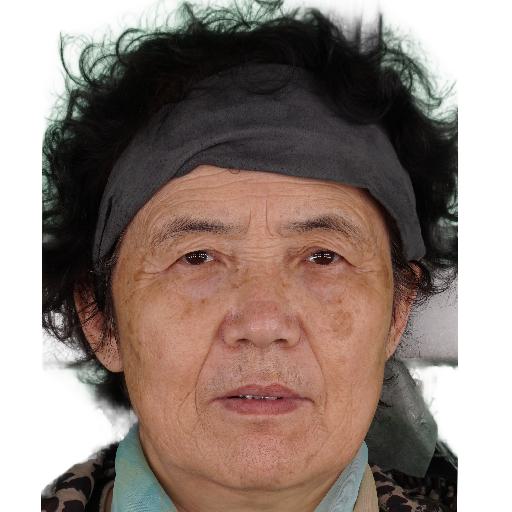}
    \includegraphics[width=0.09\textwidth]{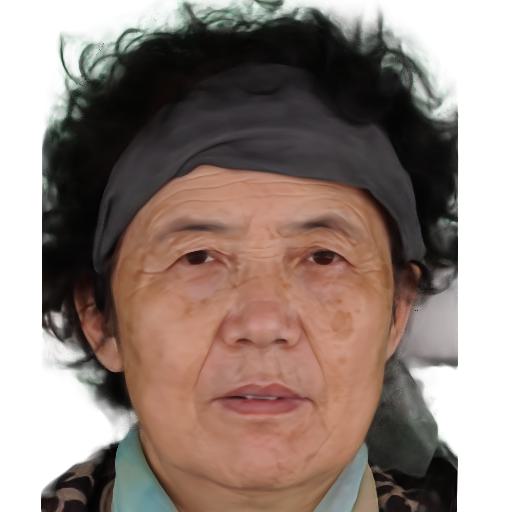}
    \includegraphics[width=0.09\textwidth]{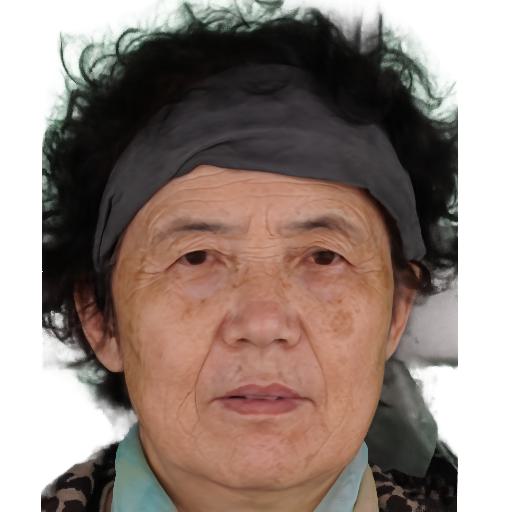}
    \includegraphics[width=0.09\textwidth]{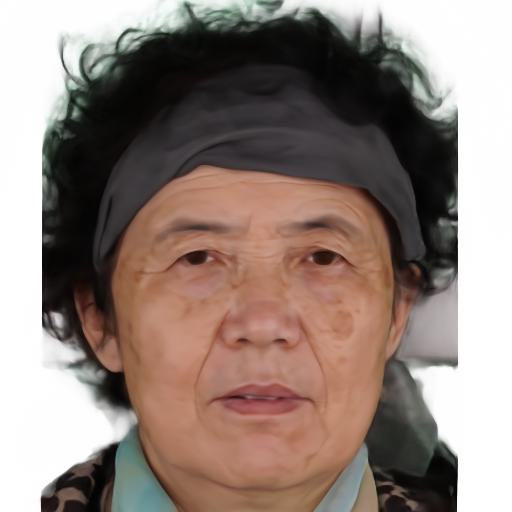}
    \includegraphics[width=0.09\textwidth]{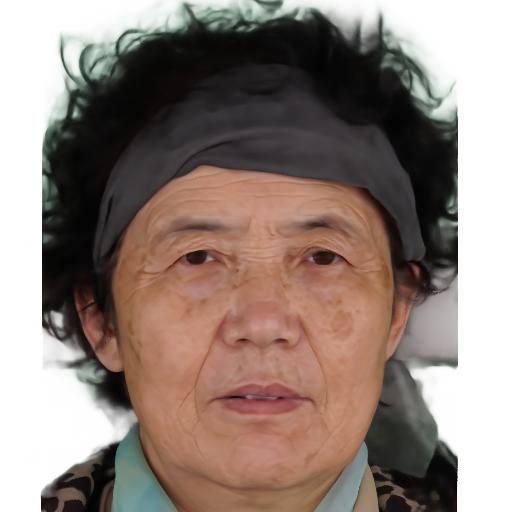}
    \includegraphics[width=0.09\textwidth]{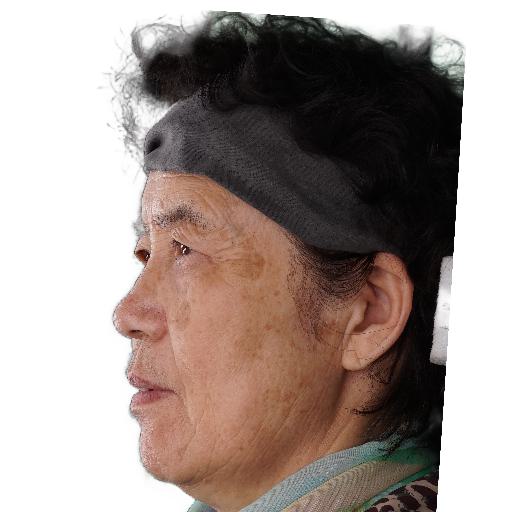}
    \includegraphics[width=0.09\textwidth]{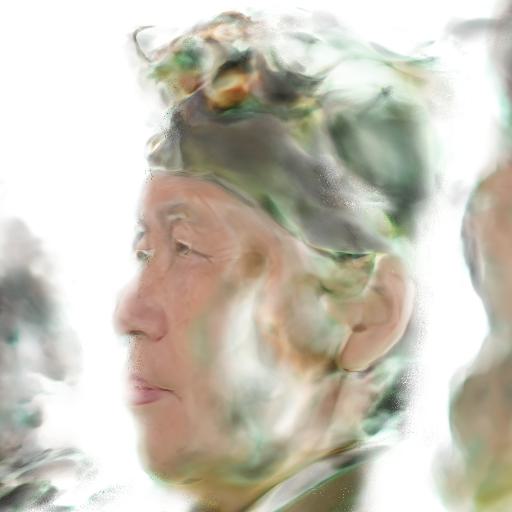}
    \includegraphics[width=0.09\textwidth]{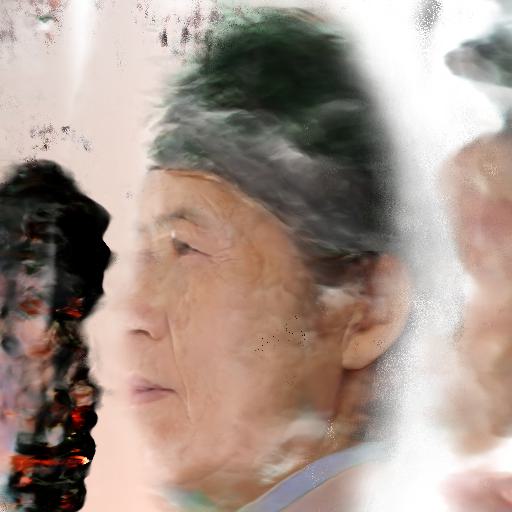}
    \includegraphics[width=0.09\textwidth]{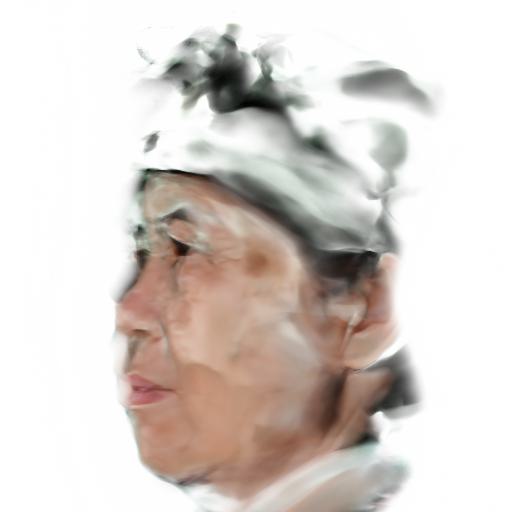}
    \includegraphics[width=0.09\textwidth]{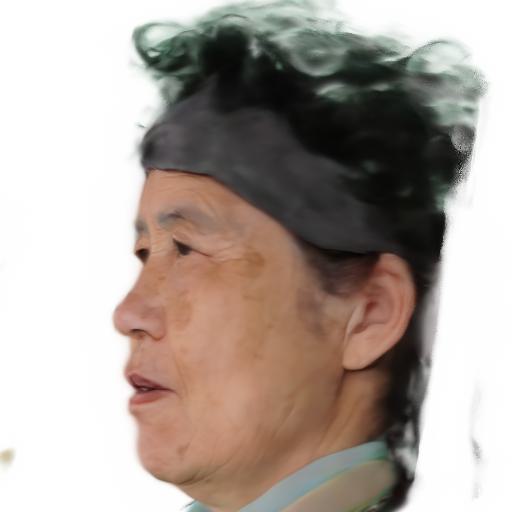}
    \rotatebox{90}{\tiny}
    \includegraphics[width=0.09\textwidth]{./results/template_effects/gt_571_blank.jpg}
    \includegraphics[width=0.09\textwidth]{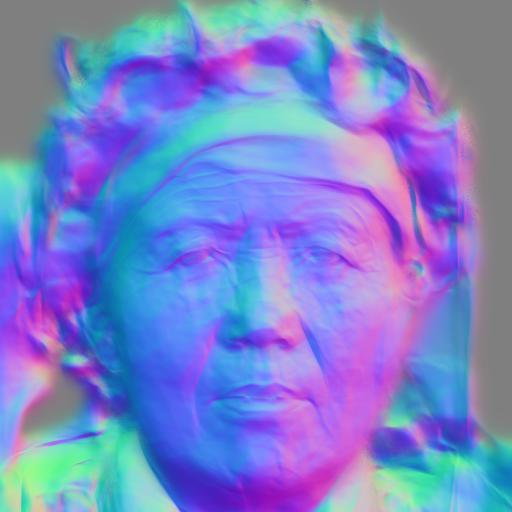}
    \includegraphics[width=0.09\textwidth]{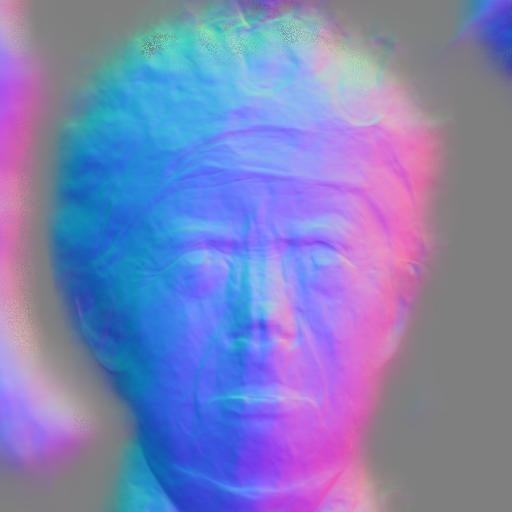}
    \includegraphics[width=0.09\textwidth]{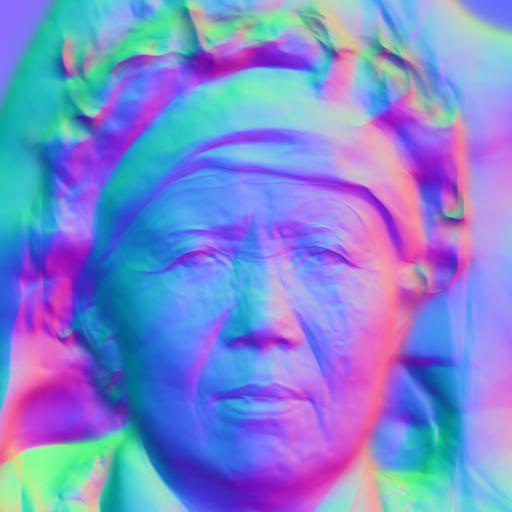}
    \includegraphics[width=0.09\textwidth]{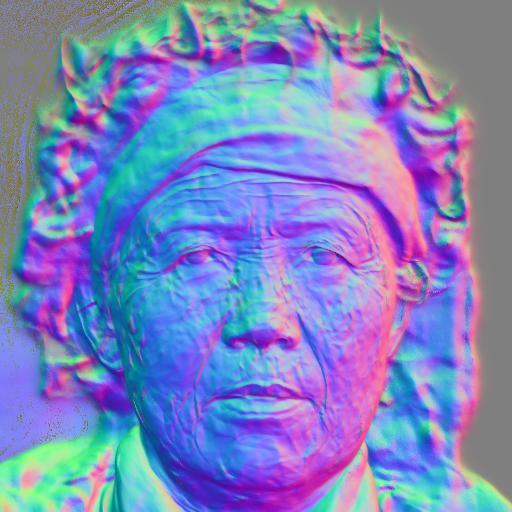}
    \includegraphics[width=0.09\textwidth]{./results/template_effects/gt_571_blank.jpg}
    \includegraphics[width=0.09\textwidth]{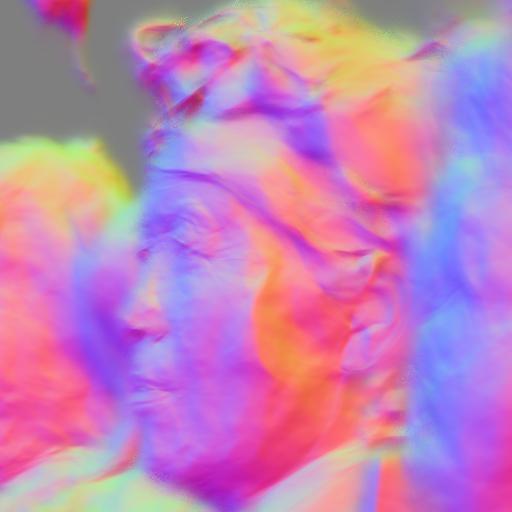}
    \includegraphics[width=0.09\textwidth]{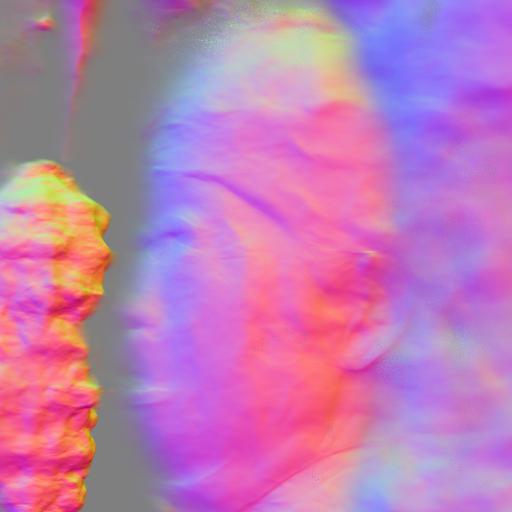}
    \includegraphics[width=0.09\textwidth]{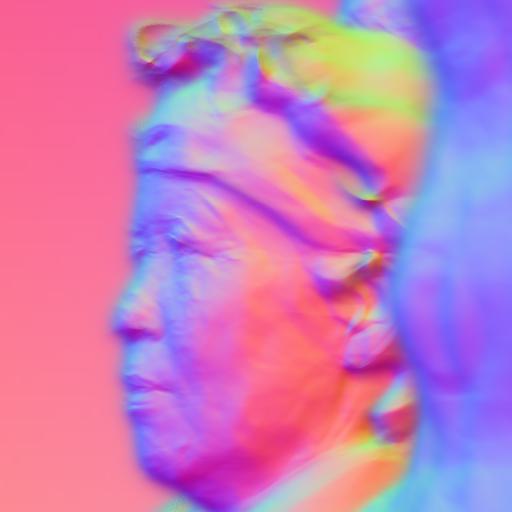}
    \includegraphics[width=0.09\textwidth]{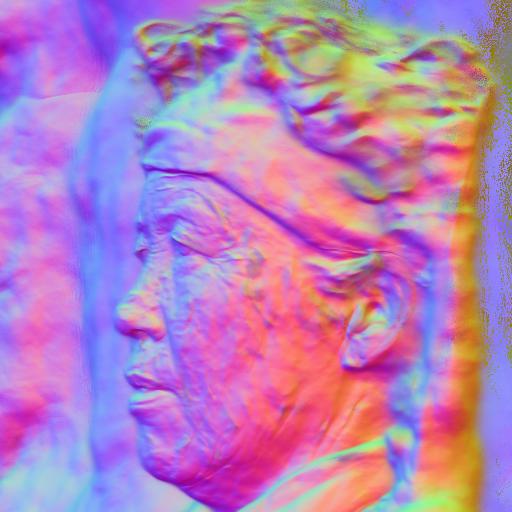}\\
    \rotatebox{90}{\textbf{491}}
    \includegraphics[width=0.09\textwidth]{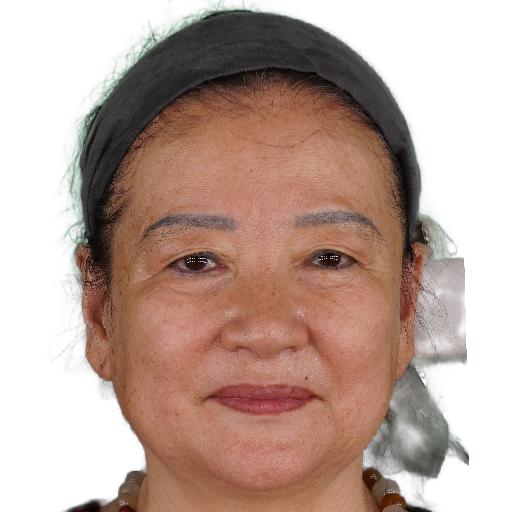}
    \includegraphics[width=0.09\textwidth]{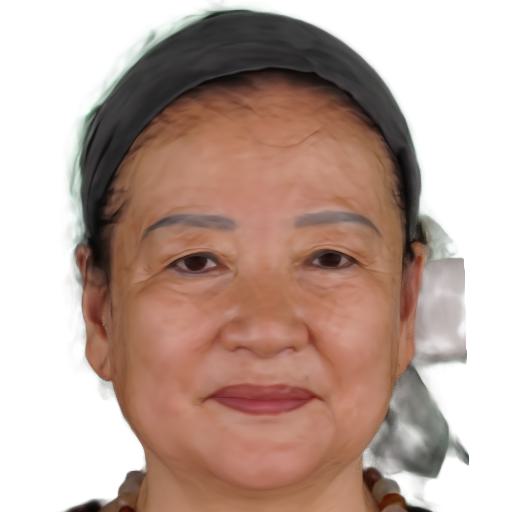}
    \includegraphics[width=0.09\textwidth]{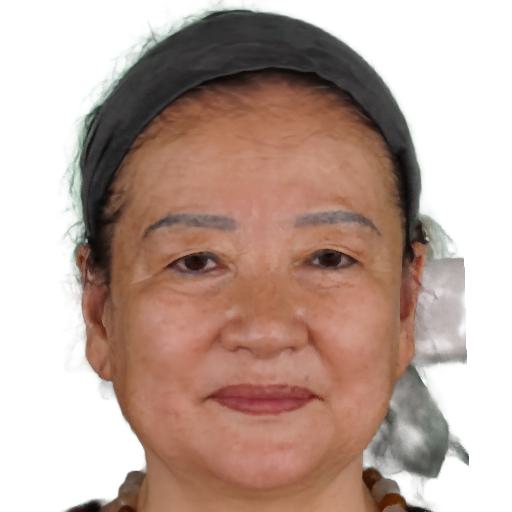}
    \includegraphics[width=0.09\textwidth]{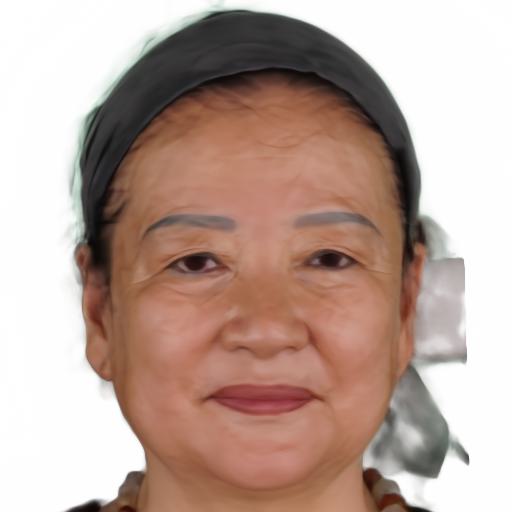}
    \includegraphics[width=0.09\textwidth]{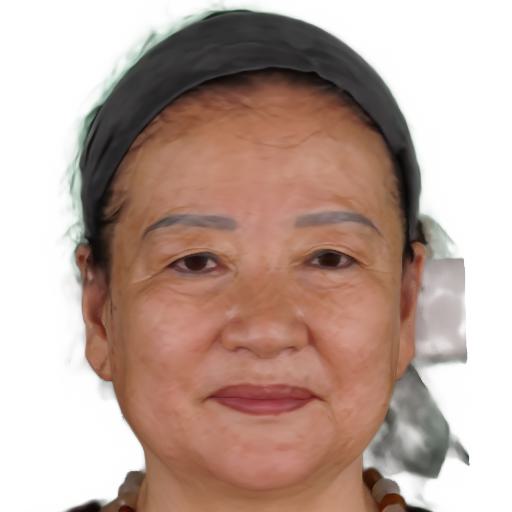}
    \includegraphics[width=0.09\textwidth]{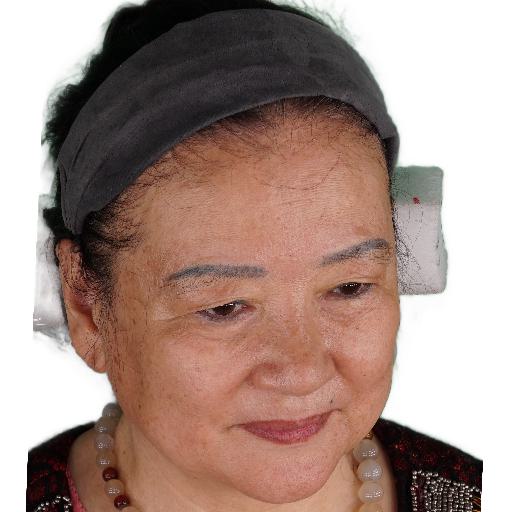}
    \includegraphics[width=0.09\textwidth]{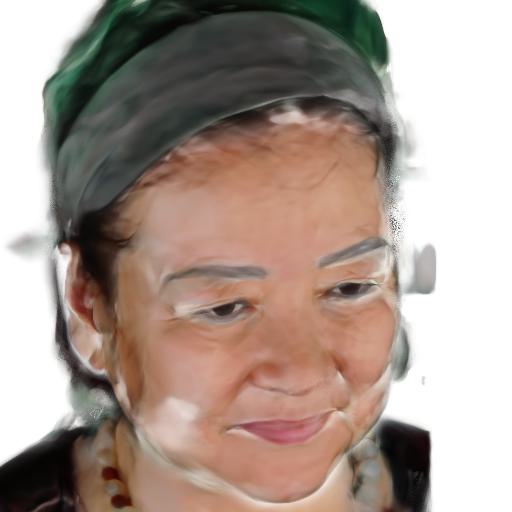}
    \includegraphics[width=0.09\textwidth]{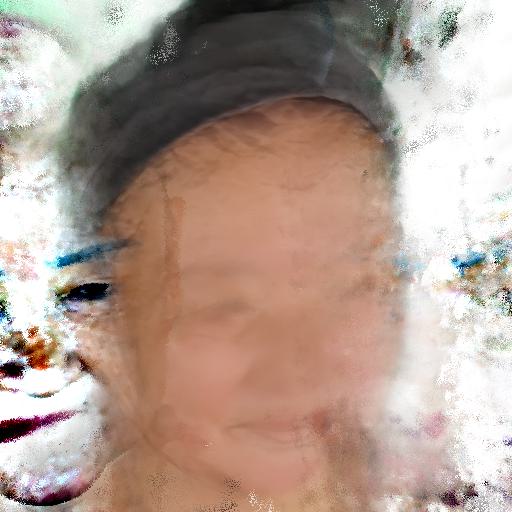}
    \includegraphics[width=0.09\textwidth]{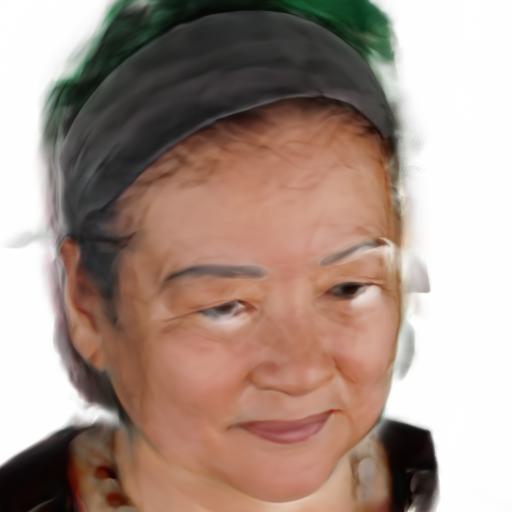}
    \includegraphics[width=0.09\textwidth]{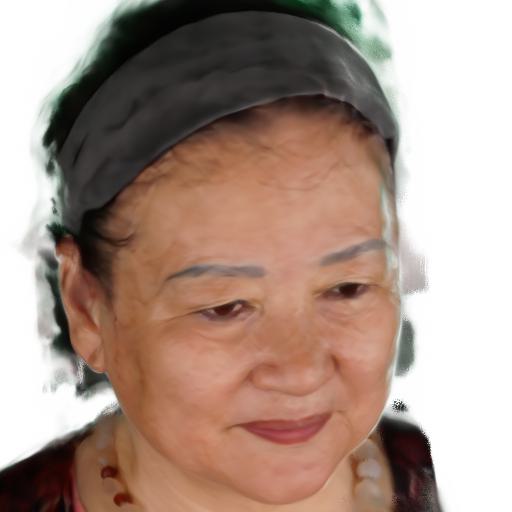}
    \rotatebox{90}{\tiny}
    \includegraphics[width=0.09\textwidth]{./results/template_effects/gt_571_blank.jpg}
    \includegraphics[width=0.09\textwidth]{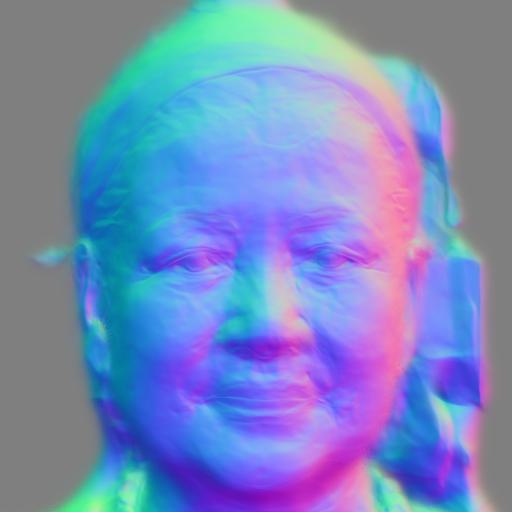}
    \includegraphics[width=0.09\textwidth]{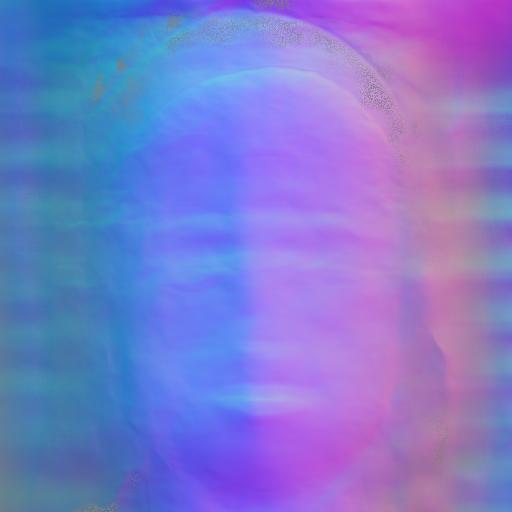}
    \includegraphics[width=0.09\textwidth]{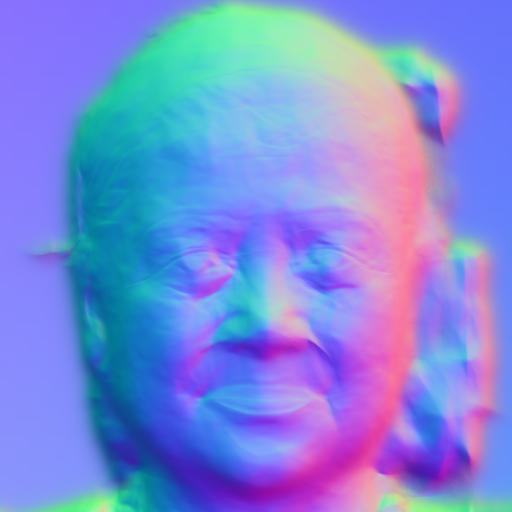}
    \includegraphics[width=0.09\textwidth]{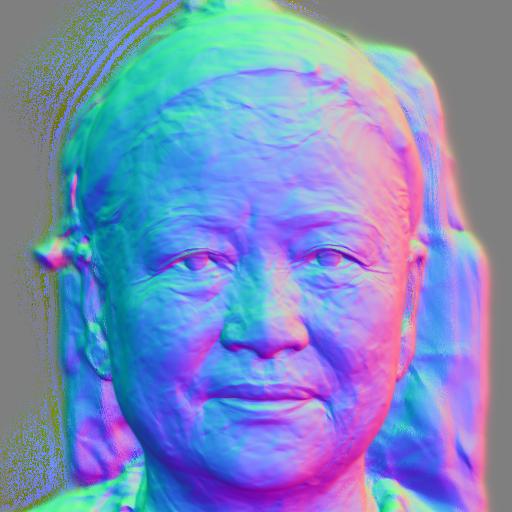}
    \includegraphics[width=0.09\textwidth]{./results/template_effects/gt_571_blank.jpg}
    \includegraphics[width=0.09\textwidth]{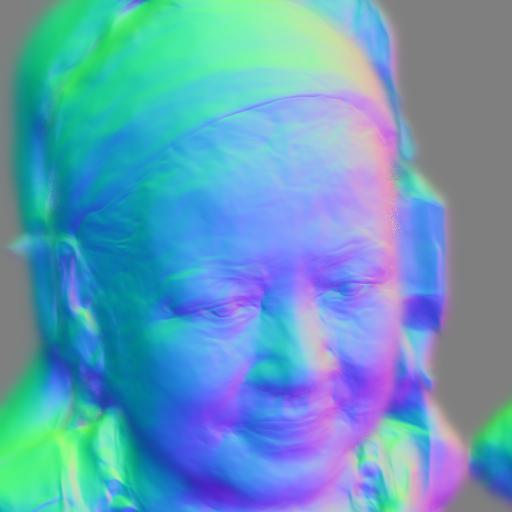}
    \includegraphics[width=0.09\textwidth]{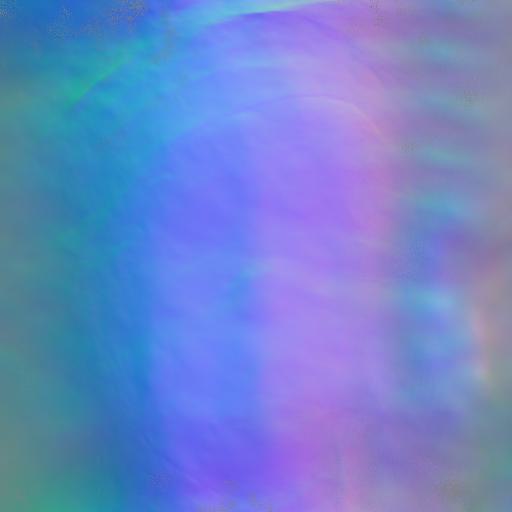}
    \includegraphics[width=0.09\textwidth]{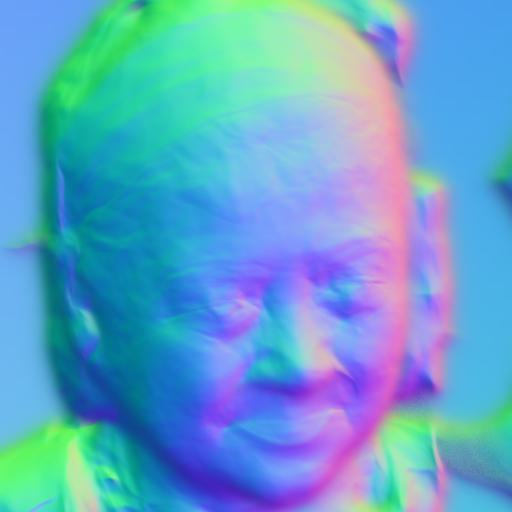}
    \includegraphics[width=0.09\textwidth]{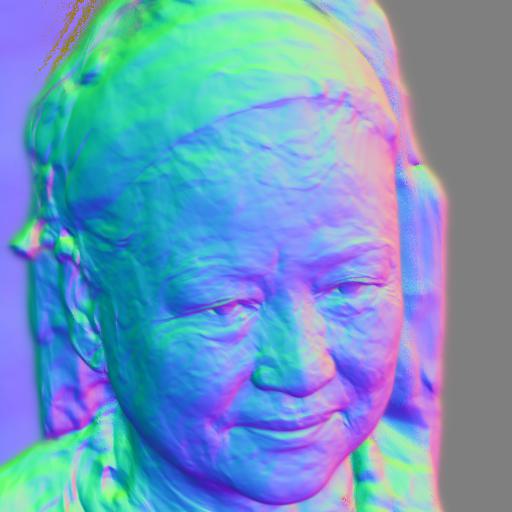}\\
    \rotatebox{90}{\textbf{548}}
    \includegraphics[width=0.09\textwidth]{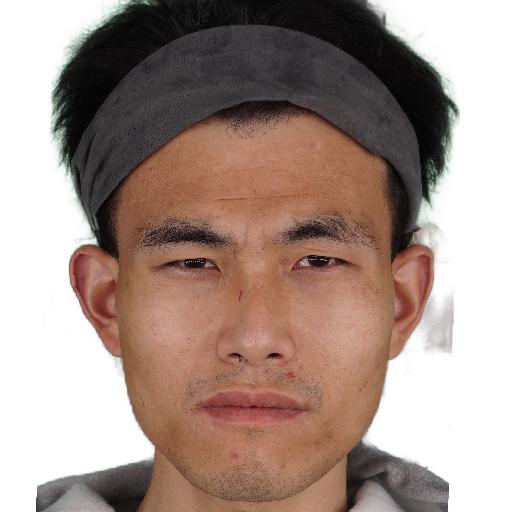}
    \includegraphics[width=0.09\textwidth]{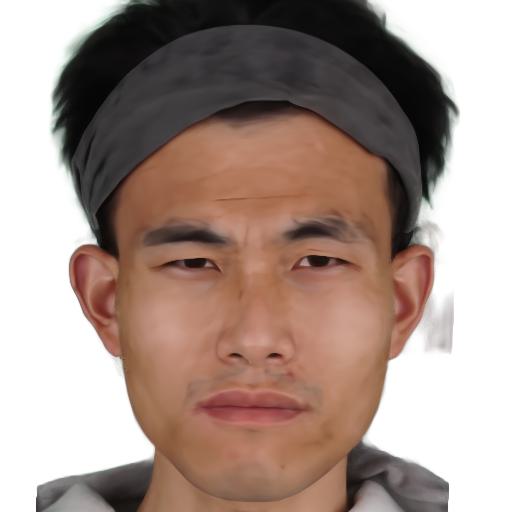}
    \includegraphics[width=0.09\textwidth]{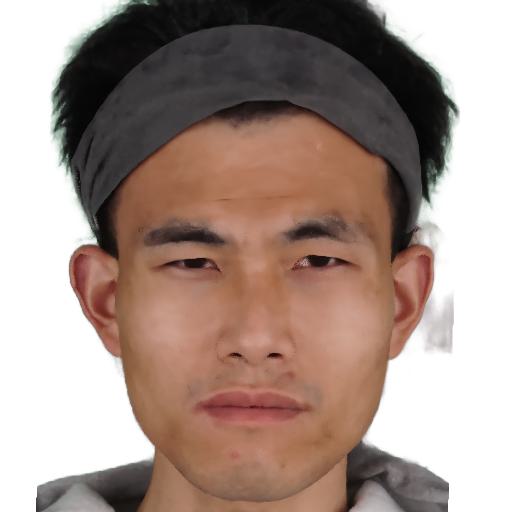}
    \includegraphics[width=0.09\textwidth]{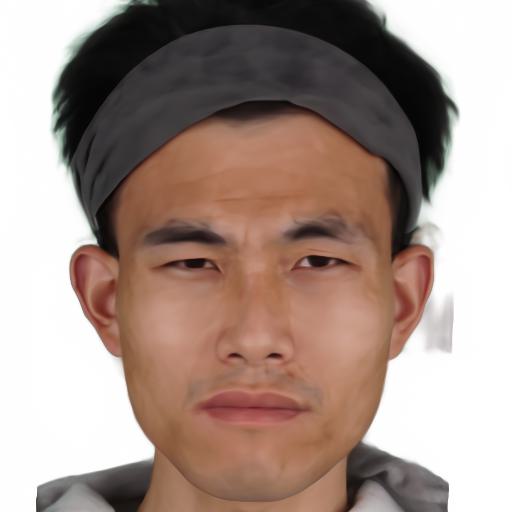}
    \includegraphics[width=0.09\textwidth]{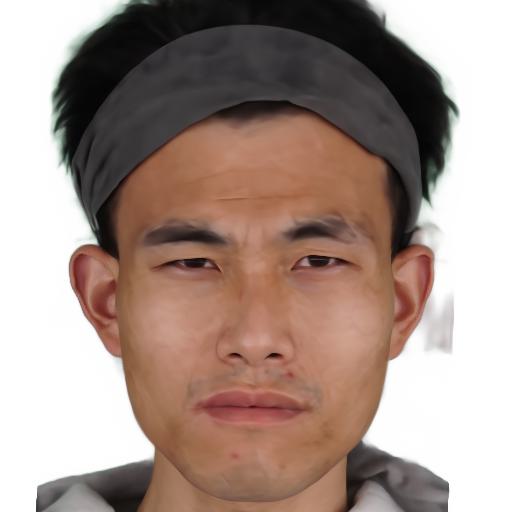}
    \includegraphics[width=0.09\textwidth]{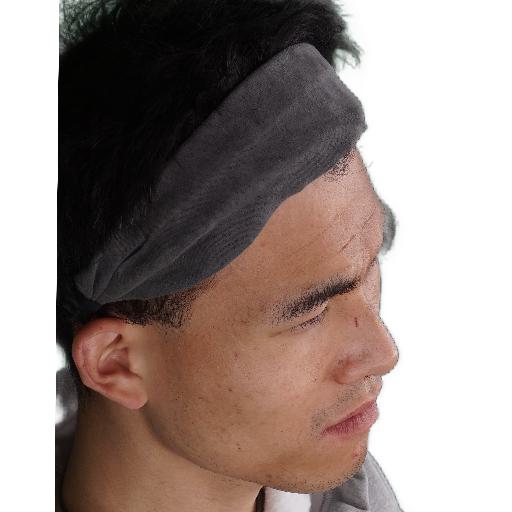}
    \includegraphics[width=0.09\textwidth]{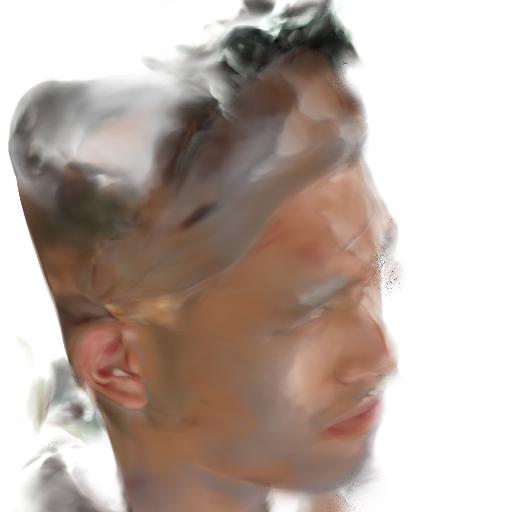}
    \includegraphics[width=0.09\textwidth]{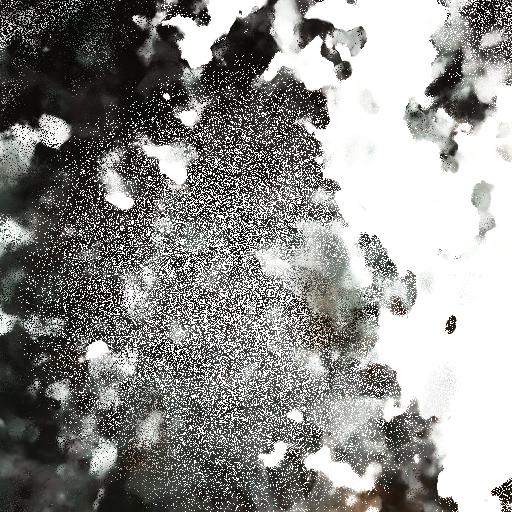}
    \includegraphics[width=0.09\textwidth]{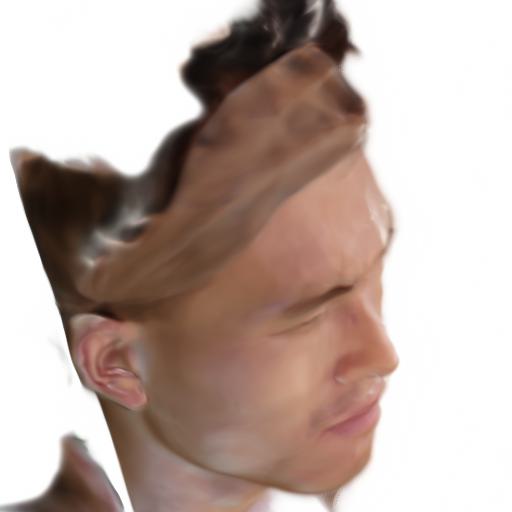}
    \includegraphics[width=0.09\textwidth]{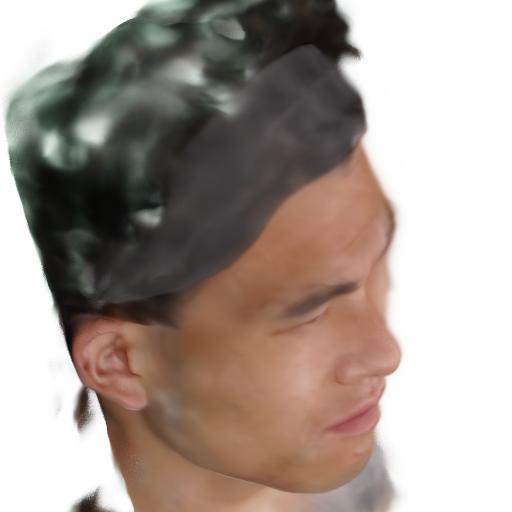}
    \rotatebox{90}{\tiny}
    \includegraphics[width=0.09\textwidth]{./results/template_effects/gt_571_blank.jpg}
    \includegraphics[width=0.09\textwidth]{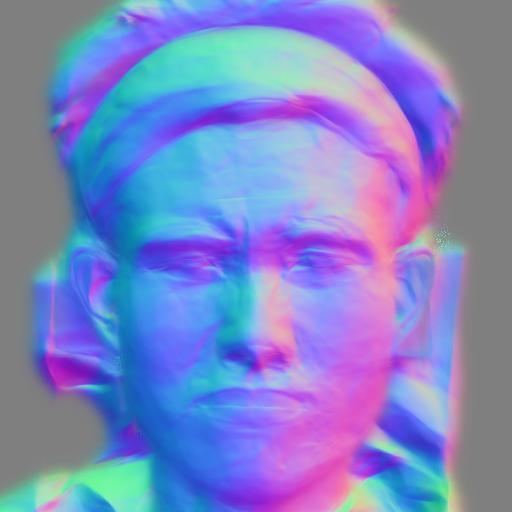}
    \includegraphics[width=0.09\textwidth]{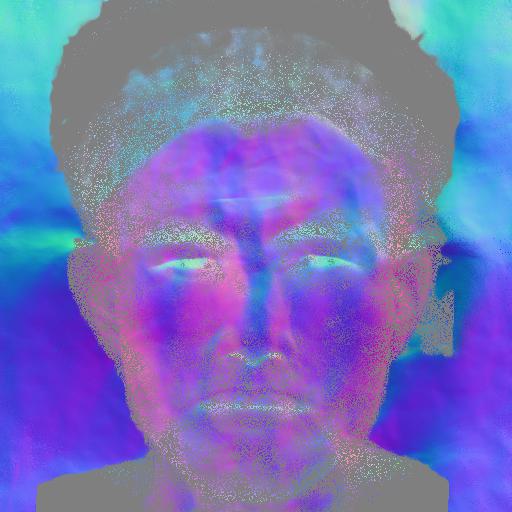}
    \includegraphics[width=0.09\textwidth]{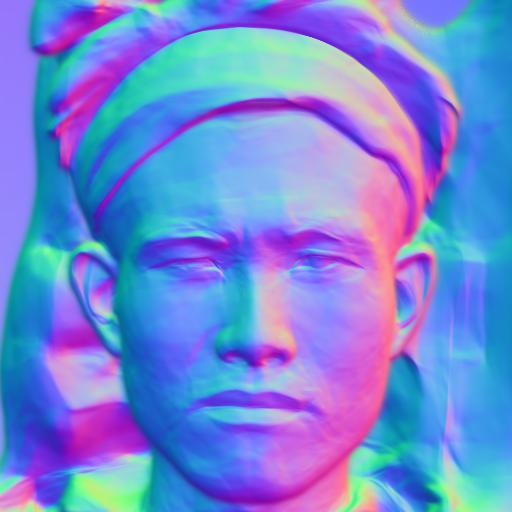}
    \includegraphics[width=0.09\textwidth]{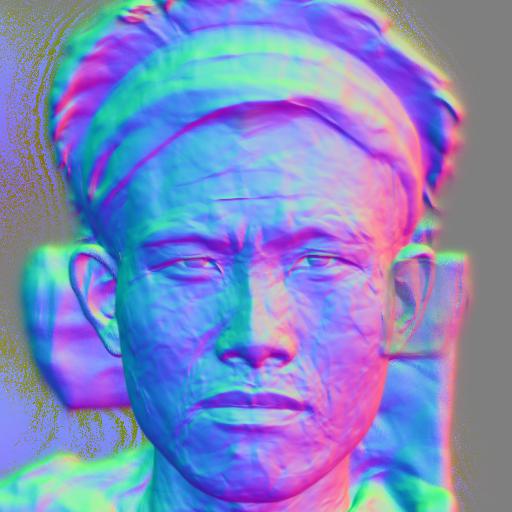}
    \includegraphics[width=0.09\textwidth]{./results/template_effects/gt_571_blank.jpg}
    \includegraphics[width=0.09\textwidth]{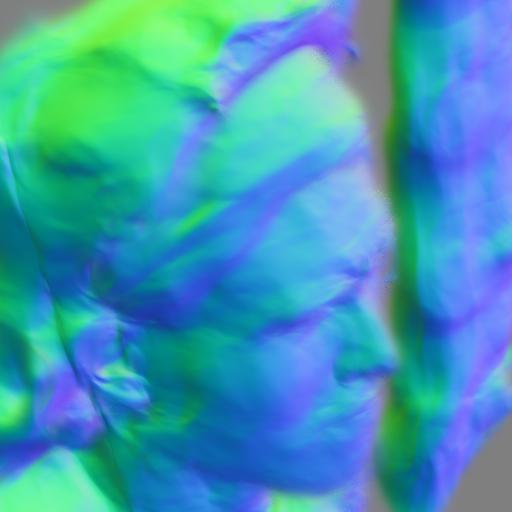}
    \includegraphics[width=0.09\textwidth]{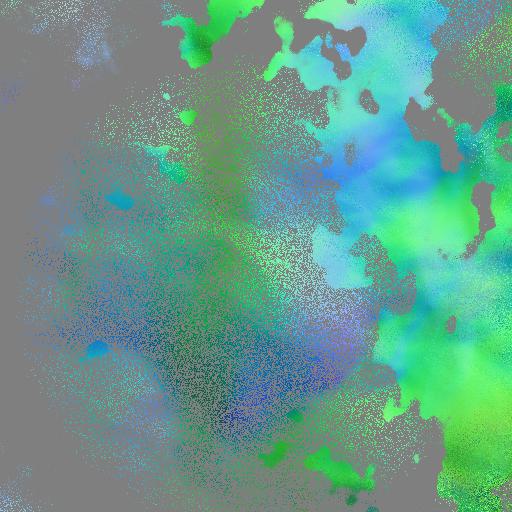}
    \includegraphics[width=0.09\textwidth]{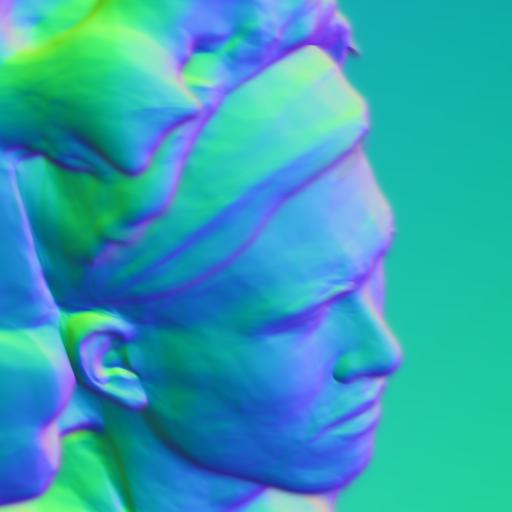}
    \includegraphics[width=0.09\textwidth]{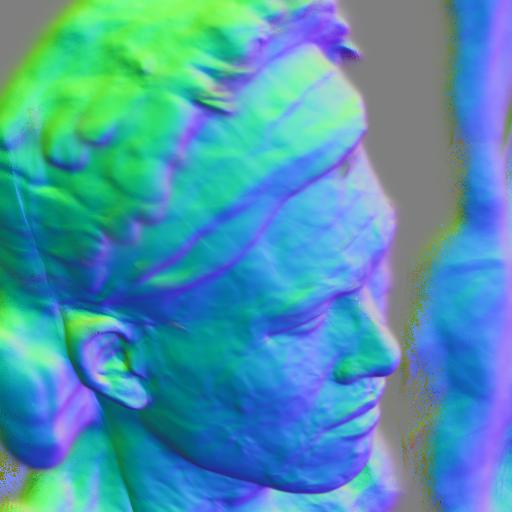}\\
    \makebox[0.09\textwidth]{GT}
    \makebox[0.09\textwidth]{NeuS}
    \makebox[0.09\textwidth]{HF-NeuS}
    \makebox[0.09\textwidth]{VolSDF}
    \makebox[0.09\textwidth]{Ours}
    \makebox[0.09\textwidth]{GT}
    \makebox[0.09\textwidth]{NeuS}
    \makebox[0.09\textwidth]{HF-NeuS}
    \makebox[0.09\textwidth]{VolSDF}
    \makebox[0.09\textwidth]{Ours}\\
    \caption{Comparison of various approaches under a 10-view setting (from Model 454 to Model 548).
    For each model, we show the results on one training view (left) and one novel view (right).}
    \label{fig:all_result4}
\end{figure*}

\end{appendices}

\end{document}